\documentclass[12pt]{article}
\pdfoutput=1

\usepackage{color}
\usepackage{graphicx}
\usepackage{amsmath}
\usepackage{amssymb}
\usepackage{bbm}
\usepackage{xspace}
\usepackage{slashed} %
\usepackage[small]{subfigure}
\usepackage[numbers,compress]{natbib}
\usepackage{epigraph}
\usepackage[normalem]{ulem} %
\usepackage{simpler-wick} %
\usepackage{physics}
\usepackage{tikz-feynman} 
\usepackage{tikz-feynhand}
\usepackage{silence}
\usepackage{etoolbox}
\makeatletter
\newlength\epitextskip
\pretocmd{\@epitext}{\em}{}{}
\apptocmd{\@epitext}{\em}{}{}
\patchcmd{\epigraph}{\@epitext{#1}\\}{\@epitext{#1}\\[\epitextskip]}{}{}
\makeatother

\usepackage[hyperfootnotes=false]{hyperref}

\usepackage{mciteplus} 

\newlength{\abstractwidth}
\renewcommand{\title}[1]{\vbox{\center\bf{\Large{#1}}}\vspace{5mm}}
\renewcommand{\author}[1]{\vbox{\center#1}\vspace{5mm}}
\newcommand{\address}[1]{\vbox{\center\em#1}}
\newcommand{\email}[1]{\vbox{\center\tt#1}\vspace{5mm}}
\newcommand{\sbreak}{
    \begin{center}
        $\blacklozenge$$\blacklozenge$$\blacklozenge$$\blacklozenge$
    \end{center}
}
\newcommand{\be}{\begin{equation}}
\newcommand{\ee}{\end{equation}}
\newcommand{\bea}{\begin{eqnarray}}
\newcommand{\eea}{\end{eqnarray}}
\newcommand{\bel}{\begin{align}}
\newcommand{\eel}{\end{align}}
\newcommand{\bi}{\begin{itemize}}
\newcommand{\ei}{\end{itemize}}
\renewcommand{\le}{\left}
\newcommand{\ri}{\right}

\definecolor{green}{rgb}{0.0, 0.64, 0.0}
\definecolor{purple}{rgb}{0.4,.2,0.7}
\definecolor{orange}{rgb}{1.0,0.65,0}
\definecolor{darkorange}{rgb}{1.0,0.40,0}

\newcommand{\nin}{N_{\text{in}}} %
\newcommand{\nf}{N} %
\newcommand{\nout}{C} %
\newcommand{\nA}{T} %
\newcommand{\nB}{\widehat{T}} %
\newcommand{\nl}{M} %
\newcommand{\nR}{N_{\text{ReLU}}} %
\newcommand{\din}{d_{\text{in}}} %
\newcommand{\dist}{\delta} %
\newcommand{\distavg}{\braket{\dist}} %

\newcommand{\fea}{\varphi} %
\newcommand{\featwo}{\psi} %
\newcommand{\feau}{\omega} %
\newcommand{\param}{\theta} %
\newcommand{\paramL}{\param^{\text{L}}} %
\newcommand{\paramNL}{\param^{\text{NL}}} %
\newcommand{\zL}{\widehat{z}^{\text{L}}} %
\newcommand{\zNL}{\widehat{z}^{\text{NL}}} %
\newcommand{\qeps}{\varepsilon} %
\newcommand{\ridge}{\gamma} %
\renewcommand{\L}{\mathcal{L}} %
\newcommand{\A}{\mathcal{A}} %
\newcommand{\B}{\mathcal{B}} %

\newcommand{\res}{q} %
\renewcommand{\Res}{Q} %
\newcommand{\resb}{\overline{q}} %
\newcommand{\Resb}{\overline{Q}} %

\newcommand{\Resbb}{\mathbb{Q}} %
\newcommand{\resbb}{\mathbbm{q}} %

\newcommand{\se}{S} %
\newcommand{\senF}{\se^F} %

\newcommand{\secF}{\Delta^{F}} %
\newcommand{\secFp}{\Delta^{F'}} %
\newcommand{\secL}{\Delta^{L}} %
\newcommand{\secLF}{\Delta^{L'}} %

\newcommand{\secG}{\Delta} %
\newcommand{\secm}{\Delta_{-1}} %
\newcommand{\nG}{\mu} %

\newcommand{\Pnf}{\widetilde P^{(4, \text{c} )}} %

\newcommand{\w}{w} %
\newcommand{\wvar}{\sigma_w^2} %

\newcommand{\fw}{u} %
\newcommand{\fwvar}{\sigma_u^2} %

\newcommand{\noise}{\epsilon} %
\newcommand{\vare}{\sigma_{\noise}^2} %

\newcommand{\covAA}{\Sigma} %
\newcommand{\covAB}{\widetilde{\Sigma}} %
\newcommand{\covBA}{\widetilde\Sigma^T} %
\newcommand{\covBB}{\widehat{\Sigma}} %
\newcommand{\covl}{\Lambda} %
\newcommand{\covf}{\Omega} %
\newcommand{\covlf}{\widetilde\Omega} %
\newcommand{\covfl}{\widetilde\Omega^T} %
\newcommand{\sqrtcovl}{v} %

\newcommand{\eig}{\lambda} %
\newcommand{\eigmin}{\lambda_-} %
\newcommand{\eigmax}{\lambda_+} %

\newcommand{\stdx}{\sigma_{x}} %
\newcommand{\varx}{\stdx^2} %

\newcommand{\IN}{I_{\nf}} %
\newcommand{\IT}{I_{\nA}} %
\newcommand{\IM}{I_{\nl}} %

\renewcommand{\bra}{\le\langle}
\renewcommand{\ket}{\ri\rangle}
\renewcommand{\braket}[1]{\bra #1 \ket} %
\renewcommand{\norm}[1]{\le|\!\le| #1 \ri|\!\ri|} %
\renewcommand{\tr}[1]{\text{tr}\!\le\{ #1 \ri\}} %
\renewcommand{\o}[1]{O\!\le(#1\ri)} %

\begin{document}

\hypersetup{pageanchor=false} %
\begin{titlepage}
\rightline{MIT-CTP/5463}
\begin{center}
\hfill \\

\title{A Solvable Model of Neural Scaling Laws}
\author{Alexander Maloney,$^{a\, \star}$ Daniel A. Roberts,$^{bc\, \star}$  and James Sully$^{de\,\star}$}
\address{$^{a}$ Department of Physics, McGill University, \\
Montr\'eal, Quebec H3A 2T8, Canada

\vspace{10pt}

$^{b}$ Center for Theoretical Physics {\it and} \\  Department of Physics, Massachusetts Institute of Technology \\ Cambridge, Massachusetts 02139, USA

\vspace{10pt}

$^{c}$ Salesforce, Cambridge, Massachusetts 02139, USA

\vspace{10pt}

$^{d}$ Department of Physics and Astronomy, University of British Columbia, \\Vancouver, BC V6T 1Z1, Canada

\vspace{10pt}

$^{e}$ Anthropic, San Francisco, California 94960, USA}

\email{alex.maloney@mcgill.ca, drob@mit.edu, jsully@anthropic.com}

\end{center}

\begin{abstract}
Large language models with a huge number of parameters, when trained on near internet-sized number of tokens, have been
empirically shown to obey \emph{neural scaling laws}: 
specifically, their
performance
behaves 
predictably
as a power law in either parameters or dataset size until bottlenecked by the other resource.
To understand this better, we first identify the necessary properties allowing such scaling laws to arise and then
propose a statistical model -- a joint generative data model and random feature model -- that captures this neural scaling phenomenology. 
By solving this model 
in the dual limit of large training set size and large number of parameters, 
we gain insight into 
\emph{(i)} the statistical structure of datasets and tasks that lead to scaling laws, 
\emph{(ii)} the way nonlinear feature maps, 
such as those provided by neural networks,
enable scaling laws when trained on these datasets, 
\emph{(iii)} the optimality of the \emph{equiparameterization} scaling of training sets and parameters,
and \emph{(iv)} 
whether such scaling laws can break down and how they behave when they do. 
Key
findings 
are
the manner in which the power laws that occur in the statistics of natural datasets are extended by \emph{nonlinear} random feature maps and then translated into power-law scalings of the test loss and how the finite extent of the data's spectral power law causes the model's performance to plateau.
\let\thefootnote\relax\footnotetext{$^{\star}$ Equal contribution.}

\end{abstract}

\end{titlepage}
\hypersetup{pageanchor=true} %

\tableofcontents

\section{Introduction}\label{sec:introduction}

Large language models (LLMs) such as GPT-3 \cite{Brown2020LanguageMA}, LaMDA \cite{thoppilan2022lamda}, and 
Palm \cite{chowdhery2022palm} 
have made fantastic advances in the generation of language, so much so that they can convincingly write text that fools humans into thinking it's written by other humans. Built from the transformer architecture \cite{attention2017}, these and similar dense LLMs \cite{rae2021scaling,zhang2022opt,megatron-nlg,hoffmann2022training} 
 are ``large'' as in \emph{size}, with 
Palm 
topping out at 540 billion parameters,
 and also ``large'' as in (big) \emph{data}, with Chinchilla \cite{hoffmann2022training} trained on 1.4 trillion tokens.
 These regime that these models operate in 
 -- jointly large parameter and large data --
differs from both
 the regime covered by classical statistical approaches to machine learning (see, e.g., \cite{hastie2009elements}) -- typically %
 an \emph{underparameterized} setting of large datasets and a fixed number of parameters and characterized by a bias-variance tradeoff -- and the regime typically studied by modern theoretical approaches to deep learning \cite{neal1996priors,lee2018deep,matthews2018gaussian,jacot2018neural,brainNTK2019,Yaida2019,PDLT-2022} -- 
 an \emph{overparameterized} setting of fixed datasets and a large number of parameters and characterized by \emph{interpolation} \cite{belkin-double} in which models memorize their training sets.

Inspired by the performance gains of the successive scaling up of LLMs, Ref.~\cite{kaplan2020scaling} comprehensively studied the test loss of such autoregressive transformer models trained on language model tasks across a large variety of model and dataset sizes. Impressively, they found that the overall performance can behave as a \emph{power law} in any of parameters, dataset size, and compute, so long as the model isn't bottlenecked by any of the other two. (See, e.g., Fig.~\ref{fig:pheno-loss-original}.) 
Moreover, by mapping the bottleneck and then jointly scaling parameters, data, and compute, practitioners can learn how to most efficiently apply their finite resources towards engineering bigger models, gathering more data, or burning their FLOPS.
Thus, given the breadth of this empirical investigation over a number of orders of magnitude, the existence of these \emph{neural scaling laws}, as they've been dubbed, have led many to believe a \emph{scaling hypothesis} \cite{branwen_2020}: performance on
language modeling tasks can be made
\emph{predictably}
good simply by taking current transformer models and continuing to scale up parameters, data, and compute.

After such a study, a number of follow ups 
appeared showing even more general applicability and more detailed understanding \cite{henighan2020scaling,hernandez2021scaling,gordon-etal-2021-data,hernandez2022scaling} -- and even improved performance scaling with data size from a power law to an exponential falloff with clever pruning~\cite{sorscher2022beyond}.\footnote{Earlier works with similar ideas include \cite{hestness2017deep}, which predicts the test loss for different deep learning scenarios and identifies a power law scaling with training set size, and \cite{rosenfeld2020a}, which models the scaling of performance with both data and model size.
}  At the same time, autoregressive generative modeling with transformers has continued to be applied to broader  AI tasks such as coding \cite{chen2021evaluating}, quantitative reasoning \cite{lewkowycz2022solving}, and even on the suite of computer vision tasks, with the advent of the Vision Transformer (ViT) \cite{dosovitskiy2020image} family of models.

Given the ever growing breadth of tasks that these models can accomplish \cite{srivastava2022beyond} and given their continuing gains in performance as we engineer ever bigger models and scrape ever bigger datasets, it 
is
increasingly important to understand the origin of these neural scaling laws. 
The set of important questions include: 
\bi
\item What are the properties of datasets and tasks that lead to scaling laws?
\item Which 
classes of models
support scaling laws when trained on these datasets?
\item How do scaling laws arise, or what mechanism leads to such predictive behavior?
\item Can this predictive behavior break down, and what happens in such regimes?
\ei
Addressing these questions can not only help us improve our AI systems practically, but 
also
 help us 
 better understand the structure of AI tasks -- such as language modeling -- that seem to require gigantic amounts of data to reach (approximate) human-level performance.\footnote{How do we understand the contrast between Chinchilla \cite{hoffmann2022training}, trained on 1.4 trillion tokens, and a human, for which the size of the training set is perhaps only of order 10 million words \cite{human-language-learning,linzen2020can}? By another estimate, LLMs may receive 1000x the linguistic data that a typical ten-year-old child might have received \cite{human-language-position}.
 }

In this paper, we will 
provide some initial answers to these questions
by jointly studying a \emph{generative data model} and \emph{random feature model} that together exhibit neural scaling laws analogous to those found in \cite{kaplan2020scaling}. This provides a theoretical framework for 
deriving
the observed phenomenology in 
a class of large-parameter and big-data models,
just like
the ``microscopic'' framework of \emph{statistical mechanics} can be used to derive the ``macroscopic'' laws of \emph{thermodynamics} in physics. We will explain how our model captures the essential statistical aspects of natural datasets and of the feature representation of nonlinear networks, 
and then we will systematically solve this joint model to compute its test loss 
as a function of \emph{both} dataset size and number of parameters.\footnote{As our analysis makes use of an exact optimization solution, it's effectively in the regime of ``infinite'' 
compute: 
thus, our framework can only teach us about 
tradeoffs between 
data and parameter resources.
} 
We will thus show how our model matches the empirically observed behavior of LLMs, 
and then we will use this setting to better understand how scaling laws arise and break down.

One of our main results is a lack of universality of scaling laws across differently structured data generation processes: datasets that lead to scaling laws have a particular 
power-law structure in their spectral statistics, which ultimately leads to a power-law scaling of the test loss when there are no resource bottlenecks present. Moreover, we find that an essential role of nonlinear feature maps 
is 
extending the
power law in the spectrum of the 
representation as a function of the number of features. 
This ability to extend the power law
differentiates the performance of different deep neural network (DNN) models and, although we don't investigate it here, is presumably an important reason  why -- from the perspective of this analysis -- transformers enable neural scaling law phenomenology. 
Finally, for generalized linear models -- i.e., linear regressions 
of potentially nonlinear feature maps -- we learn that exact \emph{equiparameterization} 
-- scaling the number of features identically with the size of the training set -- 
is optimal when some kind of regularization is applied.\footnote{
    This is consistent with the finding of \cite{hoffmann2022training}, though is slightly counter to the initial empirical results in \cite{kaplan2020scaling}. However, both of those references concern empirical investigations of LLMs, while our 
    analysis
    concerns generalized linear models and may not apply in the same way for nonlinear models that learn representations. (See \S\ref{sec:future-directions} under the subheading \emph{Representation Learning?} for further discussion.)
} 
Intuitively, for the sort of data that leads to scaling laws, each additional 
sample
can be used to learn about an additional feature in the latent feature space, and the model should have an additional parameter in order to 
represent the information from this new latent feature. %

An important insight that emerges from our analysis
is the role of a new scale 
that 
determines when the empirical behavior found by \cite{kaplan2020scaling}
breaks down. 
This scale can be understood as the size of the \emph{latent space} 
from which the data is generated
and must be much larger than \emph{both} the size of the training set and the number of parameters of the model in order to observe the power-law scaling and bottleneck behavior of Ref.~\cite{kaplan2020scaling}.\footnote{
    This is perhaps
    surprising given a general expectation that natural data should live on manifold of smaller intrinsic dimension than its embedding dimension; see \S\ref{sec:coding-theory} for further discussion.
}
If either
of these two resource scales exceed the size of the latent space, our analysis 
shows a new regime of different behaviors for the test loss that has not yet been seen in the LLM experiments. 
Since we have a generative model of the data we control this scale directly in our analysis, but it would be extremely interesting to understand this scale in natural data, such as images or text.

As the main set of tools we use to solve our joint data and feature model is random matrix theory (RMT), one of the technical contributions in this paper is the development of a diagrammatic approach -- borrowed from theoretical physics -- for computing random matrix expectations in machine learning. 
Our techniques are particularly well suited to the regime of jointly large dataset and number of parameters, in which a restriction to diagrams with a \emph{planar} topology captures the dominant contributions to RMT expectations.
While such techniques will be familiar to physicists, 
we were able to apply them here to vastly 
simplify a number of previous RMT derivations 
in machine learning; 
for example, we avoid the lengthy concentration-of-measure proofs of \cite{louart2018random}, and we also sidestep the need to use replica methods as in \cite{bordelon2020spectrum,Canatar:etal}.

While most of the previous work on scaling laws has been empirical, there have been a few papers that have sought theoretical explanations or models of neural scaling laws \cite{JMLR:v23:20-1111,bahri2021explaining,hutter2021learning,wei2022more}, and a few others that have found a power-law scaling of the test loss with dataset size \cite{spigler2020asymptotic,bordelon2020spectrum}.
The most directly relevant of these, \cite{bahri2021explaining}, considers a student-teacher model with the right phenomenology though studies it only in certain limits: in particular, the authors assume a power law of \emph{infinite} extent for the data and then import a result from \cite{Canatar:etal} to find a power-law scaling of the test loss; our work leaves the relative ratios of the size of the latent space, the size of the training set, and the number of features all finite.
Additionally, the calculation of \cite{Canatar:etal} gives the averaged test loss of a non-generalized linear regression without any random feature map to extend the power law.
Our work is more closely related to \cite{louart2018random}, which was not focused on scaling laws but finds a general expression for the test loss averaged over nonlinear random feature maps with fixed training data and labels; we extend this further to student-teacher models, finding a very simple expression for the test error and then 
are able to average it over our random data model as well.
In the process, we also find much simpler derivations of the central results in \cite{louart2018random} and \cite{Canatar:etal} using our diagrammatic techniques.

\sbreak

\noindent{}The plan of this paper is as follows: 

In \S\ref{sec:overview}, we provide a non-technical overview of the data and feature map settings that lead to neural scaling laws: 
to begin, 
we review the phenomenology of the joint parameter and dataset test loss 
behavior
discovered by \cite{kaplan2020scaling}; then, in \S\ref{sec:data-properties} and \S\ref{sec:model-properties}, we use examples from natural datasets to explain the specific spectral properties of the input data and nonlinear feature maps, respectively, needed to have both power law scaling and bottleneck behavior. 

In \S\ref{sec:derivation}, we present (\S\ref{sec:derivation-notation}) and then solve (\S\ref{sec:data-and-feature-averages}) our statistical model of neural scaling laws, consisting jointly of a generative  model for the data and random feature map, deriving a formula for the test loss as a function of the size of the dataset and number of features of the model and then showing that it precisely matches experiment. %
We also outline (\S\ref{sec:spectral-extension}) 
how we could use our same RMT tools to model 
spectral power-law extension
in nonlinear feature maps and comment (\S\ref{sec:other-methods}) on the relationship of our work to other RMT results in the machine learning literature.

In \S\ref{sec:discussion}, we interpret our calculations from the previous section and expand on our results. Most importantly, in \S\ref{sec:break-down-of-neural-scaling-laws}, we characterize the \emph{breakdown} of neural scaling law behavior 
in our model by considering our result from \S\ref{sec:derivation} 
in the limit where the size of the latent space becomes smaller than either the size of the training set or the number of features in the model. We also confirm the validity of our calculation in this limit by comparing against numerical simulations in the same regime. 
Then, in \S\ref{sec:double-descent} we explain the optimality of the \emph{equiparameterizated} regime for neural scaling, 
contrasting with the overparameterized regime and discussing the double descent phenomenon,
while in \S\ref{sec:coding-theory} we further consider our new scale that controls the size of the latent space and the breakdown of scaling laws in the context of traditional notions of dimensionality reduction.
We close in \S\ref{sec:breakdown-data-model} by discussing some limitations of our minimal power-law spectral data model
that could be improved in future analyses.

Finally, in \S\ref{sec:future-directions} we  
conclude and give an outlook towards a future research direction.
In particular, we provide a guide on how one could use the tools from \cite{PDLT-2022} to move beyond our random-feature linear regression and incorporate the type of representation learning present in nonlinear models, such as those used in realistic deep learning scenarios.

To make the paper tractable, a few additional analyses and technical details have been consigned to appendices. 
In Appendix~\ref{sec:other-models}, we present and solve a progression of 
simpler data and feature models with increasing complexity: 
first,
(\S\ref{sec:linear-mp-model})  we show that the simplest possible model, a linear model where the 
input data has independent and identically distributed Gaussian components
-- i.e. data with Marchenko-Pastur spectral statistics -- does not have scaling laws; 
then, 
(\S\ref{sec:linear-power-law-model})
we
explain why linear regression on data sampled from a more realistic data model -- but without any feature mapping -- also does not exhibit the right behavior.
Finally, in Appendix~\ref{sec:delta} we 
explain 
how to find 
analytical formulae 
for the trace of the resolvent with the covariance matrix, the quantity that controls the test loss of our model.

\section{Prerequisites for Neural Scaling}\label{sec:overview}

\begin{figure}[ht]
\begin{center}
 \includegraphics[width=0.8\linewidth]{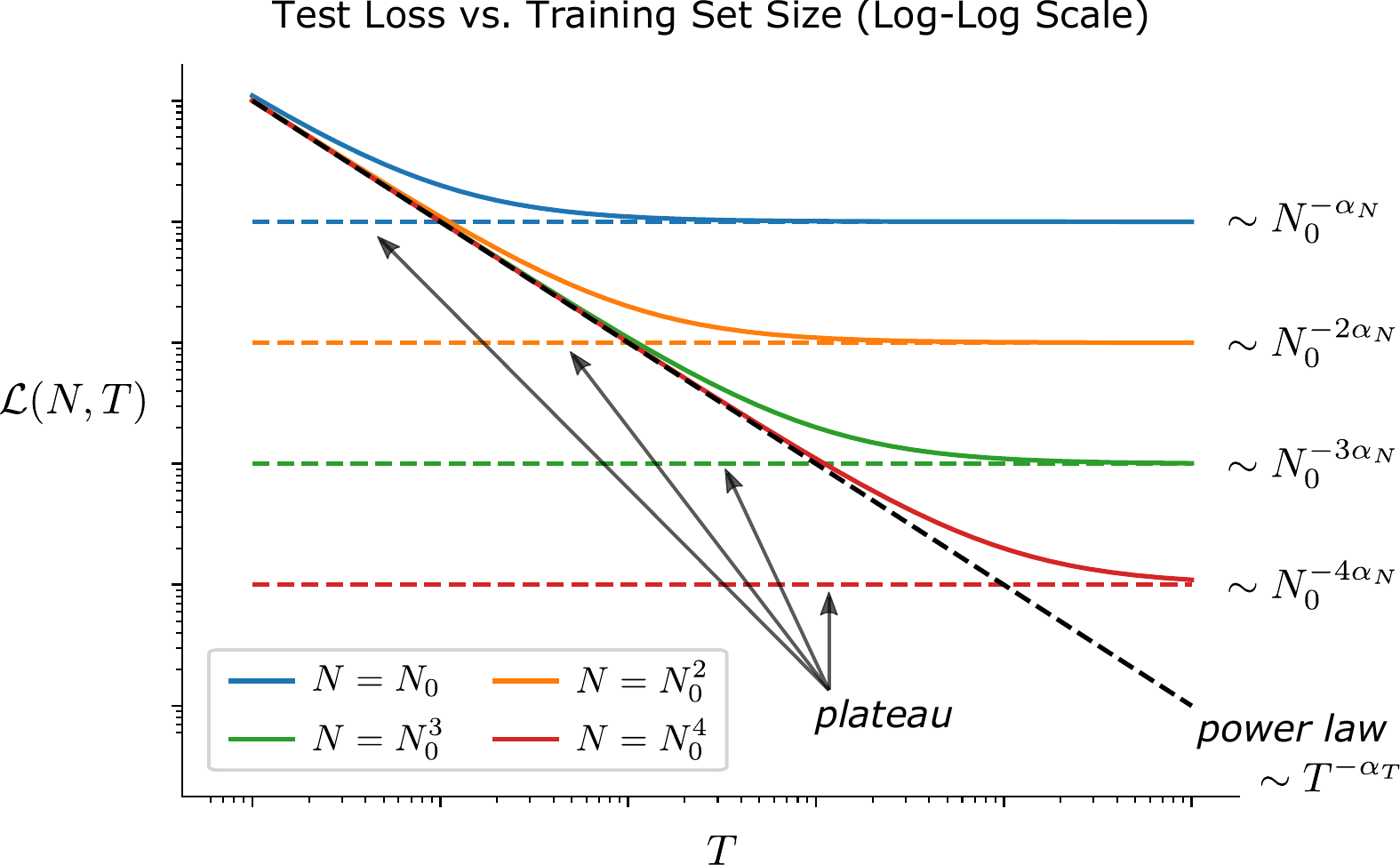}
\end{center}
\caption{
Cartoon plot of the empirical scaling laws discovered by Ref.~\cite{kaplan2020scaling} demonstrating that the test loss of LLMs trained with early stopping are predictably described by a simple phenomenological model, \eqref{eq:phenomenological-loss-original},
plotted as a function of dataset size, $\nA$, for different model sizes, $\nf = \{\nf_0, \nf_0^2,\nf_0^3,\nf_0^4 \}$:
 if the model isn't bottlenecked by the number of parameters ($\nf \to \infty$), the test loss behaves as a \emph{power law} in the training set size, $\L(\nf, \nA) \sim \nA^{-\alpha_\nA}$; otherwise, if the number of parameters is too small for a given training set, then the test loss stalls at a \emph{plateau} at a value that depends predictably on the parameters, $\L(\nf, \nA) \sim \nf^{-\alpha_\nf}$. 
Similar statements hold reversing the role of the training set and parameter resources, 
and scaling both training set and parameters jointly with relative ratio $\nf \sim \nA^{\alpha_{\nA}/\alpha_{\nf}}$ ensures the overall best performance.
}
\label{fig:pheno-loss-original}
\end{figure}

An exciting empirical observation of \cite{kaplan2020scaling} was that the test loss for large-scale transformer models \cite{attention2017} can be \emph{predicted} by fitting by a \textbf{phenomenological model} of an extremely simple form:
\be\label{eq:phenomenological-loss-original}
\L(\nf, \nA) \, = \le[ \le(\frac{\nf_c}{\nf} \ri)^{\frac{\alpha_{\nf}}{ \alpha_{\nA}} } + \frac{\nA_c}{\nA} \ri]^{\alpha_{\nA}}.
\ee
Here, $\nf$ is the number of (non-embedding) parameters characterizing the size of the model, $\nA$ is the number of datapoints in the training set characterizing how many examples the model can learn from, and  $\nf_c$, $\nA_c$, $\alpha_{\nf}$, and $\alpha_{\nA}$ are all fit constants.
A cartoon plot of \eqref{eq:phenomenological-loss-original} as a function of the training set size, $\nA$, and for a variety of different model sizes, $\nf$, is shown in Fig.~\ref{fig:pheno-loss-original}.

This formula is quite interesting for a number of reasons: 
\begin{enumerate}
\item[\emph{(i)}] On the one hand, taking the training set, $\nA$, or the model size, $\nf$, to be large the model improves as a power law in the scaled parameter:
\begin{align}\label{eq:original-power-law-scaling-law-parameters}
\L(\nf) &\equiv \L(\nf, \infty) \sim \nf^{-\alpha_{\nf}} \, , \\
\L(\nA) &\equiv \L(\infty, \nA) \sim \nA^{-\alpha_{\nA}} \, ,
\label{eq:original-power-law-scaling-law-trainingset}
\end{align}
where we now see that $\alpha_{\nf}$ is the \textbf{scaling exponent} characterizing the behavior of the loss as the number of parameters is increased, and $\alpha_{\nA}$ is the scaling exponent characterizing the its behavior as the size of the training set is increased. The first \textbf{scaling law}, \eqref{eq:original-power-law-scaling-law-parameters}, is particularly interesting in practice as often  LLMs are effectively trained in infinite data regime; in this case, we achieve predictable and continuing performance gains for increases in the size of our models.\footnote{
    The Chinchilla model and related investigation \cite{hoffmann2022training} suggest that if we train long enough, data might actually be a bottleneck, or soon will be in the future.
} If there is truly infinite data, this means that for tasks and models that have scaling law \eqref{eq:original-power-law-scaling-law-parameters}, we can become arbitrarily good at those tasks simply by engineering bigger and bigger models. More realistically, we get a scaling law like \eqref{eq:original-power-law-scaling-law-parameters} so long as the size of the model is much smaller than training set size, $\nf \ll \nA$, and we get a scaling law like \eqref{eq:original-power-law-scaling-law-trainingset} so long as the size of the training set is much smaller than the size of the model, $\nA \ll \nf$.\footnote{
    More precisely, the relative scaling to get a power law \eqref{eq:original-power-law-scaling-law-parameters} should be stated as
    $\nf \ll \nA^{\alpha_{\nA} / \alpha_{\nf} }$,
    while the relative scaling for power law \eqref{eq:original-power-law-scaling-law-trainingset} should be stated as $\nA \ll \nf^{\alpha_{\nf} / \alpha_{\nA}}$; analogously, we should more precisely have the reverse of these relations for accessing the plateau regimes in  \eqref{eq:original-pleateau-parameters} and \eqref{eq:original-pleateau-trainingset}.
}
\item[\emph{(ii)}] On the other hand, once the scaled parameter exceeds the fixed parameter, the test loss asymptotes to a \textbf{plateau} that depends on the fixed parameter.
For instance, studying the test loss as a function of the parameters for $\nf \gg \nA$ 
just gives a constant,
\be\label{eq:original-pleateau-parameters}
\L_{\text{plateau}}(\nf)\equiv\lim_{\nf \gg \nA}\L(\nf, \nA) = \le(\frac{\nA_c}{\nA}\ri)^{\alpha_{\nA}} \, ,
\ee
while analogously studying the loss as a function of the training set size for $\nA \gg \nf$ gives a similar constant,
\be\label{eq:original-pleateau-trainingset}
\L_{\text{plateau}}(\nA)\equiv\lim_{\nA \gg \nf}\L(\nf, \nA) = \le(\frac{\nf_c}{\nf}\ri)^{\alpha_{\nf}} \, .
\ee
This means that the model performance can be inhibited by a \textbf{bottleneck} when either the size of the training set or the size of the model is limited. This has practical consequences as well: 
when the performance is bottlenecked by the training data, no matter how good our engineering talent is in training larger and larger models, the loss will not be able to improve further. Of course, in this case, if instead of building a larger model we collect more training data and reinterpret \eqref{eq:original-pleateau-parameters} as a function of the training set size, e.g. as \eqref{eq:original-power-law-scaling-law-trainingset}, the loss will again improve as a power law, though it will be in the training set size, $\nA$.
\item[\emph{(iii)}] 
On our other other hand,
we can instead interpret the test loss scaling with either resource as a power law when not bottlenecked by the other in the following way: if we jointly scale both the number of parameters and the size of the training set in a particular way, then we can always achieve power-law gains in our performance and avoid the plateau behavior.
For instance, if we parameterize the size of our model in terms of the amount of training data we collected and make a power law ansatz, $\nf(\nA) \sim T^p$, 
we can then make both terms
inside the square brackets of \eqref{eq:phenomenological-loss-original} contribute equally, ensuring the loss overall decreases as a power law, 
by scaling our model size as
\be\label{eq:original-tradeoff-powerlaw}
\nf(\nA) = \nf_c \!\le(\frac{\nA}{\nA_c}\ri)^{\frac{\alpha_{\nA}}{\alpha_{\nf}}}\, ,
\ee
where the power $p \equiv \alpha_{\nA} / \alpha_{\nf}$ controls how the size of the model scales as we grow the training set.
\end{enumerate}
In principle this phenomenological model of the test loss, \eqref{eq:phenomenological-loss-original}, predicts that these LLMs -- and any related models 
that %
can be fit by this equation
-- can become arbitrarily proficient at their underlying tasks so long as we continue to jointly scale both training data and model size as \eqref{eq:original-tradeoff-powerlaw}.

Given that \eqref{eq:phenomenological-loss-original} is an empirical observation over some range of training set sizes and models sizes, and for a particular set of AI tasks and deep learning architectures, it's natural to wonder how general it actually is, and whether 
the behavior will continue
for especially large scales $\nf$ and $\nA$. In fact, there are a number of details leading to the behavior \eqref{eq:phenomenological-loss-original} that are so far implicit in this discussion and should be made explicit. For instance, 
the fit of the exponents, and the relative scaling, \eqref{eq:original-tradeoff-powerlaw}, can depend on the details of the learning algorithm;\footnote{
In the original scaling laws paper, \cite{kaplan2020scaling}, the power-law exponents for the training set and parameters were measured to be different, 
with their ratio positive, $\alpha_{\nA} / \alpha_{\nf} > 1$; 
in contrast, 
Ref.~\cite{hoffmann2022training} found equal exponents, $\alpha_{\nA} = \alpha_{\nf}$, by training their models longer and on more training data. 
}
and to find the fit \eqref{eq:phenomenological-loss-original},  the authors of \cite{kaplan2020scaling} needed to regularize their models, e.g., by using early stopping. Most importantly, in \cite{kaplan2020scaling} the authors trained on a specific natural dataset: an extended WebText dataset built from human language~\cite{radford2019language}.
Thus, by identifying the mechanism that leads to such scaling laws, we will see that they arise for far more general dataset--model combinations.

In the rest of this section we will discover  the properties of the data (\S\ref{sec:data-properties}) and model (\S\ref{sec:model-properties}) that must go into a minimal modeling scenario that contains a version of \eqref{eq:phenomenological-loss-original}, including both the \emph{scaling law} limit and the \emph{plateau} limit. 
In particular, we'll explain how the data distribution must have special statistical properties, which we'll identify in natural datasets, and how a machine learning model must transform those statistical properties in a special manner, which we'll see is generically present in nonlinear DNNs.

\subsection{Data Properties}\label{sec:data-properties}

AI tasks in different domains have very different underlying data -- the input token features of textual data used to train LLMs for natural language processing (NLP) is a priori very different than the input pixel features of image data used for computer vision (CV) applications -- yet, as we will see, both domains can exhibit the full neural scaling law phenomenology of  \eqref{eq:phenomenological-loss-original}. However, purely random data without any structure does not.\footnote{See Appendix~\ref{sec:linear-mp-model} where we analyze data with Marchenko-Pastur statistics.
}
Thus, to understand such behavior, we should try to 
identify the structure universal to these natural datasets that exhibit scaling-law behavior.

Consider a generic raw data point, $x_\alpha$, in a dataset of $\nA$ samples, for $\alpha = \{1, \dots, \nA \}$.
For our data model, we will think of each $x_\alpha$ as being sampled independently from some distribution $p(x)$ and refer to a particular one as a \textbf{sample}. 
We will denote the individual components of a sample as
\be
x_{i;\alpha} \,, \quad \text{with} \quad i= 1, \ldots, \nin \,,
\ee 
where 
$i$ indexes the $\nin$ different \textbf{input features}, and which is supposed to represent, e.g., a particular pixel or token.\footnote{
    Technically for NLP, we want to first pass our input tokens through a fixed embedding so that $x_i$ represents a component of the embedded token. 
}
Accordingly, the statistics of the dataset 
endow the data with the structure that allows for the power law and plateau in the test loss.

Intuitively, 
the correlations between the different input features, $x_{i;\alpha}$, should 
characterize
the dataset. For instance, if the $x_{i;\alpha}$ are pixels of an image, we may expect that different pixels will vary similarly across images that are similar. 
In contrast, the mean value of an input feature is uninformative, and so we will assume our data is centered in a preprocessing stage.
Thus, a object of interest for us will be the \textbf{empirical feature-feature covariance matrix} of the dataset:
\be\label{eq:estimator-of-feature-feature-covariance-matrix}
\frac{1}{\nA}\sum_{\alpha =1}^{\nA} x_{i;\alpha} x_{j;\alpha}\, .
\ee
As this covariance matrix will typically contain nonzero off-diagonal components, instead it will be simpler to consider its eigenvalues. Let us denote a particular eigenvalue as $\eig_i$, and we'll refer to all of the eigenvalues 
collectively
as the \textbf{spectrum} of the data.
Note that since 
the covariance, \eqref{eq:estimator-of-feature-feature-covariance-matrix},
is a $\nin$-by-$\nin$-dimensional matrix, 
there are $\nin$ eigenvalues
in the spectrum.
Moreover, if the size of the dataset is smaller than the number of input features,
$\nA < \nin$,
then the rank of the covariance %
is at most $\nA$, 
and at least $\nin - T$ of those eigenvalues will be zero.

\begin{figure}[ht]
\begin{center}
 \includegraphics[width=0.49\linewidth]{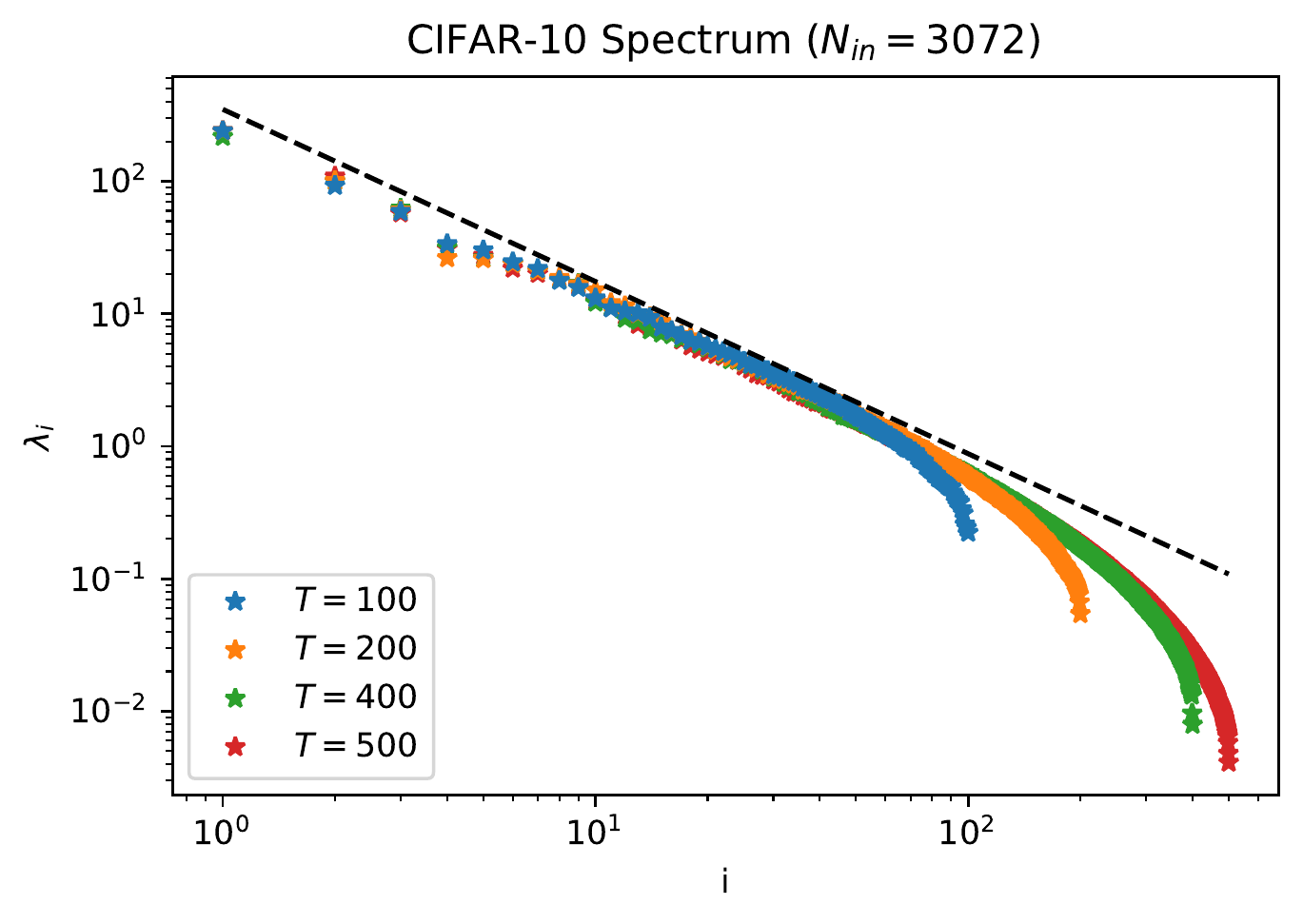}
 \includegraphics[width=0.49\linewidth]{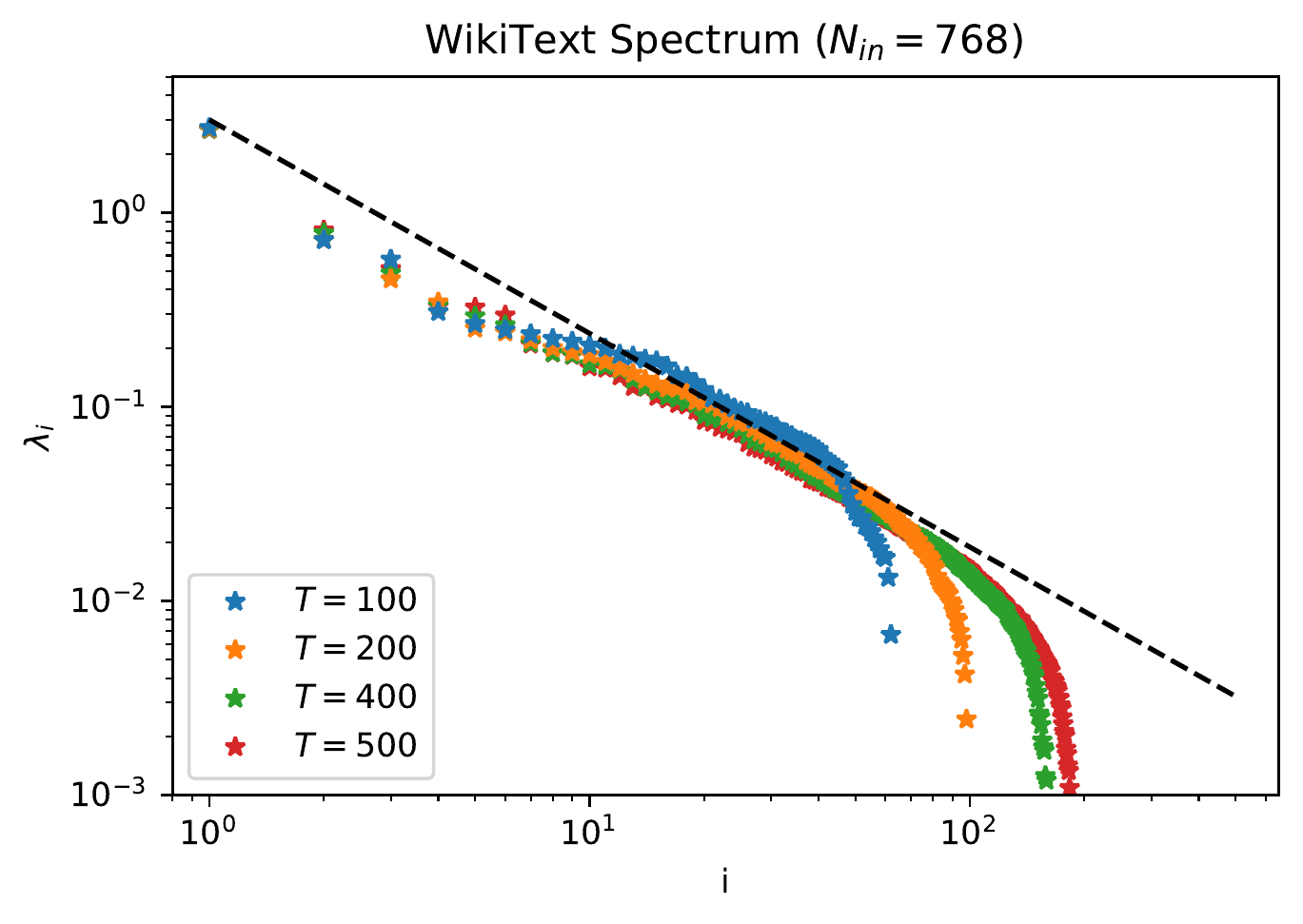}
\end{center}
\caption{Log-log plot of example spectra for different dataset sizes, $\nA$, from different data domains. Increasing the dataset size, $\nA$, increases the extent of the approximate power-law fit 
(dashed line) so long as $\nA < \nin$.  \textbf{Left:} CIFAR-10  \cite{cifar-dataset}, a CV dataset of $32 \times 32$-pixel natural color images. The $3$ color channels bring the total number of input features per image is $\nin = 3 \times 32 \times 32= 3072$.
\textbf{Right:} WikiText, an NLP dataset taken from the verified Good and Featured articles on Wikipedia \cite{merity2016pointer}. The input data was tokenized and then embedded using
Hugging Face's implementation \cite{wolf2019huggingface} of GPT-2~\cite{radford2019language}, and the embedding we use has dimension $\nin = 768$.
}
\label{fig:spectrums}
\end{figure}

\begin{figure}[ht]
\begin{center}
 \includegraphics[width=0.49\linewidth]{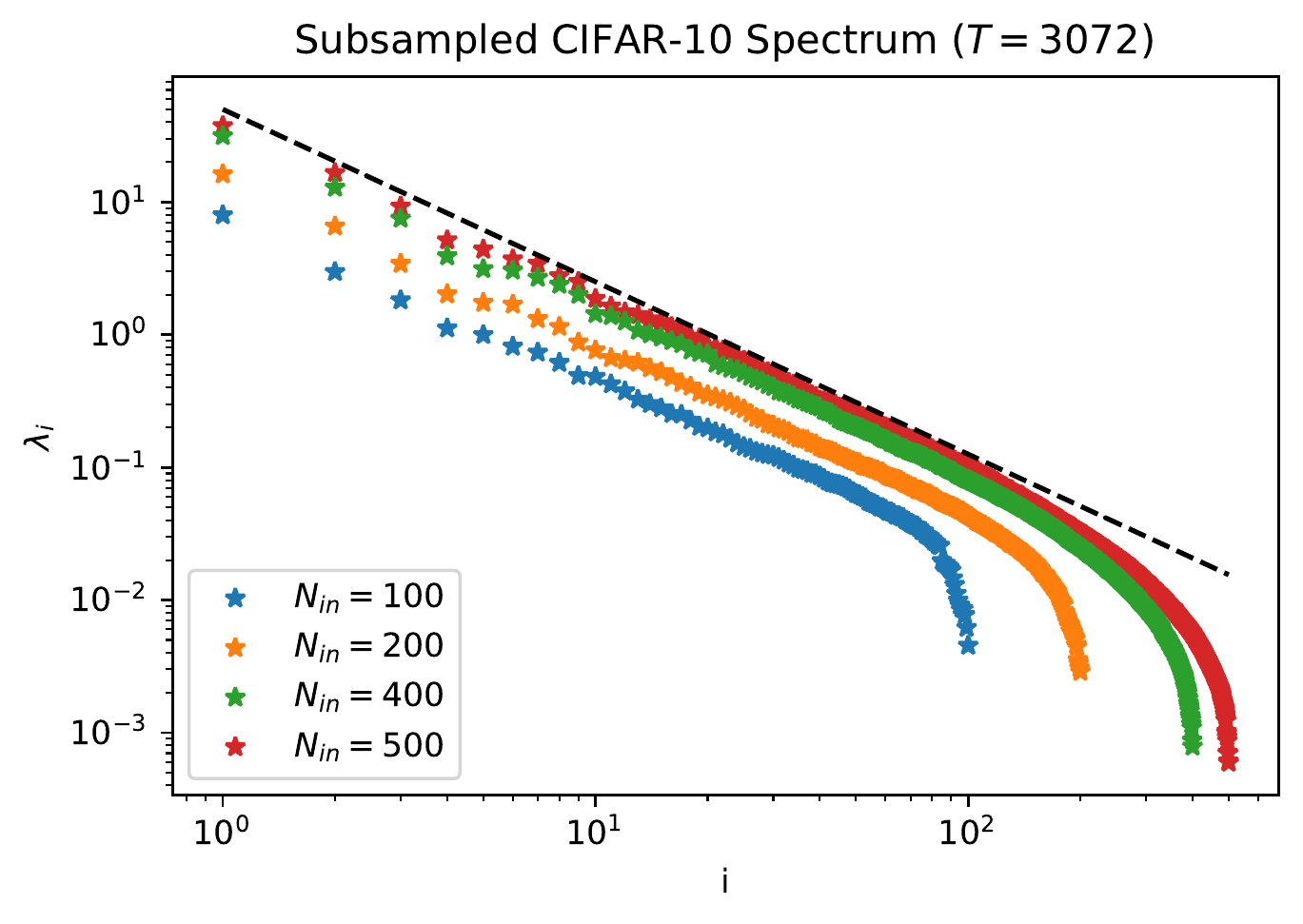}
 \includegraphics[width=0.49\linewidth]{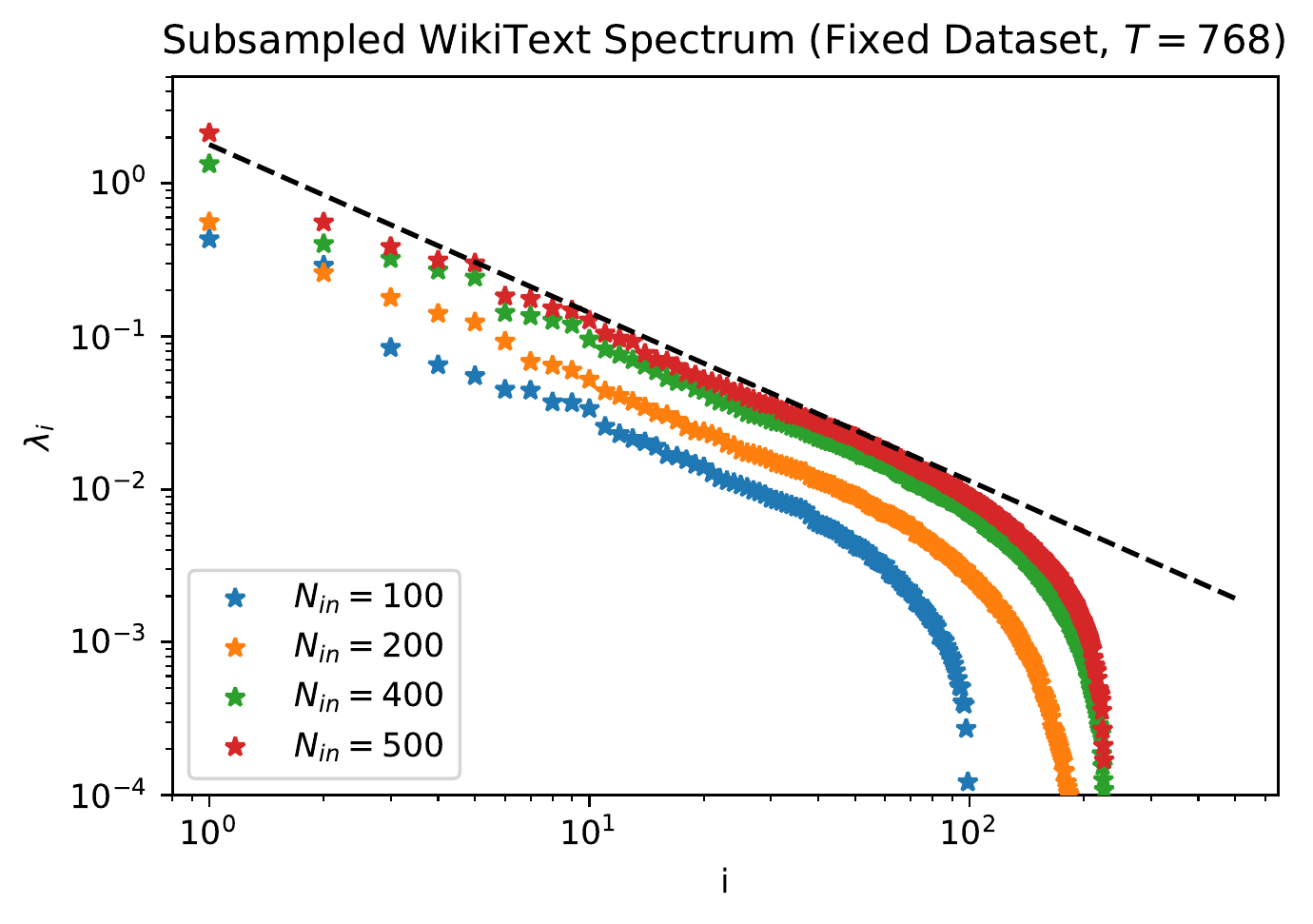}
\end{center}
\caption{Example spectra from different data domains for a fixed dataset size and subsampled input features. (For a more detailed description of the datasets, see the caption of Fig.~\ref{fig:spectrums}.) Increasing the number of input features in the subsample extends the length of the approximate power-law fit for the bulk (dashed line).  
\textbf{Left:} CIFAR-10, with pixels subsampled from the total $3072$ input features and a dataset size of $\nA=3072$.
\textbf{Right:} WikiText, 
with the components of the embedding subsampled from the total $768$-dimensional embedding vector for each token and a dataset size of $\nA=768$.
}
\label{fig:spectrums-fixed-dataset}
\end{figure}

The spectrum is a simple summary quantity %
that we can use to characterize a dataset. To see why, let's look at some spectra from real natural datasets in different domains: in Fig.~\ref{fig:spectrums} we plot example spectra for different dataset sizes, $\nA$, from a computer vision image dataset (left panel) and a tokenized and embedded natural language dataset (right panel).
Right away, we see that both these representative natural datasets have interesting and in fact, very special structure to their spectra:
\begin{enumerate}
\item[\emph{(1)}] For a few orders of magnitude beginning from around $i \approx 10$, the spectra are well fit by a power law,
\be\label{eq:eig-by-index-observation}
\eig_i \sim  \frac{1}{i^{1+\alpha} }\, ,
\ee
where here we've decomposed the exponent of decay as $1+\alpha$ with some foresight.\footnote{
    This point was also emphasized by \cite{bahri2021explaining}. Note that when the test loss has scaling law phenomenology, the exponent $\alpha$ should be a positive real number, $0 < \alpha < \infty$.
} 
\item[\emph{(2)}] For each fixed size $\nA$,  the power law 
terminates in very rapid decline in the value of the eigenvalues, $\eig_i \to 0$, 
as the index approaches the size of the dataset, $i \to \nA$, for datasets smaller than the number of input features, $\nA < \nin$, or as the index approaches the number of input features, $i \to \nin$, for larger datasets, $\nA > \nin$.
This characterizes the \textbf{tail} of the spectrum.\footnote{
    To see the tail for larger datasets, $\nA > \nin$, see Fig.~\ref{fig:spectrums-fixed-dataset}.
}
\item[\emph{(3)}] Varying the number of samples, $\nA$, 
we also vary the \emph{extent} of the power-law fit, \eqref{eq:eig-by-index-observation}. 
\end{enumerate}
As we will explain, these properties 
will translate to the power law scaling and plateau features of the test loss
\eqref{eq:phenomenological-loss-original}.\footnote{
    Moreover, 
    whether the underlying process that generates the spectrum actually comes from
    a power law distribution -- versus, e.g., a log-normal distribution -- doesn't matter; we actually only need for the spectrum to be \emph{approximately} described by a power law, as in \emph{(1)}, \emph{(2)}, \emph{(3)},
    in order for the scaling law phenomenology of the test loss to arise.
    (For further discussion of processes that give power law vs. log-normal distributions, see \cite{mitzenmacher2004brief}, and to learn more about the difficulty of identifying true power laws in nature, see \cite{clauset2009power}.)
}

To give more intuition for the spectral property \emph{(1)}, let's compare this situation to what we might have naively expected. A familiar data analysis setting in which we analyze spectra is \emph{principal component analysis} (PCA) \cite{PCA-ref}: PCA is a dimensionality-reduction tool in which the covariance matrix of a dataset is diagonalized to find linear combinations of the input features, $x_i$, that account for the majority of the variance of the data. Typically when 
PCA is useful
the spectrum has a \textbf{gap}, 
an index,  $i=\nl$, for $\nl \ll \nin$, such that these few $\nl$ large eigenvalues account for the majority of the total variance.
In that case, projecting the data onto the subspace spanned by the top $\nl$ eigenvectors is a way of reducing the naive $\nin$-dimensional \emph{input feature} space to a much smaller $\nl$-dimensional \textbf{latent feature} space. When this works, we might think of the bulk of the spectrum, $\eig_{\nl+1}, \ldots, \eig_{\nin}$, as uninformative noise, and that the true \emph{generative} process describing the distribution $p(x)$ lives on this smaller dimensional latent feature space.
However, the spectra of our natural datasets in Fig.~\ref{fig:spectrums} essentially have a \emph{continuous} spectra without any gap: this implies that the data was generated from a space without any natural cutoff for separating uninformative and informative features.\footnote{See \cite{bradde2017pca} for a renormalization group perspective on this lack of cutoff for continuous spectra.
} As such, 
especially 
in the portion of the spectrum that can be modeled by a power law, you can always do fractionally better at capturing the variance of the data by including more eigen-features.\footnote{This point, along with an extended discussion 
of
latent space dimensionality, is explored 
more in \S\ref{sec:coding-theory}.}

Considering item \emph{(2)} above, while the eigenvalues in the part of the spectrum modeled by a power law are important in capturing the variance, the \emph{tail} of the spectrum -- in which the eigenvalues rapidly approach zero -- is not. As per item \emph{(3)}, if we increase the size of the dataset up to the number of input features, $\nA \to \nin$, we can increase the \emph{useful} portion of the spectrum that participates in the power law.
This suggests a related question: if we instead fixed a dataset size $\nA$, and subsampled the input features, $\nin$, do we still get a power law and is it now limited by $\nin$? 
To answer this question, in Fig.~\ref{fig:spectrums-fixed-dataset} we plot example spectra from the same datasets as before, but this time for a fixed dataset size, $\nA$, and with the input features subsampled from the total number $\nin$;
we see that increasing the number of subsampled input features \emph{does} extend (and slightly rescale) the power law, preserving this structure in the spectrum.

Thus, we see that the inclusion of \emph{either} additional samples \emph{or} additional input features can be used to extend the spectral power law, which we expect might be useful given our discussion of continuous spectra and PCA above.
Unfortunately, the extent of the power law ultimately seems to be limited by the number of input features $\nin$. However, presumably if we \emph{increased} the number of input features, for instance if we acquired high-resolution versions of our images, we'd find an even longer power law? Relatedly, CIFAR-10 contains 50,000 images in its training set despite having only $3072$ input features: are those extra samples beyond the first $3072$ informative?

\subsection{Feature Map Properties}\label{sec:model-properties}

To answer these questions, let's try mapping the input data to a \textbf{feature space}, $\nf$, that's \emph{larger} than the input space, $\nin$. We define a collection of feature functions as 
\be
\fea_j(x) \, , \quad \text{with} \quad j=1,\dots,\nf \, ,
\ee
where $j$ indexes the $\nf$ different \textbf{features} of the \emph{representation} of the input $x$. 
At this point, the $\fea_j(x)$ could be the features of a deep neural network or they could be a simpler random feature model.
We are interested in studying the spectrum of this representation, which we can find by forming
the \emph{empirical feature-feature covariance matrix} of features,
\be\label{eq:estimator-of-random-feature-model-feature-feature-covariance-matrix}
\frac{1}{\nA} \sum_{\alpha =1}^T \fea_{i;\alpha}\fea_{j;\alpha} \, ,
\ee
and then computing its eigenvalues, $\eig_j$.
In particular, we would like to understand how the spectrum of the feature representation compares to the spectrum of the input representation for different types of feature maps.

\begin{figure}[ht]
\begin{center}
\includegraphics[width=0.49\linewidth]{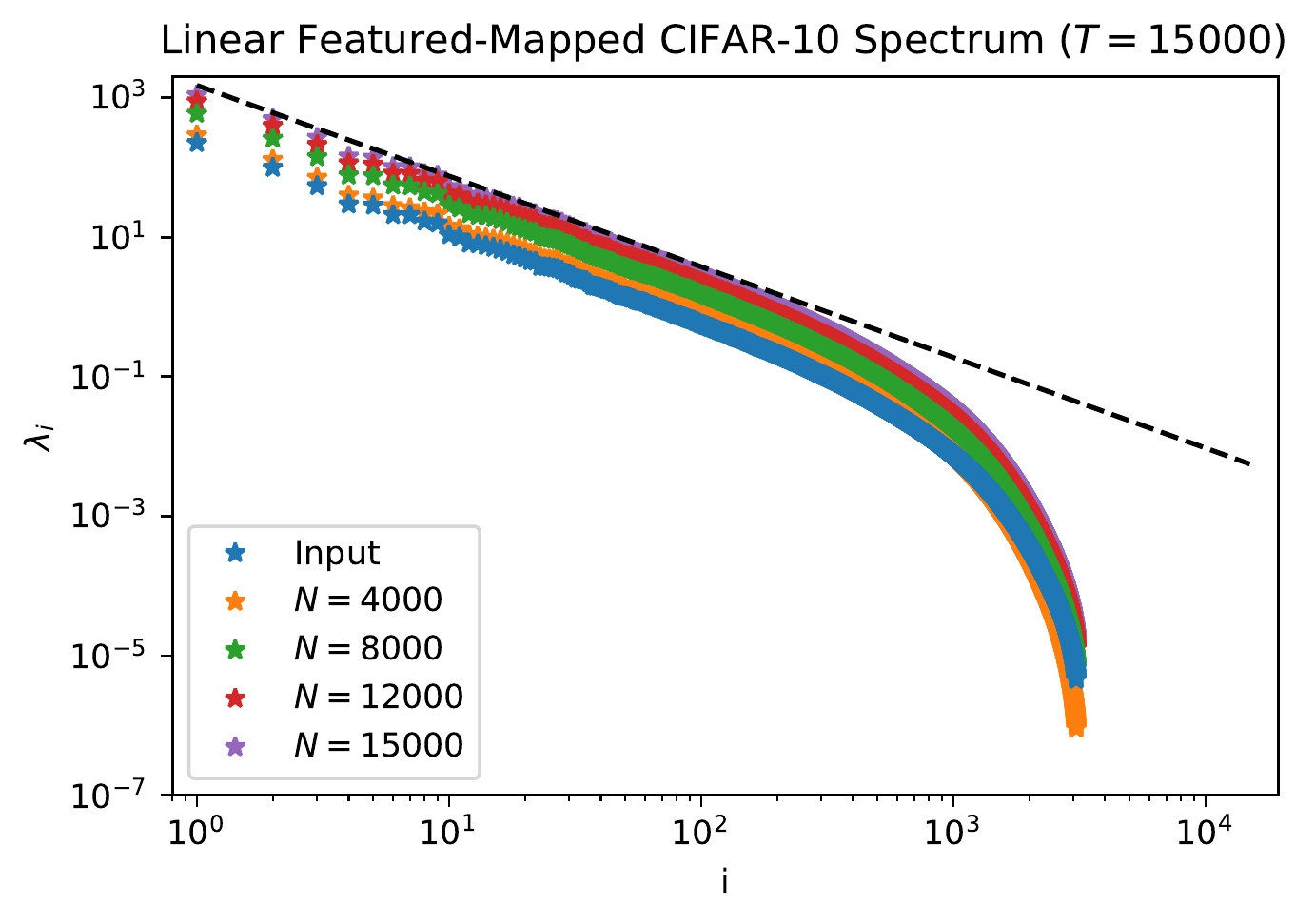}
\includegraphics[width=0.49\linewidth]{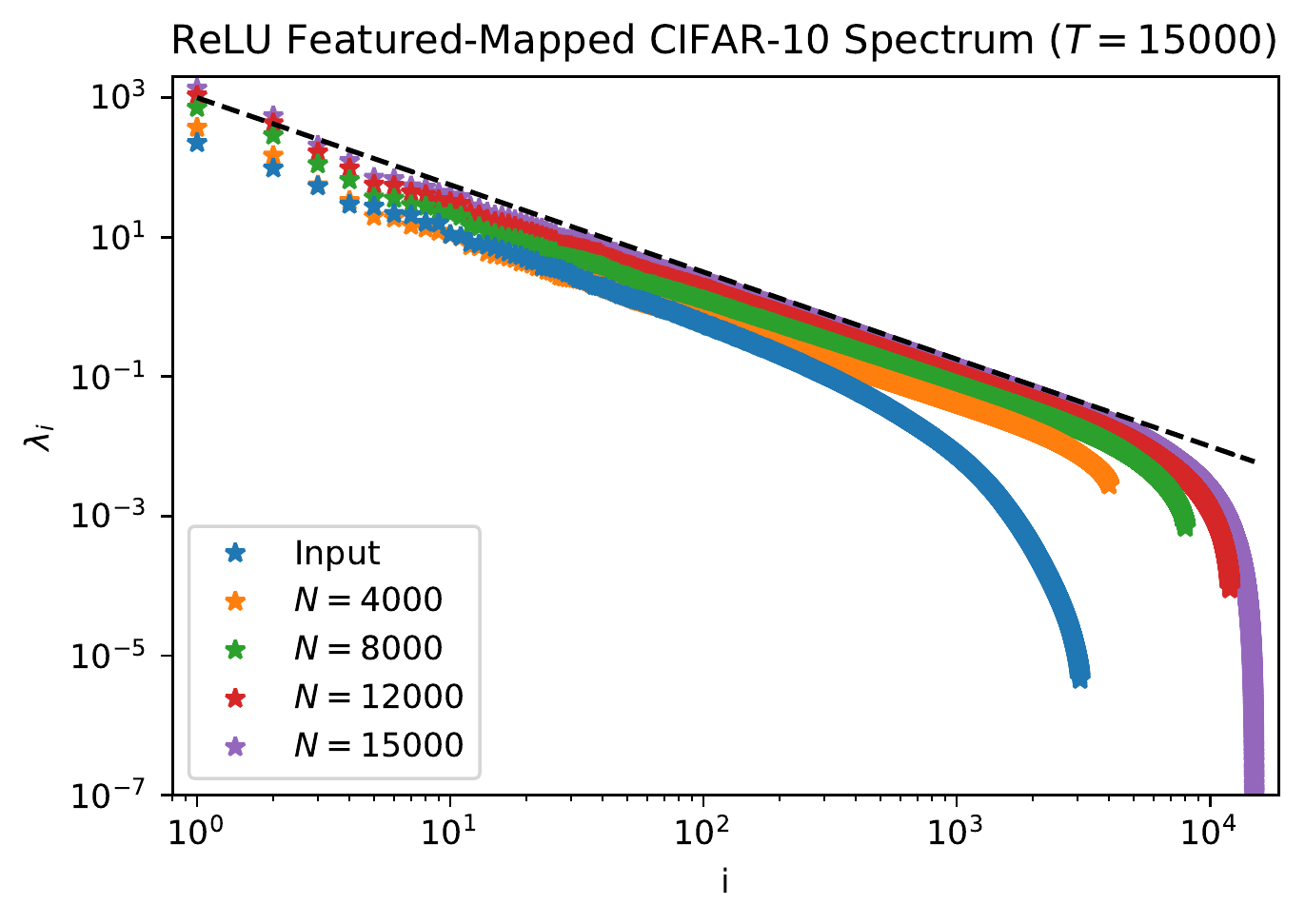}
\end{center}
\caption{Spectra of the feature representation from CIFAR-10 ($\nin = 3072$) of a fixed dataset ($\nA = 15000$), with an approximate power-law fit for the bulk (dashed line). 
\textbf{Left:}  
A linear map, \eqref{eq:linear-feature-map}, does not extend the length of the approximate power-law fit.
\textbf{Right:}  
For a nonlinear map, a
ReLU activation applied after a linear map, \eqref{eq:simplest-feature-map}, 
increasing the number of features, $\nf$, increases the extent of the approximate power-law fit. This extension is limited by the dataset size, $\nA$.
}
\label{fig:extend-linear-layer}
\end{figure}

As a naive first feature map, let's pass our input dataset through a linear transformation:
\begin{align}\label{eq:linear-feature-map}
\fea_{j;\alpha} \equiv \sum_{k=1}^{\nin} \fw_{jk} x_{k;\alpha} \,, %
\end{align}
where $\fw_{jk}$ is a $\nf$-by-$\nin$-dimensional weight matrix, which we assume to be full rank.
Concretely, we can assume each component of the weight matrix is sampled independently from a zero-mean Gaussian distribution with variance given by unity over fan-in, $1/\nin$.
In the left panel of Fig.~\ref{fig:extend-linear-layer} we've plotted the spectrum of this linear feature map applied to a fixed-sized image dataset. After inspecting this figure, we remember a basic fact about linear algebra that our linear map to the larger space can only create linearly-\emph{dependent} columns, and thus can only add zero eigenvalues to our spectrum. 
Thus, to meaningfully extend our spectrum, we will need to do something \emph{nonlinear}.

As a simple example of a nonlinear feature map, let's apply a nonlinear activation function after the linear transformation \eqref{eq:linear-feature-map}:
\begin{align}\label{eq:simplest-feature-map}
\fea_{j;\alpha} \equiv \sigma\!\le(\sum_{k=1}^{\nin} \fw_{jk} x_{k;\alpha} \ri)\,, %
\end{align}
where $\sigma$ is a scalar function that acts on each individual component $x_k$ of an input data point.
We can think of this nonlinear feature map, \eqref{eq:simplest-feature-map}, as representing the activations of a single hidden-layer neural network.
As a concrete example, let's set the activation as the ReLU,
\be\label{eq:relu}
\sigma(z) = \begin{cases}
  z\, ,  & z>0 \, ,\\
  0\,, & z \leq 0 \,,
\end{cases}
\ee
and again we will take each element of the weight matrix, $\fw_{jk}$, to be independent and initialized identically
according to a zero-mean Gaussian with variance $2/\nin$.
With these choices, \eqref{eq:simplest-feature-map} is a type of nonlinear \textbf{random feature model}. In the right panel of Fig.~\ref{fig:extend-linear-layer} we've plotted the spectrum of this nonlinear feature map applied to the same fixed-sized image dataset as before. Importantly, compared to the spectrum of the bare input data (blue stars), we see that increasing the number features in the feature map \emph{extends} the portion of the spectrum that's approximately fit by a power law.
In this way, we see that by applying a nonlinear transformation to our data we can build additional useful features when we have more samples than input features, $\nA > \nin$.

Together, this means that both the size of the model, $\nf$, and the size of the dataset, $\nA$, control 
the length of the power law in the spectrum of features:
on the one hand, 
when the model is feature limited ($\nf < \nA$) we can increase the power-law bulk by increasing the size of the model;
on the other hand,
when the model is data limited ($\nA < \nf$)
the feature-feature covariance matrix, \eqref{eq:estimator-of-random-feature-model-feature-feature-covariance-matrix}, is rank limited by the size of the dataset, and so the extra capacity is unnecessary as the extent of the power law is 
similarly limited by $\nA$. Thus,  so long as they are greater than the number of input features, $\nf, \nA > \nin$, the minimum of these two resource scales will control how many useful features there are.\footnote{
    Note that this extension effect is special for natural datasets with the properties enumerated in \S\ref{sec:data-properties} and will not be true in general.
} 
In the next section, we will construct a joint statistical model of datasets and feature maps that has this precise property, and by solving this model we will see how this is translated into the power-law scaling and performance plateau in the test loss of a trained model.

\subsubsection*{\emph{Aside:} Random Feature Maps vs. DNNs}

Before moving on to discuss our solvable model, let's just discuss more general nonlinear feature maps, namely deep neural networks. Even though they are both nonlinear models, a single-hidden-layer ReLU network is a very different model than the 540 billion parameter Palm based on the transformer architecture: in particular, DNNs have specially designed components, such as the multi-headed self-attention mechanism that powers transformers; moreover, they are not random feature models -- at least at finite width, see, e.g., \cite{PDLT-2022} -- \emph{learning} nonrandom representations of inputs; and finally, the scaling laws of \cite{kaplan2020scaling} concern the number of \emph{parameters}, but here we instead focused on the number of \emph{features}.
Let's address these concerns one by one in reverse order.

Firstly, there is often some ambiguity in the feature map of a DNN; e.g. with LLMs like BERT \cite{BERT2018}, practitioners sometimes 
use the activations of the last few layers
of the model 
as features for a downstream task. However, the distinction between parameters is sharp even for our single-layer ReLU random feature map \eqref{eq:simplest-feature-map}: as we are using it, this model has $\nf \times \nin$ parameters, but only $\nf$ features. The resolution is that the proper way to think about the features of a network
is in terms of the NTK \cite{jacot2018neural,brainNTK2019}: the NTK is a type of \emph{data-data covariance matrix} of features,
\be
\widehat{H}_{\alpha_1 \alpha_2} \equiv %
\sum_{j =1}^{\nf} \fea_{j;\alpha_1}\fea_{j;\alpha_2} \, ,
\ee
and it's easy to see that this will have the same spectral properties as the feature-feature covariance matrix that we've been considering in \eqref{eq:estimator-of-random-feature-model-feature-feature-covariance-matrix} up to an overall rescaling. 
For network architectures that have an infinite-width limit \cite{neal1996priors,lee2018deep,matthews2018gaussian}, DNNs trained by gradient descent are generalized linear models with the NTK identified as the kernel. Moreover, if $z(x_\alpha;\theta)$ is the (scalar) output of the network when evaluated on a sample $x_\alpha$, and $\theta_j$ is an $\nf$ dimensional vector that indexes \emph{all} the parameters, then the definition of the NTK \cite{jacot2018neural,brainNTK2019} tells us to identify the feature map with the derivative of the network output:
\be
\fea_{j;\alpha} \equiv \frac{dz(x_\alpha; \theta)}{d\theta_j} \,.
\ee
Thus, where this correspondence between DNN and linear models holds, then there's precise correspondence giving a feature for every parameter, and increasing the number of parameters increases the effective number of features accessible to the linear model.\footnote{For a more detailed discussion of this point and the correspondence, see \S10.4 of \cite{PDLT-2022}.} 
To this end, if we were to plot the spectrum of the features derived from the NTK, we would see a similar phenomenology to what we observed in \S\ref{sec:model-properties} for our simple nonlinear feature map \eqref{eq:simplest-feature-map}. %

Away from the infinite-width limit, at least perturbatively \cite{PDLT-2022}, the model output still depends on the NTK with a \emph{parameter}-number of features, but rather than a random feature model, the features of a finite-width network learn nontrivial representations of inputs from the data. The fact that our model in the next section exhibits the neural scaling phenomenology of power law and plateau suggests that probably feature learning isn't an essential part of scaling laws; we will address more concrete means of understanding this relationship between representation learning and scaling laws 
in the last part of \S\ref{sec:future-directions}.

Finally, what of the broader and essential differences between a one-hidden-layer ReLU network to an LLM? A standard principle of computer science is GIGO: \emph{garbage in, garbage out}. Perhaps the biggest takeaway lesson in our setting is  PIPO: \emph{power-law in, power-law out}. 
To that end, we conjecture that better DNN architectures 
are better able to preserve power law structure when transforming the spectra of input datasets and leave it to future work to understand how to translate these performance of better models into statements about 
the spectra of features.\footnote{For instance, a more careful investigation of the raw input (blue stars) in the right panel of Fig.~\ref{fig:extend-linear-layer} would show that the exponent $\alpha$ that characterizes the spectrum in \eqref{eq:eig-by-index-observation} actually decreases slightly after the ReLU layer. As we will explain in the next section, $\alpha$ ultimately will become the exponent in the power-law portion of the test loss of our model; thus, even though the power-law gets extended to give the scaling law, the slight decrease in its value ultimately leads to worse performance than if it were otherwise preserved.

A second issue worth considering when comparing a single ReLU layer to an LLM is that we haven't modeled the eigenvectors, which may need to be considered in a more detailed model. (For a discussion of scaling laws that does consider eigenvectors, see \cite{wei2022more}.)
\label{footnote-exponent-decrease}
}

\section{A Statistical Model}\label{sec:derivation}

We want to
construct a generative data model and random feature model that captures the broad empirical properties of 
real datasets composed with nonlinear feature maps 
such that
the resulting statistical model's test loss exhibits 
the scaling law phenomenology 
illustrated in Fig.~\ref{fig:pheno-loss-original}.
Recall from the previous section, our key observation is an approximate power law 
in the spectrum of the feature representation, with the extent of the power-law portion 
controlled by the minimum of the number of features, $\nf$, and the size of the training set, $\nA$.
After finding features with these properties in a simplified model, we can then use them 
in
a generalized linear regression problem 
such that the test loss exhibits our desired behavior.

Our goal will be to compute that averaged test loss analytically. We will accomplish this using tools from random matrix theory, using some simple diagrammatic techniques that can quickly and easily extract the properties of these models when $\nf$ and $\nA$ are sufficiently large.
We will begin our journey in \S\ref{sec:derivation-notation} by setting up and defining our model
as well as verifying its properties with 
numerical simulations. 
The bulk of the section will be spent in 
\S\ref{sec:data-and-feature-averages}, where we will explain how to average over our generative data model and random features in order to derive a
formula for the model's test loss. 
Then, in \S\ref{sec:spectral-extension} we outline how we could use our RMT tools to model spectral extension in nonlinear feature maps such as neural networks.
Finally, in \S\ref{sec:other-methods} we compare our methods and results to other related RMT machine-learning calculation.

\subsection{Setup and Verifying}\label{sec:derivation-notation}

Return here often as you explore the other subsections in this section.

\subsubsection*{Generative Data Model}

We will start by defining a generative model for the dataset.

Rather than generating data in the raw input space, we will generate data in a
\textbf{latent space}. 
Consider a latent data point, $x$, 
whose components are denoted
\be
x_{I} \,, \quad \text{with} \quad I= 1, \ldots, \nl \,,
\ee 
where 
$I$ indexes the $\nl$ different \emph{latent features}. To distinguish latent features from the features following a random feature map, we will use capital roman indices from the middle of the alphabet ($I, J, K, \dots$) for the former and lower-cased roman indices ($i, j, k, \dots$) 
for the latter.
Importantly, to get the right behavior we will need the dimension of the latent space to be larger than any other scales in the problem:
\be
\nl \gg \nf, \nA\, .
\ee

For each data point, we will sample components from a zero-mean Gaussian distribution with a covariance matrix $\covl$:
\be\label{eq:feature-feature-covariance-definition}
    \expval{x_I} = 0 \, , \qquad \expval{x_I x_J} = \covl_{IJ} \, ,
\ee
where we
denote expectations over random variables with the notation
\be\label{eq:expectation-definition}
\braket{f(u)} \equiv \int \!du\, p(u)\, f(u) \,,
\ee
where $u$ includes \emph{all} random variables in the expression 
$f(u)$.
If we instead want to take expectations  over only some of the random variables, we will use a subscript notation on the bracket as
\be
\braket{f(u,v)}_{u} \equiv \int \!du\, p(u|v)\, f(u,v) .
\ee
When possible we will keep our derivation generic and not make assumptions about the covariance $\covl$, other than that it is full rank. 

However, for the goal of understanding scaling laws we will be motivated to consider a class of models where the covariance spectrum has the form of a power law. 
In particular, we will assume the eigenvalues of $\covl$ are  well-approximated by a smooth number density of eigenvalues,
\be\label{eq:eig-density}
    n(\eig) \, d\eig = {\nl (\beta-1) \eigmin^{\beta-1}} \eig^{-\beta} \theta(\eig-\eigmin) \, d\eig\, ,
\ee
where $\eigmin$ is the minimum eigenvalue, $\beta$ is an exponent that characterizes 
the tail of the distribution, 
$\theta(\eig)$ is the Heaviside step function,
and the constants are chosen such that the density integrates to $\nl$. 
Alternatively, we can write the spectrum as a function of index $I$ as
\be\label{eq:exact-power-law-latent-generative}
\eig_I = \eigmax \le(\frac{1}{I}\ri)^{1+\alpha} \, .
\ee
In this form it is convenient to (hyper-)parameterize the spectrum in terms of a maximal eigenvalue,
\be\label{eq:def-eig-max}
\eigmax \equiv \eigmin \nl^{1+\alpha}\, .
\ee
The exponent $\alpha$ in (\ref{eq:exact-power-law-latent-generative}) is related to the exponent $\beta$ appearing in (\ref{eq:eig-density}) by
\be\label{eq:def-of-alpha}
\alpha \equiv \frac{2- \beta}{\beta-1} \, ,
\ee
as can be checked by integrating the density of states (\ref{eq:exact-power-law-latent-generative}).
Since $\alpha$ will ultimately be power-law exponent of the test loss, we must have $0 < \alpha < \infty$, which in turn implies $1 < \beta < 2$.\footnote{
    For power-law probability distributions of the form \eqref{eq:eig-density}, the distribution is normalizable only for $\beta > 1$ and has finite mean only for $\beta > 2$. However, if instead we fix a maximal eigenvalue $\eigmax$, then the mean (and higher moments) will exist, but the distribution will no longer be normalizable.
}
This is actually a rather compact range and implies that natural datasets have surprisingly heavy tails.
Finally, for large $\nl$ it generally does not matter whether the eigenvalues are drawn at random from a distribution of the form \eqref{eq:eig-density} or taken from a fixed spectrum of the form \eqref{eq:exact-power-law-latent-generative}, and we will generally be agnostic about this choice in our theory.\footnote{In our simulations, we find it convenient to use \eqref{eq:exact-power-law-latent-generative} and characterize the spectrum by $\nl$, $\eigmax$, and $\alpha$.}

\begin{figure}[ht]
\begin{center}
 \includegraphics[width=0.49\linewidth]{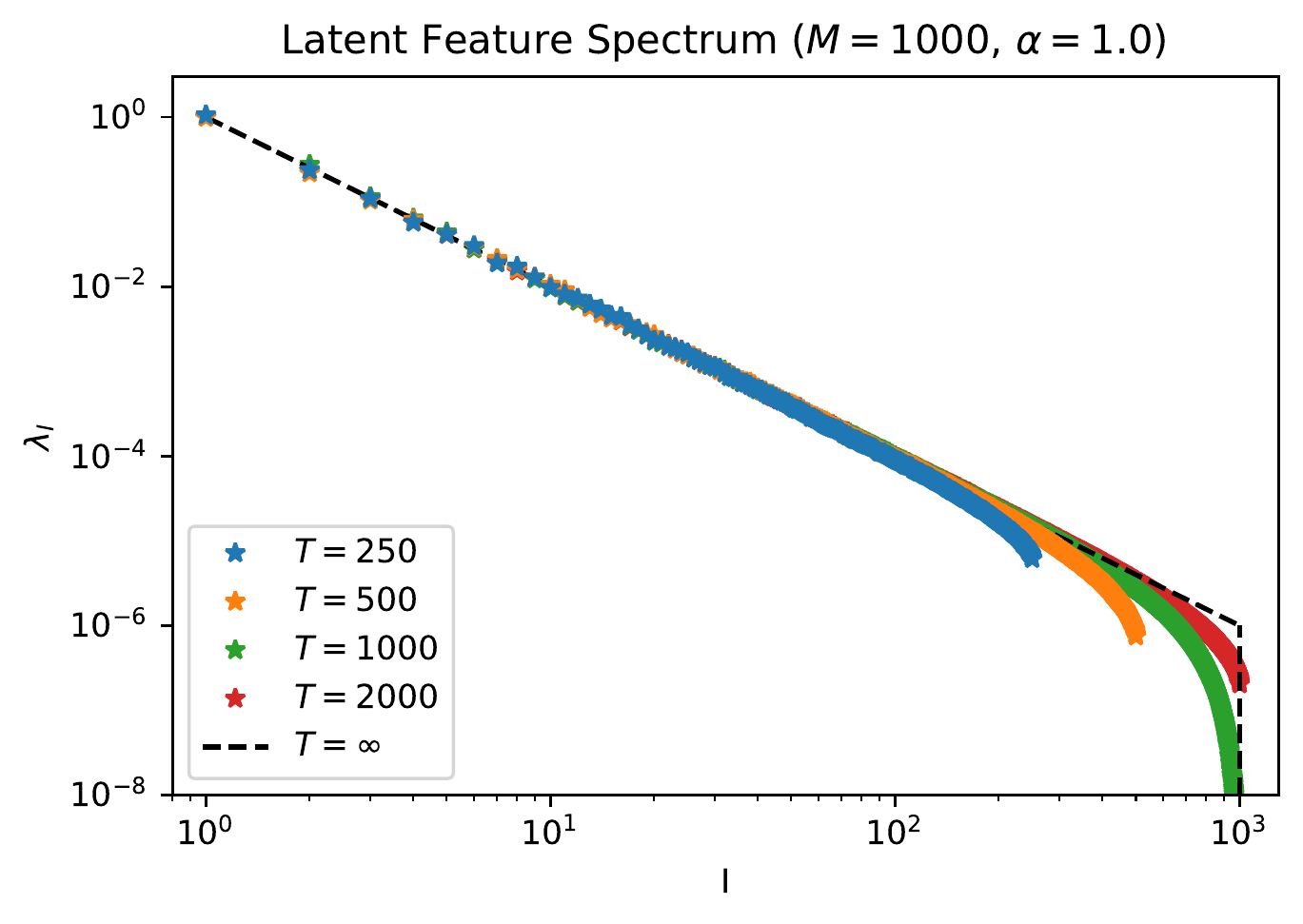}
  \includegraphics[width=0.49\linewidth]{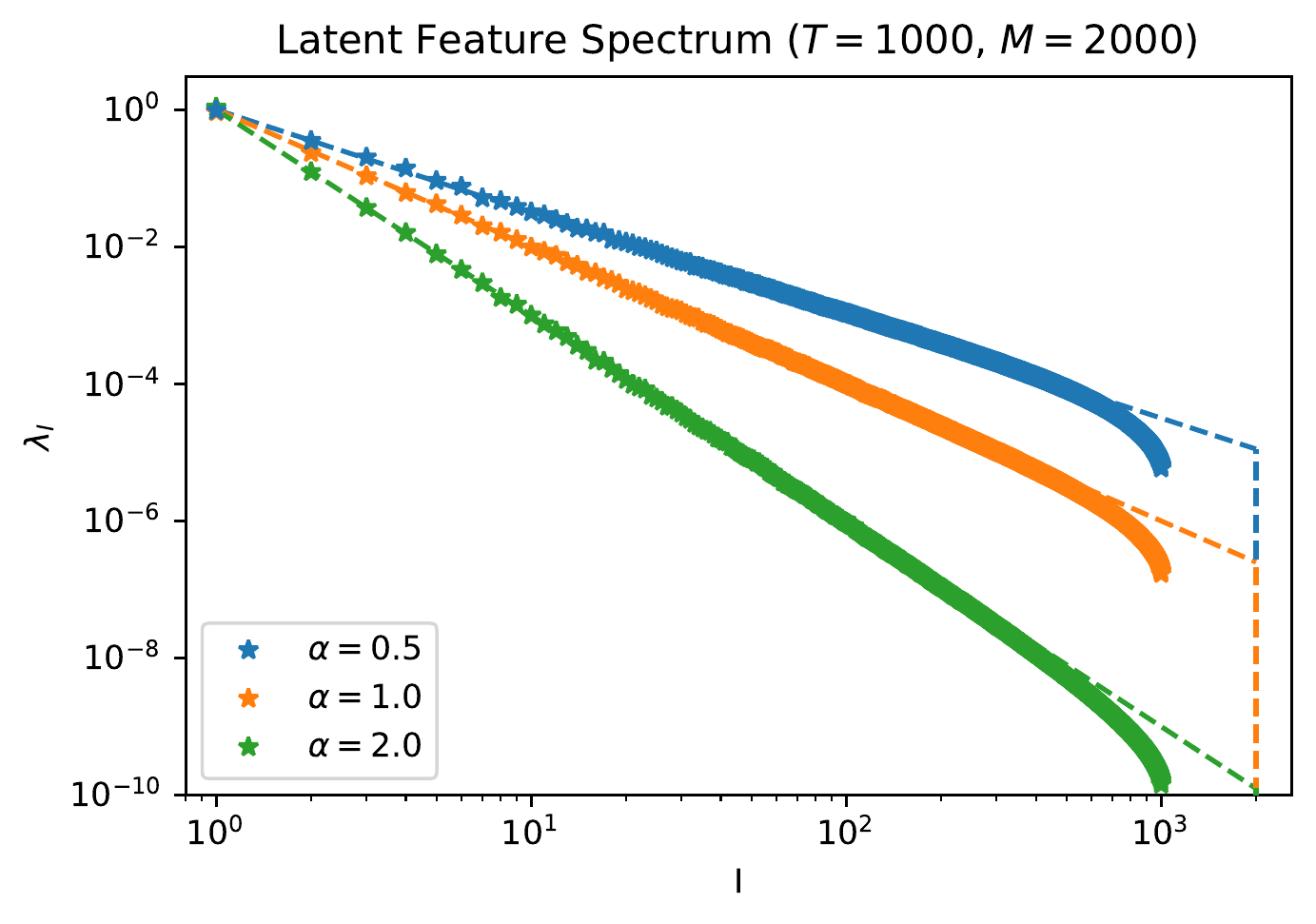}
\end{center}
\caption{Spectrum $\eig_I$ from numerical simulations (stars) of our latent data generative model, \eqref{eq:feature-feature-covariance-definition} and \eqref{eq:exact-power-law-latent-generative}, with the maximum eigenvalue fixed ($\eigmax =1$).
\textbf{Left:} The size of the dataset, $\nA$, is varied while the size of the latent space and the power-law exponent are fixed ($\nl = 1000$, $\alpha =1$). These spectra follow a pattern similar to the ones displayed in Fig.~\ref{fig:spectrums} for natural data: for dataset size smaller than the size of the latent space ($\nA <\nl$, blue and orange) the
spectrum has a bulk power law portion that terminates in a very rapid decline ($\eig_I \to 0$) as the index approaches the size of the dataset ($I \to \nA$), and the extent of the power law increases with increasing dataset; for dataset size equal to and greater than the size of the latent space ($\nA \geq \nl$, green and red), the power law terminates at the size of the latent space, but the rapid decline becomes sharper and sharper as the size of the dataset increases, forming a kink in the limit of infinite data ($\nA \to \infty$, dashed black line).
\textbf{Right:} The power-law exponent, $\alpha$, is varied as the sizes of the dataset and latent space are held fixed ($\nA=1000$, $\nl=2000$), and the spectrum for infinite data is plotted for comparison (dashed lines). As all three simulations have the same size datasets, their power laws all terminate at the same point ($\nA=1000$).
}
\label{fig:latent-spectrum}
\end{figure}

In Fig.~\ref{fig:latent-spectrum}, we numerically sample some datasets from our generative data model, \eqref{eq:feature-feature-covariance-definition} and \eqref{eq:exact-power-law-latent-generative}, for different (hyper)-parameters  $\nA$, $\nl$, and $\alpha$ to check that we have a good model of the natural datasets discussed in \S\ref{sec:data-properties}. We see that for $\nA < \nl$ the extent of the power law increases with the size of the dataset, and for $\nA \geq \nl$  increasing the size of the dataset sharpens the rapid decline towards zero but leaves the extent of the power law fixed. This confirms that we have captured the broad spectral properties of the inputs from the natural datasets we have analyzed.

Finally, for every latent datapoint $x_I$, we will also generate
a $\nout$-dimensional label
\be\label{eq:def-teacher-labels}
    y_{i} = \sum_{I=1}^{\nl} \w_{iI}x_I , \quad \text{with} \quad i= 1, \ldots, \nout\, , 
\ee
using a $\nout$-by-$\nl$-dimensional weight matrix, $\w \equiv \w_{iI}$,
whose elements we will take to be
independent and drawn from a zero-mean Gaussian distribution, so that  
\be\label{eq:teacher-weights-statistics}
    \expval{\w_{iI}} = 0\,, \qquad \expval{\w_{i_1I_1} \w_{i_2I_2}} = \frac{\wvar}{\nl} \delta_{i_1i_2} \delta_{I_1I_2} \, .
\ee
It is important that each label is allowed to
depend on all $\nl$ latent features of an input to ensure that the difficulty of the problem scales with $\nl$.
    Such a scaling 
    is
    needed 
    in order to approximate the 
    self-supervised
    generative modeling tasks that LLMs perform.

\subsubsection*{Random Feature Model}

Now let's define a random feature model that we will use to map our latent data to a feature representation. Our goal is to find a representation where the spectrum contains an approximate power-law fit that is controlled by the number of feature functions, $\nf$, in the model.

The main advantage of generating our data in a large latent space ($\nl > \nf$) rather than a smaller input space ($\nin < \nf$) is that we can use a simpler \emph{linear} map from the larger latent space to the smaller feature space rather than having to analyze a \emph{nonlinear} map from the smaller input space to the larger feature space. We will define our collection of feature functions by
\be\label{eq:statistical-model-random-feature-map}
    \fea_j(x) \equiv \sum_{I=1}^{\nl} \fw_{jI} x_I \, ,
\ee 
where $j$ indexes the $\nf$ different features of the representation of the latent input $x$, and
$\fw \equiv \fw_{jI}$ is a $\nf \times \nl$ matrix of random feature weights drawn from a zero-mean Gaussian:
\be\label{eq:random-feature-weights-statistics}
    \expval{\fw_{jI}} = 0\,, \qquad \expval{\fw_{j_1I_1} \fw_{j_2I_2}} = \frac{\fwvar}{\nl} \delta_{j_1j_2} \delta_{I_1I_2} \, .
\ee

\begin{figure}[ht]
\begin{center}
 \includegraphics[width=0.49\linewidth]{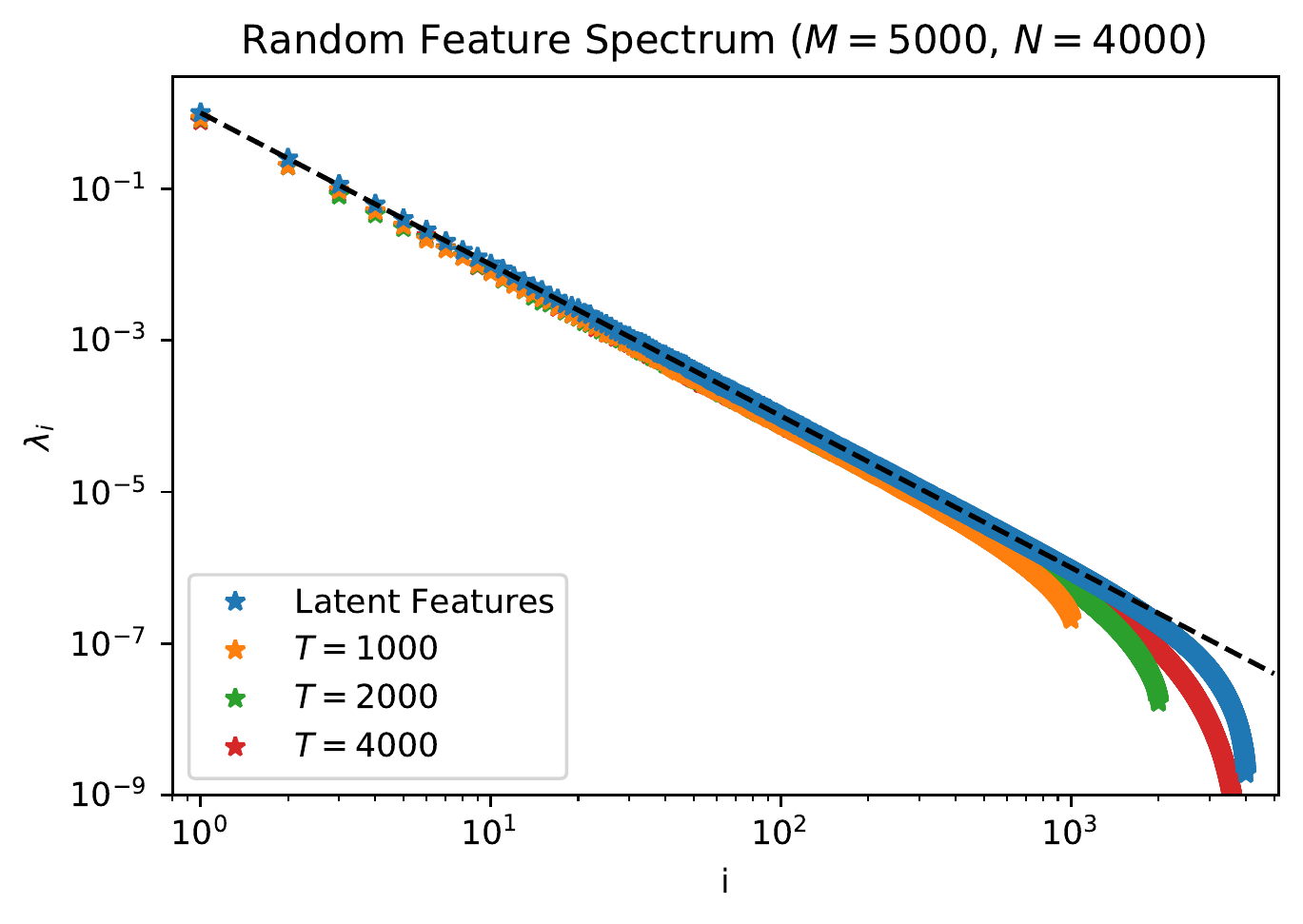}
 \includegraphics[width=0.49\linewidth]{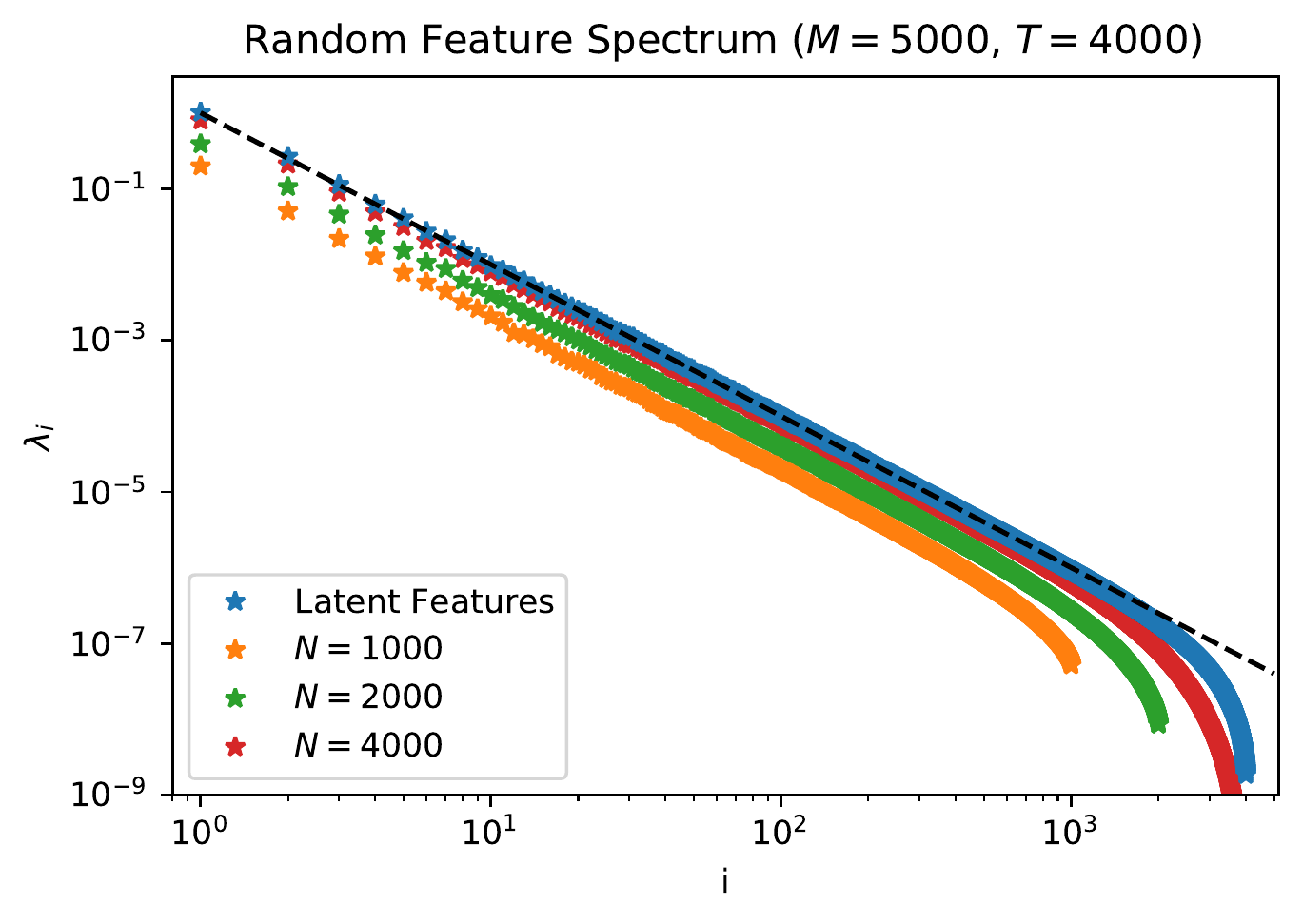}
\end{center}
\caption{Spectrum, $\eig_i$, from numerical simulations of our random feature model, \eqref{eq:statistical-model-random-feature-map}, mapping sampled data from our generative data model, 
with the size of the latent space, the maximum eigenvalue, and power-law exponent fixed ($\nl=5000$, $\eigmax =1$, $\alpha = 1$).
These spectra show that the random features of our joint model follow closely to what we observed for CIFAR-10 in
Fig.~\ref{fig:extend-linear-layer}:
the approximate power-law fit is controlled by the minimum of 
 the number of features and the size of the dataset, $\min(\nf, \nA)$.
 The latent feature spectrum is also plotted for comparison (blue) as is the power-law fit (dashed line).
\textbf{Left:} The size of the dataset, $\nA$, is varied while the number of random features is held fixed ($\nf=4000$).
\textbf{Right:} The number of random features, $\nf$, is varied while the size of the dataset is held fixed ($\nA=4000$). 
}
\label{fig:feature-spectrum}
\end{figure}

In Fig.~\ref{fig:feature-spectrum}, we take datasets sampled from our generative data model, \eqref{eq:feature-feature-covariance-definition} and \eqref{eq:exact-power-law-latent-generative}, and map them through our random feature model, \eqref{eq:statistical-model-random-feature-map}, for a fixed set of features weights, $\fw$, in order to verify that their spectra has the properties discussed in \S\ref{sec:model-properties}.  We see that the power-law portion of each spectrum is controlled by the minimum of the number of features, $\nf$, and the size of the dataset, $\nA$.
These are precisely the properties we sought to find in our simplified  joint data and feature model.

\subsubsection*{(Generalized) Linear Regression}

Now that we have features, we will ``train'' a (generalized) linear model to reproduce the 
labels, $y_i$, generated from the underlying latent features, \eqref{eq:def-teacher-labels}, by learning a linear transformation of the random features $\fea_j(x)$:
\be\label{eq:linear-model-def}
     z_{i}(x;\param) \equiv \sum_{j=1}^{\nf} \param_{ij} \fea_{j}(x) \, ,
\ee
where $\param \equiv \param_{ij}$ is a set of learnable parameters.

To fix these parameters, we will use our generative data model \eqref{eq:feature-feature-covariance-definition} to draw a collection of $\nA$ pairs of samples $\{x_I, y_i \}$ to form our training set $\A$: 
\begin{align}
    \lbrace x, y \rbrace \quad \Longleftrightarrow  \quad  \lbrace x_{I;\alpha}, y_{i;\alpha} \rbrace\, . 
\end{align}
We will typically denote our training set of latent data and labels using the matrix notation (left-hand side), although when clarity dictates we may also use the index notation (right-hand side), with $\alpha = \{1, \dots, \nA \}$ used to index particular samples in $\A$. Accordingly, we can use our feature functions, \eqref{eq:statistical-model-random-feature-map}, to construct a corresponding matrix of random features derived from the training set:
\be\label{eq:training-set-fea-def}
\fea \equiv \fea(x) \quad \Longleftrightarrow  \quad \fea_{j ; \alpha} \equiv \fea_{j}(x_\alpha) \, .
\ee

We can then use the training set to fit the parameters by optimizing a standard MSE loss function with a ridge regression term:
\begin{align}\label{eq:def-mse-loss}
    \L_\A(\param) &\equiv \frac{1}{2}\sum_{i=1}^{\nout}\sum_{\alpha=1}^{\nA}\le( z_{i;\alpha}  - y_{i;\alpha} -\noise_{i; \alpha}\ri)^2 + \frac{\ridge}{2}  \sum_{i=1}^{\nout}\sum_{j=1}^\nf \param_{ij}^2 \,   \\
    &= \frac{1}{2}\norm{\param\fea-y - \noise}^2 + \frac{\ridge}{2}\norm{\param}^2\,, \notag
\end{align}
where $\ridge$ is the ridge parameter. Here, we've also introduced the ability to 
corrupt our labels with a matrix of random noise, $\noise$, which has a separate entry for each training sample and label component, and each of which entry is drawn from another zero-mean Gaussian with statistics
\be\label{eq:def-noise-statistics}
\expval{\noise_{i;\alpha}} = 0\,, \qquad \expval{\noise_{i_1;\alpha_1} \noise_{i_2;\alpha_2}} = \vare \delta_{i_1i_2} \delta_{\alpha_1\alpha_2}  \, .
\ee
(Note that there is no normalization factor in the variance as there was in our other variances, e.g., \eqref{eq:teacher-weights-statistics} and \eqref{eq:random-feature-weights-statistics}.)

Finally, optimizing the loss \eqref{eq:def-mse-loss} with respect to $\param$ has a well-known solution:
\begin{align}
    \theta_{ij}^\star &\equiv \sum_{k=1}^{\nf} \sum_{\alpha_2=1}^{\nA}\le(\ridge \delta_{jk}+\sum_{\alpha_1=1}^{\nA}\fea_{j;\alpha_1} \fea_{k;\alpha_1} \ri)^{-1} \fea_{k;\alpha_2} (y_{i;\alpha_2}+ \noise_{i;\alpha_2}) \, \notag \\
    &=   (y+\noise)\fea^T\res \, ,
\label{eq:linear-regression-solution}
\end{align}
where on the second line we switched to matrix notation and also introduced the \emph{feature-feature resolvent} matrix
\be\label{eq:resolvent-feature-feature}
    \res(\ridge) \equiv \frac{1}{ \ridge \IN + \fea \fea^T }  \quad \Longleftrightarrow  \quad \res_{jk}(\ridge) \equiv \le(\ridge \delta_{jk} + \sum_{\alpha=1}^{\nA}\fea_{j;\alpha} \fea_{k;\alpha} \ri)^{-1} \, , 
\ee
where $\IN \equiv \delta_{ij}$ represents the identity matrix on feature space.

\subsubsection*{Computing Performance}

To evaluate our model, we will again use our generative model \eqref{eq:feature-feature-covariance-definition} to draw a collection of $\nB$ pairs of samples 
to form our training set $\B$:
\begin{align}
    \lbrace \widehat x, \widehat y \rbrace \quad \Longleftrightarrow  \quad   \lbrace\widehat{x}_{I;\beta}, \widehat{y}_{i;\beta} \rbrace\, , 
\end{align}
where we will generally use a \emph{hat} to emphasize test-set quantities, and when using indices we will use $\beta = \{1, \dots, \nB \}$ to index particular samples in $\B$. Accordingly, we can use our feature functions, \eqref{eq:statistical-model-random-feature-map}, to construct a matrix of random features derived from the test set,
\be
\widehat{\fea} \equiv \fea(\widehat{x}) \quad \Longleftrightarrow  \quad \widehat{\fea}_{j ; \beta} \equiv \fea_{j}(\widehat{x}_\beta) \, ,
\ee
and then use our
solution, \eqref{eq:linear-regression-solution}, for inference on these test examples:
\be\label{eq:test-predictions}
\widehat{z}^\star \equiv \param^\star \widehat{\fea}  \quad \Longleftrightarrow  \quad \widehat{z}^\star_{i;\beta} =  \sum_{j=1}^{\nf}\theta^\star_{ij} \widehat{\fea}_{j;\beta}\, .
\ee

The model's performance may then be measured by a test loss
\begin{align}\label{eq:def-test-loss}
    \L_\B(\param^\star) &\equiv \frac{1}{2 \nB }\norm{\widehat{z}^\star-\widehat{y}}^2 \, \\
    &=\frac{1}{2 \nB }\norm{(y+\noise)\fea^T\res  \widehat{\fea}-\widehat{y}}^2 %
    \,
    \notag\\
    &=\frac{1}{2 \nB }\norm{(\w x +\noise)\fea^T\res  \widehat{\fea}-\w \widehat{x} }^2 
    \, , \notag
\end{align}
where on the second line we substituted in for the test predictions, $\widehat{z}^\star$, using \eqref{eq:test-predictions}, and then the optimal parameters, $\param^\star$, using 
\eqref{eq:linear-regression-solution}, 
and on the final line we substituted in for the labels using \eqref{eq:def-teacher-labels}.
Note that this MSE loss has a different normalization than the training loss, \eqref{eq:def-mse-loss}, so that it represents a \emph{per sample} loss if averaged and has a nice large-test-set limit. 
Furthermore, note that by using this analytical form of the linear regression solution, we are effectively in the limit of infinite training, in which the model has been allowed to converge. This means that \emph{(a)} we will not have to worry about the way that the performance can depend on the details of the learning algorithm (see, e.g. \cite{hoffmann2022training}), but also that \emph{(b)} our statistical model will not capture capture the compute-limited scaling laws studied by Ref.~\cite{kaplan2020scaling}.

\begin{figure}[ht]
\begin{center}
 \includegraphics[width=0.49\linewidth]{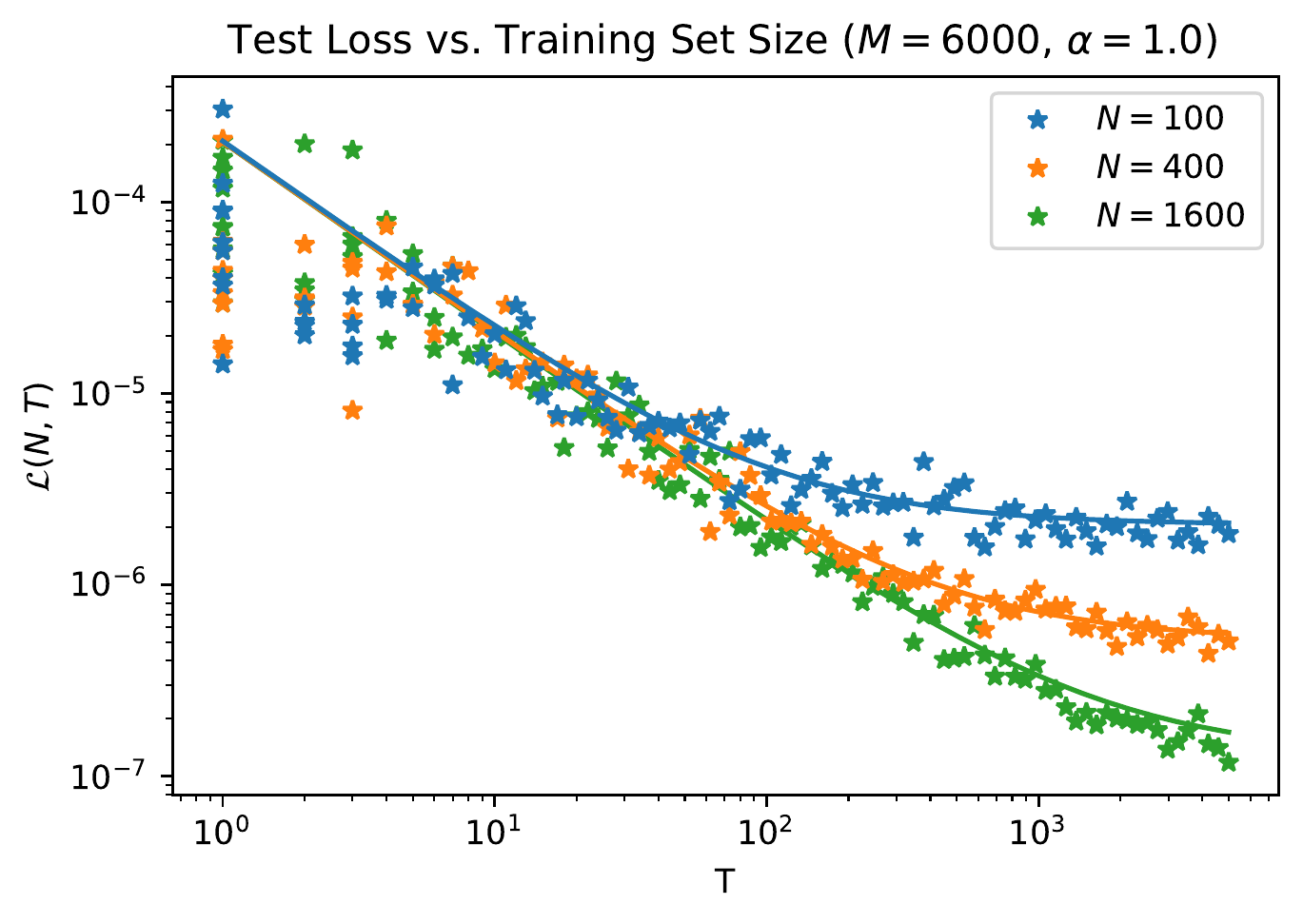}
  \includegraphics[width=0.49\linewidth]{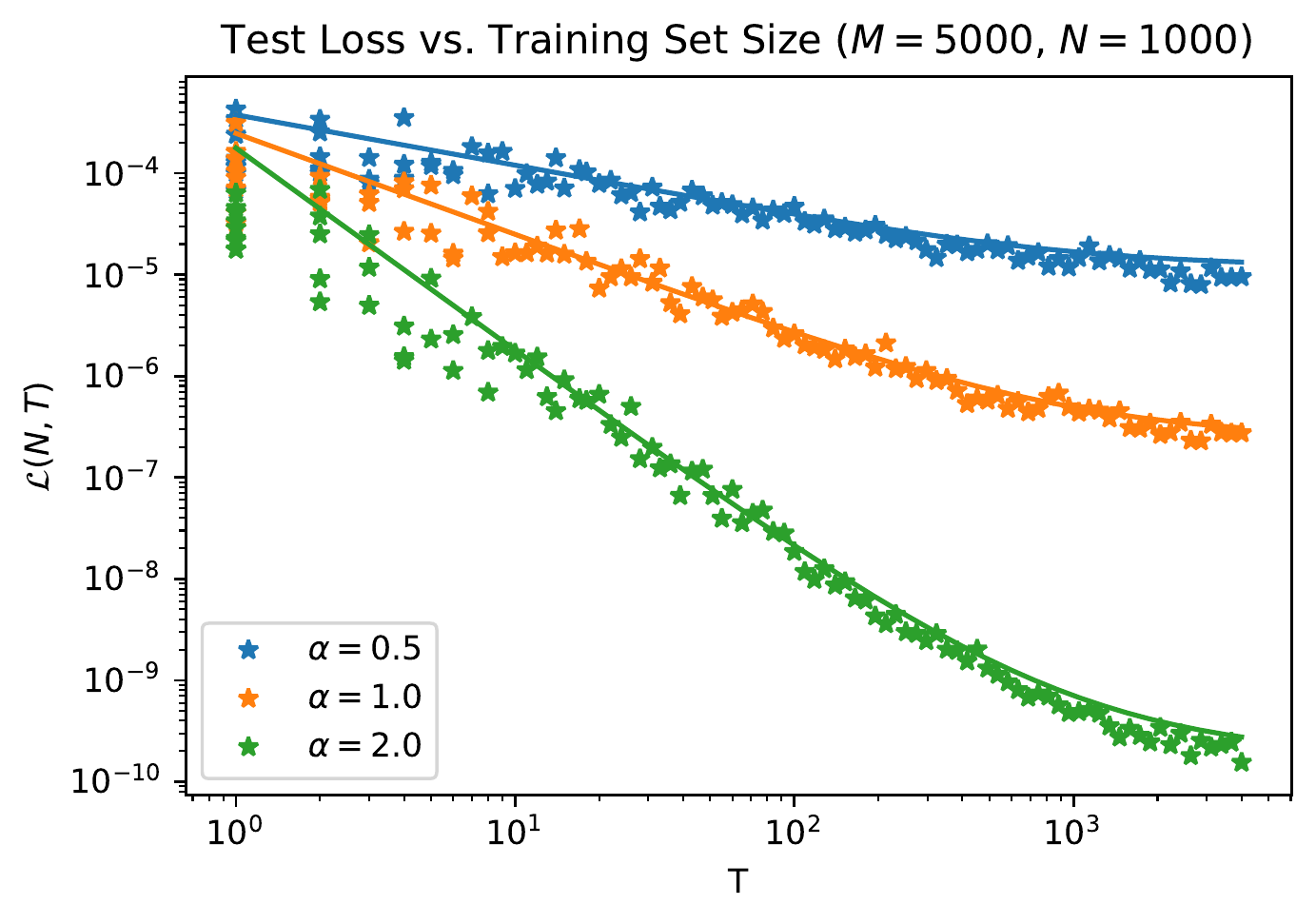}
\end{center}
\caption{Test loss from numerical simulations (stars) of our optimally regularized ($\ridge=\ridge^\star)$ joint statistical model ($\fwvar = 1$, $\eigmax=1$, $\wvar =1$, $\vare = 0$) 
as a function of training set size
demonstrating the 
same rich behavior as LLMs
and a simple fit (solid lines) given by \eqref{eq:phenomenological-model-simplified}: 
analogous to Fig.~\ref{fig:pheno-loss-original},
if the model isn't bottlenecked by the number of parameters, $\nf \to \infty$, the test loss behaves as a \emph{power law} in the training set size, $\L(\nf, \nA) \sim \nA^{-\alpha}$; otherwise, if the number of parameters is too small for a given training set, then the test loss stalls at a \emph{plateau} at a value that depends predictably on the parameters, $\L(\nf, \nA) \sim \nf^{-\alpha}$.
Similar statements would hold if we plotted the test loss as a function of the number of features.
(For smaller values of $\nA$, the variance of any particular realization is large, and so we've plotted multiple simulations for $\nA \lesssim 10$.)
\textbf{Left:} The size of the training set, $\nA$, is varied for a few different sized models, $\nf$, while the size of the latent space and the power-law exponent is held fixed ($\nl=6000$, $\alpha=1.0$).
\textbf{Right:} The size of the training set, $\nA$, is varied for a few different power-law exponents, $\alpha$, while the size of the latent space and the size of the model is held fixed  ($\nl=6000$, $\nf=1000$).
}
\label{fig:model-numerics}
\end{figure}

One advantage of a joint model of data \emph{and} features is that we are able to numerically simulate it for different (hyper)-parameters to confirm that it has the right behavior. In  Fig.~\ref{fig:model-numerics}, we plot the test loss \eqref{eq:def-test-loss} for a variety of model sizes, $\nf$, as a function of the training set size, $\nA$: first, we generate latent training and test sets by sampling from  our generative data model, \eqref{eq:feature-feature-covariance-definition} and \eqref{eq:exact-power-law-latent-generative}; then, we map both sets through random feature models of different sizes, \eqref{eq:statistical-model-random-feature-map}; next, we use the linear regression solution \eqref{eq:linear-regression-solution} to compute test-set predictions using \eqref{eq:test-predictions} for different values of the ridge parameter, $\ridge$, and evaluate the test loss \eqref{eq:def-test-loss} as a function of $\ridge$; finally, we optimize the ridge parameter ($\ridge=\ridge^\star$) and plot the test loss that gives the best performance.\footnote{For the most stable results, we use the form \eqref{eq:linear-regression-solution} of the linear regression solution when the model is underparameterized ($\nf < \nA$), and we use \eqref{eq:linear-regression-solution-data-data-resolvent} when the model is overparameterized  ($\nf > \nA$).
} The figure illustrates the way our statistical model exhibits the 
same
\emph{power-scaling law} and \emph{plateau} regions
as the early-stopped LLMs studied in Ref.~\cite{kaplan2020scaling}, cf. Fig.~\ref{fig:pheno-loss-original}:
in particular,  our numerical simulations are 
predicted
by a phenomenological model of an extremely simple form,
\be\label{eq:phenomenological-model-simplified}
\L(\nf ,\nA) = L_0 \le(\frac{1}{\nf} + \frac{1}{\nA} \ri)^\alpha\, ,
\ee 
where $\alpha$ is the power-law exponent that parameterizes the spectrum defined in \eqref{eq:def-of-alpha}, and $L_0$ is a constant that we will compute explicitly in the following subsection but for now can be thought of as a constant to be fit. This 
equation, \eqref{eq:phenomenological-model-simplified}, follows from the original phenomenological model of the test loss, \eqref{eq:phenomenological-loss-original}, by setting
\be
\alpha \equiv \alpha_{\nf} = \alpha_{\nA} \, , \qquad  L_0 \equiv \nf_0^\alpha = \nA_0^\alpha  \, ,
\ee
and seems to be the appropriate simplification for an optimally-regularized random feature model used for linear regression.%

\subsubsection*{Average Goals}

Having confirmed that our statistical model has the right properties,  
our goal now is to analytically compute the expected value of the test loss,  $\expval{\L_\B(\param^\star)}$, averaged according to the statistics of our generative data model and random feature maps, in order to understand the full phenomenology of neural scaling laws. This will involve: 
averaging over different realizations of the latent training inputs, $x$, and latent test inputs, $\widehat x$; 
averaging over different label weights, $\w$, used to compute training labels, $y$, and test labels, $\widehat{y}$;
averaging over realizations of the noise, $\noise$, added to the training labels;
and averaging over different feature weights $\fw$, 
that determine the training features, $\fea$, and the test features, $\widehat \fea$, given the latent inputs, $x$ and $\widehat x$, respectively.
Since these random variables are all matrices, the computation of the expected test loss is a problem in random matrix theory.   

Some of these averages are very easy to perform and can be evaluated immediately:
\bi
\item The test loss, \eqref{eq:def-test-loss}, is quadratic in the random label noise, $\noise$. Expanding in $\noise$ and using its statistics, \eqref{eq:def-noise-statistics}, we find:
\begin{align}\label{eq:test-loss-noise-average}
    \expval{\L_\B(\param^\star)}_\noise &=\frac{1}{2 \nB }\norm{w (x \fea^T\res  \widehat{\fea}-  \widehat{x})}^2 + \frac{\nout \vare }{2 \nB }\norm{\fea^T\res  \widehat{\fea}}^2  \,.
\end{align}
The first term is independent of the noise, %
and the second term is independent of the labels but depends on the random features $\fea, \widehat \fea$.
Therefore, we will refer to these two terms as the \emph{label term} and the \emph{noise term}, respectively. 
\item The label term now involves a simple square of the label weights, $\w$, which can be averaged over using its statistics, \eqref{eq:teacher-weights-statistics}, to find:
\begin{align}\label{eq:test-loss-averaged-noise-weights}
    \expval{\L_\B(\param^\star)}_{\noise,\w} &=\frac{\nout\wvar}{2 \nB \nl}\norm{x \fea^T\res  \widehat{\fea}-  \widehat{x}}^2 + \frac{\nout \vare }{2 \nB }\norm{\fea^T\res  \widehat{\fea}}^2  \, .
\end{align}
\ei
Since the output dimension $\nout$ just gives an overall scaling of the test loss, %
we will simply set $\nout = 1$ for the rest of the paper without loss of generality.

Two of the three remaining averages, over the latent input training set, $x$, and over the random feature weights, $\fw$, will be more challenging to compute.\footnote{The average over the latent test inputs, $\widehat x$, is relatively easy but we will find it convenient to defer this computation until later.} 
They will be carried out in the remainder of the section, with some of the less conceptual and more mechanical details relegated to Appendix~\ref{sec:delta}. Before moving on to the details, let us meditate on the mechanics of these averages.

The feature functions are defined as the product of two zero-mean Gaussian variables, $x$ and $\fw$, cf. \eqref{eq:statistical-model-random-feature-map}.  
Holding either $x$ or $\fw$ fixed and averaging over the other is straightforward given their statistics, \eqref{eq:feature-feature-covariance-definition} and \eqref{eq:random-feature-weights-statistics}.
From this it follows that the features are centered,
\be\label{eq:feature-mean-any-averaged}
\expval{\fea_{j;\alpha}}= 0 \, , \qquad \expval{\widehat \fea_{j;\beta}}= 0 \, ,
\ee
but their covariances are non-trivial.

It will be convenient to decompose these covariances in terms of matrices that have either sample indices or feature indices, but not both. %
For instance, when averaging over the random features, using \eqref{eq:random-feature-weights-statistics} we find
\be\label{eq:feature-feature-averaged-u-decomposition}
\expval{\fea_{j_1; \alpha_1}\fea_{j_2; \alpha_2}}_{\fw} = \covAA_{\alpha_1 \alpha_2}\, \delta_{j_1 j_2} \, , \quad  \expval{\fea_{j_1; \alpha}\widehat \fea_{j_2; \beta}}_{\fw} = \covAB_{\alpha \beta}\, \delta_{j_1 j_2} \, , \quad \expval{\widehat{\fea}_{j_1;\beta_1}\widehat\fea_{j_2;\beta_2}}_{\fw} = \covBB_{\beta_1 \beta_2} \,\delta_{j_1 j_2}\, ,
\ee
where we have defined the matrices
\be\label{eq:def-Sigmas}
\covAA  \equiv \frac{\fwvar}{\nl} x^T x   \, , \qquad  
\covAB  \equiv \frac{\fwvar}{\nl} x^T \widehat x \, , \qquad \covBB \equiv \frac{\fwvar}{\nl} \widehat x^T \widehat x \, ,
\ee
which have \emph{sample indices} only.  Note that, as they depend on $x$ and $\widehat x$, these matrices are themselves random variables.
Relatedly, when averaging either over the training inputs or over the test inputs, using \eqref{eq:feature-feature-covariance-definition} we find
\be\label{eq:feature-feature-averaged-x-or-xhat-decomposition}
\expval{\fea_{j_1;\alpha_1}\fea_{j_2;\alpha_2}}_{x} = \covf_{j_1 j_2} \delta_{\alpha_1 \alpha_2} \, , \qquad \expval{\widehat\fea_{j_1;\beta_1}\widehat\fea_{j_2;\alpha_2}}_{\widehat x} = \covf_{j_1 j_2} \delta_{\beta_1 \beta_2} \, ,
\ee
where we have defined the random matrix
\be\label{eq:def-Omega}
\covf \equiv  \fw \covl \fw^T   \quad \Longleftrightarrow  \quad  \covf_{j_1j_2} \equiv \sum_{I_1,I_2=1}^{\nl} \fw_{j_1I_1} \fw_{j_2I_2}\covl_{I_1I_2} \, ,
\ee
which has \emph{random feature indices} only and is essentially a projection of the latent-space covariance matrix, $\covl$, onto our model's random feature space. 
Lastly, again using \eqref{eq:feature-feature-covariance-definition} to average over training inputs or test inputs, there's a nontrivial cross-correlation between the latent inputs and the training or test features, 
\be\label{eq:feature-latent-averaged-x-or-xhat-decomposition}
\expval{x_{I;\alpha_1}\fea_{j;\alpha_2}}_{x} = \covlf_{Ij} \, \delta_{\alpha_1 \alpha_2} \, , 
\qquad
\expval{\widehat x_{I;\beta_1}\widehat \fea_{j;\beta_2}}_{\widehat x} = \covlf_{Ij} \,\delta_{\beta_1 \beta_2}\,,
\ee
where we have defined a final random matrix, 
\be\label{eq:def-Omega-tilde}
\covlf \equiv  \covl \fw^T  \quad \Longleftrightarrow  \quad  \covlf_{Ij} \equiv \sum_{J=1}^{\nl}  \fw_{jJ}\covl_{IJ} \,.
\ee
Note that this matrix has a feature index and a latent index but does not depend on samples, and,  as a partial projection, the relation
\be
\covf =  \fw \covlf  
\ee
follows from the above definitions.

Finally, we note that the following 
training-set-averaged covariances vanish, %
\be
 \expval{ x_{I;\alpha}\widehat\fea_{j; \beta}}_{x} = \expval{\widehat x_{I;\beta}\fea_{j;\alpha}}_{x}  =\expval{\fea_{j_1; \alpha} \widehat\fea_{j_2; \beta}}_{x} = 0 \, ,
\ee
and the analogous set of test-set-averaged covariances vanish,
\be
 \expval{x_{I;\alpha} \widehat\fea_{j; \beta}  }_{\widehat x} = \expval{\widehat x_{I;\beta}\fea_{j;\alpha}}_{\widehat x}  =\expval{\fea_{j_1; \alpha} \widehat\fea_{j_2; \beta}}_{\widehat x} = 0 \, ,
\ee
together
indicating no cross-correlation between train and test latent or random features; 
this is as expected, since each sample is drawn independently. We also note that any mixed covariance between random features and latent will vanish when averaged over the random feature weights,
\be
 \expval{ x_{I;\alpha_1} \fea_{j;\alpha_2} }_{\fw} 
 = 
 \expval{  x_{I;\alpha} \widehat \fea_{j;\beta} }_{\fw} 
 =
 \expval{ \widehat x_{I;\beta} \fea_{j;\alpha} }_{\fw} 
 = 
 \expval{  \widehat x_{I;\beta_1} \widehat \fea_{j;\beta_2}}_{\fw} 
 =  0 \, ,
\ee
since these expressions are linear $\fw$. %

\subsubsection*{Final Definitions}

The main difficulty of our analysis is that the resolvent, $\res(\ridge)$,
involves inverses of the feature functions, \eqref{eq:resolvent-feature-feature}.
Roughly speaking, our approach to compute the averaged loss involves %
expanding
it
around $\ridge \to\infty$, thus the expanding factors of $\res(\ridge)$ it contains, and then using the data-averaged covariances, 
\eqref{eq:feature-feature-averaged-x-or-xhat-decomposition} and \eqref{eq:feature-latent-averaged-x-or-xhat-decomposition},
to evaluate the resulting infinite sum of Gaussian expectations. This leads to an implicit equation for a quantity that ultimately determines the test loss, which can be solved in certain limits as well as averaged %
over the random features.\footnote{This is an important difference from Ref.~\cite{bahri2021explaining}, where the authors performed averages over random training data \emph{only} using the results of a replica calculation from Refs.~\cite{bordelon2020spectrum,Canatar:etal}. In Appendix~\ref{sec:other-models}, we discuss such (non-generalized) linear models using the simpler techniques of this paper. In particular, in \S\ref{sec:linear-power-law-model} we explain how models with the right generative data model, but without any random feature maps, behave qualitatively differently than the LLMs observed in Ref.~\cite{kaplan2020scaling}.
}

One wrinkle is that the above only works well in the underparameterized regime with $\nf < \nA$. To analyze the overparameterized regime, $\nf > \nA$, it will be useful to rewrite the linear regression solution, \eqref{eq:linear-regression-solution},
in terms of a \emph{data-data resolvent} matrix, defined as
\be\label{eq:resolvent-data-data}
\Res(\ridge) \equiv \frac{1}{\ridge \IT + \fea^T \fea} \quad \Longleftrightarrow  \quad \Res_{\alpha_1\alpha_2}(\ridge) \equiv \le( \ridge \delta_{\alpha_1 \alpha_2} + \sum_{j=1}^{\nf}\fea_{j;\alpha_1} \fea_{j;\alpha_2} \ri)^{-1} \, ,
\ee
where here $\IT\equiv \delta_{\alpha_1 \alpha_2} $ represents the identity matrix  on sample space. 
To see how to rewrite the linear regression solution, note that
\begin{align}\label{eq:commutation-relation}
\res \fea  %
&= \le[ \sum_{s=0}^{\infty}\frac{1}{\ridge}\le(-\frac{\fea \fea^T}{\ridge}\ri)^s\ri] \fea  \,  \\
&= \frac{\fea}{\ridge} \le[ \IT - \frac{\fea^T \fea}{\ridge} + \le(\frac{\fea^T \fea}{\ridge}\ri)^2 + \dots \ri] \, \notag \\
&= \fea \Res \notag \,,
\end{align}
where on the first line we used the definition of $\res$, \eqref{eq:resolvent-feature-feature}, to expand the resolvent around $\ridge\to\infty$, on the second line we pulled out a $\fea$ from the sum to the left and put the $\fea$ from the right into the sum, and on the third line we resummed the geometric series and used the definition \eqref{eq:resolvent-data-data}.
Using the transpose of this commutation relation, $ \fea^T  \res = \Res \fea^T $,
we can rewrite the linear regression solution,
\eqref{eq:linear-regression-solution}, as
\be\label{eq:linear-regression-solution-data-data-resolvent}
    \theta^\star = (y+\noise)\Res\fea^T \,. 
\ee

Now, let us state
two simple identities that we can use to drastically simplify our calculations.
First, from the definitions of the two resolvents, \eqref{eq:resolvent-feature-feature} and  \eqref{eq:resolvent-data-data} note that %
\be\label{eq:identity-squared}
    \res(\ridge)^2 = - \frac{\partial}{\partial \ridge} \res(\ridge) \, , \qquad \Res(\ridge)^2 = - \frac{\partial}{\partial \ridge} \Res(\ridge) \, .
\ee
With these, we can simplify the averaging by 
eliminating powers of $\res$ and $\Res$ from various expressions.
Second, we also see from these definitions that %
\be\label{eq:identity-inverse}
    (\ridge \IN + \fea \fea^T ) \res = \IN \, ,  \qquad (\ridge \IT + \fea^T \fea ) \Res = \IT \, ,
\ee
which will similarly be used to eliminate factors of $\fea \fea^T$ and $\fea^T \fea$. %

Finally, %
we will find it convenient to adopt the following notation: %
\be\label{eq:def-resolvent-overline}
    \overline{\res} \equiv \expval{\res}_{x} \qquad \overline{\Res} \equiv \expval{\Res}_{x} \, ,
\ee
where the \emph{overline} notation represents a training set average of the resolvent.

\tikzfeynmanset{compat=1.1.0}
\tikzfeynmanset{doubled/.style={
/tikz/double,
/tikz/decoration={name=none}
}}

\subsection{Data and Feature Averaging}\label{sec:data-and-feature-averages}
 
We now begin with the more challenging part of the calculation, the dataset and random feature averages. As we will explain, the expectation of the test loss cannot be computed
for all values of $\nl, \nA, \nf$: instead we will have to settle for expressions that are valid in the limit where %
$\nl, \nA, \nf \gg 1$, though their ratios, $\nl/ \nA$, $\nl / \nf$, and $\nA /\nf$, may still take any value. This is not a problem: the neural scaling laws that LLMs exhibit in practice 
arise for very large data and models sizes, where our solutions are extremely accurate.%
\footnote{From our numerics, cf. Fig.~\ref{fig:main-result}, we will see that subleading corrections are only important for very small training set sizes and numbers of features, $\nA, \nf \lesssim 10$.
} 
And, although we will not need to do so, the techniques that we describe below can be used to systematically compute the subleading corrections to the loss, which %
are suppressed by inverse powers of $\nl$, $\nA$, and $\nf$.

\subsubsection{The Noise Term}
\label{sec:the-noise-term}

We begin with the \emph{noise term}
 in \eqref{eq:test-loss-averaged-noise-weights}.
As the calculation that we
perform
here can 
be repurposed and significantly generalized when we analyze the label term, 
this section will also serve as a gentle introduction to our techniques.

Considering the noise term  in the partially-averaged test loss,
\eqref{eq:test-loss-averaged-noise-weights}, 
and taking expectations over both datasets, $x$, $\widehat x$, and the random feature weights, $\fw$, we get:
\begin{align}\label{eq:test-loss-noise-term}%
     \frac{ \vare }{2 \nB }\expval{\norm{\fea^T\res  \widehat{\fea}}^2}_{\widehat{x},x,\fw}  &=\frac{ \vare }{2 \nB }\expval{ \tr{\fea \fea^T \res  \widehat{\fea}\widehat{\fea}^T \res} }_{\widehat{x},x,u} \, .
     \cr
     &=     \frac{ \vare }{2}\expval{ \tr{\fea \fea^T \res \covf \res }}_{x,\fw} 
     \cr
     &=  \frac{ \vare }{2 } \expval{\left(1 + \ridge \frac{\partial}{\partial \ridge} \right) \tr{\covf \resb}}_{\fw} \, ,
\end{align}
Here, in the first line we expanded the square and expressed it as a trace,  and in the second line we used our expression for the test-set covariance, \eqref{eq:feature-feature-averaged-x-or-xhat-decomposition}, to perform the average over the test set.
Finally, in the third line, we first used our first identity, \eqref{eq:identity-inverse}, to eliminate the $\fea \fea^T$; we second used our second identity, \eqref{eq:identity-squared}, to exchange the $\res^2$ term for a derivative; and we third used the definition,  \eqref{eq:def-resolvent-overline}, to replace $\res$ with $\resb$. In this way, we've reduced the computation to three steps: \emph{(i)} compute the quantity
\be\label{eq:fea-selfenergy-correction}
    \secF \equiv \tr{\covf \resb} \, ,
\ee
which involves evaluating the training-set average of the resolvent, $\resb$, and
then
\emph{(ii)} apply the differential operator, and finally
\emph{(iii)} evaluate its random feature average. %

The only nontrivial step will be \emph{(i)}, which we will now describe:
we need to compute the training-set-averaged resolvent, \eqref{eq:resolvent-feature-feature},
\be\label{eq:resolvent-feature-feature-reprint}
\resb \equiv \expval{\res}_x\,, \qquad   \res \equiv\frac{1}{ \ridge \IN +\fea \fea^T}\, , 
\ee
using the fact that, for fixed $\fw$, the training-set average of 
$\fea$ is given by 
\be\label{eq:varfea}
    \expval{\fea_{j;\alpha}}_x = 0 \,, \qquad  \expval{\fea_{j_1;\alpha_1}\fea_{j_2;\alpha_2}}_{x} = \covf_{j_1 j_2} \delta_{\alpha_1 \alpha_2} \, , \qquad \covf \equiv  \fw \covl \fw^T \,,
\ee
cf. \eqref{eq:feature-mean-any-averaged} and \eqref{eq:feature-feature-averaged-x-or-xhat-decomposition}.
To evaluate $\resb$, we will take the limit of large training set and large number of random features, $\nA, \nf \to \infty$, with their ratio fixed.\footnote{We are also implicitly taking the size of the latent space to be large, $\nl \to \infty$, though it may also have fixed ratios with $\nA$ and $\nf$.
} This computation 
is a classic result of random matrix theory that goes back to the original work of Marchenko and Pastur \cite{Mar_enko_1967} and was further studied in later works (see e.g. \cite{silverstein1995empirical}).\footnote{In particular, the quantity $\tr{\resb(\ridge)}$ 
is the Stieltjes transform of the 
eigenvalue distribution of the matrix $\fea \fea^T$: 
$\tr{\res(\ridge)}$ is a meromorphic function with poles given by the (negative of the) eigenvalues of $\fea \fea^T$; after averaging over $x$ to get $\tr{\resb(\ridge)}$ these poles condense to a branch cut, and so the  discontinuity across this branch cut determines the 
eigenvalue density. %
In the case where 
$\covf$ is proportional to the identity, this reproduces the famous Marchenko-Pastur distribution. 
 \label{footnote:density}
}
Here we give a simple derivation using \emph{Feynman diagram} techniques that can be easily generalized to the other averages we will consider later.\footnote{A similar diagrammatic derivation of $\resb$ can be found in \cite{BURDA2004295}. For other applications of Feynman diagrams in machine learning, see also \cite{dyer2019asymptotics}.
}

To begin, note from its definition, \eqref{eq:resolvent-feature-feature-reprint}, that we can expand the resolvent in a power series in the ridge parameter:
\begin{equation}\label{eq:resolvent-power-series}
    \res(\ridge) = \ridge^{-1}\sum_{s=0}^{\infty} (-\ridge)^{-s} \left(\fea \fea^T\right)^s \, .
\end{equation}
Each term in this expansion is a power of the elements of the matrix $\fea$.  For fixed $\fw$, the elements of the matrix $\fea$ are Gaussian random variables with statistics \eqref{eq:varfea}. The higher-order moments of a zero-mean Gaussian distribution are determined entirely by the covariance matrix, $\covf_{j_1 j_2} \delta_{\alpha_1 \alpha_2}$, and can be determined by a repeated application of \eqref{eq:varfea}.
Let us enumerate the first few terms, which will make our overall strategy clear:
\bi
\item The $s=0$ term is trivial and simply given by $\ridge^{-1}$.
\item The $s=1$ term is also trivial and can be evaluated directly using the middle equation in \eqref{eq:varfea}, giving
\be\label{eq:s-equal-1-term}
-\ridge^{-2}\expval{\fea \fea^T}_x = - \ridge^{-2}\nA \covf \, ,
\ee
where the factor of $\nA$ comes from the sum over the training set.
\item The $s=2$ term involves the average of the quartic matrix $\fea \fea^T \fea \fea^T$. 
Generally, %
the expectation value of a product of four arbitrary $\fea$ components is a sum 
of the three different ways that these features can be ``paired up" together, with each pairing weighted by the appropriate covariance:\footnote{This an example of a general fact about Gaussian distributions: the expectation value of a product of Gaussian random variables is always a sum over the different ways that the variables can be paired up. This is known as \emph{Isserlis' theorem} in probability theory and \emph{Wick's theorem} in the physics literature. Note that if the statistics of $\fea$ were non-Gaussian, there could be an additional contribution to the right-hand side of \eqref{4ptwick} related to the fourth cumulant of the distribution.} 
\begin{align}\label{4ptwick}
\braket{\fea_{j_1; \alpha_1} \fea_{j_2;\alpha_2}\fea_{j_3;\alpha_3} \fea_{j_4;\alpha_4}}_x = &\braket{\fea_{j_1; \alpha_1} \fea_{j_2;\alpha_2}}_x \braket{\fea_{j_3;\alpha_3} \fea_{j_4;\alpha_4}}_x +\, \\
&\braket{\fea_{j_1; \alpha_1} \fea_{j_3;\alpha_3}}_x \braket{\fea_{j_2;\alpha_2} \fea_{j_4;\alpha_4}}_x +\,  \notag \\
&\braket{\fea_{j_1; \alpha_1} \fea_{j_4;\alpha_4}}_x \braket{\fea_{j_2;\alpha_2} \fea_{j_3;\alpha_3}}_x \, , \notag 
\end{align}
which can be evaluated using \eqref{eq:varfea} to give
\be
\covf_{j_1 j_2} \covf_{j_3 j_4}  \delta_{\alpha_1 \alpha_2}\delta_{\alpha_3 \alpha_4}
+ \covf_{j_1 j_3} \covf_{j_2 j_4} \delta_{\alpha_1 \alpha_3} \delta_{\alpha_2 \alpha_4}
+\covf_{j_1 j_4} \covf_{j_2 j_3} \delta_{\alpha_1 \alpha_4} \delta_{\alpha_2 \alpha_3} \,. 
\ee
We can use this to evaluate $\braket{\fea \fea^T \fea \fea^T}_x$ by setting 
$\alpha_1=\alpha_2$, $\alpha_3=\alpha_4$, $j_2=j_3$ and summing, which altogether gives for the $s=2$ term:
\be\label{4ptcontracted}
\ridge^{-3} \expval{\left(\fea \fea^T\right)^2}_x = \ridge^{-3}  \Big(\nA^2 \covf^2 + \nA \covf^2 + \nA  \covf \, \tr{\covf}\Big)\, .
\ee
The three terms in this expression correspond directly to the three terms in~\eqref{4ptwick}.
\ei
In principle, it is straightforward -- though tedious -- to proceed order-by-order in this way, evaluating the average of each term. 
The main challenge %
is 
that
the super-exponential proliferation of pairings that arise for larger $s$ makes it difficult to
keep track of how the indices in the resulting expressions 
are summed over: as is already evident in 
\eqref{4ptcontracted}, many different types of terms can arise, with different powers of $\nA$, $\covf$, and $\tr{\covf}$.

The method of \textbf{Feynman diagrams} is simply a diagrammatic technique that we can use to keep track of these various terms. 
For example, we represent the expectation value of the power series for the averaged resolvent, \eqref{eq:resolvent-power-series}, as the following diagrammatic expression: %
\begin{align}\label{eq:feyn-example}
{\resb} \quad &\equiv \quad
\begin{tikzpicture}[baseline=-3pt]
\begin{feynhand}
\vertex (a) at (0,0); \vertex[ringblob](b) at (1,0) {${\resb}$} ; \vertex(c) at (2,0);
\propag (a) to (b); \propag (b) to (c);
\end{feynhand}
\end{tikzpicture}
\\
&=
~~\,\gamma^{-1}~
\begin{tikzpicture}[baseline=-2pt]
\begin{feynhand}
\vertex (a) at (0,0); \vertex(b) at (2.0,0) ;
\propag (a) to (b);
\end{feynhand}
\end{tikzpicture} 
\quad
\notag
\\
&
~~~-\gamma^{-2}~\begin{tikzpicture}[baseline=-3pt]
\begin{feynhand}
\vertex (a) at (0,0); \vertex (b) at (1.5,0);
\vertex (c) at (.1,0); \vertex (d) at (1.4,0);
\vertex (e) at (.1,0); \vertex (f) at (1.4,0);
\vertex (g) at (-.5,0); \vertex (h) at (0,0);
\vertex (i) at (1.5,0); \vertex (j) at (2,0);
\propag (a) to [in=90, out=90, looseness=1.75] (b);
\propag[sca] (c) to [in=90, out=90, looseness=1.75] (d);
\propag[sca] (e) to (f);
\propag (g) to (h);
\propag (i) to (j);
\end{feynhand}
\end{tikzpicture}
\quad
\notag\\
&~~~+\gamma^{-3}~\begin{tikzpicture}[baseline=-3pt]
\begin{feynhand}
\vertex (a) at (0,0); \vertex (b) at (1.5,0);
\vertex (c) at (.1,0); \vertex (d) at (1.4,0);
\vertex (e) at (.1,0); \vertex (f) at (1.4,0);
\vertex (g) at (-.5,0); \vertex (h) at (0,0);
\vertex (i) at (1.5,0); \vertex (j) at (2,0);
\vertex (a2) at (2,0); \vertex (b2) at (3.5,0);
\vertex (c2) at (2.1,0); \vertex (d2) at (3.4,0);
\vertex (e2) at (2.1,0); \vertex (f2) at (3.4,0);
\vertex (i2) at (3.5,0); \vertex (j2) at (4,0);
\propag (a) to [in=90, out=90, looseness=1.75] (b);
\propag[sca] (c) to [in=90, out=90, looseness=1.75] (d);
\propag[sca] (e) to (f);
\propag (g) to (h);
\propag (i) to (j);
\propag (a2) to [in=90, out=90, looseness=1.75] (b2);
\propag[sca] (c2) to [in=90, out=90, looseness=1.75] (d2);
\propag[sca] (e2) to (f2);
\propag (i2) to (j2);
\end{feynhand}
\end{tikzpicture}
~+\gamma^{-3}~
\begin{tikzpicture}[baseline=-3pt]
\begin{feynhand}
\vertex (a) at (0,0); \vertex (b) at (0.5,0);
\vertex (c) at (.6,0); \vertex (d) at (1.4,0);
\vertex (e) at (1.5,0); \vertex (f) at (2.5,0);
\vertex (g) at (2.6,0); \vertex (h) at (3.4,0);
\vertex (i) at (3.5,0); \vertex (j) at (4,0);
\vertex (k) at (0.5,0); \vertex (l) at (2.5,0);
\vertex (m) at (0.6,0); \vertex (n) at (2.6,0);
\vertex (o1) at (1.4,0); \vertex (p1) at (1.85,.85);
\vertex (q1) at (1.5,0); \vertex (r1) at (1.85,.75);
\vertex (o2) at (2.3, 1); \vertex (p2) at (3.4,0);
\vertex (q2) at (2.3, .9); \vertex (r2) at (3.5,0);
\propag (a) to (b);
\propag[sca] (c) to (d);
\propag (e) to (f);
\propag[sca] (g) to (h);
\propag (i) to (j);
\propag (k) to [in=90, out=90, looseness=1.75] (l);
\propag[sca] (m) to [in=90, out=90, looseness=1.75] (n);
\propag[sca] (o1) to [in=210, out=90, looseness=.875] (p1);
\propag (q1) to [in=210, out=90, looseness=.875] (r1);
\propag (q2) to [in=90, out=30, looseness=.875] (r2);
\propag[sca] (o2) to [in=90, out=30, looseness=.875] (p2);
\end{feynhand}
\end{tikzpicture}
~+\gamma^{-3}~
\begin{tikzpicture}[baseline=-3pt]
\begin{feynhand}
\vertex (w) at (2.5,0); \vertex (x) at (3,0);
\vertex (y) at (-.5,0); \vertex (z) at (0,0);
\vertex (a) at (0,0); \vertex (b) at (2.5,0);
\vertex (c) at (.1,0); \vertex (d) at (2.4,0);
\vertex (e) at (.1,0); \vertex (f) at (0.4,0);
\vertex (g) at (2.1,0); \vertex (h) at (2.4,0);
\vertex (i) at (.4,0); \vertex (j) at (2.1,0);
\vertex (k) at (.5,0); \vertex (l) at (2,0);
\vertex (m) at (.5,0); \vertex(n) at (1.25,0) ;  \vertex (o) at (2,0);
\propag (a) to [in=90, out=90, looseness=1.75] (b);
\propag[sca] (c) to [in=90, out=90, looseness=1.75] (d);
\propag[sca] (e) to (f);
\propag[sca] (g) to (h);
\propag[sca] (i) to [in=90, out=90, looseness=1.75] (j);
\propag (k) to [in=90, out=90, looseness=1.75] (l);
\propag (m) to (n); \propag (n) to (o);
\propag (w) to (x);
\propag (y) to (z); 
\end{feynhand}
\end{tikzpicture}
\notag
~
\\\notag
\\
&~~~+\dots\notag
\end{align}
In this expression,
each diagram on the right-hand side represents a term in the expansion and can be interpreted using the following rules: 
\bi
\item Each \emph{single horizontal solid line} represents a \emph{random-feature} $j$-type index as
\be\label{eq:def-random-feature-rule}
\delta_{j_1 j_2}
\quad \equiv \quad 
_{j_1}\,
\begin{tikzpicture}[baseline=-3pt]
\begin{feynhand}
\vertex (a) at (0,0); 
\vertex(c) at (2,0);
\propag (a) to (c);
\end{feynhand}
\end{tikzpicture}
\,
_{j_2}
\quad
.
\ee
\item Each \emph{single horizontal dashed line} represents a \emph{training-sample} $\alpha$-type index as
\be\label{eq:def-training-sample-rule}
\delta_{\alpha_1 \alpha_2}
\quad \equiv \quad 
_{\alpha_1}\,
\begin{tikzpicture}[baseline=-3pt]
\begin{feynhand}
\vertex (a) at (0,0); 
\vertex(c) at (2,0);
\propag[sca] (a) to (c);
\end{feynhand}
\end{tikzpicture}
\,
_{\alpha_2}
\quad
.
\ee
\item Then, we can represent $\fea \equiv \fea_{j;\alpha}$ as the \emph{vertex} of a horizontal solid line and a horizontal dashed line as
\be\label{eq:def-phi-rule}
\fea_{j;\alpha} 
\quad \equiv \quad 
_{j}\,
\begin{tikzpicture}[baseline=-3pt]
\begin{feynhand}
\vertex (a) at (0,0); 
\vertex (b) at (.95,0) ; 
\vertex(c) at (1.05,0);
\vertex(d) at (2,0);
\propag (a) to (b);
\propag[sca] (c) to (d);
\end{feynhand}
\end{tikzpicture}
\,
_{\alpha}
\quad ,
\ee
with $\fea^T$ drawn as the mirror image, with the dashed line preceding the solid line.
\item Each \emph{set of curved double lines} represents the \emph{random-feature covariance matrix}
as
\be\label{eq:def-covariance-omega-rule}
\covf_{j_1 j_2}\delta_{\alpha_1\alpha_2}
\quad \equiv \quad 
_{j_1;\alpha_1}
\begin{tikzpicture}[baseline=-3pt]
\begin{feynhand}
\vertex (a) at (0,0); \vertex (b) at (1.5,0);
\vertex (c) at (.1,0); \vertex (d) at (1.4,0);
\propag (a) to [in=90, out=90, looseness=1.75] (b);
\propag[sca] (c) to [in=90, out=90, looseness=1.75] (d);
\end{feynhand}
\end{tikzpicture}
_{j_2;\alpha_2}
\quad
.
\ee
Note that it doesn't matter whether the dashed line is above or below the solid line, and in some diagrams they may even twist around each other.
\item If two lines are connected, the corresponding indices are set equal and summed over.
\item If a closed loop appears, the corresponding index is summed over. 
\ei
Note also that for each of the diagrams in the expansion, we have written explicitly the factors of $\ridge$ which accompany each of the terms in the sum. %
Please convince yourself that with these rules the second, third, and fourth lines in the diagrammatic expression \eqref{eq:feyn-example} correspond to the $s=0$, $s=1$, and $s=2$ terms, respectively, in the expansion computed above.  

For example, 
consider the $s=1$ term in the third line: first, 
we write $\covf_{j_1 j_2}\delta_{\alpha_1\alpha_2}$ for the curved double lines; then, we set $\alpha_1 = \alpha_2$ for the horizontal dashed line connecting to the curved dashed line; lastly, we sum over its index to give a factor of $\nA$.
Including the correct factors of 
the ridge parameter and the overall sign, we find $-\ridge^{-2}\nA \covf$, matching the $s=1$ term, \eqref{eq:s-equal-1-term}. Similarly, one can check that
 the three diagrams on the final line correspond to each of the three terms in the $s=2$ contribution, \eqref{4ptcontracted}.\footnote{Note that you can also read off the form of the term before any averaging by considering the horizontal straight lines along the bottom of a diagram: 
 the matrix $\fea\equiv \fea_{j;\alpha}$ is represented by a solid line followed by a dashed line, and its transpose,  $\fea^T\equiv \le(\fea_{j;\alpha}\ri)^T$, is represented a dashed line followed by a solid line. The curved lines emerging from these vertices then determine the different pairings that result when averaging over $x$.
 }

This Feynman diagram representation of the expansion is useful because it helps us select the terms that dominate in the large-$\nA$
 and large-$\nf$ limit. 
 Each closed loop in a Feynman diagram represents a trace:
a loop of a dashed line will give a factor of $\nA$ coming from the trace of the identity matrix $\IT$, while a loop of a solid line  will give a factor of $\tr{\covf^p}$ for some integer $p$, which will overall scale linearly with $\nf$ in the large-$\nf$ limit.\footnote{
    This latter statement depends on the assumption that the elements $\covf_{j_1 j_2}$ remain of order one as $\nf$ is increased. This is explicitly true for the data models considered in our paper. 
}

With these rules, it turns out
that a simple way to evaluate the overall scaling of a diagram is to consider its \emph{topology}:
the diagrams that dominate at large $\nA$ and $\nf$ are \emph{planar}, i.e. 
can be drawn on the plane without any lines crossing one another.   %
For example, inspecting \eqref{eq:feyn-example}, the only non-planar diagram  is the middle diagram on the final line; this reflects the fact that the middle term in \eqref{4ptcontracted}  scales only \emph{linearly} with  $\nA$, while the first and last terms scale \emph{quadratically}, as $\nA^2$ and $\nA \tr{\covf}  \sim \nA\nf$, respectively. 
Indeed, when we take the limit of large $\nA$ and $\nf$ with the fixed ratio $\nA/\nf$, then there is a simple way to organize our calculation as an expansion in $1/\nf$: %
diagrams that can only be drawn on 
a surface of a more complicated topology
are suppressed relative to the leading contributions by powers 
of $\nf$.\footnote{In physics, this is known as the 't Hooft large-$N$ expansion \cite{tHooft:1973jz}, and the limit of large $\nA$ and $\nf$ is often referred to as the \emph{planar limit}. 
In mathematics, this observation is the starting point for the study of \emph{free probability} (see, e.g., \cite{mingo2017free}): in free probability theory, we replace the usual cumulants, 
which are sums over partitions, by sums only over \emph{non-crossing} partitions. 
The non-crossing criterion is, in the language of Feynman diagrams, just the statement that we include only planar diagrams.
For example, in free probability we would label the three terms in \eqref{4ptwick} by the three possible partitions of the index set $\{1,2,3,4\} $ into pairs, as $ \{1,2\}\cup\{3,4\}$, $ \{1,3\}\cup\{2,4\}$ and $ \{1,4\}\cup \{2,3\}$.  The first and third are non-crossing partitions, while the second ``crossing" partition is excluded in free probability.  Comparing to \eqref{eq:feyn-example}, this is just the planar limit.} 
As we are only interested in the leading contribution, we will only need to consider planar diagrams in our calculation.  By focusing on the leading contribution, we can thus make a significant reduction in the number of terms that we have to consider.

Even so, there are \emph{still} an infinite number of diagrams that contribute $\resb$. To simplify things further, we will organize the expansion of the resolvent, \eqref{eq:resolvent-power-series}, in a slightly different way.  First, let us  write the \emph{averaged} resolvent as %
\be\label{eq:def-self-energy-features}
 \resb(\ridge) \equiv \frac{1}{\ridge \IN + \senF(\ridge)}\,, %
\ee
which we take as the definition of some $\nf\times\nf$ matrix, $\senF$. Note importantly that $\senF$ is independent of $x$, as $x$ has already been averaged over in this formula. Comparing the original definition $\res$, \eqref{eq:resolvent-feature-feature-reprint}, with the definition of $\resb$, \eqref{eq:def-self-energy-features}, we see that the entire effect of averaging has been packaged into a substitution of the random matrix  $\fea \fea^T$ for the fixed matrix $\senF$.\footnote{In the physics literature, quantities like $\senF$ that appear in the denominators of averaged resolvents or propagators of interacting systems are often referred to as \emph{self-energies}.  As an $\nf$-by-$\nf$-dimensional matrix that in principle depends on the random feature weights, $\fw$, one might therefore call $\senF$ the \emph{self-energy of the features}. We only mention this name as an explanation for our choice of symbol, $\senF$. 
} 
Next, expanding the definition \eqref{eq:def-self-energy-features} for large $\ridge$, we can express $\resb(\ridge)$ as series,
\begin{equation}\label{eq:resolvent-power-series-self-energy}
    \res(\ridge) = \ridge^{-1}\sum_{s=0}^{\infty} (-\ridge)^{-s} \le(\senF\ri)^s \, ,
\end{equation}
and use our Feynman rules to represent 
the terms in that series diagrammatically as
\begin{equation}\label{eq:feyn-self}
\begin{tikzpicture}[baseline=-3pt]
\begin{feynhand}
\vertex (a) at (0,0); \vertex[ringblob](b) at (1,0) {${\resb}$} ; \vertex(c) at (2,0);
\propag (a) to (b); \propag (b) to (c);
\end{feynhand}
\end{tikzpicture}
\quad
=
\quad
\gamma^{-1}~
\begin{tikzpicture}[baseline=-2pt]
\begin{feynhand}
\vertex (a) at (0,0); \vertex(b) at (2.0,0) ;
\propag (a) to (b);
\end{feynhand}
\end{tikzpicture}
~~-~~\gamma^{-2}~
\begin{tikzpicture}[baseline=-0.1cm]
\begin{feynhand}
\vertex (a) at (0,0); \vertex[ringblob](b) at (1,0) {$\senF$} ; \vertex(c) at (2,0);
\propag (a) to (b); \propag (b) to (c);
\end{feynhand}
\end{tikzpicture}
~~+~~\gamma^{-3}~
\begin{tikzpicture}[baseline=-0.1cm]
\begin{feynhand}
\vertex (a) at (0,0); \vertex[ringblob](b) at (1,0) {$\senF$} ; \vertex[ringblob](c) at (2.5,0) {$\senF$}; \vertex(d) at (3.5,0); 
\propag (a) to (b); \propag (b) to (c); \propag (c) to (d);
\end{feynhand}
\end{tikzpicture}
~~
-
~~
\dots \,,
 \end{equation}
where the circle labeled with ``$\senF$'' represents an insertion of the matrix $\senF \equiv \senF_{j_1 j_2}$. %
Of course, this is only a reorganization of our original expansion \eqref{eq:feyn-example}: %
all that we've done is assumed that $\resb$ can be resummed in a particular way and then
repackaged our ignorance into the as-yet unknown matrix $\senF$.

To make this useful, let's develop
a Feynman diagram expansion for %
$\senF$.  This expansion must have the property that when inserted into the $\resb$ expansion, \eqref{eq:feyn-self}, it reproduces our original expansion, 
\eqref{eq:feyn-example}.  %
In the planar limit,
it turns out that the correct expansion is:
\begin{align}\label{eq:ip1-expansion}
\begin{tikzpicture}[baseline=-3pt]
\begin{feynhand}
\vertex[ringblob](a) at (0,0) {${\senF}$}; 
\propag (a) to (a);
\end{feynhand}
\end{tikzpicture}
\quad
\equiv 
\quad
\begin{tikzpicture}[baseline=-3pt]
\begin{feynhand}
\vertex (a) at (0,0); \vertex (b) at (1.5,0);
\vertex (c) at (.1,0); \vertex (d) at (1.4,0);
\vertex (e) at (.1,0); \vertex (f) at (1.4,0);
\propag (a) to [in=90, out=90, looseness=1.75] (b);
\propag[sca] (c) to [in=90, out=90, looseness=1.75] (d);
\propag[sca] (e) to (f);
\end{feynhand}
\end{tikzpicture}
\quad
-
\quad
\begin{tikzpicture}[baseline=-3pt]
\begin{feynhand}
\vertex (a) at (0,0); \vertex (b) at (2.5,0);
\vertex (c) at (.1,0); \vertex (d) at (2.4,0);
\vertex (e) at (.1,0); \vertex (f) at (0.4,0);
\vertex (g) at (2.1,0); \vertex (h) at (2.4,0);
\vertex (i) at (.4,0); \vertex (j) at (2.1,0);
\vertex (k) at (.5,0); \vertex (l) at (2,0);
\vertex (m) at (.5,0); \vertex[ringblob](n) at (1.25,0) {${\resb}$} ;  \vertex (o) at (2,0);
\propag (a) to [in=90, out=90, looseness=1.75] (b);
\propag[sca] (c) to [in=90, out=90, looseness=1.75] (d);
\propag[sca] (e) to (f);
\propag[sca] (g) to (h);
\propag[sca] (i) to [in=90, out=90, looseness=1.75] (j);
\propag (k) to [in=90, out=90, looseness=1.75] (l);
\propag (m) to (n); \propag (n) to (o);
\end{feynhand}
\end{tikzpicture}
\quad
+
\quad
\begin{tikzpicture}[baseline=-3pt]
\begin{feynhand}
\vertex (a) at (0,0); \vertex (b) at (4.5,0);
\vertex (c) at (.1,0); \vertex (d) at (4.4,0);
\vertex (e) at (.1,0); \vertex (f) at (0.4,0);
\vertex (g1) at (2.1,0); \vertex (h1) at (2.4,0);
\vertex (i1) at (.4,0); \vertex (j1) at (2.1,0);
\vertex (k1) at (.5,0); \vertex (l1) at (2,0);
\vertex (m1) at (.5,0); \vertex[ringblob](n1) at (1.25,0) {${\resb}$} ;  \vertex (o1) at (2,0);
\vertex (g2) at (4.1,0); \vertex (h2) at (4.4,0);
\vertex (i2) at (2.4,0); \vertex (j2) at (4.1,0);
\vertex (k2) at (2.5,0); \vertex (l2) at (4,0);
\vertex (m2) at (2.5,0); \vertex[ringblob](n2) at (3.25,0) {${\resb}$} ;  \vertex (o2) at (4,0);
\propag (a) to [in=90, out=90, looseness=1.25] (b);
\propag[sca] (c) to [in=90, out=90, looseness=1.25] (d);
\propag[sca] (e) to (f);
\propag[sca] (g1) to (h1);
\propag[sca] (i1) to [in=90, out=90, looseness=1.75] (j1);
\propag (k1) to [in=90, out=90, looseness=1.75] (l1);
\propag (m1) to (n1); \propag (n1) to (o1);
\propag[sca] (g2) to (h2);
\propag[sca] (i2) to [in=90, out=90, looseness=1.75] (j2);
\propag (k2) to [in=90, out=90, looseness=1.75] (l2);
\propag (m2) to (n2); \propag (n2) to (o2);
\end{feynhand}
\end{tikzpicture}
\quad
-
\quad
\dots
\,.
\end{align}
Note that \emph{(a)}, 
that rather than using solid horizontal lines instead we use subdiagrams labeled with
``$\resb$'', which represents an insertion of the \emph{averaged} $\resb$ matrix, 
and 
\emph{(b)} 
that
the signs alternate between diagrams.\footnote{
    In the physics literature,
    $\resb$ is sometimes referred to as a \emph{dressed} propagator, while, by contrast, the matrix corresponding to the solid horizontal line, $\IN$, is known as a \emph{bare} propagator.  
}

To see why this is the correct expansion, 
note that diagrams in this expansion have the property that they cannot be split into two parts by cutting a single or double line in one particular place.\footnote{In the physics literature, such diagrams are known as one-particle irreducible (1PI).} 
Contrast that with the first diagram on the fourth line in \eqref{eq:feyn-example}: there, if we cut the middle solid line connecting the two rising suns, then the diagram will be split into two separate pieces. This underscores the reason why organizing the computation in terms of $\senF$ is helpful and suggests the strategy for its computation: first, write down an expansion for \emph{all} the diagrams that have the property that they cannot be split by cutting a solid line according to our conditions \emph{(a)}--\emph{(b)}; then, by packaging them back into the expansion 
\eqref{eq:feyn-self},
we will generate \emph{all} the needed diagrams, including those that fall into two disconnected pieces when a line is cut. %
To check that this claim is correct, you can insert this definition, \eqref{eq:ip1-expansion}, for $\senF$ into the expansion \eqref{eq:feyn-self}; this will give exactly the same expansion as our original one, \eqref{eq:feyn-example}, provided that 
every time either $\resb$ or $\senF$ appears 
you also
recursively insert expansions \eqref{eq:feyn-self} or \eqref{eq:ip1-expansion}, respectively.
The utility of this is that drawing all planar irreducible-by-cutting diagrams is much simpler than trying to draw all diagrams overall, and so by restricting to this set of diagrams -- or equivalently, the terms in the expansion represented by these diagrams -- we can make the computation of $\resb$ not only tractable, but simple.
Note that for this to work, the large-$\nf$ limit restriction to planar diagrams
was absolutely crucial: at finite $\nf$, simple recursive representations like this do not exist, although in many cases it is possible to systematically compute subleading $1/\nf$ corrections.

All that remains is to actually evaluate $\senF$.
Evaluating the diagrams in the expansion \eqref{eq:ip1-expansion} according to our rules for 
Feynman diagrams,
we see that the pattern is 
\be\label{eq:self-energy-before}
    \senF = \nA \covf  - \nA \covf\, \tr{\covf \resb} + \nA \covf \,\tr{\covf \resb}^2 + \dots + \nA \covf \le(-\tr{\covf \resb}\ri)^s + \dots \, .
\ee
Recognizing yet
another geometric series, 
we can sum it up to get
\be\label{eq:self-energy-solution}
\senF(\ridge) = \frac{\nA \covf}{1 + \secF(\ridge)} \, ,
\ee
where we've recalled our earlier definition, \eqref{eq:fea-selfenergy-correction}.\footnote{Analogous to $\senF$, you might call $\secF$ the \emph{Delta of the features}.} %
Together \eqref{eq:def-self-energy-features} and \eqref{eq:self-energy-solution} form
a system of two coupled matrix equations that can be solved to determine the $x$-averaged resolvent $\resb$, and the quantity that we ultimately need, $\secF$.\footnote{In the physics literature, \eqref{eq:feyn-self} and \eqref{eq:ip1-expansion} 
    are sometimes known as Schwinger-Dyson equations~\cite{DysonEq,Schwinger452}.
Note also that there is no simple closed form solution of these equations, except in certain special cases.
One 
scenario in which it is possible to jointly solve these equations is when the covariance proportional to the identity matrix: $\covf = \sigma^2 \IN$. In this case, it's easy to check that the system, \eqref{eq:feyn-self} and \eqref{eq:ip1-expansion}, reduces to a simple quadratic equation for $\tr{\resb}$.
 We discuss this case explicitly in \S\ref{sec:linear-mp-model} of the appendix.
}

To finally finish item \emph{(i)} in our three-step plan for calculating the noise term, 
let us 
rewrite our system of equations as an implicit equation for $\secF$ as a function of $\covf$: 
\be\label{eq:self-consistent}
    \secF(\ridge) = \tr{\frac{\covf}{ \ridge \IN + \frac{\nA \covf}{1+\secF(\ridge)}}} \, .
\ee
Here, we have used \eqref{eq:self-energy-solution} to eliminate $\senF$ from \eqref{eq:def-self-energy-features}, multiplied $\resb$ by $\covf$, and then took the trace of both sides. This equation can be solved numerically as a function of $\covf$, $\nA$, and $\ridge$, or can be analyzed analytically in the ridgeless limit ($\ridge \to 0$). %
In this paper, our analysis will mostly be concerned with the ridgeless limit. Now, let's continue with our plan with step \emph{(ii)}, %
applying the differential operator,
$\le(1+\ridge \frac{\partial}{\partial \ridge}\ri)$, and step \emph{(iii)}, averaging over the random features, $\fw$.

To apply step \emph{(ii)} in the ridgeless limit, we should
expand $\secF$ near $\ridge\to0$:\footnote{The justification for such an expansion is 
provided
in Appendix~\ref{sec:delta}, but intuitively 
note the possibility of a $1/\ridge$ divergence as $\ridge \to 0$ by considering the self-consistent equation \eqref{eq:self-consistent} as $\secF$  becomes large.}
\be\label{eq:Delta-expansion-ansatz-maintext}
\secF(\ridge) \equiv \frac{\secF_{-1}}{\ridge} + \secF_{0} + \secF_{1} \ridge + \dots \, .
\ee
Applying our differential operator to this expansion eliminates the leading term %
as
\begin{align}\label{eq:ridge-limit-noise}
\lim_{\ridge \rightarrow 0} \left(1+\ridge\frac{\partial}{\partial \ridge} \right) \secF(\ridge)  = \secF_{0} \, ,
\end{align} 
which means that in this limit the noise term, \eqref{eq:test-loss-noise-term}, has a very simple expression:
\be\label{eq:second-to-final-noise-term}
\frac{ \vare }{2 \nB }\expval{\norm{\fea^T\res  \widehat{\fea}}^2}_{\widehat{x},x,\fw} = \frac{ \vare }{2 } \expval{\secF_{0}}_{\fw}\,,
\ee
where 
$\secF_0$ is understood to be determined by solving the self-consistent equation \eqref{eq:self-consistent}.
  It turn out the solution for $\secF_0$ depends on
whether the model is underparameterized ($\nA > \nf$) or overparameterized ($\nA < \nf$) and also on whether the model is in the regime valid for neural scaling behavior ($\nf < \nl$) or in a regime where that behavior breaks down ($\nl > \nf$). 
The computation of $\secF_0$ in these four cases is somewhat technical and not particularly illuminating, so is relegated to Appendix~\ref{sec:delta-zero} %
(under the final subheading labeled \emph{$\secG_0$ for the Truncated Power-Law Spectrum}).

Finally, for step \emph{(iii)} we note that the solution for $\secF_0$ given in \eqref{eq:delta-zero-complete-projected} 
does not explicitly depend on the fixed random feature weights, $\fw$. Thus, the average over $\fw$ is trivial, and, substituting  \eqref{eq:delta-zero-complete-projected} for $\secF_0$ in  \eqref{eq:second-to-final-noise-term}, 
we find a final answer for the noise term:
\begin{equation}\label{eq:noise-term-final}
 \L_\noise \equiv \frac{ \vare }{2 \nB }\expval{\norm{\fea^T\res  \widehat{\fea}}^2}_{\widehat{x},x,\fw} = \frac{ \vare }{2 }
    \begin{cases}
        \alpha + \frac{1}{\nf/\nA-1} \,,   & \nA < \nf \,, \;\; \nf<\nl \,, \\
        \frac{1}{\nA/\nf-1} \,,            & \nA > \nf \,, \;\; \nf<\nl \, ,\\
        \alpha + \frac{1}{\nl/\nA-1} \,,   & \nA < \nl \,, \;\; \nf>\nl \,, \\
        \frac{1}{\nA/\nl-1} \,,            & \nA > \nl \,, \;\; \nf>\nl \, .
    \end{cases}
\end{equation}

\subsubsection{The Label Term}\label{sec:derivation-label}

We now move on to the more complicated \emph{label term} in \eqref{eq:test-loss-averaged-noise-weights}. Starting from the partially-averaged expression and taking expectations over both datasets, $x$, $\widehat x$, and the random feature weights, $\fw$, we expand the square to express the term as a sum of traces:
\begin{align}\label{eq:label-term-expanded}
   \L_\w &\equiv \frac{\wvar}{2 \nB \nl}\expval{\norm{x \fea^T\res  \widehat{\fea}-  \widehat{x}}^2 }_{\widehat{x},x,\fw} \,  \\
 &=  \frac{\wvar}{2 \nB \nl} \expval{\Big(\tr{\widehat{x}\widehat{x}^T} - 2\tr{\widehat{x}\widehat{\fea}^T \res \fea x^T } + \tr{\widehat{\fea}\widehat{\fea}^T \res \fea x^T x \fea^T \res} \Big)}_{\widehat{x},x,\fw} \, \notag \\
 &=  \frac{\wvar}{2 \nl}  \le(\tr{\covl} - 2\expval{\tr{\covlf \res \fea x^T} }_{x,\fw} + \Big\langle\tr{\covf \res \fea x^T x \fea^T \res}\Big\rangle_{x,\fw} \ri)
 \notag
 \,.
\end{align} 
In the final line, we used the random feature covariance, \eqref{eq:feature-feature-averaged-x-or-xhat-decomposition}, and the 
mixed latent-feature-random-feature covariance, \eqref{eq:feature-latent-averaged-x-or-xhat-decomposition},
to perform the averages over $\widehat x$. Thus, we have two expectations to evaluate.

To calculate these $x$ and $\fw$ expectations, we will need to separately consider the cases of underparameterization ($\nf < \nA$) and overparameterization ($\nf > \nA$).
These computations are somewhat technical and follow very
similar techniques to those discussed in the previous subsection; 
if you  feel you've already got the hang of these techniques and are now ready for the final result
please feel free to 
skip ahead to \S\ref{sec:derivation-results}.

\subsubsection*{The Underparameterized Regime} %

We start with the case of $\nf < \nA$, 
proceeding in two steps: 
\emph{(i)} first, we'll compute the average over the training set, $x$, 
and 
\emph{(ii)} then, we'll perform the average over the random feature weights, $\fw$.
\\ %

\noindent\emph{(i)} \textbf{Training Set Average}\\

To perform the training set average, we 
will apply the same Feynman diagram techniques considered earlier to 
more involved expectations.  The essential difference is that we now must consider both the random and latent features, $\fea$ and $x$, as Gaussian random matrices, with covariances  given by
\be\label{eq:label-statistics-needed-for-data-averaging}
\braket{ x_{I_1;\alpha_1} x_{I_2;\alpha_2}}_x = \covl_{I_1I_2} \delta_{\alpha_1 \alpha_2}\,,
\qquad
 \braket{\fea_{j_1;\alpha_1} \fea_{j_2;\alpha_2}}_x = \covf_{j_1 j_2} \delta_{\alpha_1\alpha_2}\,, 
 \qquad 
 \braket{x_{I;\alpha_1} \fea_{j;\alpha_2}}_x = \covlf_{Ij} \delta_{\alpha_1 \alpha_2}\,,
\ee
cf. \eqref{eq:feature-feature-covariance-definition}, \eqref{eq:feature-feature-averaged-x-or-xhat-decomposition}, and
\eqref{eq:feature-latent-averaged-x-or-xhat-decomposition}
.
To account for these statistics, we will have to augment our Feynman rules from before: 
\bi
\item In addition to the horizontal solid line, which carried a \emph{random-feature} $j$-type index, \eqref{eq:def-random-feature-rule}, and the horizontal dashed line, which carried a \emph{training sample} $\alpha$-type index, \eqref{eq:def-training-sample-rule}, we will need to introduce the horizontal \emph{wavy line}, which will carry a \emph{latent-feature} $I$-type index as
\be\label{eq:def-latent-feature-rule}
\delta_{I_1 I_2}
\quad \equiv \quad 
_{I_1}\,
\begin{tikzpicture}[baseline=-3pt]
\begin{feynhand}
\vertex (a) at (0,0); 
\vertex(c) at (2,0);
\propag[bos] (a) to (c);
\end{feynhand}
\end{tikzpicture}
\,
_{I_2}
\quad .
\ee
\item Then, we can represent $x\equiv x_{I;\alpha}$ as the \emph{vertex} of a horizontal wavy line and a horizontal dashed line as
\be\label{eq:def-x-rule}
x_{I;\alpha} 
\quad \equiv \quad 
_{I}\,
\begin{tikzpicture}[baseline=-3pt]
\begin{feynhand}
\vertex (a) at (0,0); 
\vertex (b) at (.925,0) ; 
\vertex(c) at (1.075,0);
\vertex(d) at (2,0);
\propag[bos] (a) to (b);
\propag[sca] (c) to (d);
\end{feynhand}
\end{tikzpicture}
\,
_{\alpha}
\quad .
\ee
$x^T$ is drawn as the mirror image, with the dashed line preceding the wavy line. %
\item Similarly, in addition to the curved dashed-and-solid double lines in \eqref{eq:def-covariance-omega-rule},
which represents the covariance 
$\langle \fea_{j_1;\alpha_1} \fea_{j_2;\alpha_2}\rangle_x$, 
we must introduce two additional double lines which represent the other covariances in 
  \eqref{eq:label-statistics-needed-for-data-averaging}.
The \emph{curved dashed-and-wavy double line} will represent the covariance of the latent features, 
$ \langle x_{I_1;\alpha_1} x_{I_2;\alpha_2}\rangle_x$, 
as
\be\label{eq:def-covariance-lambda-rule}
\covl_{I_1 I_2}\delta_{\alpha_1\alpha_2}
\quad \equiv \quad 
_{I_1;\alpha_1}
\begin{tikzpicture}[baseline=-3pt]
\begin{feynhand}
\vertex (a) at (0,0); \vertex (b) at (1.5,0);
\vertex (c) at (.1,0); \vertex (d) at (1.4,0);
\propag[bos] (a) to [in=90, out=90, looseness=1.75] (b);
\propag[sca] (c) to [in=90, out=90, looseness=1.75] (d);
\end{feynhand}
\end{tikzpicture}
_{I_2;\alpha_2}
\quad
,
\ee
and the \emph{curved dashed-and-solid-turns-into-wavy} double lines represents the mixed covariance of random and latent features, 
$\langle \fea_{j;\alpha_1} x_{I;\alpha_2}\rangle_x $
, as
\be\label{eq:def-covariance-mixed-feature-rule}
\covlf_{Ij} \delta_{\alpha_1 \alpha_2}
\quad \equiv \quad 
_{I;\alpha_1}
\begin{tikzpicture}[baseline=-3pt]
\begin{feynhand}
\vertex (a1) at (0,0) ;
\vertex (b1) at (.75,.75);
\vertex [dot] (a2) at (.75,.75) {}; \vertex (b2) at (1.5,0);
\vertex (c) at (.15,0); \vertex (d) at (1.35,0);
\propag[bos] (a1) to [in=180, out=90, looseness=1] (b1);
\propag (a2) to [in=90, out=0, looseness=1] (b2);
\propag[sca] (c) to [in=90, out=90, looseness=1.75] (d);
\end{feynhand}
\end{tikzpicture}
_{j;\alpha_2}
\quad
.
\ee
Here, the dot on the \emph{solid-and-wavy-line} reminds us that $\covlf \equiv \covlf_{Ij}$ transforms a latent feature index, $I$, into a random feature index, $j$. 
\ei
As before, we will work in the limit where all the scales in the problem are all large ($\nl, \nf, \nA \to \infty$) but with fixed ratios. Thus, we can continue to restrict our attention  to only planar diagrams.

With these rules defined, let's now look at 
the expectations appearing in \eqref{eq:label-term-expanded}. Consider the first,
\be\label{eq:label-term-first-expectation}
\expval{\tr{\covlf\, \res \fea x^T} }_{x, \fw} = \expval{\tr{\covlf \braket{\res \fea x^T}_x } }_{\fw} \, .
\ee
Here we have used the fact that $\covlf$ is independent of the training set to move it outside the $x$ average.
The evaluation of $\braket{\res \fea x^T}_x$ is similar to that of $\braket{\res}_x$, except that now the factors of $(\fea\fea^T)^s$ appearing in the expansion of $\res$, \eqref{eq:resolvent-power-series}, can have correlations with $\fea$ \emph{and} $x$, cf.~\eqref{eq:label-statistics-needed-for-data-averaging}. 

To proceed, we expand $\res$ order-by-order in $\ridge$ as in \eqref{eq:resolvent-power-series}. Representing the terms in the expansion of $\braket{\res \fea x^T}_x$ using our Feynman rules, we find: %
\begin{align}\label{eq:middle-term-naive-diagrams}
\notag
\braket{\res \fea x^T}_x
\quad = \quad
&\quad\gamma^{-1}~
\begin{tikzpicture}[baseline=-3pt]
\begin{feynhand}
\vertex (a1) at (0,0) ;
\vertex (b1) at (.75,.75);
\vertex [dot] (a2) at (.75,.75) {}; \vertex (b2) at (1.5,0);
\vertex (c) at (.15,0); \vertex (d) at (1.35,0);
\vertex (e) at (.15,0); \vertex (f) at (1.35,0);
\vertex (g) at (-.5,0); \vertex (h) at (0,0);
\vertex (i) at (1.5,0); \vertex (j) at (2,0);
\propag (a1) to [in=180, out=90, looseness=1] (b1);
\propag[bos] (a2) to [in=90, out=0, looseness=1] (b2);
\propag[sca] (c) to [in=90, out=90, looseness=1.75] (d);
\propag[sca] (e) to (f);
\propag (g) to (h);
\propag[bos] (i) to (j);
\end{feynhand}
\end{tikzpicture}
\\
-&\quad\gamma^{-2}~
\begin{tikzpicture}[baseline=-3pt]
\begin{feynhand}
\vertex (a) at (0,0); \vertex (b) at (1.5,0);
\vertex (c) at (.1,0); \vertex (d) at (1.4,0);
\vertex (e) at (.1,0); \vertex (f) at (1.4,0);
\vertex (g) at (-.5,0); \vertex (h) at (0,0);
\vertex (i) at (1.5,0); \vertex (j) at (2,0);
\vertex (a2) at (2,0); \vertex (b2) at (2.75,.75);
\vertex [dot] (a3) at (2.75,.75) {}; \vertex (b3) at (3.5,0);
\vertex (c2) at (2.15,0); \vertex (d2) at (3.35,0);
\vertex (e2) at (2.15,0); \vertex (f2) at (3.35,0);
\vertex (i2) at (3.5,0); \vertex (j2) at (4,0);
\propag (a) to [in=90, out=90, looseness=1.75] (b);
\propag[sca] (c) to [in=90, out=90, looseness=1.75] (d);
\propag[sca] (e) to (f);
\propag (g) to (h);
\propag (i) to (j);
\propag (a2) to [in=180, out=90, looseness=1] (b2);
\propag[bos] (a3) to [in=90, out=0, looseness=1] (b3);
\propag[sca] (c2) to [in=90, out=90, looseness=1.75] (d2);
\propag[sca] (e2) to (f2);
\propag[bos] (i2) to (j2);
\end{feynhand}
\end{tikzpicture}
\quad
-
\quad
\gamma^{-2}~
\begin{tikzpicture}[baseline=-3pt]
\begin{feynhand}
\vertex (w) at (2.5,0); \vertex (x) at (3,0);
\vertex (y) at (-.5,0); \vertex (z) at (0,0);
\vertex (a1) at (0,0); \vertex (b1) at (1.25,1.25);
\vertex [dot](a2) at (1.25,1.25) {}; \vertex (b2) at (2.5,0);
\vertex (c) at (.15,0); \vertex (d) at (2.35,0);
\vertex (e) at (.15,0); \vertex (f) at (0.4,0);
\vertex (g) at (2.1,0); \vertex (h) at (2.35,0);
\vertex (i) at (.4,0); \vertex (j) at (2.1,0);
\vertex (k) at (.5,0); \vertex (l) at (2,0);
\vertex (m) at (.5,0); \vertex(n) at (1.25,0) ;  \vertex (o) at (2,0);
\propag (a1) to [in=180, out=90, looseness=1] (b1);
\propag[bos] (a2) to [in=90, out=0, looseness=1] (b2);
\propag[sca] (c) to [in=90, out=90, looseness=1.75] (d);
\propag[sca] (e) to (f);
\propag[sca] (g) to (h);
\propag[sca] (i) to [in=90, out=90, looseness=1.75] (j);
\propag (k) to [in=90, out=90, looseness=1.75] (l);
\propag (m) to (n); \propag (n) to (o);
\propag[bos] (w) to (x);
\propag (y) to (z); 
\end{feynhand}
\end{tikzpicture}
\notag 
\\~\notag\\
+&\quad\dots \quad,
\end{align}
Since this is more involved than our previous expansions, it is worth summarizing what each of these diagrams represents. Each line represents a different order in the %
expansion \eqref{eq:resolvent-power-series} of $\res$:
\bi
\item The first line represents $\braket{\ridge^{-1} \fea x^T}_x$, the term resulting from the $s=0$ part of the expansion of $\res(\ridge)$.
To interpret the diagram, we first read along the bottom horizontal lines: the feature $\fea$ is represented by the vertex that connects a solid line and a dashed line, and then the latent feature $x^T$ is represented by the %
the vertex that connects a dashed line and a wavy line. %
The average of these two matrices is given by \eqref{eq:label-statistics-needed-for-data-averaging}, 
\be
\braket{\fea x^T}_x = T \covfl \,;
\ee
the double curved lines represent this covariance
as per our rule \eqref{eq:def-covariance-mixed-feature-rule}. Note that our rules also tell us to include the factor of $\nA$ due to the closed dashed-line loop.
\item The second line represents $\braket{-\ridge^{-2} \fea\fea^T \fea x^T}_x$, which arises from the $s=1$ part of the expansion of $\res(\ridge)$. Reading along the bottom of either diagram, we see a pattern of solid, dashed, and wavy lines that represents the quantity to be averaged, $\fea\fea^T \fea x^T$. There are three ways to pair up these four matrices, but one of these would result in a non-planar diagram that is subleading. The two planar diagrams are depicted in the second line, and represent the pairings 
\be
\braket{\fea\fea^T}_x\braket{\fea x^T}_x = \nA^2 \covf \covfl\,, \qquad 
\braket{\fea x^T}_x\braket{\fea^T \fea}_x = T \covfl \tr{\covf} \,, 
\ee
respectively. These diagrams both contribute at the same order in our large-$(\nf,\nA)$ counting, since the first diagram scales like $\nA^2$ and the second diagram scales like $\nA \tr{\covf} \sim \nA \nf$. 
\ei
We could continue on expanding in this manner, but there is a better way. 
In fact, this far is enough to let us notice 
a pattern that allows us to immediately write down the answer in terms of quantities we've already computed.

To explain this, 
we will need 
to define one more object.
Consider the data-data resolvent, $\Res(\ridge)$, defined in \eqref{eq:resolvent-data-data}.
We can compute its data average, $\Resb \equiv \braket{\Res}_x$, via a diagrammatic expansion just as we did for $\resb$:
\begin{align}\label{eq:diagrams-Qbar}
{\Resb} \quad &\equiv \quad
\begin{tikzpicture}[baseline=-3pt]
\begin{feynhand}
\vertex (a) at (0,0); \vertex[ringblob](b) at (1,0) {${\Resb}$} ; \vertex(c) at (2,0);
\propag[sca] (a) to (b); \propag[sca] (b) to (c);
\end{feynhand}
\end{tikzpicture}
\\
&=
~~\,\gamma^{-1}~
\begin{tikzpicture}[baseline=-2pt]
\begin{feynhand}
\vertex (a) at (0,0); \vertex(b) at (2.0,0) ;
\propag[sca] (a) to (b);
\end{feynhand}
\end{tikzpicture} 
\quad-\quad\gamma^{-2}~
\begin{tikzpicture}[baseline=-3pt]
\begin{feynhand}
\vertex (a) at (0,0); \vertex (b) at (1.5,0);
\vertex (c) at (.1,0); \vertex (d) at (1.4,0);
\vertex (e) at (.1,0); \vertex (f) at (1.4,0);
\vertex (g) at (-.5,0); \vertex (h) at (0,0);
\vertex (i) at (1.5,0); \vertex (j) at (2,0);
\propag[sca] (a) to [in=90, out=90, looseness=1.75] (b);
\propag (c) to [in=90, out=90, looseness=1.75] (d);
\propag (e) to (f);
\propag[sca] (g) to (h);
\propag[sca] (i) to (j);
\end{feynhand}
\end{tikzpicture}
\notag\\
&~~~+\gamma^{-3}~
\begin{tikzpicture}[baseline=-3pt]
\begin{feynhand}
\vertex (a) at (0,0); \vertex (b) at (1.5,0);
\vertex (c) at (.1,0); \vertex (d) at (1.4,0);
\vertex (e) at (.1,0); \vertex (f) at (1.4,0);
\vertex (g) at (-.5,0); \vertex (h) at (0,0);
\vertex (i) at (1.5,0); \vertex (j) at (2,0);
\vertex (a2) at (2,0); \vertex (b2) at (3.5,0);
\vertex (c2) at (2.1,0); \vertex (d2) at (3.4,0);
\vertex (e2) at (2.1,0); \vertex (f2) at (3.4,0);
\vertex (i2) at (3.5,0); \vertex (j2) at (4,0);
\propag[sca] (a) to [in=90, out=90, looseness=1.75] (b);
\propag (c) to [in=90, out=90, looseness=1.75] (d);
\propag (e) to (f);
\propag[sca] (g) to (h);
\propag[sca] (i) to (j);
\propag[sca] (a2) to [in=90, out=90, looseness=1.75] (b2);
\propag (c2) to [in=90, out=90, looseness=1.75] (d2);
\propag (e2) to (f2);
\propag[sca] (i2) to (j2);
\end{feynhand}
\end{tikzpicture}
\quad+\quad\gamma^{-3}~
\begin{tikzpicture}[baseline=-3pt]
\begin{feynhand}
\vertex (w) at (2.5,0); \vertex (x) at (3,0);
\vertex (y) at (-.5,0); \vertex (z) at (0,0);
\vertex (a) at (0,0); \vertex (b) at (2.5,0);
\vertex (c) at (.1,0); \vertex (d) at (2.4,0);
\vertex (e) at (.1,0); \vertex (f) at (0.4,0);
\vertex (g) at (2.1,0); \vertex (h) at (2.4,0);
\vertex (i) at (.4,0); \vertex (j) at (2.1,0);
\vertex (k) at (.5,0); \vertex (l) at (2,0);
\vertex (m) at (.5,0); 
\vertex (o) at (2,0);
\propag[sca] (a) to [in=90, out=90, looseness=1.75] (b);
\propag (c) to [in=90, out=90, looseness=1.75] (d);
\propag (e) to (f);
\propag (g) to (h);
\propag (i) to [in=90, out=90, looseness=1.75] (j);
\propag[sca] (k) to [in=90, out=90, looseness=1.75] (l);
\propag[sca] (m) to (o); 
\propag[sca] (w) to (x);
\propag[sca] (y) to (z); 
\end{feynhand}
\end{tikzpicture}
\quad
-
\quad
\dots
\quad
\notag\\
~\notag\\
&=
~\gamma^{-1}~
\begin{tikzpicture}[baseline=-2pt]
\begin{feynhand}
\vertex (a) at (0,0); \vertex(b) at (2.0,0) ;
\propag[sca] (a) to (b);
\end{feynhand}
\end{tikzpicture} 
~-\gamma^{-1}~
\begin{tikzpicture}[baseline=-3pt]
\begin{feynhand}
\vertex (a) at (0,0); \vertex (b) at (1.8,0);
\vertex (c) at (.1,0); \vertex (d) at (1.7,0);
\vertex (e1) at (.1,0); \vertex[ringblob] (f) at (.9,0) {$\resb$}; \vertex (e2) at (1.7, 0);
\vertex (g) at (-.5,0); \vertex (h) at (0,0);
\vertex (i) at (1.8,0); \vertex (j) at (2.3,0);
\propag[sca] (a) to [in=90, out=90, looseness=1.75] (b);
\propag (c) to [in=90, out=90, looseness=1.75] (d);
\propag (e1) to (f);
\propag (f) to (e2);
\propag[sca] (g) to (h);
\propag[sca] (i) to (j);
\end{feynhand}
\end{tikzpicture}
~+\gamma^{-1}~
\begin{tikzpicture}[baseline=-3pt]
\begin{feynhand}
\vertex (a) at (0,0); \vertex (b) at (1.8,0);
\vertex (a1) at (2.3,0); \vertex (b1) at (4.1,0);
\vertex (c) at (.1,0); \vertex (d) at (1.7,0);
\vertex (c1) at (2.4,0); \vertex (d1) at (4,0);
\vertex (e1) at (.1,0); \vertex[ringblob] (f) at (.9,0) {$\resb$}; \vertex (e2) at (1.7, 0);
\vertex (e3) at (2.4,0); \vertex[ringblob] (f1) at (3.2,0) {$\resb$}; \vertex (e4) at (4, 0);
\vertex (g) at (-.5,0); \vertex (h) at (0,0);
\vertex (g2) at (4.1,0); \vertex (h2) at (4.6,0);
\vertex (i) at (1.8,0); \vertex (j) at (2.3,0);
\propag[sca] (a) to [in=90, out=90, looseness=1.75] (b);
\propag[sca] (a1) to [in=90, out=90, looseness=1.75] (b1);
\propag (c) to [in=90, out=90, looseness=1.75] (d);
\propag (c1) to [in=90, out=90, looseness=1.75] (d1);
\propag (e1) to (f);
\propag (f) to (e2);
\propag (e4) to (f1);
\propag (f1) to (e3);
\propag[sca] (g) to (h);
\propag[sca] (g2) to (h2);
\propag[sca] (i) to (j);
\end{feynhand}
\end{tikzpicture}
~
-
~
\dots\,.
\notag
\end{align}
Here, to reach the final line, we recognized a pattern:
the diagrams appearing in $\Resb$ contain as subdiagrams the same diagrams that appeared in the expansion for $\resb$.\footnote{It is easiest to see this by using the \eqref{eq:feyn-self} form of the $\resb$ expansion and plugging in for $\senF$ with its diagrams, \eqref{eq:ip1-expansion}. %
One could also define a self-energy matrix like $\senF$ as an intermediate step in the computation of $\Resb$; the diagrammatic expansion for that object makes the result \eqref{eq:Qbar-result} almost immediately apparent.  
In fact, from 
\eqref{eq:Qbar-result}, we see that the analog of $\senF$ for $\Resb$ is just the matrix $\ridge \secF I_T$.} 
Note also that all but one of the factors of $\ridge^{-1}$ that appear in the expansion of $\Resb$ are accounted for by the factors of $\resb$ appearing in the final line.
Evaluating these diagrams on the final line, 
we can sum the resulting geometric series:
\begin{align}\label{eq:Qbar-result}
\Resb &= \ridge^{-1} \IT - \ridge^{-1}\tr{\covf\resb} \IT + \ridge^{-1}\le(\tr{\covf\resb}\ri)^2 \IT - \dots 
\,\\
&=\frac{\IT}{\ridge ( 1 + \secF)} \,,
\notag
\end{align}
where to get the resulting expression we used the definition $\secF \equiv \tr{\covf\resb}$.  As a sanity check, note that $\Resb$ must be proportional to the identity because none of our covariance matrices have nontrivial training-set indices after averaging over the data, cf. \eqref{eq:label-statistics-needed-for-data-averaging}.

Returning now to \eqref{eq:middle-term-naive-diagrams}, we can similarly see a pattern in the expansion:
\be\label{eq:label-first-expectation-final-diagram}
\braket{\res \fea x^T}_x
\quad = 
~\gamma~%
\begin{tikzpicture}[baseline=-3pt]
\begin{feynhand}
\vertex (a1) at (0,0) ;
\vertex (b1) at (.90,.90);
\vertex [dot] (a2) at (.90,.90) {}; \vertex (b2) at (1.8,0);
\vertex (c) at (.15,0); \vertex (d) at (1.65,0);
\vertex (e1) at (.15,0); \vertex[ringblob] (f) at (.90,0) {$\Resb$}; \vertex (e2) at (1.65, 0) ;
\vertex (g1) at (0,0); \vertex[ringblob](h) at (-.75,0) {$\resb$} ; \vertex(g2) at (-1.5,0);
\vertex (i) at (1.8,0); \vertex (j) at (2.3,0);
\propag (a1) to [in=180, out=90, looseness=1] (b1);
\propag[bos] (a2) to [in=90, out=0, looseness=1] (b2);
\propag[sca] (c) to [in=90, out=90, looseness=1.75] (d);
\propag[sca] (e1) to (f);
\propag[sca] (e2) to (f);
\propag (g1) to (h);
\propag (h) to (g2);
\propag[bos] (i) to (j);
\end{feynhand}
\end{tikzpicture}
\quad
,
\ee
which can be easily checked by recursively substituting in the diagrams for $\resb$ and $
\Resb$.  
Note that there is an explicit factor of $\ridge$ in this expression to account for the factors of $\ridge^{-1}$ that are included in \emph{both} $\resb$ and $\Resb$.
Using our rules to evaluate this diagram, we find the first $x$ expectation in the label term, \eqref{eq:label-term-first-expectation} is:
\be\label{eq:first-term-label-result}
\expval{\tr{\covlf\, \res \fea x^T} }_{x,} = \le(\frac{\nA}{1+\secF}\ri) \tr{\covfl \covlf\, \resb }   \, ,
\ee
where, to get this expression, we used here the cyclicity of the trace and substituted for the trace of \eqref{eq:Qbar-result}, %
\be\label{eq:trace-Qbar-gamma}
\ridge\, \tr{\Resb} = \frac{\nA}{1+\secF}\,.
\ee

Now let's consider the final expectation in the label term \eqref{eq:label-term-expanded}:
\be\label{eq:label-term-second-expectation}
 \Big\langle\tr{\covf \res \fea x^T x \fea^T \res}\Big\rangle_{x,\fw} = 
 \Big\langle\tr{\covf \braket{ \res \fea x^T x \fea^T \res}_x }\Big\rangle_{\fw} \, 
\ee
where we've used the fact that $\covf$ is independent of the training set to take it outside the $x$ average. This will be the most complicated expectation to evaluate, 
since terms like $(\fea\fea^T)^s$ resulting from the expansion of \emph{either} $\res$ in the expression can pair with any of the other four matrices: $\fea, x^T, x, \fea^T$.

To proceed, we will use a similar diagrammatic strategy as before: we will consider the various diagrams that result from the dual expansion of the resolvents in $\braket{ \res \fea x^T x \fea^T \res}_x$ and try to recognize patterns in the subdiagrams. 
Before we begin, in this case note that the different terms arising in the expectation must be determined by the different ways in which
 the 
 $x$ and $x^T$ are paired. 
In particular, there are four different possibilities to consider: 
\emph{(1)} the $x$ and $x^T$ are paired together; 
\emph{(2)}  the $x$ is paired with a $\fea^T$, and the $x^T$ is paired with a $\fea$ to their left; 
\emph{(3)}
the $x$ is paired with a $\fea^T$, and the $x^T$ is paired with a $\fea$ to their right; 
and 
\emph{(4)}
the $x^T$ is paired with a $\fea$ to the left and the $x^T$ is paired with a $\fea^T$ to the right. As you can check, any other pairings will involve crossings, leading to non-planar diagrams, and therefore subleading contributions in the large-$\nf$ limit.

Next, when expanding both $\res$'s and making these pairings, you'll notice that for the first three pairings, \emph{(1)--(3)}, the expectation will factorize into a part with a common factor, while for the fourth pairing, \emph{(4)}, there will be two terms, one that factorizes the same way and one that does not. 
Altogether, the structure will be
\be\label{eq:final-term-factorized-pre}
\braket{ \res \fea x^T x \fea^T \res}_x = \braket{\res \fea  \le(\widetilde{P}^{(1)}+ \widetilde{P}^{(2)}  + \widetilde P^{(3)} + \widetilde P^{(4)} \ri) \fea^T \res}_x  + \Pnf \, ,
\ee
where the various $\widetilde P^{(a)} \equiv \widetilde P^{(a)}_{\alpha_1 \alpha_2}$ are $(\nA \times \nA)$-dimensional \emph{$x$-independent} matrices that represent the four factorized contributions from the four different pairings, and $\Pnf \equiv \Pnf_{j_1j_2}$ is  an $(\nf \times \nf)$-dimensional \emph{$x$-independent} matrix representing the so-called \emph{connected} contribution that does not factorize the same way as the other.
This decomposition, \eqref{eq:final-term-factorized-pre}, can be shown by expanding in both $\res$'s and recognizing that the resulting diagrammatic expansion can be reorganized into sets of \emph{disconnected} diagrams and \emph{connected} diagrams.
Importantly, in doing so you will see that the overall factor for the disconnected diagrams 
is the same for each pairing.

Moreover, as we already explained above after \eqref{eq:Qbar-result} when evaluating $\Resb$, the only matrix with training-data indices that can appear after $x$-averaging is the identity matrix $\IT$. Thus, each of these factorized pairings can be written as $\widetilde P^{(a)} \equiv  P^{(a)} \IT$, for scalar $P^{(a)}$, which lets us further simplify \eqref{eq:final-term-factorized-pre} as
\be\label{eq:further-simplification}
\braket{ \res \fea x^T x \fea^T \res}_x = \braket{\res \fea  \fea^T \res}_x  \le(P^{(1)}+ P^{(2)}  +  P^{(3)} +  P^{(4)} \ri) +  \Pnf\, .
\ee
Finally, the factor, $\braket{q \fea \fea^T q}_x$, can be expressed in terms of already known quantities 
as
\be\label{eq:disconnected-piece}
\braket{q \fea \fea^T q}_x = \le(1 + \ridge \frac{\partial}{\partial\ridge} \ri) \resb\,,
\ee
where we've used our identities,  \eqref{eq:identity-inverse} and then \eqref{eq:identity-squared}, and finally our definition $\resb \equiv \braket{\res}_x$.

Now, let's re-enumerate the four possible pairings of $x$ and $x^T$ to evaluate the four $\widetilde P^{(a)}$ and $\Pnf$:
\bi
    \item[\emph{(1)}] The simplest 
    case is when 
    $x$ and $x^T$
    are paired together:
    \be
        \widetilde{P}^{(1)} \equiv \braket{x^T x}_x 
 \quad = \quad
\begin{tikzpicture}[baseline=-3pt]
\begin{feynhand}
\vertex (a) at (0,0); \vertex (b) at (1.5,0);
\vertex (c) at (.1,0); \vertex (d) at (1.4,0);
\propag[sca] (a) to [in=90, out=90, looseness=1.75] (b);
\propag[bos] (c) to [in=90, out=90, looseness=1.75] (d);
\propag[bos] (c) to (d);
\end{feynhand}
\end{tikzpicture}
\quad
=
\quad
\tr{\covl} \IT \,.
    \ee
This is trivial, but it's still nice to practice our drawing.

    \item[\emph{(2)}] The next contribution is when both 
    $x$ and $x^T$
    pair with a $\fea^T$ and  a $\fea$ to their left:
    \be\label{eq:P2}
    \widetilde{P}^{(2)} \equiv -\braket{ \fea^T \braket{ \res \fea x^T}_x x  }_x
\quad
=
\quad-~
\ridge
~
\begin{tikzpicture}[baseline=-3pt]
\begin{feynhand}
\vertex (a1) at (0,0) ;
\vertex (b1) at (.90,.90);
\vertex [dot] (a2) at (.90,.90) {}; \vertex (b2) at (1.8,0);
\vertex (c) at (.15,0); \vertex (d) at (1.65,0);
\vertex (e1) at (.15,0); \vertex[ringblob] (f) at (.90,0) {$\Resb$}; \vertex (e2) at (1.65, 0) ;
\vertex (g1) at (0,0); \vertex[ringblob](h) at (-.75,0) {$\resb$} ; \vertex(g2) at (-1.5,0);
\vertex (g3) at (-1.65, 0); \vertex (g4) at (-2.15, 0);
\vertex (g5) at (-2.25, 0); \vertex (g6) at (-2.75, 0);
\vertex (i) at (1.8,0); \vertex (j) at (3.3,0);
\vertex (i2) at (3.45,0); \vertex (j2) at (3.95,0);
\vertex (i3) at (4.05,0); \vertex (j3) at (4.55,0);
\vertex[dot] (x) at (.90, 2) {};
\propag (a1) to [in=180, out=90, looseness=1] (b1);
\propag[bos] (a2) to [in=90, out=0, looseness=1] (b2);
\propag[sca] (c) to [in=90, out=90, looseness=1.75] (d);
\propag[sca] (e1) to (f);
\propag[sca] (e2) to (f);
\propag (g1) to (h);
\propag (h) to (g2);
\propag[bos] (i) to (j);
\propag[sca] (i2) to (j2);
\propag[sca] (g3) to (g4);
\propag[sca] (g3) to [in=90, out=90, looseness=1.45] (i2); 
\propag (g2) to [in=180, out=90, looseness=1] (x);
\propag[bos] (x) to [in=90, out=0, looseness=1] (j);
\end{feynhand}
\end{tikzpicture}
\quad
.
\ee
Here we have written this term as a set of nested expectation values 
such
that $x^T$ and $x$ can both be contracted to the left without any crossings.  
To understand why the diagram is the correct way to represent these nested expectations,
note that the inner expectation, $\braket{\res \fea x^T}_x$, was already analyzed  in \eqref{eq:label-first-expectation-final-diagram} and appears as
 a subdiagram here, while the outer expectation is evaluated using our Feynman rule \eqref{eq:def-covariance-mixed-feature-rule}.
To understand why the overall factor of $-\ridge$ appears, 
let's go back to how this term would naturally arise in our resolvent expansion: to find this term, we would expand the first $\res$ in $\braket{ \res \fea x^T x \fea^T \res}_x$ as $\res = \ridge^{-1} \sum_s(-\ridge)^{-s}\left(\fea\fea^T\right)^s$ and then rewrite 
it
in terms of two sums for a pair of resolvents $\res$ and $\Res$;  
the point of this reorganization would be to  
extract a factor of $\fea$ to pair up with the $x^T$ and a factor of $\fea^T$ to pair up with the $x$.  Since each factor of $\fea\fea^T$ in the expansion comes with a factor of $-\ridge^{-1}$, when we pull out only a factor $\fea\fea^T$ in this way, we must also multiply by an explicit factor of $-\ridge$ to compensate.  
Evaluating the diagram, we get:
\be\label{eq:P2is}
P^{(2)} = -\ridge \, \tr{\Resb} \tr{\covfl \covlf \resb} = 
 -\le(\frac{\nA}{1+\secF}\ri) \tr{\covfl \covlf \resb}\,,
\ee
where in the last expression we used \eqref{eq:trace-Qbar-gamma} to substitute for $\ridge \, \tr{\Resb}$.  

\item[\emph{(3)}] Similarly, both $x$ and $x^T$ can pair with a $\fea^T$ and $\fea$ to their right:
\be
 \widetilde{P}^{(3)} \equiv -\braket{ x^T \braket{x\fea^T \res  }_x \fea  }_x 
\ee
As this is just the transpose of $\widetilde{P}^{(2)}$, the scalar trace is identical: 
\be
P^{(3)} = P^{(2)}  = -\le(\frac{\nA}{1+\secF}\ri) \tr{\covfl \covlf \resb} \,.
\ee
(The diagram representing this would be the mirror image of \eqref{eq:P2}.)

    \item[\emph{(4)}] Finally, $x^T$ can pair with a $\fea$ to the left, and $x$ can pair with a $\fea^T$ to the right:
    \be\label{eq:P4}
    \widetilde{P}^{(4)} \equiv \braket{\fea^T \braket{\res \fea x^T}_{x} \braket{x \fea^T \res}_x \fea}_x 
=~
\ridge^2
~
\begin{tikzpicture}[baseline=-3pt]
\begin{feynhand}
\vertex (a1) at (0,0) ;
\vertex (b1) at (.90,.90);
\vertex [dot] (a2) at (.90,.90) {}; \vertex (b2) at (1.8,0);
\vertex (c) at (.15,0); \vertex (d) at (1.65,0);
\vertex (e1) at (.15,0); \vertex[ringblob] (f) at (.90,0) {$\Resb$}; \vertex (e2) at (1.65, 0) ;
\vertex (aa) at (-1.6,0); \vertex (bb) at (-1.9,0);
\vertex (cc) at (6.1,0); \vertex (dd) at (6.4,0);
\vertex (g1) at (0,0); \vertex[ringblob](h) at (-.75,0) {$\resb$} ; \vertex(g2) at (-1.5,0);
\vertex (i) at (1.8,0); \vertex (j) at (2.1,0);
\vertex (a1p) at (4.2, 0) ;
\vertex (b1p) at (3.3, .90);
\vertex [dot] (a2p) at (3.3,.90) {}; \vertex (b2p) at (2.4,0);
\vertex (cp) at (2.55,0); \vertex (dp) at (4.05,0);
\vertex (e1p) at (2.55,0); \vertex[ringblob] (fp) at (3.3,0) {$\Resb$}; \vertex (e2p) at (4.05, 0) ;
\vertex (g1p) at (4.2,0); \vertex[ringblob](hp) at (5.1,0) {$\resb$} ; \vertex(g2p) at (6.0,0);
\vertex (ip) at (2.1,0); \vertex (jp) at (2.4,0);
\propag (a1) to [in=180, out=90, looseness=1] (b1);
\propag[bos] (a2) to [in=90, out=0, looseness=1] (b2);
\propag[sca] (c) to [in=90, out=90, looseness=1.75] (d);
\propag[sca] (e1) to (f);
\propag[sca] (e2) to (f);
\propag (g1) to (h);
\propag (h) to (g2);
\propag[bos] (i) to (j);
\propag (a1p) to [in=0, out=90, looseness=1] (b1p);
\propag[bos] (a2p) to [in=90, out=180, looseness=1] (b2p);
\propag[sca] (cp) to [in=90, out=90, looseness=1.75] (dp);
\propag[sca] (e1p) to (fp);
\propag[sca] (e2p) to (fp);
\propag (g1p) to (hp);
\propag (hp) to (g2p);
\propag[bos] (ip) to (jp);
\propag[sca] (aa) to (bb);
\propag[sca] (cc) to (dd);
\propag[sca] (aa) to [in=90, out=90, looseness=.86]  (cc);
\propag (g2) to [in=90, out=90, looseness=.85]  (g2p);
\end{feynhand}
\end{tikzpicture}
~
,
\ee
To see why the pairing given by the middle expression is correct, note that this is the appropriate and only way to contract the $x$ and $x^T$ in different directions without any crossings. 
Similar to before, we can see why this is the correct diagram to represent these nested expectations by using our Feynman rule, \eqref{eq:def-covariance-omega-rule}, for the outer part, and then by using \eqref{eq:label-first-expectation-final-diagram} and its mirror as subdiagrams to represent  the two middle expectations.\footnote{The factor of $\ridge^2$ appears because 
in this case we've extracted two factors of $\fea \fea^T$ when reorganizing our expansion, one from each of the $\res$'s in $\braket{ \res \fea x^T x \fea^T \res}_x$, and so we must multiply by $(-\ridge)^2$ to compensate.  }
 Evaluating this diagram, we get:
 \be\label{eq:P4is}
 P^{(4)} = \le(\ridge \, \tr{\Resb}\ri)^2 \tr{\resb \covfl \covlf \resb \covf} = 
 \le(\frac{\nA}{1+\secF}\ri)^2 \tr{\resb \covfl \covlf \resb \covf}\,,
\ee
where, as is our recent \emph{M.O.}, 
we used \eqref{eq:trace-Qbar-gamma} to substitute for $\ridge \, \tr{\Resb}$. 

\item[\emph{(4, c)}] Finally prime, again $x^T$ can pair with a $\fea$ to the left, and $x$ can pair with a $\fea^T$ to the right:
\be
\Pnf \equiv \braket{q\fea x^T}_x \braket{x \fea^T q}_x  
~= 
~
\ridge^2
~
\begin{tikzpicture}[baseline=-3pt]
\begin{feynhand}
\vertex (a1) at (0,0) ;
\vertex (b1) at (.90,.90);
\vertex [dot] (a2) at (.90,.90) {}; \vertex (b2) at (1.8,0);
\vertex (c) at (.15,0); \vertex (d) at (1.65,0);
\vertex (e1) at (.15,0); \vertex[ringblob] (f) at (.90,0) {$\Resb$}; \vertex (e2) at (1.65, 0) ;
\vertex (g1) at (0,0); \vertex[ringblob](h) at (-.75,0) {$\resb$} ; \vertex(g2) at (-1.5,0);
\vertex (i) at (1.8,0); \vertex (j) at (2.1,0);
\vertex (a1p) at (4.2, 0) ;
\vertex (b1p) at (3.3, .90);
\vertex [dot] (a2p) at (3.3,.90) {}; \vertex (b2p) at (2.4,0);
\vertex (cp) at (2.55,0); \vertex (dp) at (4.05,0);
\vertex (e1p) at (2.55,0); \vertex[ringblob] (fp) at (3.3,0) {$\Resb$}; \vertex (e2p) at (4.05, 0) ;
\vertex (g1p) at (4.2,0); \vertex[ringblob](hp) at (5.1,0) {$\resb$} ; \vertex(g2p) at (6.0,0);
\vertex (ip) at (2.1,0); \vertex (jp) at (2.4,0);
\propag (a1) to [in=180, out=90, looseness=1] (b1);
\propag[bos] (a2) to [in=90, out=0, looseness=1] (b2);
\propag[sca] (c) to [in=90, out=90, looseness=1.75] (d);
\propag[sca] (e1) to (f);
\propag[sca] (e2) to (f);
\propag (g1) to (h);
\propag (h) to (g2);
\propag[bos] (i) to (j);
\propag (a1p) to [in=0, out=90, looseness=1] (b1p);
\propag[bos] (a2p) to [in=90, out=180, looseness=1] (b2p);
\propag[sca] (cp) to [in=90, out=90, looseness=1.75] (dp);
\propag[sca] (e1p) to (fp);
\propag[sca] (e2p) to (fp);
\propag (g1p) to (hp);
\propag (hp) to (g2p);
\propag[bos] (ip) to (jp);
\end{feynhand}
\end{tikzpicture}
\quad
,
\ee 
and we can use \eqref{eq:label-first-expectation-final-diagram} and \eqref{eq:trace-Qbar-gamma} to evaluate this as
\be\label{eq:Pnf-evaluated}
\Pnf = \le(\frac{\nA}{1+\secF}\ri)^2 \resb \covfl \covlf \resb \,.
\ee
\ei
Assembling all five terms back into \eqref{eq:final-term-factorized-pre}, and then substituting into \eqref{eq:label-term-second-expectation}, we find
\begin{align}\label{eq:second-term-label-result}
 \Big\langle\tr{\covf \res \fea x^T x \fea^T \res}\Big\rangle_{x} &= \le(\frac{\nA}{1+\secF}\ri)^2 \tr{ \covf \resb \covfl \covlf \resb } + \le[\le(1 + \ridge \frac{\partial}{\partial\ridge} \ri) \secF \ri]  \times \,\\
 &\quad~\le[
  \tr{\covl} 
  - 2\le(\frac{\nA}{1+\secF}\ri) \tr{\covfl \covlf \resb}  
  + \le(\frac{\nA}{1+\secF}\ri)^2 \tr{\resb \covfl \covlf \resb \covf} 
  \ri] \notag \,,
  \notag
\end{align}
where the first term on the first line 
is the connected contribution, $\Pnf$, \eqref{eq:Pnf-evaluated},
and where the second term gives 
the four factorized contributions \emph{(1)--(4)}, for which we used $\secF  \equiv \tr{\covf \resb}$, with the $\resb$ coming from the factor in \eqref{eq:disconnected-piece}.

Finally, we can now collect the contribution to the $x$ average of the label term in the underparameterized regime. Starting from our expression \eqref{eq:label-term-expanded} and substituting in both evaluated expectations, 
\eqref{eq:first-term-label-result} and \eqref{eq:second-term-label-result},
we get
\begin{align}
\L_\w = \frac{\wvar}{2  \nl}&\bigg\langle\le[1+\le(1 + \ridge \frac{\partial}{\partial\ridge} \ri) \secF \ri] \times \, \\ 
&\qquad\le[
  \tr{\covl} 
  - 2\le(\frac{\nA}{1+\secF}\ri) \tr{\covfl \covlf \resb}  
  + \le(\frac{\nA}{1+\secF}\ri)^2 \tr{\resb \covfl \covlf \resb \covf} 
  \ri]\bigg\rangle_{\fw} \,, \notag
\end{align}
where all the other terms just conspired to give ``1'' inside the first square brackets. (Note that the terms in the second square brackets could be written as a square, if desired.)
Lastly, note that from combining the Schwinger-Dyson equations,
\eqref{eq:def-self-energy-features} and \eqref{eq:self-energy-solution}, we can write
\be\label{eq:third-identity}
\le(\frac{\nA }{1 + \secF}\ri) \covf \resb = \IN - \ridge \resb \, ,
\ee
where to get this we first inverted \eqref{eq:def-self-energy-features} to get an expression for $\senF$, and then we it equal to the right-hand side of \eqref{eq:self-energy-solution} and multiplied both sides by $\resb$.
Using this new identity, \eqref{eq:third-identity}, and remembering that $\covf$ and $\resb$ commute and that the trace is cyclic, we can finally rewrite $\L_\w$ compactly as  
\begin{align}\label{eq:final-x-averaged-Lw}
    \L_\w =  \frac{\wvar}{2 \nl}\braket{\le[1+\le(1 + \ridge \frac{\partial}{\partial\ridge}  \ri) \secF   \ri] \le( \tr{\covl} -\le(\frac{\nA}{1+\secF}\ri) \tr{(\IN+\ridge \resb)  \resb \covfl \covlf }\ri)}_{\fw} \, .
\end{align}
\\ %

\noindent\emph{(ii)} \textbf{Random Feature Average}\\

We now turn to the average over the random features, $\fw$.
This will be simpler than the $x$ averages we just performed:
our strategy will be to massage $\L_\w$ into a form where we can apply the same (simpler) machinery that was used in \S\ref{sec:the-noise-term} for the noise term.

We begin by noting that the latent feature covariance matrix, $\covl$, is symmetric and positive semi-definite, by definition.
This implies that it is possible to find a symmetric, positive semi-definite matrix, $\sqrtcovl$, that is the ``square root" of $\covl$, 
such that\footnote{Since $\sqrtcovl$ is symmetric we could equally well have written this equation as $\covl\equiv\sqrtcovl^2$. We prefer to use the form \eqref{eq:def-vv} because it makes manifest that $\covl$ is symmetric.} %
\be\label{eq:def-vv}
    \covl \equiv \sqrtcovl \sqrtcovl^T \, , \qquad \sqrtcovl=\sqrtcovl^T \, .
\ee
Recalling definitions \eqref{eq:def-Omega} and \eqref{eq:def-Omega-tilde}, we can  express our other two covariances as
\be\label{eq:def-uv}
\covlf = \sqrtcovl \sqrtcovl^T \fw^T\,, \qquad \covf = \fw \sqrtcovl \sqrtcovl^T \fw^T \, ,
\ee
and, using 
\eqref{eq:def-self-energy-features} and \eqref{eq:self-energy-solution}, express our $x$-averaged resolvent as
\be
\resb(\ridge) = \frac{1}{\ridge\IN + \le(\frac{\nA}{1+\secF}\ri)  \fw \sqrtcovl \sqrtcovl^T \fw^T} \,.
\ee
As we will soon see, the utility of this is that as our expression for $\L_\w$ had various products of $\resb$ and $\covlf$,  with our new notation we can now rewrite these using 
\be\label{eq:commutation-to-latent}
     \resb \fw\sqrtcovl  = \fw\sqrtcovl  \Resbb \,,
\ee
where we have %
defined the \emph{latent-feature resolvent}:
\be\label{eq:resolvent-latent-feature}
    \Resbb(\ridge) \equiv \frac{1}{\ridge \IM + \le(\frac{\nA}{1+\secF}\ri) \sqrtcovl^T \fw^T \fw \sqrtcovl } \, .
\ee
This is an $(\nl \times \nl)$ matrix $\Resbb\equiv\Resbb_{I_1I_2}$ that operates on our latent space.\footnote{To derive \eqref{eq:commutation-to-latent} we can follow the same manipulations we used to derive the commutation relation \eqref{eq:commutation-relation}, with the substitutions $\res \to \resb$ and $\fea \to  \fw \sqrtcovl$.}
Contrasting carefully this new definition with the one for our (random) feature-feature resolvent $\res$, \eqref{eq:resolvent-feature-feature}, and the data-data resolvent $\Res$, \eqref{eq:resolvent-data-data}, we note that those resolvents are defined in terms of the matrix $\fea$, which depend on \emph{both} the random training set $x$ and random feature weights $\fw$, while here $\Resbb$ 
depends \emph{only} on the random feature weights $\fw$.

Much like our other resolvents, $\Resbb$ satisfies identities
\be\label{eq:latent-space-resolvent-identities}
\Resbb^2(\ridge) = - \frac{\partial}{\partial \gamma} \Resbb(\ridge) \,, \qquad
\le[\ridge \IM + \le(\frac{\nA}{1+\secF}\ri) \sqrtcovl^T\fw^T \fw \sqrtcovl\ri] \Resbb = \IM \,.
\ee
Multiplying the second identity by $\covl$, taking the trace, and rearranging, we find
\be\label{eq:first-relation}
 \tr{\covl}-\le(\frac{\nA}{1+\secF}\ri) \tr{\resb \covfl \covlf } = \ridge \tr{\covl \Resbb } \, , %
\ee
where in the second term on the left-hand side we used \eqref{eq:def-uv}, 
\eqref{eq:commutation-to-latent}, and the fact that $\sqrtcovl=\sqrtcovl^T$. %
Taking $-\ridge \frac{\partial}{\partial \ridge}$ of this equation gives
\be\label{eq:second-relation}
    -\ridge \le(\frac{\nA}{1+\secF}\ri)  \tr{\resb^2 \covfl \covlf } = -\ridge \tr{\Resbb \covl} - \ridge^2 \frac{\partial}{\partial \ridge}\tr{\covl \Resbb}\,,
\ee
where we used the identity
\eqref{eq:identity-squared} 
to rewrite the derivative of $\resb$ in terms of its square.
We now note that the sum of the left hand sides of \eqref{eq:first-relation} and \eqref{eq:second-relation} is exactly the combination that appears in  $\L_\w$, allowing us to rewrite \eqref{eq:final-x-averaged-Lw} as 
\begin{align}\label{eq:finalish}
    \L_\w = \frac{\wvar}{2 \nl}\braket{\le[1+\le(1 + \ridge \frac{\partial}{\partial\ridge}  \ri) \secF \ri] \le(- \ridge^2 \frac{\partial}{\partial \ridge} \tr{\covl \Resbb}\ri)}_\fw \, .
\end{align}
This is progress, since we have written $\L_\w$ entirely in terms of the expectation value of the new resolvent, $\Resbb$. 

To proceed let's consider the ridgeless limit $\ridge \to 0$, where this expression simplifies considerably. 
In \S\ref{sec:delta-zero}, we explain that
in the underparameterized regime ($\nf < \nA$), we have an expansion that begins with the constant term,\footnote{
    To see this from our discussion in \S\ref{sec:delta-zero} under the subheading \emph{$\nG > \nl$}, make the substitutions $\nG \to \nA$ and $\nl \to \nf$. This expansion holds regardless of the form of the covariance.
}
\be
\secF=\secF_0+ \secF_1 \ridge + \dots \,,
\ee 
and thus in the ridgeless limit we have:
\be\label{eq:limit-of-delta-F-ridgeless}
\lim_{\ridge \rightarrow 0} \secF = \secF_0 \,.
\ee
Similarly, the leading term in the square brackets has the limit
\begin{align}\label{eq:ridge-limit-noise-reprint}
\lim_{\ridge \rightarrow 0} \left(1+\ridge\frac{\partial}{\partial \ridge} \right) \secF(\ridge)  = \secF_{0}   \,.
\end{align} 
Moreover, as we discussed at the conclusion of \S\ref{sec:the-noise-term} and in more detail in \S\ref{sec:delta-zero} under the final subheading labeled \emph{$\secG_0$ for the Truncated Power-Law Spectrum}, $\secF_{0}$ does not explicitly depend on the fixed random feature weights, $\fw$, and is given by \eqref{eq:noise-term-final}. In the underparameterized regime for $\nf < \nl$, this is simply
\be\label{eq:noise-term-underparam}
\secF_0 = \frac{1}{\nA/\nf - 1} \, .
\ee

We are thus left with evaluating the term in round brackets in \eqref{eq:finalish}:
\be\label{eq:wow}
\lim_{\ridge \rightarrow 0} \braket{-\ridge^2 \frac{\partial}{\partial\ridge}\tr{\covl \Resbb}}_\fw  = \lim_{\ridge \rightarrow 0} \le( -\ridge^2 \frac{\partial}{\partial\ridge}\tr{\covl \braket{\Resbb}_\fw} \ri).
\ee
The quantity $\tr{\covl \braket{\Resbb}_\fw}$ is quite similar to the quantity $\secF \equiv \tr{\covl \resb}$ that was computed back in \S\ref{sec:the-noise-term}.
To proceed, we define a new random variable
\be\label{eq:def-rescaled}
\feau \equiv \le(\frac{\nl}{\fwvar}\ri)^{1/2} \sqrtcovl^T \fw^T
 \quad \Longleftrightarrow  \quad
\feau_{Ij}
\equiv \le(\frac{\nl}{\fwvar}\ri)^{1/2} \sum_{J=1}^{\nl} \fw_{jJ} \sqrtcovl_{JI}\,,
\ee
which, since $\sqrtcovl$ is a fixed matrix, has 
the statistics of zero-mean Gaussian when averaged over $\fw$:
\be\label{eq:statistics-rescaled}
\braket{\feau_{Ij}}_\fw = 0 \, , \qquad \braket{\feau_{I_1j_1} \feau_{I_2j_2}}_\fw = \covl_{I_1I_2} \delta_{j_1j_2}  \,,
\ee
where to evaluate the covariance in \eqref{eq:statistics-rescaled} we used statistics \eqref{eq:random-feature-weights-statistics} and the definition \eqref{eq:def-vv}. %
In terms of this variable $\feau$, $\Resbb$ takes the form 
\be\label{eq:resolvent-latent-feature-again}
    \Resbb(\ridge) = \frac{1}{\ridge \IM +  \zeta\, \feau\feau^T } \,.
\ee
where we have defined 
\be
\zeta\equiv \frac{\fwvar}{\nl}\le(\frac{\nA}{1+\secF} \ri) \, .
\ee
Naively it would be a problem that the factor of $\secF$ appearing in $\zeta$ itself depends on $\fw$, which would make it rather difficult to compute $\braket{\Resbb}_\fw$  in general.\footnote{While $\secF_0$ is independent of $\fw$, is not necessarily true that $\secF$ is similarly independent.
} 
However, as we noted above in \eqref{eq:limit-of-delta-F-ridgeless}, $\secF$ is independent of $\fw$ in the ridgeless limit. So as long as we are working in this limit we may treat $\zeta$ as a constant which is independent of $\fw$.

At this point $\fw$-averaging \eqref{eq:resolvent-latent-feature-again} would be form-identical to the $x$-averaging of $\res$ we already performed, but for the constant $\zeta$. It's not to hard to see that if we were to follow the same method we used to solve for $\resb$ in \S\ref{sec:the-noise-term} -- i.e. by rescaling $\covl$ --  we'd find
\be\label{eq:Resbbzeta}
\braket{\Resbb}_\fw = \frac{1}{\ridge \IM + \frac{\zeta\nf \covl }{1+\secL}} \,,
\ee
where $\secL \equiv \tr{\covl \Resbb}$ solves the self-consistent equation\footnote{
    The ``L'' in $\secL$ stands for \emph{latent features}.
}
\be\label{eq:self-consistent-main-text-L}
    \secL = \tr{\frac{\zeta\covl}{ \ridge \IM + \frac{\zeta \nf \covl}{1+\secL} } } \, .
\ee
As per our discussion at the beginning of \S\ref{sec:delta-minus-one}, for $\nf < \nl$ we have an expansion in the $\ridge\to0$ limit as 
\be\label{eq:another-expansion}
\secL = \secL_{-1}\ridge^{-1} +  \secL_{0} +\dots \, ,
\ee
and plugging this expansion back into the self-consistent equation \eqref{eq:self-consistent-main-text-L} tells us that $\secL_{-1}$ solves the non-linear equation
\be\label{eq:nonlin}
1=\tr{\frac{\zeta\covl}{\secL_{-1} \IM + \zeta \nf \covl}}\, ,
\ee
while plugging the expansion into the averaged resolvent \eqref{eq:Resbbzeta} gives
\be\label{eq:Rlin}
\braket{\Resbb}_\fw  = \frac{\secL_{-1} \IM}{\secL_{-1}\IM +\zeta N\Lambda}\gamma^{-1} + \dots \, .
\ee

In fact, although it is not immediately obvious from \eqref{eq:Rlin}, it turns out that the leading part of $\braket{\Resbb}_\fw$ as $\ridge\to0$ written in this formula is actually independent of $\zeta$.  To see this, we can take the derivative of \eqref{eq:nonlin} with respect to $\zeta$ to obtain
\be
0=\tr{\frac{ \covl}{(\secL_{-1} \IM + \zeta \nf \covl)^2}}\left(\secL_{-1}- \zeta \frac{\partial \secL_{-1}}{\partial \zeta} \right) \quad \implies \quad  \secL_{-1}- \zeta\frac{\partial \secL_{-1}}{\partial \zeta}=0\,.
\ee
We can then use this to evaluate the derivative with respect to $\zeta$ of the coefficient of $\ridge^{-1}$ term in the $\braket{\Resbb}_\fw$ expansion in \eqref{eq:Rlin}:
\be
\frac{d}{d\zeta} \left(\frac{\secL_{-1}\IM}{\secL_{-1}\IM +\zeta N\Lambda}\right)=
\frac{N\Lambda}{(\secL_{-1}\IM +\zeta N\Lambda)^2}
\left(\zeta \frac{\partial \secL_{-1}}{\partial \zeta} - \secL_{-1}\right)=0\,.
\ee
We thus conclude that this term is independent of $\zeta$.

This implies that, as long so we are concerned only with the $\ridge\to 0$ limit, we may set $\zeta=1$ in our computation of $\braket{\Resbb}_\fw$, a computation which is now 
manifestly the same as the one we already performed.
In other words, using limits \eqref{eq:ridge-limit-noise-reprint} and \eqref{eq:wow}, substituting for the constant \eqref{eq:noise-term-underparam},
we obtain our
our final result for the label term, \eqref{eq:finalish},
in the under-parameterized regime $(\nf < \nA)$, and $(\nf < \nl)$: %
\be\label{eq:label-term-underparam}
    \L_\w 
    = \frac{\wvar}{2\nl}  \le(\frac{\secL_{-1}}{1-\nf/\nA} \ri)  \, ,
\ee
where  $\secL_{-1}$ is now determined by solving the $\zeta$-independent non-linear equation
\be\label{eq:nonlin2}
1=\tr{\frac{\covl}{\secL_{-1}\IM +  \nf \covl}} \,.
\ee
Curiously, this $\zeta$-independence  means that the variance of the random feature weights, $\fwvar$, drops out of the solution \eqref{eq:label-term-underparam} as well.

\subsubsection*{The Over-Parameterized Regime}\label{sec:derivation-over}

We can now consider the over-parameterized regime where $\nf > \nA$.  In this regime, we will need to average over the random features $\fw$ \emph{first}, and then analyze the resulting expression in ridgeless limit to determine the $x$ average.
Don't despair: luckily, there is a clever trick that we can use to find the answer with nearly no additional work.

Starting from pre-$\widehat x$-averaged form of the label term -- the middle line in \eqref{eq:label-term-expanded} -- let us rewrite this term as
\be\label{eq:label-term-expanded-over}
    \L_\w =  \expval{\frac{\wvar}{2 \fwvar \nB } \left(\tr{\covBB} - 2\tr{   \covBA \Res  \fea^T \widehat \fea } + \tr{\covAA \Res \fea^T \widehat{\fea} \widehat{\fea}^T \fea \Res }  \right)}_{\widehat{x},x,\fw}
\ee
To get this expression, we've used the definitions of the matrices $\covBB$, $\covAB$ and $\covAA$, cf. \eqref{eq:def-Sigmas}, used the commutation relation $\res \fea = \fea \Res$, cf. \eqref{eq:commutation-relation}, and used, on the middle term, the fact that the trace is invariant under transpose. 
Now, note that this form of the pre-averaged label term, \eqref{eq:label-term-expanded-over}, is in a very similar form to the post-$\widehat x$-averaged form of the label term -- the final line in \eqref{eq:label-term-expanded} -- with the matrix substitutions:
\be\label{eq:duality}
\covBB \to \covl \,, \quad \covBA \to \covlf \, , \quad \covAA \to \covf \, ,\quad \Res \to \res \, , \quad \fea^T \to \fea\, , \quad \widehat \fea \to x^T \, , 
\ee
and the overall rescaling 
\be
 \fwvar \nB   \to \nl \,.
\ee
Moreover, considering the definition of the training set features $\fea$, cf.  \eqref{eq:statistical-model-random-feature-map} and \eqref{eq:training-set-fea-def},
\be\label{eq:statistical-model-random-feature-map-reprint}
   \fea \equiv \fw x \quad \Longleftrightarrow \quad \fea_{j;\alpha} \equiv  \sum_{I=1}^{\nl} \fw_{jI} x_{I;\alpha} \, ,
\ee 
we see that transformation $\fea^T \to \fea$ actually is enacting a \emph{training-data--random-feature duality}:
\be\label{eq:data-feature-duality}
x^T \to \fw \, ,\qquad \fw^T \to x\, ,
\ee
which also implies
\be\label{eq:data-feature-scales-transform}
\nA \to \nf \, , \qquad \nf \to \nA\,.
\ee
What this means is that the $\fw$ average of the label term in the form \eqref{eq:label-term-expanded-over} is really the same calculation as the $x$ average we performed in the underparameterized regime so long as we make the transformations \eqref{eq:duality} and \eqref{eq:data-feature-duality}.\footnote{This remarkable simplification is due to a very general duality transformation for certain linear models. Such a duality is not completely new and has been noted before in some settings \cite{bahri2021explaining}. 
}
Ultimately this means that we will find the same final answer \eqref{eq:label-term-underparam} so long as we exchange $\nf \to \nA$ and $\nA \to \nf$.

Let us see how this works in slightly more detail. If you are already happy with the application of the duality, you may skip to the end of the section.
To proceed carefully, let us
apply the inverse of these transformations \eqref{eq:duality} and \eqref{eq:data-feature-scales-transform} to rewrite our computation of  the $x$-averaged label term, \eqref{eq:final-x-averaged-Lw}:
\begin{align}\label{eq:final-u-averaged-Lw}
    \L_\w =  \frac{\wvar}{2  \fwvar \nB } &\bigg\langle \le[1+\le(1 + \ridge \frac{\partial}{\partial\ridge}  \ri) \secFp   \ri]  \times \,\\ 
    \notag
    &\qquad\le( \tr{\covBB} -\le(\frac{\nf}{1+\secFp}\ri) \tr{\le(\IT+\ridge \braket{\Res}_\fw \ri)  \braket{\Res}_\fw \covAB \covBA }\ri) \bigg\rangle_{x, \widehat x} \, .
\end{align}
Here, the $\fw$-average of $\Res$ is given by
\be\label{eq:data-data-resolvent-averaged}
\braket{\Res}_\fw = \frac{1}{\ridge\IT + \le(\frac{\nf}{1+\secFp}\ri)  \covAA} \,,
\ee
where 
we've define $\secFp$, the dual of $\secF$, as this quantity satisfied by another self-consistent equation:
\be\label{eq:self-consistent-prime}
    \secFp(\ridge) \equiv \tr{\frac{\covAA}{ \ridge \IT + \frac{\nf \covAA}{1+\secFp(\ridge)}}} \, .
\ee
Now, to compute the data averages over $x$ and $\widehat x$, we follow similar steps as before: define
\be\label{eq:resolvent-latent-feature-again-again}
    \resbb(\ridge) \equiv \frac{1}{\ridge \IM + \le(\frac{\nf}{1+\secFp}\ri) \frac{\fwvar}{\nl} x x^T } \, ,
\ee
which satisfies $\braket{\Res}_\fw x^T = x^T \resbb$, and where we have used the definition of $\covAA$, \eqref{eq:def-Sigmas}. The same manipulations as before, cf. \eqref{eq:finalish}, gives
\begin{align}\label{eq:finalish-prime}
    \L_\w = \frac{\wvar}{2  \fwvar \nB}\braket{\le[1+\le(1 + \ridge \frac{\partial}{\partial\ridge}  \ri) \secFp \ri] \le(- \ridge^2 \frac{\partial}{\partial \ridge} \tr{\covBB \resbb}\ri)}_{x, \widehat x} \, .
\end{align}
Now, note that the only test-set dependence is in $\covBB$, since $\secFp$ is independent of $\widehat x$. This means we can easily  perform the average over the test data $\widehat x$:
\begin{align}\label{eq:finalish-prime-prime}
    \L_\w = \frac{\wvar}{2  \nl }\braket{\le[1+\le(1 + \ridge \frac{\partial}{\partial\ridge}  \ri) \secFp \ri] \le(- \ridge^2 \frac{\partial}{\partial \ridge} \tr{\covl \resbb}\ri)}_{x} \, ,
\end{align}
where to get this we used definition \eqref{eq:def-Sigmas} and the latent data statistics \eqref{eq:feature-feature-covariance-definition}. 

Finally, the $x$ average proceeds now the same as the $\fw$ average did before. Starting from \eqref{eq:finalish-prime-prime} and taking the ridgeless limit:
\bi
\item The factor in the first square brackets simplifies as before using
\begin{align}\label{eq:ridge-limit-noise-reprint-prime}
\lim_{\ridge \rightarrow 0} \left(1+\ridge\frac{\partial}{\partial \ridge} \right) \secFp(\ridge)  = \secFp_{0}   \,;
\end{align} 
this is quantity independent of $x$ and in the overparameterized regime is given by
\be\label{eq:noise-term-overparam}
\secFp_0 = \frac{1}{\nf/\nA - 1} \, ,
\ee
cf. \eqref{eq:noise-term-underparam} and then exchange $\nf \leftrightarrow \nA$.
\item The factor in the round brackets gives
\be\label{eq:maintext-another-limit-wow-prime}
\lim_{\ridge \rightarrow 0} \le(-\ridge^2 \frac{\partial}{\partial\ridge} \ri)\secLF = \secLF_{-1}  \, ,
\ee
where 
$\secLF_{-1}$ is defined by the solution to the non-linear equation
\be\label{eq:nonlin2-prime}
1\equiv \tr{\frac{\covl}{\secLF_{-1}\IM +  \nA \covl}} \,.
\ee
\ei
Thus, the final answer for the label term takes a very similar form to the underparameterized regime, with the number of random features exchanged with the training set size ($\nf \leftrightarrow \nA$):
\be\label{eq:label-term-overparam}
    \L_\w  = \frac{\wvar}{2\nl}  \le(\frac{\secLF_{-1}}{1-\nA/\nf} \ri)  \, .
\ee

\subsubsection{The Result}\label{sec:derivation-results}

Assembling our results from the different regimes together, \eqref{eq:label-term-underparam}and \eqref{eq:label-term-overparam}, we find for the label term:
\begin{align}\label{eq:statistical-model-answer}
    \L(\nf, \nA; \nl) \equiv \L_\w 
    =  \frac{\wvar}{2\nl}
        \begin{cases}
         \le(1-\nf/\nA\ri)^{-1} \secG_{-1}(\nf, \nl) \,, & \nf < \nA \, ,\\
         \le(1-\nA/\nf\ri)^{-1} \secG_{-1}(\nA, \nl) \,, & \nf > \nA \, ,
    \end{cases}
\end{align}
where $\secG_{-1}(\nG, \nl)$, for $\nG = \{\nf, \nA\}$, is the solution of the non-linear equation
\be\label{eq:secGG}
1=\tr{\frac{ \covl}{\secG_{-1}(\nG, \nl) \, \IM+ \nG \covl}}\, .
\ee
This quantity, $\L(\nf, \nA; \nl)$, represents the test loss in the noiseless limit ($\vare \to 0$) and is symmetric under the interchange of $\nA \leftrightarrow \nf$, a natural manifestation of the aforementioned training-data--random-feature duality.
Moreover, this formula  holds for a random feature linear regression trained on latent data with completely general covariance matrices $\covl$; at no point did we assume anything about the latent feature covariance, $\covl$, to derive this result.
Finally, note that \eqref{eq:statistical-model-answer} is only valid in the regime of $\nf, \nA < \nl$; we will give a formula for the case of a small latent space when we discuss the breakdown of neural scaling laws in \S\ref{sec:break-down-of-neural-scaling-laws}.

Importantly, the factors of $\secG_{-1}(\nG,\nl)$ appearing in \eqref{eq:statistical-model-answer} have an  intuitive explanation.  
In  \S\ref{sec:derivation-label}, we saw that $\secG_{-1}(\nf,\nl)$ arose when we studied a resolvent $\Resbb$, whose average encodes the statistical properties of the matrix %
$(\fw\sqrtcovl)^T \fw \sqrtcovl$.  This is an $\nl\times \nl$ matrix that lives in the latent space and  essentially describes the ``lift" of our empirical covariance $\covf = \fw \sqrtcovl \sqrtcovl^T \fw^T$, cf. \eqref{eq:def-uv}, from the $\nf$-dimensional feature space to $\nl$-dimensional latent space.  Since $\nl>\nf$ this lifted matrix is not full rank and (typically) has a null space of dimension $\nl-\nf$; this null space describes the latent features which our model cannot access.
Indeed, the matrix given by
the ridgeless limit of $\braket{\Resbb}_\fw$,
\eqref{eq:Rlin}, can be interpreted as the projector onto this null space.  
The result is that $\secG_{-1}(\nf,\nl) \equiv \tr{\covl \lim_{\ridge\to0}\left(\ridge\braket{\Resbb}_\fw\right)}$ is the projection of the latent-space covariance matrix onto the null space.  \emph{In other words: $\secG_{-1}(\nf,\nl)$ measures how much of the latent covariance matrix is missed because we have too few features.}  Similarly, $\secG_{-1}(\nA,\nl)$ measures how much of the latent covariance matrix is missed because we have too few samples.  Both effects -- poor performance due to too few samples and poor performance  due to too few features -- can occur in our model, but which one dominates in the ridgeless limit depends on whether $\nf<\nA$ or $\nf>\nA$.

Interestingly, 
even when the label noise is turned off, the factors $\le(1-\nf/\nA\ri)^{-1}$ and $\le(1-\nA/\nf\ri)^{-1}$  indicate a spectrum-independent \emph{universal} noise-like divergence at
 the point of equiparameterization: $\nf \approx \nA$. This divergence occurs because our random features are still noisy samples of the underlying latent features, and our model drastically overfits to this kind of random noise at $\nf = \nA$.  However, as we will discuss in a few paragraphs and in more detail in \S\ref{sec:double-descent}, this divergence is a relic of the ridgeless limit and disappears when the model is optimally regularized.

\begin{figure}
    \begin{center}
     \includegraphics[width=0.49\linewidth]{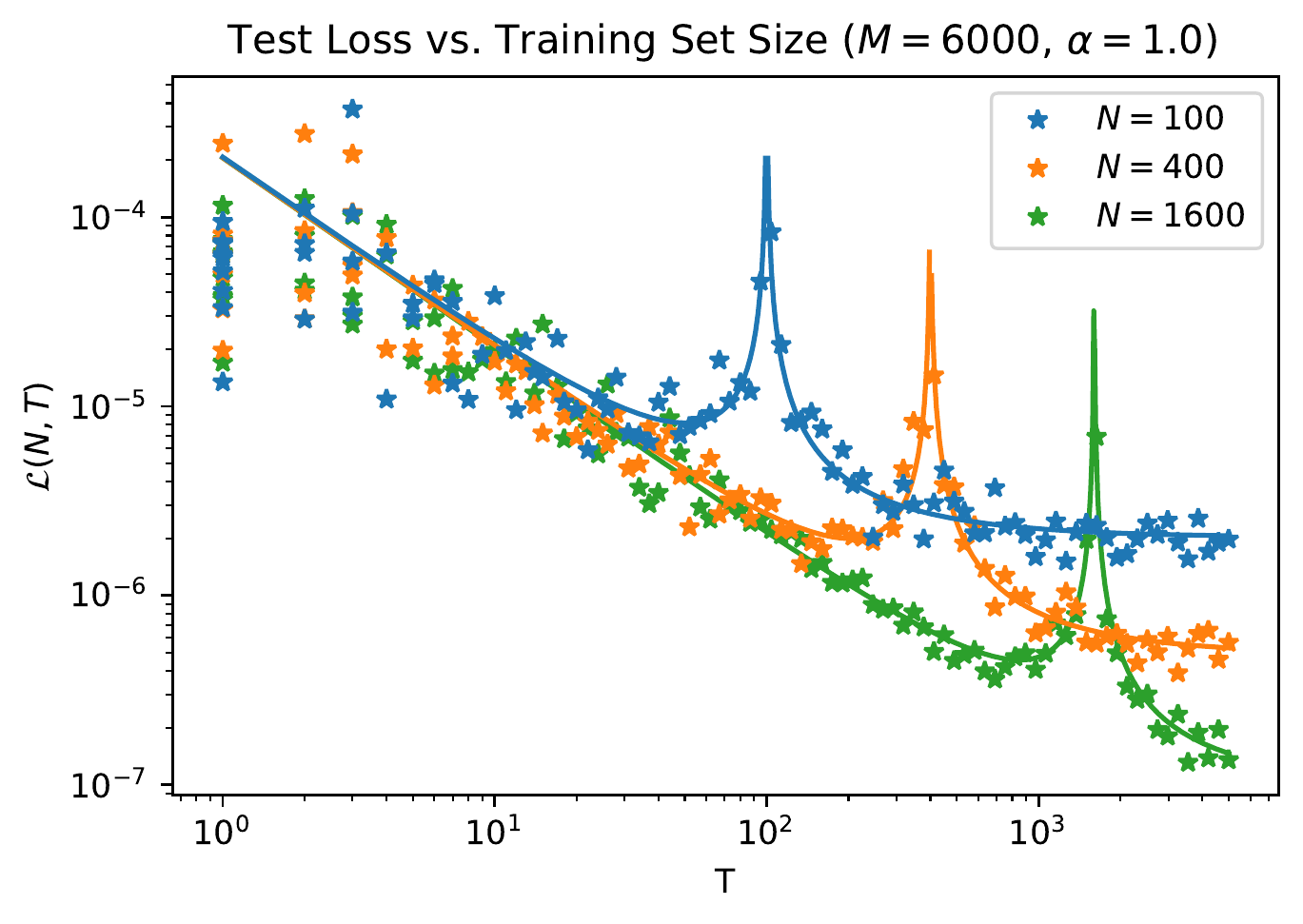} 
     \includegraphics[width=0.49\linewidth]{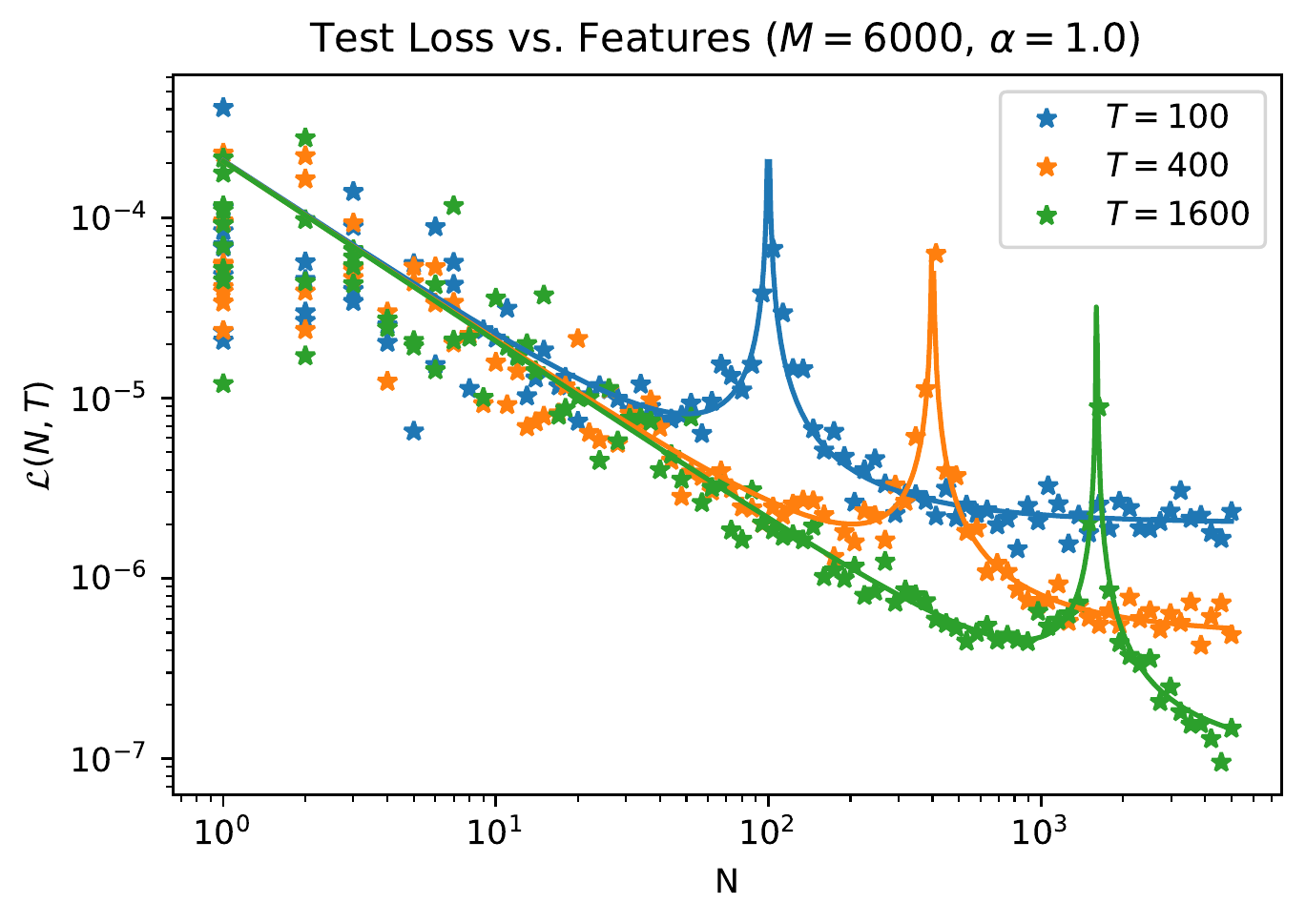} \\
     \includegraphics[width=0.49\linewidth]{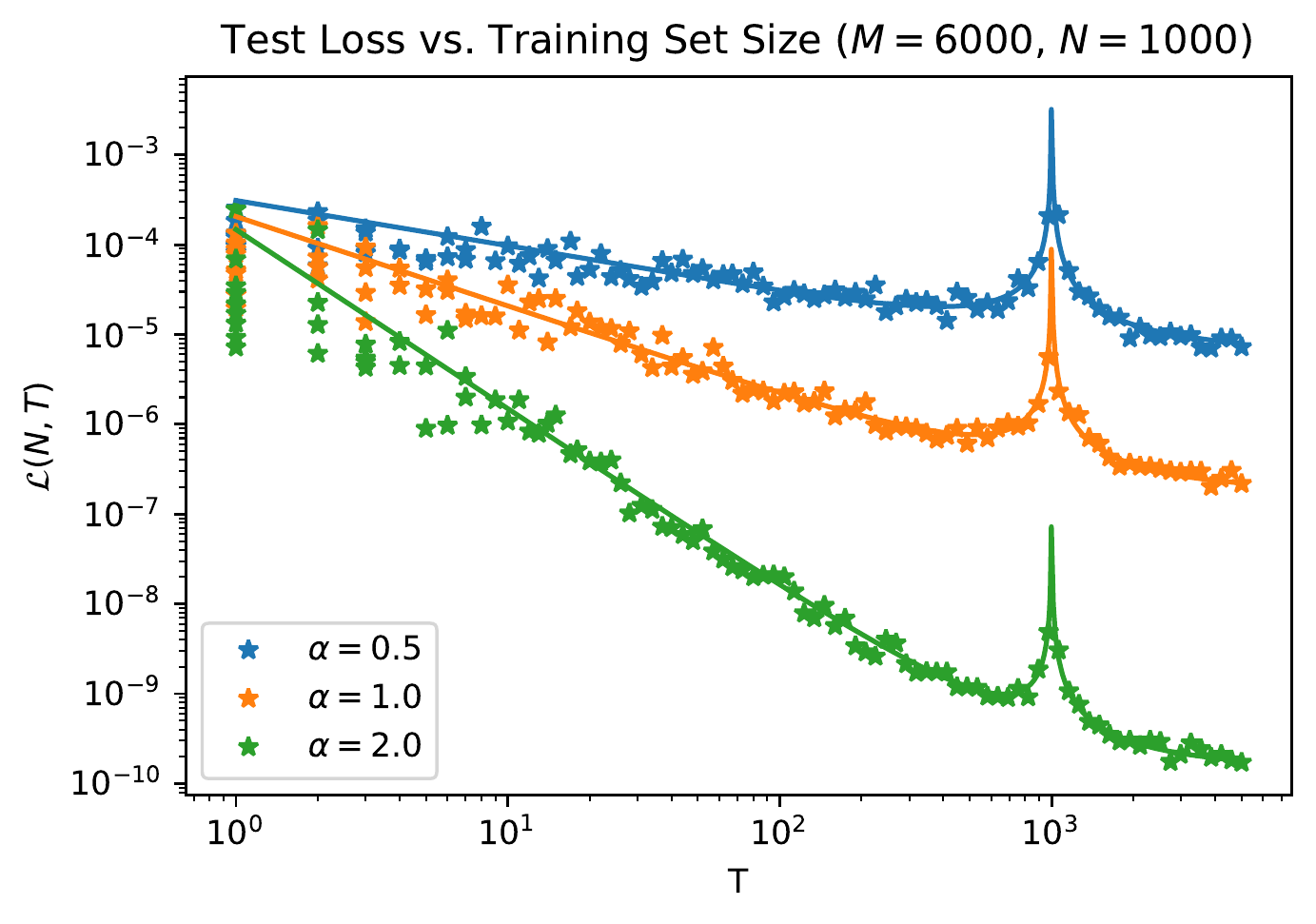}
     \includegraphics[width=0.49\linewidth]{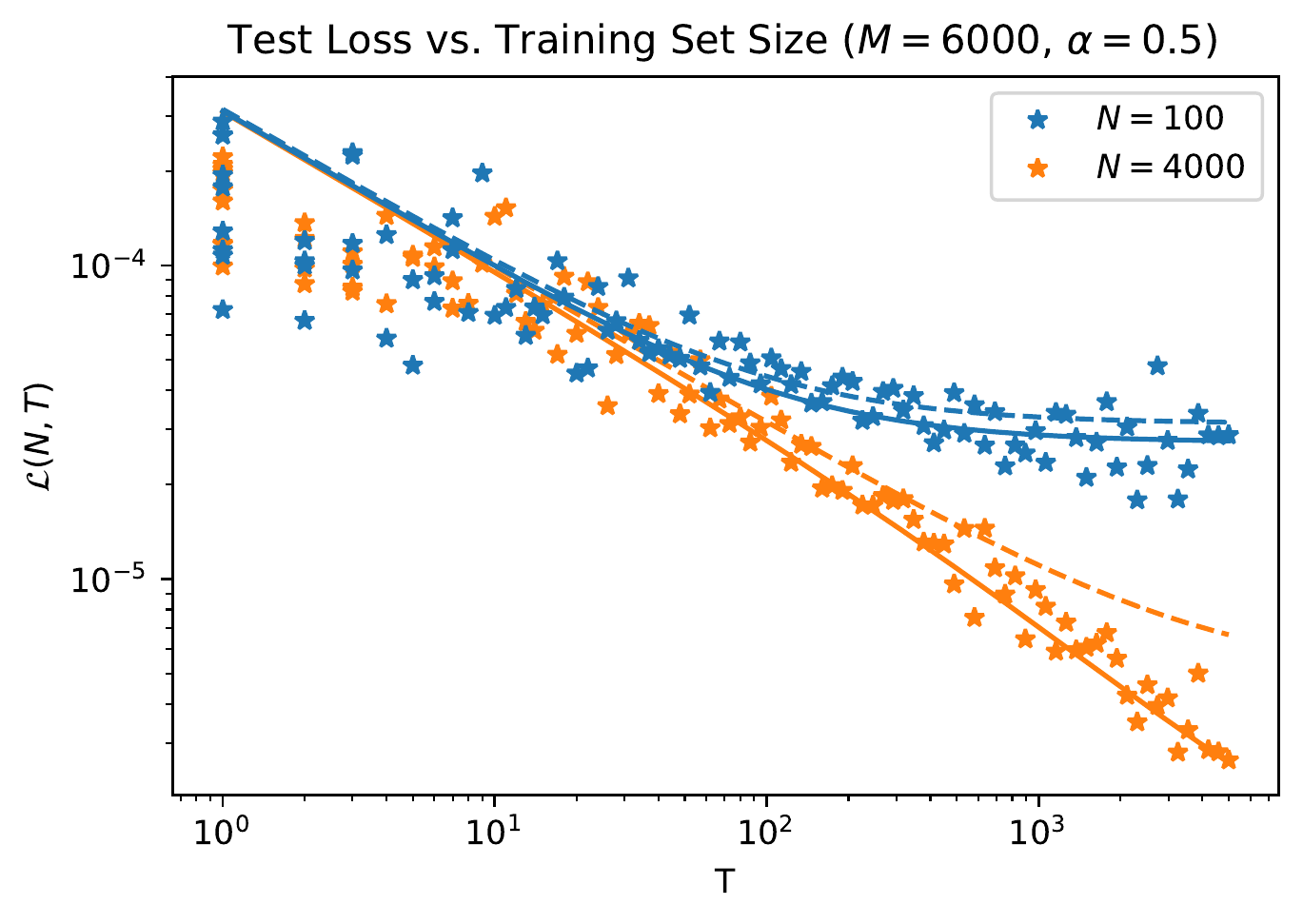}
    \end{center}
    \caption{Test loss from numerical simulations (stars) of our joint statistical model ($\fwvar = 1$, $\eigmax=1$, $\wvar =1$, $\vare = 0$)
compared to the very close fit (solid lines, top and bottom left panels) of the result from our RMT calculation, \eqref{eq:statistical-model-answer}.
For the top and bottom left panels, no regularization is used ($\ridge=0$), while for the bottom right panel the ridge parameter is optimized ($\ridge = \ridge^\star$).
The top panels illustrate the symmetry of the test loss under the exchange $\nA \leftrightarrow \nf$. 
(As before, for smaller values of $\nA, \nf$, the variance of any particular realization is large, and so we've plotted multiple simulations for $\nA, \nf \lesssim 10$.)
\textbf{Top Left:} 
The size of the training set, $\nA$, is varied for a few different sized models, $\nf$, while the size of the latent space and the power-law exponent is held fixed ($\nl=6000$, $\alpha=1.0$).
\textbf{Top Right:} The number of features, $\nf$, is varied for a few different training set sizes, $\nA$, while the size of the latent space and the power-law exponent is held fixed ($\nl=6000$, $\alpha=1.0$).
\textbf{Bottom Left:}
The size of the training set, $\nA$, is varied for a few different power-law exponents, $\alpha$, while the size of the latent space and the size of the model is held fixed  ($\nl=6000$, $\nf=1000$).
\textbf{Bottom Right:} 
When the ridge parameter is optimized ($\ridge = \ridge^\star$), 
the original phenomenological model, \eqref{eq:phenomenological-model-simplified-redux} is a poor fit (dashed lines) as the number of features approaches the size of the latent space ($\nf \to \nl$), 
but our improved model that incorporates the size of the latent space, \eqref{eq:phenomenogical-model-better},  is a much better fit (solid lines).
}
\label{fig:main-result}
\end{figure}

Now, in order to learn about neural scaling phenomenology, let's specialize to the case where 
the latent features have a power-law spectrum, \eqref{eq:exact-power-law-latent-generative}. 
In this case, the self-consistent equations for $\secG_{-1}$ can be easily solved numerically or approximated analytically, the details of which are given in 
\S\ref{sec:delta-minus-one}, starting from \eqref{eq:leading-generic-self-energy-equation-pretrace}. 
The result is \eqref{eq:delta-minus-one-explicit-form-both-regimes}:
\be
\label{eq:delta-minus-one-maintext-1}
\secm(\nG, \nl) = \begin{cases} 
\frac{\eigmax}{\nl^\alpha}\le\{
    k \le[\le(\frac{\nl}{\nG}\ri)^\alpha -1 \ri] + \le[2+ \alpha(1-k)\ri] \le(1 - \frac{\nG}{\nl} \ri)
\ri\} \, , & \nG < \nl \, , \\
0 \,, &  \nG > \nl \, ,
\end{cases}
\ee
with the $\alpha$-dependent constant $k$ defined as 
\be\label{eq:def-k-maintext}
k \equiv \left[\frac{\frac{\pi}{1+\alpha } }{\sin\!\left(\frac{\pi }{1+\alpha}\right)}\right]^{1+\alpha} \,.
\ee
In Fig.~\ref{fig:main-result}, we plot numerical simulations of the test loss, \eqref{eq:def-test-loss} in a variety of ways to demonstrate the fit of our RMT result, \eqref{eq:statistical-model-answer} with the power-law spectral solution \eqref{eq:delta-minus-one-maintext-1}. The simulations were performed analogously to those shown in  Fig~\ref{fig:model-numerics}: we generate latent training and test sets by sampling from  our generative data model, \eqref{eq:feature-feature-covariance-definition} and \eqref{eq:exact-power-law-latent-generative}; then, we map both sets through random feature models of different sizes, \eqref{eq:statistical-model-random-feature-map}; finally, we use the linear regression solution \eqref{eq:linear-regression-solution} to compute test-set predictions using \eqref{eq:test-predictions}. The top left, top right, and bottom left panels of the figure demonstrate the close fit of our RMT result, \eqref{eq:statistical-model-answer}, in the limit of no regularization $(\ridge = 0)$, and the top panels illustrate that the test loss of this statistical model \emph{is} symmetric under the exchange $\nA \leftrightarrow \nf$, as we expected from our discussion of the duality \eqref{eq:duality}.\footnote{However, this exact duality is broken in the presence of noise,
as per our result for the noise term of the loss, \eqref{eq:noise-term-final}.
}

From the top two panels of the figure, we can see that our model captures the
phenomenological behavior of the LLM test loss, e.g., Fig.~\ref{fig:pheno-loss-original}. On the one hand, taking either $\nA$ or $\nf$ (and $\nl$) large in \eqref{eq:delta-minus-one-maintext-1} and plugging into  \eqref{eq:statistical-model-answer}, we find a \emph{power law scaling law}, cf. \eqref{eq:original-power-law-scaling-law-parameters} and~\eqref{eq:original-power-law-scaling-law-trainingset}:
\begin{align}\label{eq:model-power-law-scaling-law-parameters}
\L(\nf) &\equiv \lim_{\nA, \nl \to \infty} \L(\nf, \nA; \nl)   \sim \nf^{-\alpha} \, , \\
\L(\nA) &\equiv \lim_{\nf, \nl \to \infty} \L(\nf, \nA; \nl) \sim \nA^{-\alpha} \, .
\label{eq:model-power-law-scaling-law-trainingset}
\end{align}
On the other hand, once the scaled resource parameter exceeds the large-but-fixed parameter, we see that the test loss asymptotes to a \emph{plateau} that depends on that fixed parameter, cf. \eqref{eq:original-pleateau-parameters} and~\eqref{eq:original-pleateau-trainingset}:
\begin{align}\label{eq:model-pleateau-parameters}
\L_{\text{plateau}}(\nf)&\equiv\lim_{\nf \gg \nA}\L(\nf, \nA; \nl) = 
\frac{\wvar}{2\nl} \secG_{-1}(\nA, \nl) 
\, , \\
\label{eq:model-pleateau-trainingset}
\L_{\text{plateau}}(\nA)&\equiv\lim_{\nA \gg \nf}\L(\nf, \nA; \nl) =
\frac{\wvar}{2\nl} \secG_{-1}(\nf, \nl) 
\, .
\end{align}

Finally, the bottom right panel of Fig.~\ref{fig:main-result} simulates a model that is optimized over different values of the ridge parameter ($\ridge=\ridge^\star$), similar to the regularization used in Ref.~\cite{kaplan2020scaling} that led to the phenomenological model \eqref{eq:phenomenological-loss-original} and the test loss curves in Fig.~\ref{fig:pheno-loss-original}. In this panel, we compare the version of this model appropriate to our setup ($\alpha_{\nA} = \alpha_{\nf}$, $\nf_0^\alpha = \nA_0^\alpha$), cf.~ \eqref{eq:phenomenological-model-simplified}, 
\be\label{eq:phenomenological-model-simplified-redux}
\L_{\text{reg}}(\nf ,\nA) = L_0 \le(\frac{1}{\nf} + \frac{1}{\nA} \ri)^\alpha\, ,
\ee 
with an improved model that incorporates the size of the latent space,
\be\label{eq:phenomenogical-model-better}
\L_{\text{reg}}(\nf ,\nA; \nl) = L_0\le[ \le(\frac{1}{\nf} + \frac{1}{\nA} \ri)^\alpha - \le(\frac{1}{\nl}\ri)^\alpha\ri]\, ,
\ee 
with the constant $L_0$ extracted from our $\ridge=0$ result as
\be
L_0 \equiv \frac{\eigmax \wvar k }{2\nl} \, ,
\ee
and $k$ defined above in \eqref{eq:def-k-maintext}. We see that as the number of features approaches the size of the latent space ($\nf \to \nl$), the initial fit, \eqref{eq:phenomenological-model-simplified-redux}, ignores the finiteness of the latent space and begins to diverge from the simulations, while the improved fit, \eqref{eq:phenomenogical-model-better}, continues to fit the experiments closely. 
This breakdown of the empirical model from Ref.~\cite{kaplan2020scaling} suggests that for LLMs observed in practice, we still haven't seen effects resulting from finite $\nl$. As LLMs become even bigger, it would be interesting to try to fit \eqref{eq:phenomenogical-model-better} and measure the value of $\nl$ for natural language data.

\subsection{(Towards) Modeling Spectral Extension}
\label{sec:spectral-extension}

In our statistical model, the fact that our $\nf$ random features were linearly-related to the $\nl$ latent features made it possible to understand how the scaling laws worked analytically.  However, our minimal statistical model is really an approximation of a more complicated phenomenon: rather than having a large latent space ($\nl > \nf$) with a \emph{linear} transformation,
in a more realistic model the random features are instead \emph{nonlinear} functions of the smaller $\nin$-dimensional input space ($\nin < \nf$).

In \S\ref{sec:model-properties}, we explained that the essential property of such nonlinear feature maps -- which include one-hidden-layer ReLU networks, \eqref{eq:simplest-feature-map}, or the transformer networks that power LLMs -- is the way they \emph{extend} the portion of the spectrum that's approximately fit by a power law. From this we observed that the minimum of the data set size, $\nA$, and the number of features $\nf$, controlled the extent of the power law; the fact that the statistical model we described in \S\ref{sec:derivation-notation} also had a power law controlled by the minimum of these two resource scales is what enabled it to capture the behavior we wanted to model.
In a more realistic model, we can view the large number of useful features that we access from increasing $\nA$ and $\nf$ as being decoded from a non-linear transformation of the input features.

Thus, we would like to better understand how a nonlinear map for generating random features can 
extent the power law in the spectrum of input data.
To be more concrete,  let's consider a simple one-hidden-layer network with a nonlinear activation as in \eqref{eq:simplest-feature-map}:
\begin{align}\label{eq:simplest-feature-map-reprint}
\fea_{j;\alpha} \equiv \sigma\!\le(\sum_{k=1}^{\nin} \fw_{jk} x_{k;\alpha} \ri)\,, %
\end{align}
with $\sigma$ a scalar activation function that acts on each individual component of the input data $x_k$, and $\fw \equiv \fw_{jk}$ is a $\nf \times \nin$ matrix of random feature weights drawn from a zero-mean Gaussian:
\be\label{eq:random-feature-weights-statistics-reprint}
    \expval{\fw_{jk}} = 0\,, \qquad \expval{\fw_{j_1k_1} \fw_{j_2k_2}} = \frac{\fwvar}{\nin} \delta_{j_1j_2} \delta_{k_1k_2} \, .
\ee
As long as the input dimension is sufficiently large, these random features will be approximately Gaussian,\footnote{
    By adding an additional layer or by considering preactivations instead of activations we can remove the requirement of a large input dimension. 
} 
and due to that Gaussianity 
this nonlinear model 
is fully-described by the same random matrix techniques described in the preceding section.\footnote{The fact that one can use RMT tools to 
to study nonlinear random feature maps was a key point in Ref.~\cite{louart2018random}. Note also that these techniques go beyond other approaches in the literature, such as Ref.~\cite{adlam2019random}: while the RMT techniques used here are ``linear'' in so far as they 
treat the features as Gaussian,
they still use the full \emph{nonlinear} kernel, $\covAA$. If you further linearized the calculation of $\covAA$ in terms of the inputs, as was done in \cite{adlam2019random}, you would not see the power-law extension we are interested in finding; this further approximation is excellent at studying the largest eigenvalues, but fails to adequately capture the spectral density for small eigenvalues.
}
In particular, the spectrum of a fixed dataset can be computed from the $\fw$-averaged resolvent:\footnote{
    This was explained in footnote~\ref{footnote:density}: the eigenvalue density can be determined in terms of the discontinuity across the branch cut along the real axis.
}
\be\label{eq:last-resolvent-we-need-question}
    \expval{\Res(\ridge)}_\fw = \frac{1}{ \ridge \IT + \frac{\nf \covAA}{1 + \secFp} } \, ,
\ee
with the quantity $\secFp$ the solution to the self-consistent equation,
\be\label{eq:self-consistent-prime-reprint}
    \secFp(\ridge) \equiv \tr{\frac{\covAA}{ \ridge \IT + \frac{\nf \covAA}{1+\secFp(\ridge)}}} \, ,
\ee
cf.~\eqref{eq:data-data-resolvent-averaged} and \eqref{eq:self-consistent-prime},
and $\covAA$ the data-data covariance matrix of the random features,
\be
\covAA  \equiv \frac{1}{\nf}\braket{\fea^T \fea}_\fw  \quad \Longleftrightarrow \quad \covAA_{\alpha_1 \alpha_2} \equiv \frac{1}{\nf} \sum_{j=1}^{\nf} \braket{\fea_{j;\alpha_1} \fea_{j;\alpha_2}}_\fw
\, .
\ee 
Importantly, when the features are Gaussian this means that the spectrum
is solely determined from knowledge of $\covAA$.\footnote{
    In the context of the theoretical literature \cite{neal1996priors,lee2018deep,matthews2018gaussian,jacot2018neural,brainNTK2019,Yaida2019,PDLT-2022}, this matrix is often known as the \emph{kernel}. Sometimes it's helpful to think of this as arising from a random feature model with infinite features ($\nf \to \infty$).
}

As noted in \cite{louart2018random} and in many other places in the literature, the kernel of a single-hidden-layer network, $\covAA$, is explicitly calculable and well-known for many choices of activation function as a function of the input data. As an explicit example, for the ReLU activation \eqref{eq:relu}, one has that  %
\be\label{eq:covariance-relu}
    \covAA_{\alpha_1 \alpha_2 } = \frac{\sigma_\fw^2}{2 \pi}\|x_{\alpha_1}\|\|x_{\alpha_2}\|\left(\angle(\alpha_1, \alpha_2) \operatorname{acos}\big(-\angle(\alpha_1, \alpha_2)\big)+\sqrt{1-\angle(\alpha_1, \alpha_2)^2}\right) \, ,
\ee
where the norms and inner products are defined as
\be
\norm{x_\alpha}^2 \equiv \sum_{k=1}^{\nin} x_{k;\alpha}^2 \, ,\qquad
\angle(\alpha_1, \alpha_2) \equiv \frac{ 1 }{ \norm{x_{\alpha_1}} \norm{x_{\alpha_2}}} \sum_{k=1}^{\nin} x_{k;\alpha_1}  x_{k;\alpha_2} \, .
\ee
If we could find the spectrum of  this infinite-random-feature kernel, \eqref{eq:covariance-relu}, for a given data model or natural dataset, we could use that
in conjunction with the resolvent, \eqref{eq:last-resolvent-we-need-question}, to learn about the spectral extension for the finite-random-feature kernel evaluated on power-law data. Such a rigorous analytical derivation would be interesting to work out and perhaps give insights into in the interaction between datasets and activation functions. %

\subsection{Comparing to Other Methods}
\label{sec:other-methods}

In this section, we compare our results to other RMT derivations in the machine learning literature, in particular to Refs.~\cite{louart2018random} and \cite{bordelon2020spectrum,Canatar:etal}. In particular, we discuss the uses and limitations of our Feynman diagrammatic methods and contrast with other methods.

First, random matrix theory techniques used in this paper have many similarities with the approach taken in \cite{louart2018random}, which was an influential starting point for the present work. Our intermediate result,  \eqref{eq:final-u-averaged-Lw}, is essentially the same result that appears as the random feature-averaged test loss in Conjecture 1 of their paper, up to combining some terms with the use of our averaging over the random teacher weights. While our diagrammatic derivation may then appear less general than \cite{louart2018random} because of the random teacher assumption for the labels, our derivation nevertheless can be used in either setting. 

In particular, in following our derivation in \S\ref{sec:derivation}, one can see that the covariance matrices appearing at the beginning of each term in  \eqref{eq:label-term-expanded-over} could be substituted for arbitrary matrices without changing the structure of the following calculation. The formula from Ref.~\cite{louart2018random}, without the random teacher averaging, then follows  (up to trivial constants) by replacing 
\begin{equation}
    \covBB \to \widehat{y}^T\widehat{y}\,,  \qquad  \covAB \to {y}^T\widehat{y}\,,  \qquad  \covAA \to {y}^T {y} \,,
\end{equation}
at the beginning of each term and following these substitutions through the calculation of the random feature $\fw$ average. The resulting expression is then consistent with what appears in \cite{louart2018random}.  

Nevertheless, there's also some notable differences between our methods and results. First,
our expression is somewhat simpler than the one that appears in \cite{louart2018random} due to the fact that we rewrote the prefactor in terms of a differential operator. Second, we also believe our diagrammatic techniques give a more transparent and extensible methodology for computing similar averaged expressions. Finally, our work  generalizes the result in \cite{louart2018random}, both by deriving a dual expression for averaging over training samples instead of random features, \eqref{eq:final-x-averaged-Lw}, and by calculating the test loss when both random feature and data averages are taken at the same time, \eqref{eq:statistical-model-answer}. This final \emph{double} average is entirely novel to our work and does not otherwise appear in the literature.

Relatedly, our work also concerns similar types of averages to Refs.~\cite{bordelon2020spectrum,Canatar:etal}. In these papers, the authors used replica methods to compute 
the expected test loss of a kernel regression. One way to interpret their work is in terms of a single average over the data $x$. We find similar results using our methods in Appendix~\ref{sec:other-models} where we study (non-generalized) linear models and compute a single $x$ average over the model's features, $x_j$, since, in these models, there are no random features to average over. Although they are in different forms, our computation, \eqref{eq:result-appendix-a}, matches the main result of Ref.~\cite{Canatar:etal}, cf. their Eq. 4. 

One reason we were able to simplify the calculation of their very nice result is, by rewriting the loss in terms of differential operators, we immediately see that the test loss is determined entirely by the averaged quantity $\secG \equiv \tr{ \covl \resb}$. The computation of this quantity for a general spectrum $\covl$ is a classic result of random matrix theory that goes back to the original work of Marchenko and Pastur \cite{Mar_enko_1967} and has been studied or rediscovered many times since. 
In particular, this quantity $\secG$, or equivalently, the $x$ averaged resolvent, $\resb$, can be computed by many means, e.g., by the Feynman diagrammatic approach of \S\ref{sec:the-noise-term} and, of course, also by the replica methods in Refs.~\cite{bordelon2020spectrum,Canatar:etal}. 

On the one hand, we think that the  computation using Feynman diagrams has a few advantages over replica techniques:  in addition to being more straightforward,
one also does not need to prove that there is a unique analytic continuation in the replica index.  Moreover, although we have no need for it here, the Feynman diagrammatic techniques can be easily adapted to compute subleading $1/\nf$ corrections to averages, if desired.  

On the other hand, even for Gaussian distributions, some quantities cannot be readily computed using Feynman diagrams.  This generally occurs when one computes the expectation value of a non-analytic function, such as the logarithm of a function of $\fea$.\footnote{This occurs in the information theory context in the study of von Neumann entropies and in the physics context in the study of quenched disorder.} In such a case, one can use replica methods.\footnote{This requires one to instead consider $n$ copies of the original random variable and compute the expectation value of a logarithm by analytically continuing in $n$ using $\log z = \lim_{n\to 0}\left(z^n-1\right)/n$.  This method only works for certain classes of probability distributions (and quantities being averaged) such that they are so strongly constrained that there is a unique analytic continuation in $n$. 
 }

On our third and final hand, other more advanced techniques to study matrix integrals at large $N$ include the use of loop equations and the method of orthogonal polynomials. These techniques are often necessary to treat more complicated (non-Gaussian) probability distributions.\footnote{For a useful review of these methods, see \cite{Eynard:2015aea}.}  These methods allow one to effectively sum up  infinite classes of Feynman diagrams but are typically overkill for Gaussian matrix expectations.  To our knowledge, these techniques have not yet been used in a machine learning context.

\section{Discussion of Results}\label{sec:discussion}

Now that we have a statistical model of scaling laws that we understand for jointly large-but-finite model size $\nf$, training set size $\nA$, and latent space size $\nl$, in this section we have a discussion of what we can learn from our statistical model of scaling laws.

In \S\ref{sec:break-down-of-neural-scaling-laws}, we interpret our results from \S\ref{sec:derivation} in the limit that the model size or training set size approaches the size of the latent space, $\nf, \nA \sim \nl$, and the neural scaling law phenomenology of \S\ref{eq:phenomenological-loss-original} breaks down.

In \S\ref{sec:double-descent}, we discuss how the regime of scaling laws, of large training set size and large model size, pushes resource efficient  and properly regularized models towards the equiparameterization regime, and how the phenomenon of double descent is not really relevant for such models. 

In \S\ref{sec:coding-theory}, we try to reconcile the large latent space, $\nl$, required for datasets that allow for neural scaling laws with the traditional idea that input datasets are embedded in high-dimensional spaces, $\nin$, and can be compressed to a latent space with much smaller intrinsic dimension, $\din$. We note that there are a number of notions of dimensionality, and the particular power-law structure of the datasets that give rise to scaling laws makes different notions meaningful for different questions.

Finally, in \S\ref{sec:breakdown-data-model}, 
we identify some limitations of our generative data model that could be improved in future analyses.

\subsection{The Breakdown of Neural Scaling Laws}
\label{sec:break-down-of-neural-scaling-laws}

In this subsection, we would like to understand whether 
the empirically-derived phenomenological model of the test loss, \eqref{fig:pheno-loss-original}, can break down.\footnote{In some parts of the literature, e.g. in \cite{JMLR:v23:20-1111}, ``breakdown of scaling laws'' is 
    used to mean the transition from the power law to the plateau 
    in the test loss;
    as should be clear, this is not what we mean here. 
} 
In particular, we would like to know \emph{(a)} when is such a formula no longer predictive of performance and \emph{(b)} what is the behavior in the new regime? 

To give insight into the answer of these questions, we can use our statistical model from \S\ref{sec:derivation}.
One hint towards an answer is that model \eqref{fig:pheno-loss-original} only depends on two of the scales in the problem, $\nf$ and $\nA$, while the formula we derived, \eqref{eq:statistical-model-answer}, also depends on size of the latent space of the data distribution, $\nl$.
In particular, in our regularized $(\ridge > 0)$ simulations of the test loss of our model
(Fig.~\ref{fig:model-numerics}), to fit \eqref{fig:pheno-loss-original} it was important that the size of the latent space be the largest scale in the problem, $\nf, \nA \ll \nl$. Here we will explore what happens when this condition is violated.

To proceed, it is convenient to organize our discussion around both whether or not there's noise added to the training labels and whether or not there's a nonzero ridge parameter regularizing the regression.

\subsubsection*{No noise}

\begin{figure}[ht]
\begin{center}
 \includegraphics[width=0.49\linewidth]{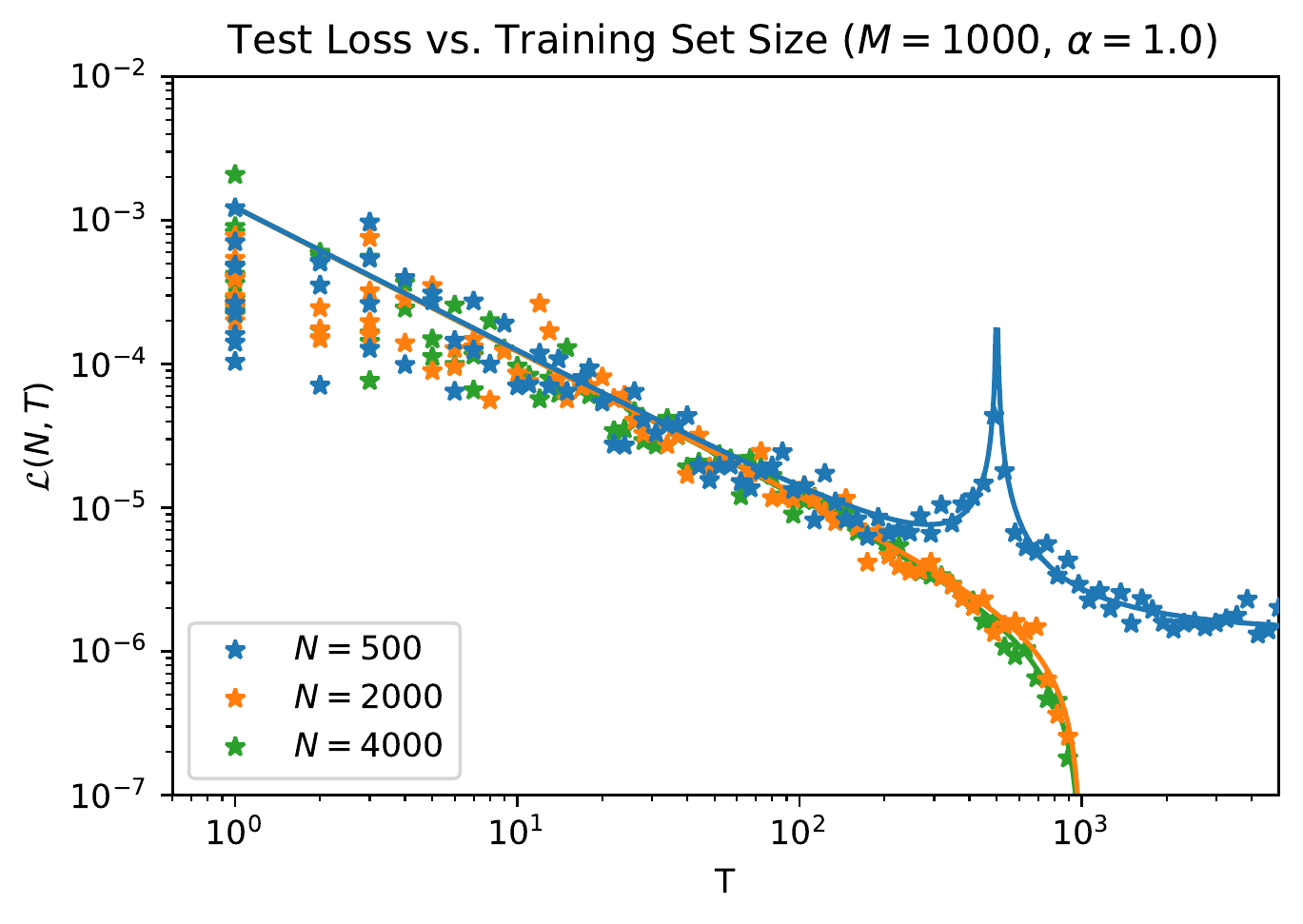} \includegraphics[width=0.49\linewidth]{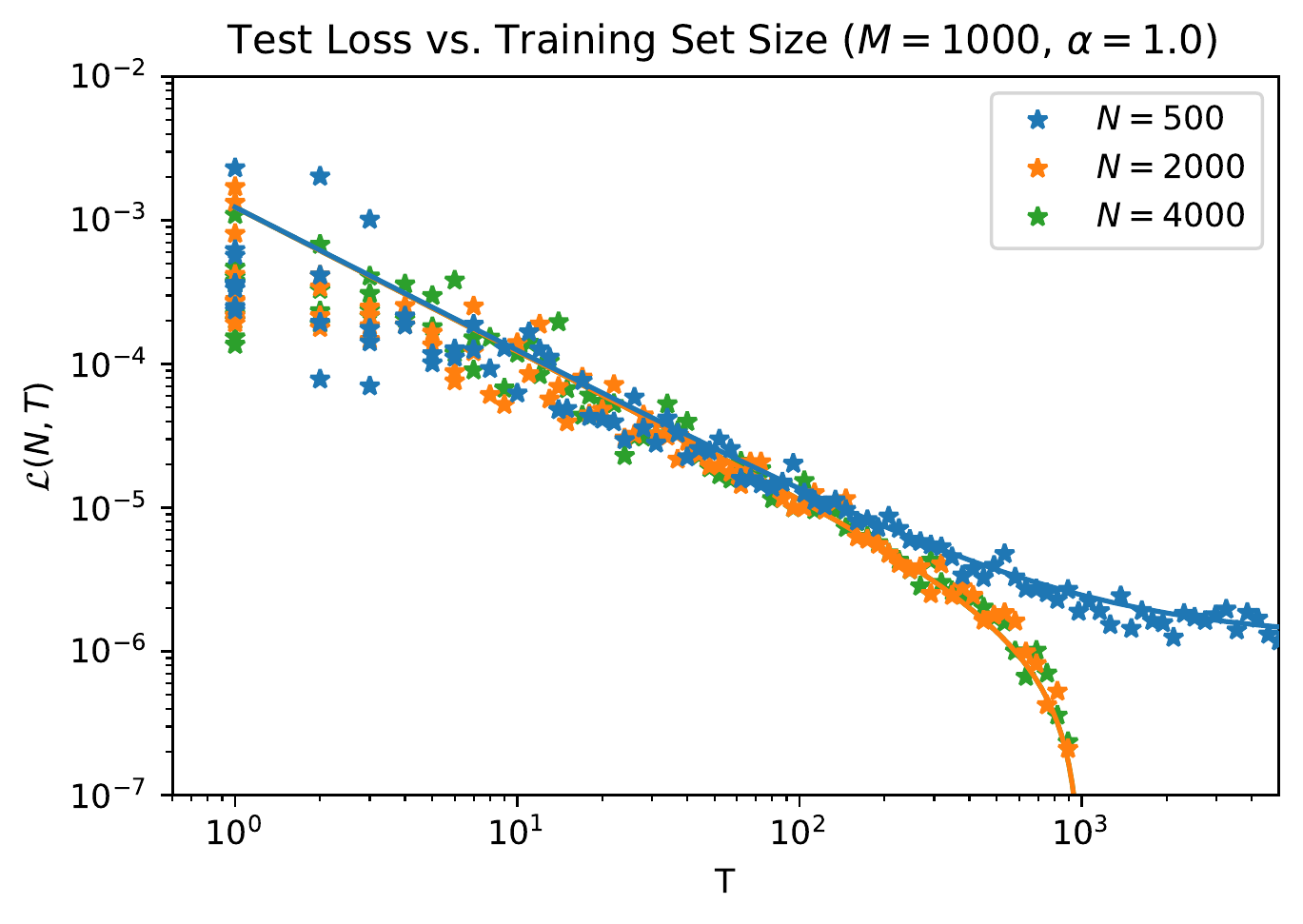}
\end{center}
\caption{Simulation (stars) and fit (solid) of the scaling-law model from \S\ref{sec:derivation} with exponent $\alpha = 1$, latent-space size $\nl = 1000$, and the following entirely inconsequential choices: $\eigmax=\wvar =\fwvar =1$. The orange and green curves represent models in different large-latent-space regimes $(\nA, \nl < \nf)$ that demonstrate novel behavior, while the blue curves represent models in the usual regime $(\nA, \nf < \nl)$, including both power law and plateau behavior, and is included for purposes of comparison. As the size of the training set is increased to the size of the latent space, the test loss very rapidly approaches zero.
\textbf{Left:} No regularization is used ($\ridge=0$), and the fit is the result of our RMT calculation \eqref{eq:statistical-model-answer}. %
\textbf{Right:} The ridge parameter is optimized ($\ridge = \ridge^\star$), which is intended to model the early stopping regularization used in Ref.~\cite{kaplan2020scaling}, and the  semi-heuristic fit is given by \eqref{eq:breakdown-no-noise-regularized}.
}
\label{fig:breakdown-no-noise}
\end{figure}

First, let's consider the case without noise. In the left panel of Fig.~\ref{fig:breakdown-no-noise} we plot 
a simulation of our statistical model with no regularization ($\ridge=0$)
and also plot our RMT calculation of the test loss, 
\eqref{eq:statistical-model-answer},
for 
models for which the number of features is much larger than the size of the latent space and the number of samples in the training set $(\nf > \nA, \nl)$; in the right panel, we use optimal regularization ($\ridge = \ridge^\star$) in the simulation and also plot a new fit that we will discuss below.
In both panels, we learn that ``breakdown'' of scaling laws -- without noise -- is a lot like a \emph{singularity}: all of the sudden at $\nA = \nl$ the test loss drops very rapidly to zero!

To understand these results, let's turn to our calculation:
if we consider a model in the
large-latent-space
regime, $\nA, \nl \ll \nf \to \infty$, from \eqref{eq:statistical-model-answer} and \eqref{eq:delta-minus-one-maintext-1} we see that the test loss of our model simplifies as
\begin{align}\label{eq:breakdown-no-noise}
\lim_{\nf \to\infty} \L(\nf, \nA; \nl) &= \frac{\wvar}{2\nl} \secm(\nA,\nl) \, \\
&=  
\begin{cases}
         \frac{\wvar\eigmax }{2\nl^{\alpha+1} }\le\{
    k \le[\le(\frac{\nl}{\nA}\ri)^\alpha -1 \ri] + \le[2+ \alpha(1-k)\ri] \le(1 - \frac{\nA}{\nl} \ri) 
\ri\}\,, & \nA < \nl \, ,\\
         0 \,, & \nA > \nl \, . \notag
\end{cases}
\end{align}
Recalling our analysis in \S\ref{sec:delta-minus-one}, we see that the \emph{scaling} regime controls the test loss when the training set is smaller than the size of the latent space, $\nA \ll \nl$, and we find the same power law scaling law as before: 
\be
\lim_{\nA \ll \nl} \lim_{\nf \to\infty} \L(\nf, \nA; \nl) \sim \nA^{-\alpha} \, .
\ee
However, when the size of the training set approaches the size of the latent space, $\nA \to \nl$, the test loss is instead controlled by \emph{coincident} regime, and we have
\be
\lim_{\nA \nearrow \nl} \lim_{\nf \to\infty} \L(\nf, \nA; \nl) = \frac{2+\alpha}{2}  \le(\frac{\wvar\eigmax}{\nl^{\alpha+1} }\ri) \le(1- \frac{\nA}{\nl}\ri) \, ,
\ee
which vanishes as $\nA$ approach $\nl$ from below. 

Adding and optimizing the ridge parameter doesn't really change this as regularization only really matters around the point of equiparameterization ($\nA \approx \nf$), and here we're interested in the limit of $\nf \to \infty$. 
For this reason, it's easy to construct the fit to the simulated in the right panel of Fig.~\ref{fig:breakdown-no-noise}. The blue curve is included for comparison and shows a similar regime as 
in the experiments plotted in bottom right panel of Fig.~\eqref{fig:main-result}
$(\nf < \nl)$, for which 
we are using the improved fit
\eqref{eq:phenomenogical-model-better}. For the new 
large-latent-space
regime (orange and green curves) with $\nf > \nl$, we can simply use the unregularized large feature formula, \eqref{eq:breakdown-no-noise}. Altogether, we can write our no-noise optimal regularization fit to the simulations in this plot as
\begin{align}\label{eq:breakdown-no-noise-regularized}
\L_{\text{reg}}(\nf, \nA; \nl) = \frac{\wvar\eigmax}{2\nl}
    \begin{cases}
         k \le[\le(\frac{1}{\nf} + \frac{1}{\nA}\ri)^{\alpha} - \frac{1}{\nl^\alpha}\ri]& \nA, \nf \leq \nl\, , \\
    k \le(\frac{1}{\nA^\alpha} -\frac{1}{\nl^{\alpha}} \ri) + \frac{1}{\nl^{\alpha}}\le[2+ \alpha(1-k)\ri] \le(1 - \frac{\nA}{\nl} \ri) \,, & %
    \nA\leq \nl < \nf \,, \\
    0\,, & \nA, \nf > \nl \,.
    \end{cases} 
\end{align}
In other words, in the right panel of Fig.~\ref{fig:breakdown-no-noise}, we use the top line for the blue curve, cf.~\eqref{eq:phenomenogical-model-better},  and the middle and bottom lines for the green and orange curves.

Finally, given the duality that exchanges the training set size and number of features, 
$\nf \leftrightarrow \nA$,
similar formulas will hold in a regime where we have a super-sized training set, $\nA > \nl$: in the unregularized case as we increase the model size we will see behavior described by \eqref{eq:breakdown-no-noise} but with the swapping of number of features with the size of the training set, $\nf \leftrightarrow \nA$, and an analogous statement will hold in the case of regularization and \eqref{eq:breakdown-no-noise-regularized}.

\subsubsection*{Noise}
\begin{figure}[ht]  
\begin{center}
 \includegraphics[width=0.49\linewidth]{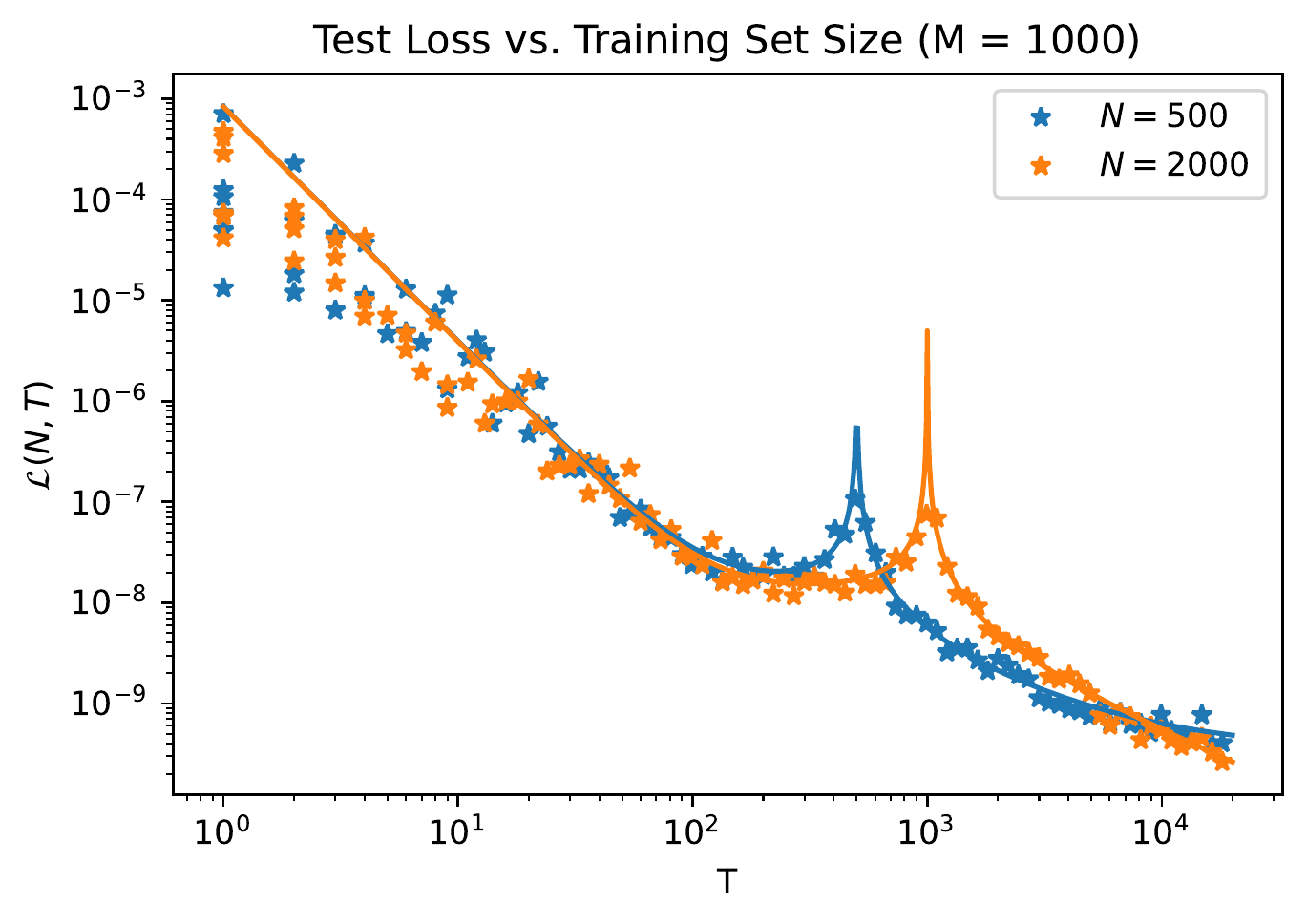} \includegraphics[width=0.49\linewidth]{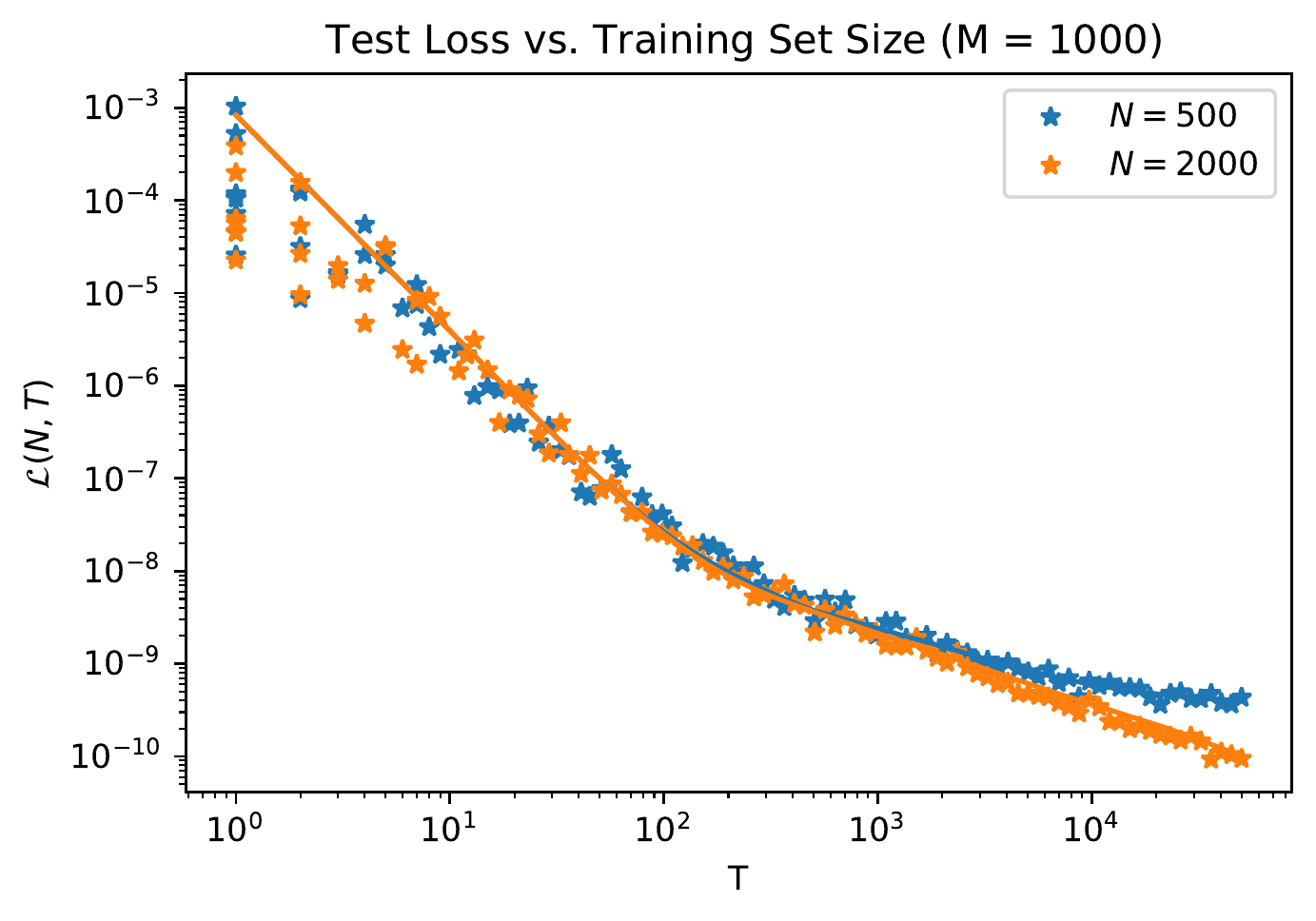}
\end{center}
\caption{
Simulation (stars) and fit (solid) of the scaling-law model from \S\ref{sec:derivation} with  
exponent
$\alpha = 2.33$, latent-space size $\nl = 1000$, $\vare=10^{-8}$, and the following choices that are made so that there's good signal to noise in the data: $\eigmax=\wvar =\fwvar =1$. 
The orange curves represent models in different large-latent-space regimes $(\nl < \nf)$ that demonstrate novel behavior, while the blue curves represent models in the usual regime $(\nf < \nl)$, including both power law and plateau behavior, and is included for purposes of comparison.  %
\textbf{Left:} No regularization is used ($\ridge=0$) and the fit is our RMT calculation $\L(\nf, \nA; \nl) + \L_\noise$, cf. \eqref{eq:breakdown-no-noise} and \eqref{eq:noise-breakdown-formula}. 
\textbf{Right:} Optimal regularization ($\ridge = \ridge^\star$) and the fit is to our phenomenological formula $\L_{\text{reg}}(\nf, \nA; \nl) + \L_{\noise ,\text{reg} }$, cf. \eqref{eq:breakdown-no-noise-regularized} and \eqref{eq:regularization-breakdowon-noise}, 
shows the transition from the power-law regime with $\L \sim \nA^{-\alpha}$ scaling to the   noise-dominated  $\L \sim \nA^{-1}$ scaling (orange) and to resource-bottlenecked plateau  (blue) for the large-latent-space regime and the usual regime, respectively.
}
\label{fig:breakdown-noise}
\end{figure}

Now, let's turn on the label noise ($\vare > 0$). In addition to the label term that made up the loss immediately above, we now also must include the contribution from the noise term, \eqref{eq:noise-term-final}, when $\nf>\nl$:
\begin{equation}\label{eq:noise-breakdown-formula}
    \L_\noise = \frac{ \vare }{2 }
       \begin{cases}
        \alpha + \frac{1}{\nl/\nA-1} \,,   & \nA < \nl \,, \;\; \nf>\nl \,, \\
        \frac{1}{\nA/\nl-1} \,,            & \nA > \nl \,, \;\; \nf>\nl \, .
       \end{cases}
\end{equation}
In the left panel of Fig.~\ref{fig:breakdown-noise}, we see that the inclusion of this term 
gives an excellent fit to noisy numerical simulations.

Although we are still unable to analytically optimize over the ridge parameter, $\ridge$, we can again write a phenomenological formula for the regularization of the noise term by finding a simple solution that interpolates between the asymptotics of the above formula. Such a formula could be 
\begin{equation}
    \frac{ \vare }{2 } \le[\frac{1}{\nl/(\nA+\alpha\nl) + \nA/\nl}\ri] \, .
\end{equation}
However, we can do slightly better when both terms in the denominator are of comparable size by normalizing with a factor of $1/2$ using a bump function:
\begin{equation}\label{eq:regularization-breakdowon-noise}
    \L_{\noise ,\text{reg} }= \frac{ \vare }{2} \le[ \frac{1}{\nl/(\nA+\alpha\nl) + \nA/\nl}\ri] \le[\frac{1}{1+ e^{-\left(\nl / \nA + \nA / \nl \right)/c}}\ri] \, ,
\end{equation}
for some constant $c$ that we can choose to fit the experiment. 
We see in the right panel of Fig.~\ref{fig:breakdown-noise} that this formula, in conjunction with the label term \eqref{eq:breakdown-no-noise-regularized}, gives an excellent fit to the simulations, although we emphasize that the exact functional form near equiparameterization is chosen in a rather ad hoc way and should not be understood to be the definitive functional form of the optimized solution.

Most notably, for large $\nA$ we see that both the unregularized case, \eqref{eq:noise-breakdown-formula}, and the regularized case, \eqref{eq:regularization-breakdowon-noise}, %
there's a \emph{universal} $\sim 1/\nA$ falloff of the test loss when the model size and training set size jointly exceed the size of the latent space ($\nf, \nA > \nl$). %
If such a transition in powers appears in a model at an otherwise undefined scale, it could be suggestive of a breakdown associated with having reached the size of the latent space.\footnote{However, note that this universal $\sim 1/\nA$ behavior depends on 
the label noise not being correlated from sample to sample, as in our noise model \eqref{eq:def-noise-statistics}; it's not obvious to us whether this is an appropriate assumption for self-supervised generative modeling tasks. 
}
Interestingly, the consequences of this for the practitioner now depend on the size of the power-law exponent $\alpha$: for $\alpha > 1$, this transition would limit the model's performance gains with increasing resources $\nf$ and $\nA$, while for exponents $\alpha < 1$, it would enhance such gains.

\subsection{The Battle of the Parameterizations}
\label{sec:double-descent}

Despite common wisdom 
coming from classical statistical intuition,
it's been long observed in the modern deep learning setting that very large models trained to near zero training error can generalize well without \emph{overfitting}, even with minimal to no regularization \cite{belkin-double,opper1995statistical,opper2001learning,advani2020high,spigler2018jamming,geiger2019jamming,nakkiran2021deep}.\footnote{Such regularization includes both \emph{explicit} regularization, such as a nonzero ridge parameter $\ridge$, or \emph{implicit} regularization, such as early stopping the
gradient-based optimization of an objective function.} 
This behavior is a feature of the overparameterized regime, in which the number of parameters of the model is greater than the size of the training set, $\nf > \nA$. 
In this regime,
inference is said to succeed by \emph{interpolating} between the learned training points, and
 the performance of such fully-trained models will continue to improve as model sizes increase for a fixed training set.

In contrast, in the underparameterized regime in which the size of the training set exceeds the model size, $\nA > \nf$, the \textbf{bias-variance tradeoff} of classical statistics dictates performance of under-regularized models (see, e.g., \cite{hastie2009elements}).
For a fixed training set, a model may \emph{underfit} if it's too small, 
exhibiting a \emph{bias} 
as
it's not expressive enough to learn the target function; it may overfit if it's too big and unhelpfully learns the noisy fluctuations in 
the
training set due to the \emph{variance} between different realizations of the data: 
thus, there's an 
extremal
size balancing the bias and variance at which performance peaks. 

Typically, this point is only a local minimum of the test error, 
and
 -- for a fixed training set size -- increasing the model size into the overparameterized regime 
can
 eventually lead to models that outperform the optimal underparameterized model.
This overall description of the test error as a function of parameters has been dubbed \textbf{double descent} \cite{belkin-double} 
and a large amount of effort has been spent on understanding the setting for and the mechanism of this behavior.\footnote{See, e.g., \cite{nakkiran2021deep} and additional references therein.}  

\begin{figure}[ht]
\begin{center}
 \includegraphics[width=0.75\linewidth]{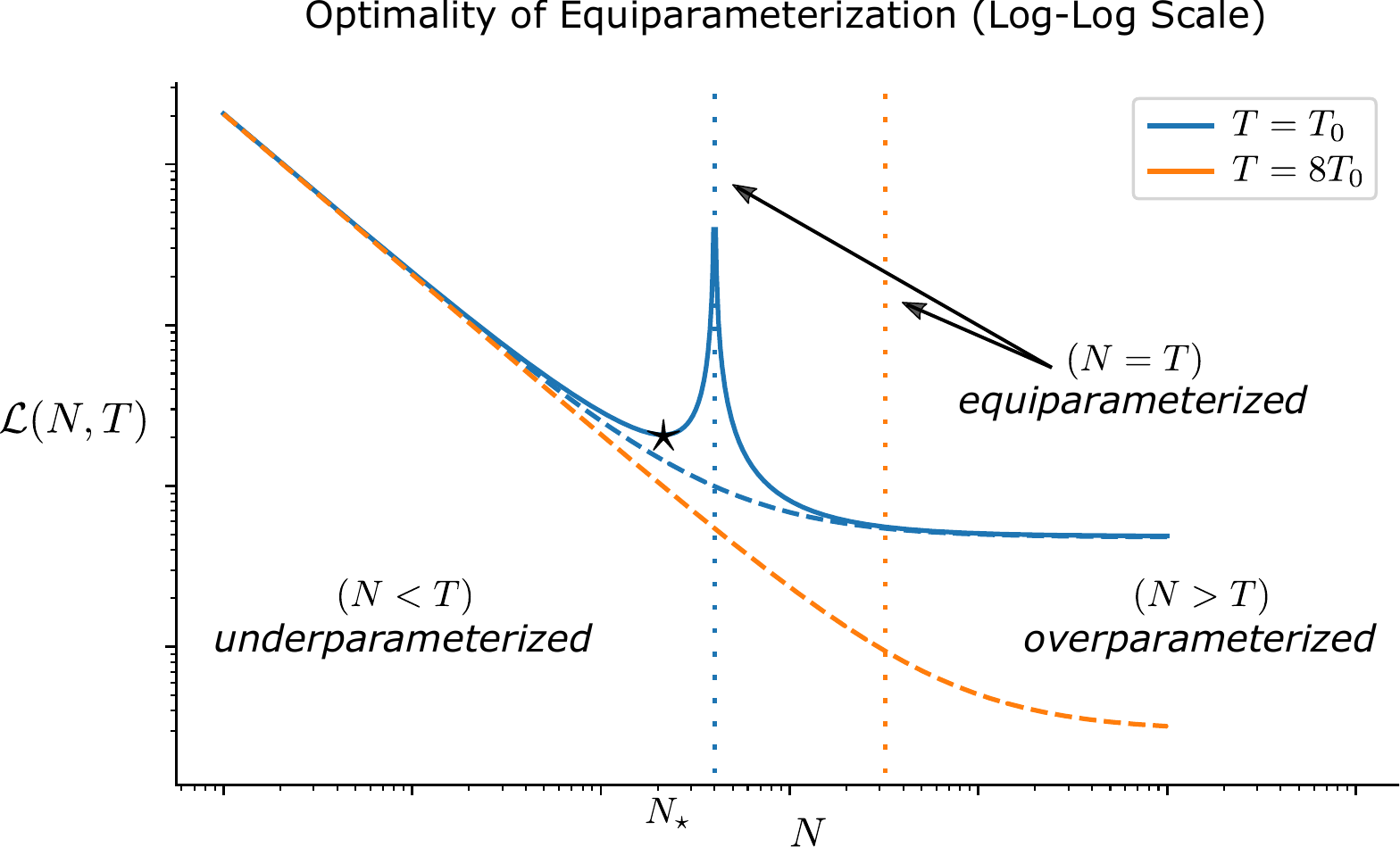}
\end{center}
\caption{Sketch 
of test losses of our statistical model from \S\ref{sec:derivation} on a log-log scale for different fixed training set sizes for both unregularized or ridgeless (solid) and optimally regularized (dashed) models. The solid blue curve exhibits the double descent phenomenon, with a local minimum of performance in the underparameterized region (black star, $\nf =\nf_\star$) and with performance further improving asymptotically in the overparameterized region. 
The two blue curves illustrate how the double-descent peak is an artifact of the ridgeless limit ($\ridge=0$), with performance monotonically improving through the point of equiparameterization (vertical dotted lines) when the models are properly regularized.
Comparison of the dashed blue ($\nA=\nA_0$) and orange ($\nA=8\nA_0$) curves illustrates the optimality of near-equiparameterization when using regularization properly: the best performance boost results from scaling 
the model jointly with the size of the training set $(\nf \sim \nA)$.
}
\label{fig:double-descent}
\end{figure}

In Fig.~\ref{fig:double-descent} we exhibit 
this phenomenon (solid blue curve)  based on \eqref{eq:statistical-model-answer}, the now-familiar result of our RMT calculation of the test loss of our statistical model presented in \S\ref{sec:derivation}.\footnote{
    Technically, we should distinguish between the test loss and the generalization error (the difference between the test loss and training loss) in the underparameterization regime since the model can underfit. However, this difference is not important for the qualitative point we want to make here.
}
The plot depicts the performance of our model for a fixed training set size ($\nA=\nA_0$) in the ridgeless limit ($\ridge=0$): 
in the underparameterized region to the left of the \emph{equiparameterization} peak at $\nf=\nA$, the test loss 
has a local minimum
($\nf = \nf_{\star}$), which optimally trades off the bias of too small a model with the variance of too large a model;
in the overparameterized region to the right of the peak, the test loss 
continues to improve past the local minimum at $\nf = \nf_{\star}$ after overcoming the non-analytic peak, though this improvement is slow and asymptotic.\footnote{Of course, from our discussion in \S\ref{sec:break-down-of-neural-scaling-laws}, 
we know that there eventually will be a qualitative change to this behavior when
the model size and the training set size exceed the size of the latent space, $\nf, \nA > \nl$.
}

However, from the LLM scaling law phenomenology of \cite{kaplan2020scaling} and the explicit realization of the test loss \eqref{eq:phenomenological-loss-original} by our statistical model, we 
 know that the classical ``U'' shape of the underparameterization region can be modified with proper regularization such that the test loss can be made analytic, monotonically decreasing as models become larger.\footnote{The observation that double descent can be mitigated by optimal regularization was first made by~\cite{nakkiran2021optimal}.}
The dashed curves in Fig.~\ref{fig:double-descent} -- generated from \eqref{eq:phenomenogical-model-better} -- depict this behavior, showing that
optimally regularized models exhibit monotonically decreasing test loss with model size and do not have any special behavior around the point of equiparameterization (vertical dotted lines).
Further, with this regularization it's apparent that the slow asymptotic improvement for very large models at fixed training set size is simply the plateau region of \eqref{eq:phenomenological-loss-original}.
Furthermore, comparison of the unregularized and regularized models 
-- from our numerical simulations and exhibited in the figure by the blue solid and  blue dashed curves, respectively --
we see that regularization only seems to be important around the equiparameterization peak.\footnote{
    A related point: early stopping, which can implicitly play the same role of a ridge parameter \cite{friedman2003gradient,raskutti2014early}, only helps models in the regime of near-equiparameterization \cite{nakkiran2021deep}. 
} 
From this perspective, the double descent phenomenon 
is an
artifact of not using regularization in a small region around $\nf \sim \nA$.

Thus, when using proper regularization, we can achieve the best test loss by \emph{jointly} scaling 
the sizes of the model and the size of the training set. Comparison of the two regularized (dashed) curves in  Fig.~\ref{fig:double-descent} illustrates the way 
that 
increasing model size alone 
eventually encounters a plateau, 
while increasing the size of the training set 
extends the power-law portion of the performance gains.
As originally pointed out by \cite{kaplan2020scaling}, the reason for the optimality of this joint scaling is avoiding resource bottlenecks: each additional sample in the training set is informative about one additional eigen-feature in the power-law portion of 
the spectrum of data's latent spectrum 
\eqref{eq:exact-power-law-latent-generative}, and we need an additional feature in our model to represent that eigen-feature.\footnote{
    Cf. our discussion of PCA in \S\ref{sec:data-properties}: if the latent data had a gap instead of a continuous spectrum without a natural cutoff, then the performance would not continue to improve substantially by jointly increasing model and training-set sizes.
}

As such, given finite resources and an ability to both scale models as well as gather training points, an optimal allocation involves 
a kind of joint 
near-equiparameterization
scaling: for generalized linear models such as our statistical model, there is only a single exponent, $\alpha$, that controls the test loss power-law behavior in both the number of features and size of the training set, and for this model scaling the number of features of the model to equal the size of the training set, $\nf(\nA) = \nA$, will avoid the plateau region; for 
other models
such as the LLMs discussed in \cite{kaplan2020scaling}, we may have a more general scaling relation, $\nf(\nA) \sim \nA^p$, 
as in \eqref{eq:original-tradeoff-powerlaw}.\footnote{However, recall that \cite{hoffmann2022training} gives evidence that 
the scaling for LLMs should actually be linear, $\nf(\nA) \sim \nA$, if trained sufficiently.
Note 
also 
that 
this doesn't necessarily mean that 
the number of parameters must equal the size of the training set, but only the relative ratio should always remain constant.
}
Even in this more general case, jointly scaling both training data and model size pushes the performance away from the tails of the test loss curves and back towards the termination of the power-law region -- back towards the non-analytic peak of the unregularized model
-- which is the region of large-data and large-parameter \emph{equiparameterization}. Altogether, we conclude that this regime with proper regularization -- and not the overparameterization regime -- is the practical setting of interest for deep learning.\footnote{
    Interestingly, a curated dataset of machine learning systems taken from highly-cited and highly-influential papers from 1952 to 2021 \cite{sevilla_villalobos_2021} gives strong evidence that skilled practitioners have always been implicitly 
    working in
    this jointly large-training-set-and-large-model-size equiparameterized regime: plotting the parameter counts vs. training set sizes of the models in this dataset on a log-log scale gives a linear fit with a slope extremely close to unity (see, \cite{adlam}). We thank Ben Adlam for bringing this point to our attention.
}

\subsection{The Battle of the Dimensions}
\label{sec:coding-theory}

Typically, we expect that natural data lives on a data manifold of smaller \textbf{intrinsic dimension}, $\din$, than its embedding dimension, $\nin$.
The representation that lives on the latent data manifold is supposed to be a type of encoded representation of the input, and we imagine there exists a transformation that decompresses the latent representation into the higher-dimensional embedding or input space, where, 
for example,
the pixels of an image would live. Usually we 
are interested in the
inverse of such a transformation, which can
be
used as a dimensionality-reduction technique for data.\footnote{A very simple example of such a technique is PCA.
}

However, this seems to be in tension with the equiparameterization picture for natural data discussed in the last subsection: the power-law scaling of the test loss arises from a model 
learning increasingly more eigen-features from the latent-space representation of the data. For this to work, 
the 
\textbf{latent-space dimension},
$\nl$, must be the largest scale in the problem: $\nin, \nf, \nA  \ll \nl$. 
In this picture,
rather than decompressing data, 
the transformation from the latent representation to the input representation is a type of \emph{projection},
and many samples 
are needed to reconstruct the true underlying description.

If we accept both of these pictures, then it must be the case that the intrinsic dimension, $\din$, is a different quantity than the  size of the latent space, $\nl$.

\subsubsection*{Defining and Measuring Intrinsic Dimension}

There are multiple sensible ways to estimate the intrinsic dimension, $
\din$, of a dataset \cite{CAMASTRA201626}. 
A nice method considers the typical (Euclidean) distance 
between neighboring points \cite{MLE-dim-estimate,two-NN}: if we define $x'(x_\alpha)$ to be the nearest neighbor (NN) of a sample, $x_\alpha$, then the NN-distance, $\dist_\alpha$, is defined by
\be
\dist^2_\alpha \equiv \sum_{i=1}^{\nf}  \le[x'_i(x_\alpha) - x_{i;\alpha}\ri]^2 \, ,
\ee
and the
typical distance $\distavg$ in a dataset of $\nA$ points is given by the sample average,
\be
\distavg \equiv \frac{1}{\nA}\sum_{\alpha=1}^{\nA}\dist_\alpha \,.
\ee 
So, 
if we increase the density of points,  we expect the average NN-distance to decrease:
\be\label{eq:def-intrinsic-dimension}
\frac{T}{V} \sim \distavg^{-\din} \,,
\ee
where $V$ is the volume of the space. We will use this equation to \emph{define} the intrinsic dimension $\din$:
intuitively, 
the power that relates the linear-dimensional NN-distance, $\distavg$, to the density of points $T/V$, is operationally what we mean by the dimension of a space.\footnote{
    For instance, if we place points uniformly on a Euclidean lattice, $\distavg$ would be the fixed lattice size, $\din$ would be the usual Euclidean dimension of the lattice, and the density of points would be exactly $T/V = \distavg^{-\din}$.
}
Moreover, this definition, \eqref{eq:def-intrinsic-dimension}, gives a simple way to measure $\din$: plot the scaling of the average NN-distance versus the size of the dataset and look at its slope on a log-log plot.\footnote{There are much more sophisticated ways of measuring $\din$, see, e.g., \cite{MLE-dim-estimate,two-NN,JMLR:v23:20-1111}, but \eqref{eq:def-intrinsic-dimension} is sufficient for the qualitative point we wish to make here.}

\subsubsection*{Natural (Power-Law) Data}

Now, let's apply this definition to the types of natural datasets we've considered in this paper. Recall that such datasets have power-law spectra (\S\ref{sec:data-properties}), and we expect that 
models trained on such data will have power-law scalings for their test losses when not otherwise bottlenecked by the number of parameters.
Recalling this power law, %
\eqref{eq:model-power-law-scaling-law-trainingset},
and rearranging our definition \eqref{eq:def-intrinsic-dimension},  we note curiously that both the test loss and the NN-distance are power laws in the size of the training set:
\be\label{eq:curiosity}
\distavg \sim \nA^{-1/\din}\,, \qquad  \L(\nA) \sim \nA^{-\alpha} \, .
\ee

Naively, there's no reason to relate these two quantities. However, Ref.~\cite{JMLR:v23:20-1111} argued that the test loss should be monotonically related to $\distavg$, the typical linear size of a subregion occupied by a training point: the smaller the typical region, the greater that any test point will fall near a training point and inference can succeed via interpolation, the smaller the test loss.\footnote{
     The connection between dataset size, the intrinsic dimension, and scaling laws was also made by \cite{spigler2020asymptotic}.
}
With this proposal, we relate the two equations in \eqref{eq:curiosity} and identify
\be\label{eq:dimension-is-alpha}
\din =  \frac{\#}{\alpha} \, ,
\ee
where the order-one numerical factor $\#$ can depend on the details of the manifold \cite{JMLR:v23:20-1111}.\footnote{The numerical factor suggests that $\L(\nA) = \distavg^p$, for some power $p$. 
}

\begin{figure}[ht]
\begin{center}
 \includegraphics[width=0.49\linewidth]{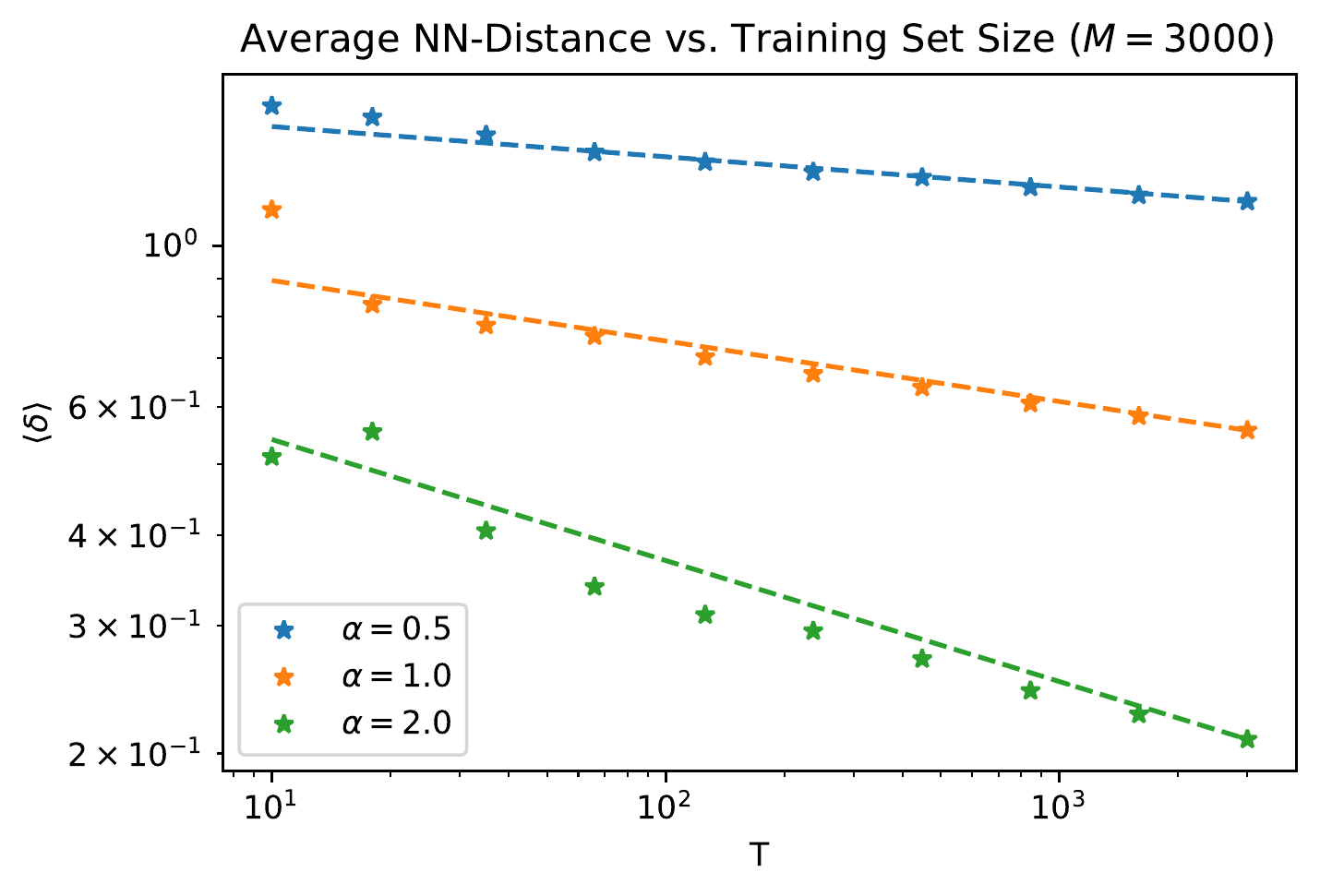}
  \includegraphics[width=0.49\linewidth]{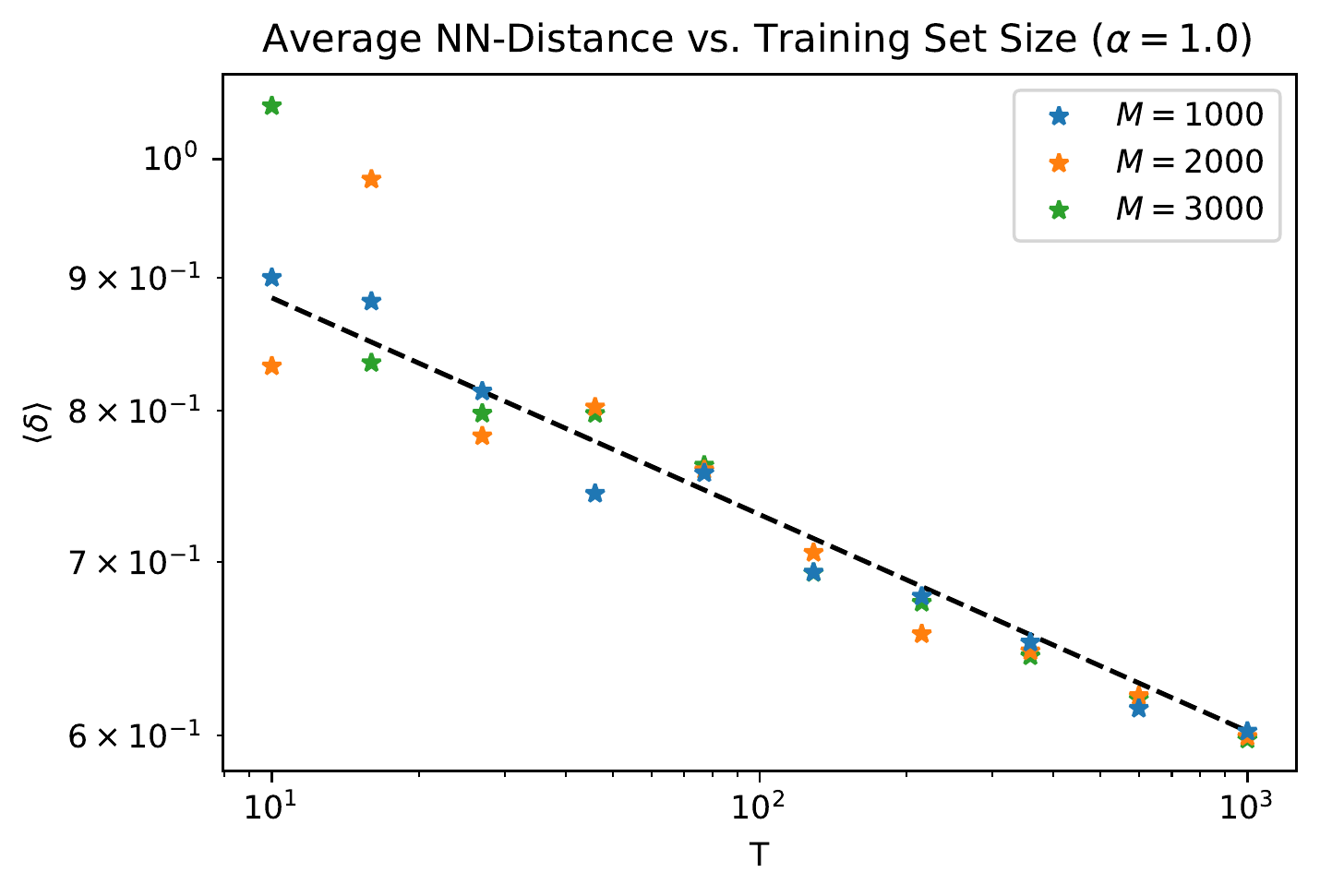}
\end{center}
\caption{Typical NN-distance from numerical simulations (stars) of our latent data generative model, with the maximum eigenvalue fixed ($\eigmax = 1$), fit (dashed lines) by a simple power law, $\distavg \sim T^{-\alpha/12}$. These experiments support the identification of the intrinsic dimension with the inverse of the power-law exponent: $\din = \#/\alpha$.
\textbf{Left:} The size of the dataset, $\nA$, is varied for a few different power-law exponents, $\alpha$, while the dimension of the latent space is held fixed ($\nl=3000$).
\textbf{Right:} The size of the dataset, $\nA$, is varied for a few different latent-space dimensions, 
while power-law exponent is held fixed ($\alpha=1.0$). These experiments suggest that $\din$ does not vary with $\nl$.
}
\label{fig:intrinsic-dimension}
\end{figure}

To verify this relationship, we can use the scaling of the NN-distance with the dataset size, \eqref{eq:def-intrinsic-dimension},  to measure $\din$ for our generative data model and see how it varies with the (hyper-)parameters of the distribution: $\alpha$ and $\nl$. In Fig.~\ref{fig:intrinsic-dimension}, we numerically simulate latent datasets of different sizes according to \eqref{eq:feature-feature-covariance-definition} and \eqref{eq:exact-power-law-latent-generative}, varying the power-law exponent $\alpha$ (left panel), and the size of the latent space $\nl$ (right panel). These experiments support the relationship \eqref{eq:dimension-is-alpha} between 
$\din$ and $\alpha$
and further suggest that $\din$  is entirely unrelated to $\nl$. Thus, it must be the case that, while both dimensions $\din$ and $\nl$ characterize 
the latent data manifold, they represent very different intrinsic properties.

To understand this better, recall our discussion of PCA in \S\ref{sec:data-properties}: when
the spectrum of the data has a gap, the few principal components above the gap are taken to span the intrinsic data manifold, the other components are thought to be random noise,
and the number of principal components gives an estimate of $\din$.\footnote{Note, 
PCA can overestimate the intrinsic dimension of manifolds with large curvature \cite{CAMASTRA201626,JMLR:v23:20-1111}.
}
However, for continuous spectra arising from natural data and modeled by our latent data generative model, there's no natural cutoff for separating informative and uninformative features. Instead, the exponent characterizing the spectrum, $\alpha$, provides a way of estimating $\din$: intuitively, if the exponent is large, $\alpha \gg 1$, then the spectrum very rapidly decreases to zero,
and much of the variance of the data is captured by the first few principal components. Thus, we might expect that there's a way to compress such data onto a manifold with small intrinsic dimension $\din \sim 1/\alpha$.\footnote{
   Note that in our work, $\alpha$ and $\nl$ are independent (hyper-)parameters of our latent data model: $\alpha$ can be any positive real number, and so we can create datasets with any $\din$.
In contrast, the scaling exponents in LLMs \cite{kaplan2020scaling} and images \cite{rosenfeld2020a,spigler2020asymptotic} are small, $\alpha<1$, but large enough such that the intrinsic dimension is smaller than the input dimension ($\din < \nin$).
 It's unclear what the interpretation would be of a natural dataset if $\alpha$ is so small such that $\din > \nin$. 
}

To conclude,
on the one hand, for power-law data
there's no natural cutoff in the spectrum,
and you can always do fractionally better by modeling additional eigen-features: this is why there are neural scaling laws.
On the other hand, 
although 
such data 
has $\nl$ independent components, the 
power-law structure is a type of constraint: the data manifold is not very typical 
among
$\nl$-dimensional manifolds, and the intrinsic dimension, $\din$, is determined by $\alpha$, not $\nl$.

\subsubsection*{Latent Features vs. Trained Features}

The argument of Ref.~\cite{JMLR:v23:20-1111} that led to the identification \eqref{eq:dimension-is-alpha} implicitly relied on not just the input data, but also the labels: in principle, only the features relevant for the underlying task will be useful for reducing the test loss. To account for that, the authors actually considered the intrinsic dimension of the final-layer activations of a trained network.

In general, we should expect our estimate of $\din$ to depend on whether we consider trained features or raw input data (or the latent features of our statistic model): if the task is designed so that the label depends on only a few of the input features, then the trained network will learn to ignore the rest of the features, and we might expect the trained features to have a much smaller intrinsic dimension. For example, consider input features 
that represent the $\nin$ pixels of a black and white image: if the label is simply whether the first pixel is on or off, then we should expect a much smaller intrinsic for the trained features 
than if the label 
had depended on correlations among the $\nin$ pixels of the input, as in a typical image classification task; %
in the former case, we might expect $\din \sim 1$, while in the latter case, if there's $\nout \ll \nin$ classification categories, we might instead expect $\din \sim \nout$.

In our statistical model, there are two reasons that make this distinction between latent features and trained features unimportant: 
\emph{(1)} 
our labels, $y$, depend on \emph{all} 
latent features of an input, $x$, cf. \eqref{eq:def-teacher-labels};
and 
\emph{(2)} 
our model uses \emph{random} features -- not trained features -- 
and 
those random features cannot depend on any label information.\footnote{Additionally, we expect that the $\nf$ random features would have a similar intrinsic dimension as 
the $\nl$
latent features, 
given experiment in the right panel of Fig.~\ref{fig:intrinsic-dimension}. 
showing that the size of the latent space doesn't affect the intrinsic dimension.
}
This first point is necessary to see the full set of observed neural scaling law phenomenology -- the power-law and plateau regions -- and so %
this suggests it's an important
property
of self-supervised generative modeling tasks, such as those that LLMs perform. 
As to the second point, it would be interesting to 
find
a statistical model for which the features can learn nontrivial representations of the inputs based on the tasks. We discuss some ideas towards this end at the end of \S\ref{sec:future-directions}.

\subsection{The Breakdown of our Data Model}
\label{sec:breakdown-data-model}

The main goal of this paper has been to show how a simple generative data model combined with a simple random feature linear regression model could capture the rich neural scaling
phenomenology observed in a broad class of real world large-scale models.
An essential part of our data model was that the spectrum of the input data's covariance matrix had many decades of power-law eigenvalues: when we jointly increased the number of random features or training samples, we were able to see a larger and larger fraction of the underlying latent eigenvalues in the empirical covariance matrix; this ultimately led to a power-law scaling of the test loss when neither features nor training data were constrained.
If the data model were any simpler -- e.g., that purely Gaussian data with an isotropic spectrum that we study in \S\ref{sec:linear-mp-model} -- we would not find power-law scaling of the test loss. 
However, there are certain limitations to the simplicity of the data model, which we will discuss in the rest of this subsection.

In particular, one might wonder whether we can use our random feature and generative data model from \S\ref{sec:derivation}
as a model for \emph{input data}:
\bi
\item If we take a dataset sampled from this random feature model \eqref{eq:statistical-model-random-feature-map} and interpret the size of the random feature space as the data's input dimension ($\nf \to \nin$), and if we then act on the dataset with a \emph{nonlinear} random feature map, e.g. \eqref{eq:simplest-feature-map-reprint}, with dimension $\nR > \nf$, will the power law of the feature spectrum extend? %
\item If the spectrum does extend, will it also extend the power law in the test loss?
\ei
To answer these questions, we can numerically study our data model with a single-hidden-layer ReLU network and compare against a natural dataset.

\begin{figure}[ht]
\begin{center}
 \includegraphics[width=0.49\linewidth]{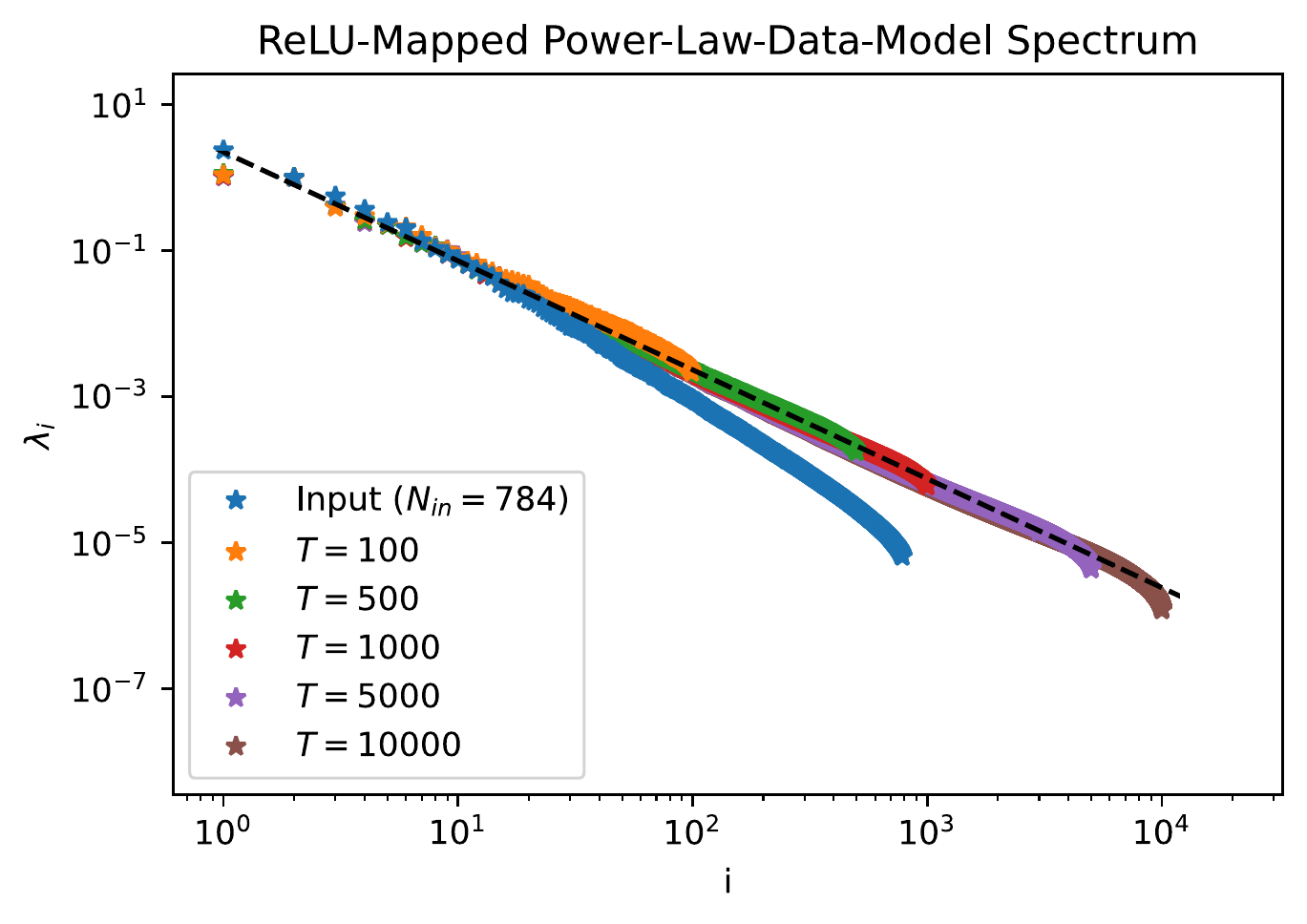}
 \includegraphics[width=0.49\linewidth]{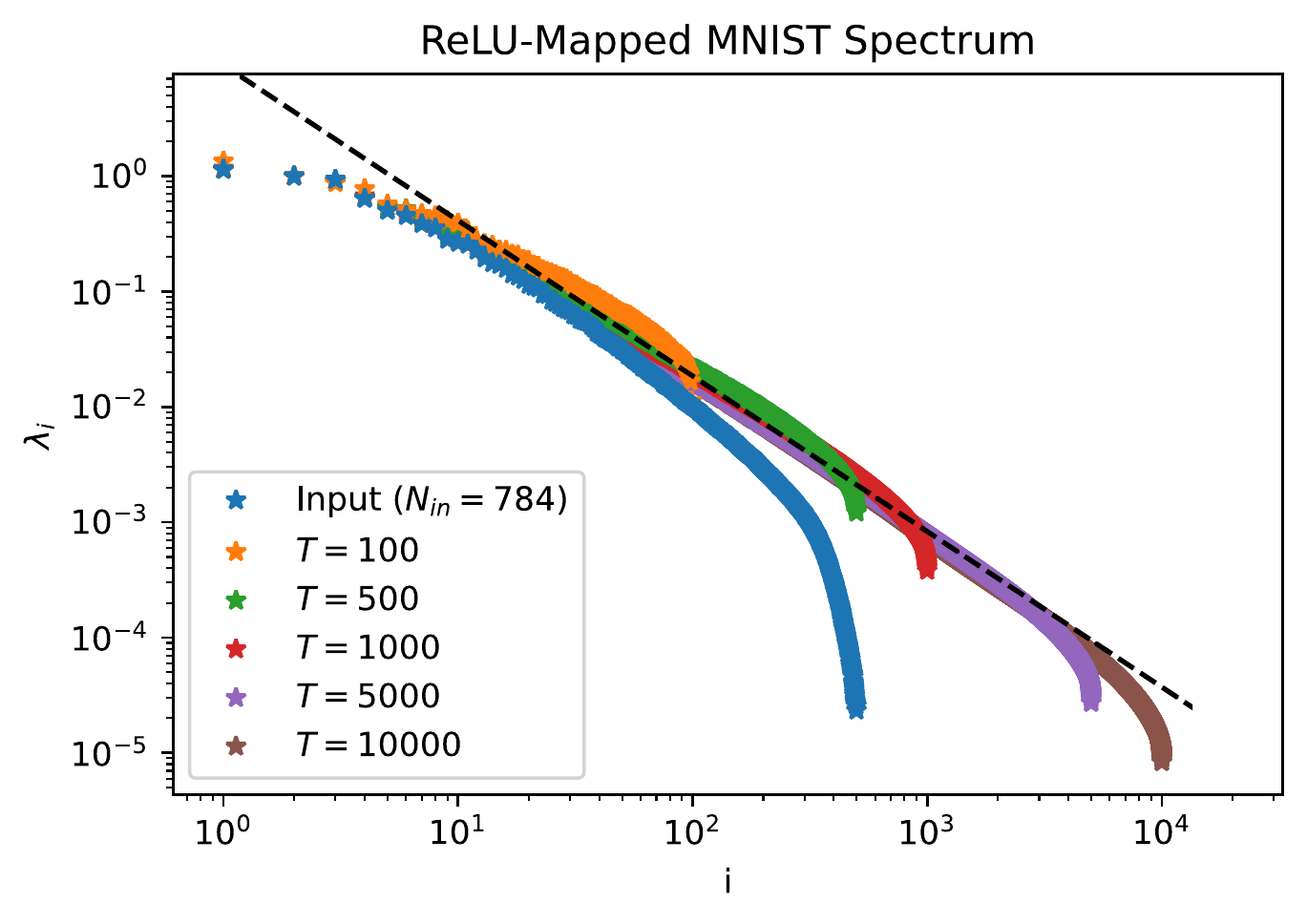} 
\end{center}
\caption{Spectra of the activations of a single-hidden-layer random ReLU network obtained by diagonalizing the 
data-data covariance matrices, $\covAA$, defined in \eqref{eq:covariance-relu}, which can be thought of as an
infinite-dimensional feature model ($\nR \to \infty$). In both panels, we see that the power-law portion of the spectrum extends beyond $\nin$ as we increase the dataset size, $\nA$; we also see in both cases that passing the data through the ReLU feature map \emph{decreases} the exponent $\alpha$ of the power law fit.
(Note also that
the spectra have  been rescaled by their largest eigenvalue so that the curves lie on top of one another.)
\textbf{Left:} The input dataset is obtained by sampling latent data points, $x_{I;\alpha}$, with statistics \eqref{eq:feature-feature-covariance-definition} and power-law eigenvalues \eqref{eq:exact-power-law-latent-generative}  ($\nl = 20000,\alpha=1, \eigmax=1$) and then subsampling these latent features with the linear map \eqref{eq:statistical-model-random-feature-map} ($\fwvar =1$) to obtain input features with 
 $\nin = 784$.
\textbf{Right:} The input dataset is obtained from samples of  MNIST dataset \cite{mnist-dataset},  a CV dataset of  natural black and white images with input dimension $\nin = 784$. 
}
\label{fig:extended-power-law-mnist-gauss}
\end{figure}

 In Fig.~\ref{fig:extended-power-law-mnist-gauss} we answer this first question. In these plots we show the spectrum of random features 
 for our data model
 (left panel) with the input features given by an explicit power law and 
 for a common computer vision (CV) dataset  (right panel). To generate these plots, we used the explicit form of the single-layer-ReLU-network kernel, \eqref{eq:covariance-relu} and then explicitly diagonalized the kernel matrix. Recalling that such a kernel
 can be thought of as an infinite-dimensional feature model ($\nR \to \infty$), we can thus study a fixed set of input features $\nin = 784$ and see how the spectrum extends as we vary the size of the dataset, $\nA$. As we see from the figure, the answer to our first bulleted point above is \emph{yes}: our data model behaves identically to the natural CV dataset. In particular, since the number of features is implicitly infinite and not a bottleneck, we see that the extent of the power law is determined by the size of the limiting resource $\nA$.
This property is quite universal, and we have observed it across a wide range of natural datasets and activation functions.

In both panels of this figure, we also see another interesting effect, one that was already pointed out in footnote~\ref{footnote-exponent-decrease} and present in the right panel of Fig.~\ref{fig:extend-linear-layer}: when data with a power-law spectrum passes through the ReLU layer -- for both our data model and natural datasets -- the exponent $\alpha$ seems to decrease slightly. Since this exponent also characterizes the power-law behavior of the model's test loss, this decrease ultimately leads to worse performance for a given set of resources than if it were otherwise preserved. It would be interesting to use our RMT tools, e.g. as discussed in \S\ref{sec:spectral-extension}, to try to understand the origin of this effect.

\begin{figure}
\begin{center}
\includegraphics[width=0.49\linewidth]{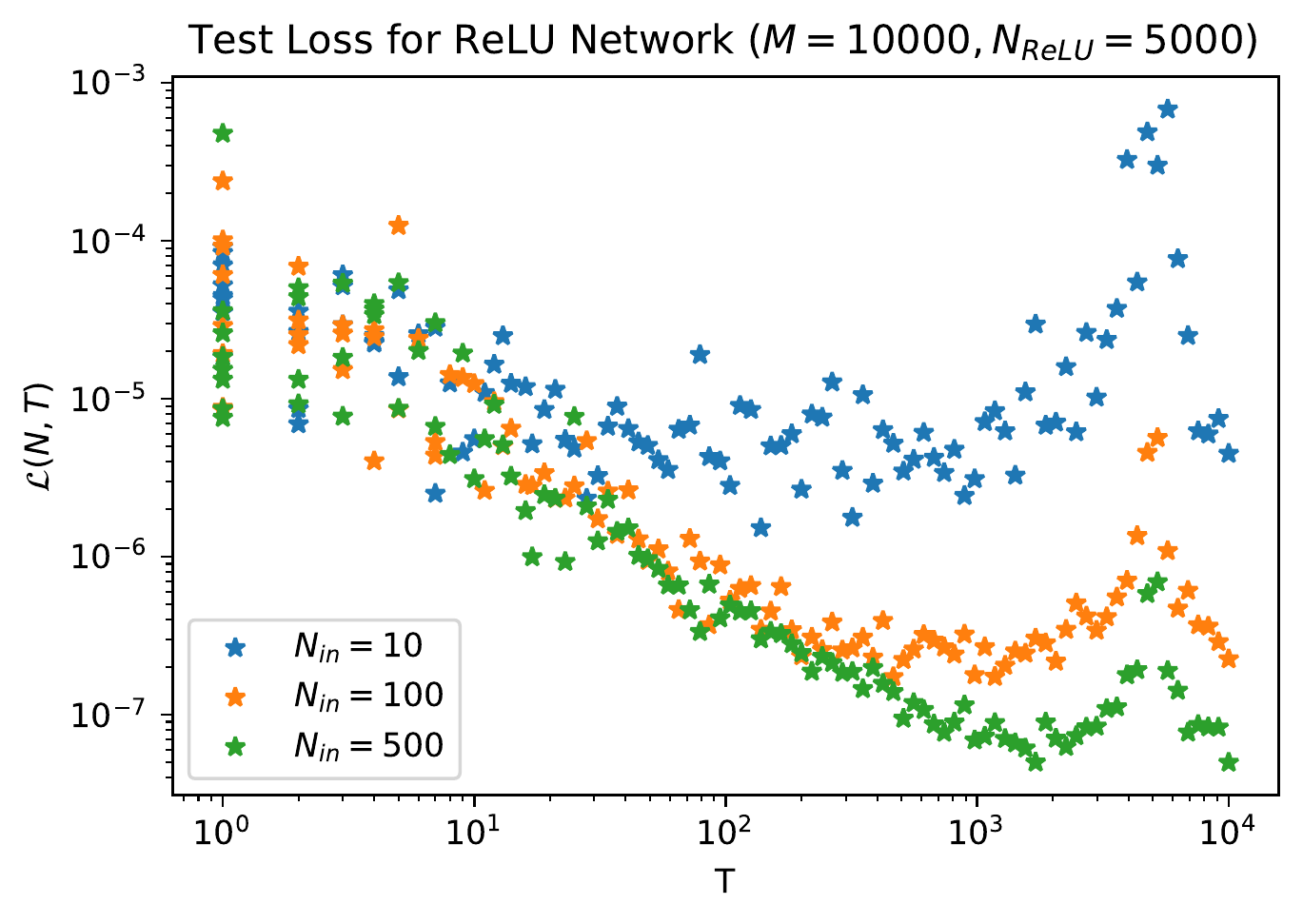} 
\includegraphics[width=0.49\linewidth]{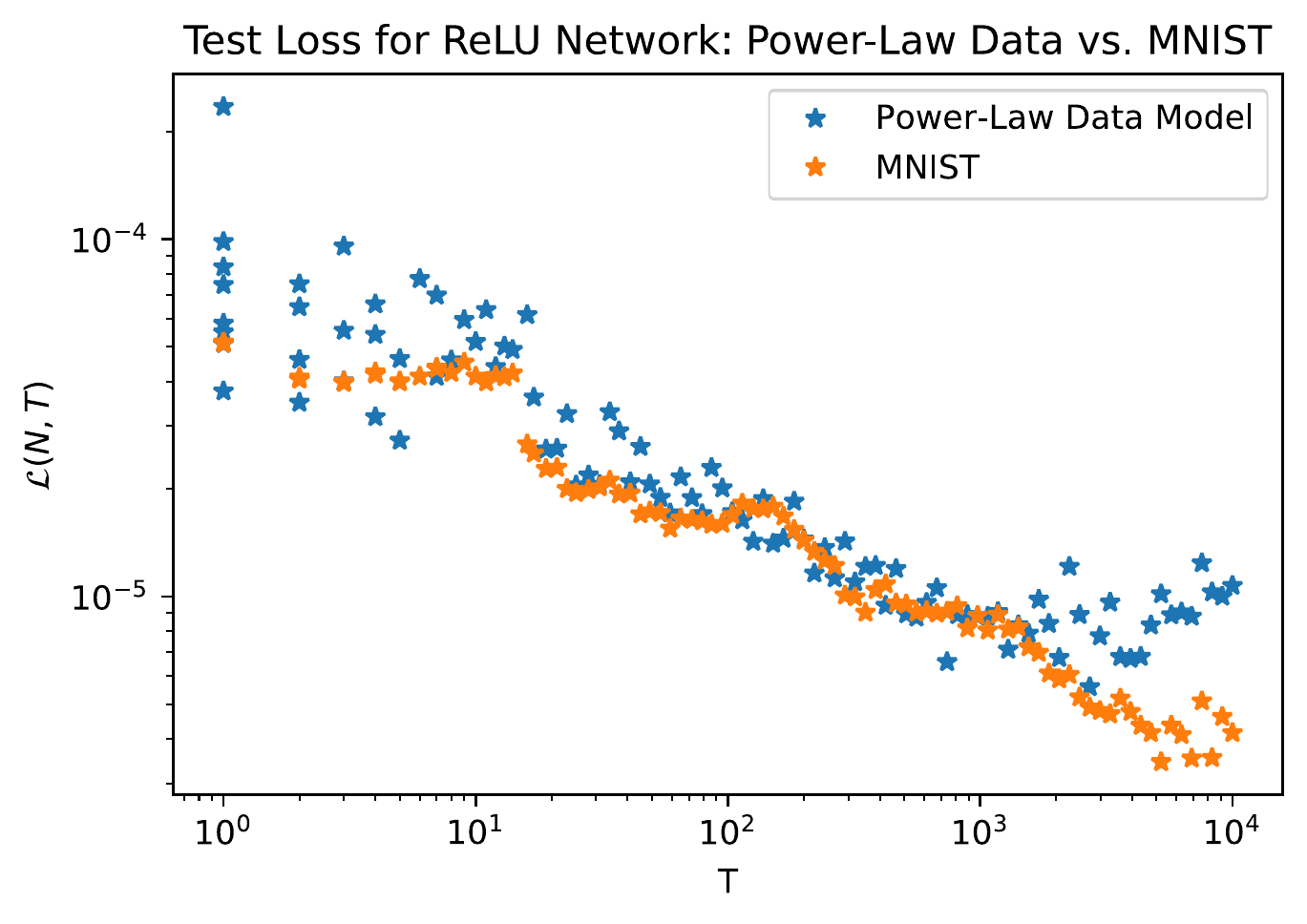} 
\end{center}
\caption{Model performance for our data model, with a comparison to MNIST, when the random features are the activations of a single-hidden-layer ReLU network of width $\nR > \nin$.
\textbf{Left:} 
Test loss from numerical simulations of our data model with $\nl = 10000$, $\alpha = 1.5$, and $\eigmax = 1$  as a function of training set size $\nA$: first, we generate latent data samples, $x_{I;\alpha}$, with statistics \eqref{eq:feature-feature-covariance-definition} and power-law eigenvalues \eqref{eq:exact-power-law-latent-generative} 
and subsample them with the linear map \eqref{eq:statistical-model-random-feature-map}, with $\fwvar = 1$, to obtain input features with different
$\nin$; then, we pass the inputs through a single-layer random ReLU network of width $\nR = 5000$; finally, we compute the test loss via \eqref{eq:def-test-loss} with our generated labels \eqref{eq:def-teacher-labels}.
We see that the power-law scaling ends and the loss plateaus after the size of the training set, $\nA$, reaches the number of input features, $\nin$, rather than the number of ReLU features, $\nR$.
\textbf{Right:}
We compare our data model (blue) with $\nl=20000$, $\alpha=0.4$, $\eigmax=1$, $\fwvar = 1$, and input features of size $\nin=784$, against a simplified MNIST classification problem (orange), where we only include \texttt{7}s and \texttt{9}s. In both cases, we pass the data
through a single-hidden-layer ReLU network of width $\nf=20000$ and compute the test loss via \eqref{eq:def-test-loss}. 
We see that while our data model plateaus for training set sizes $\nA$ greater than $\nin$, the MNIST performance continues to follow a power-law falloff up to higher scales. (Note:  the MNIST loss is rescaled to lay on top of the power-law data's loss, and the exponent of our data model was chosen to match that of MNIST in order to facilitate the comparison.)
}
\label{fig:relu-gaussian-mnist-gauss}
\end{figure}

Next, in Fig.~\ref{fig:relu-gaussian-mnist-gauss}, we answer the 
second bulleted question above. 
To generate plots, we map the $\nin$ input features through a single-hidden-layer ReLU network with a much larger number of random features ($\nR > \nin$), and then we train a linear regression model on these random features and plot the test loss. In the left panel, we see the results for our random-feature-and-label model from \S\ref{sec:derivation-notation}. Note that the performance plateaus as soon as the number of training samples, $\nA$, surpasses the size of the input dimension, $\nin$, rather than at the much larger random-feature dimension, $\nR$: in other words, our random feature model interpreted as a data model does not itself give a power-law scaling of the test loss when the spectral extension is due to a nonlinear network. This means that
the answer to our second bulleted point above is \emph{no}:
 despite the fact that the spectrum extends when passed through the ReLU (cf. Fig.~\ref{fig:extended-power-law-mnist-gauss}), this extension is simply not enough to translate to extending the power law scaling of the test loss.

 For the sake of comparison, in the right panel of Fig.~\ref{fig:relu-gaussian-mnist-gauss}  we compare to the natural CV dataset trained on a simplified classification problem. We can see clearly that, while our model has a plateau at $\nin$, the natural dataset extends the power-law scaling of the test loss past $\nin$, indicating that the spectral extension \emph{was} translated to the test loss.\footnote{N.B. in this plot, 
 the particular
 maximum value of $\nA$
 was chosen 
 because
 there are only 12,214 images in the training set of MNIST for the ``7'' and ``9'' digits; we expect that the power-law (orange curve) would continue further if  more training samples were available.}

A heuristic explanation for this discrepancy is clear: when we down-sample the latent space to generate the random features in \eqref{eq:statistical-model-random-feature-map}, we are throwing away many of the latent eigenvalues \emph{and eigenvectors} present in the latent data $x_{I}$ necessary to predicting the label $y_{I}$, cf. \eqref{eq:def-teacher-labels}. 
While the nonlinear feature map \eqref{eq:simplest-feature-map-reprint} does extend the power-law eigenvalues (cf. Fig.~\ref{fig:extended-power-law-mnist-gauss}), there is no reason to expect these nonlinear features to be aligned with the latent eigenvectors as they were determined at random. In other words, this information was fundamentally lost when our down-sampling projected out these directions. Hence, the $\nin$ inputs do not generate any more useful features when passed through a ReLU network.

However, unlike our generative data model,
natural datasets \emph{do} seem to contain sufficient information in their limited input features to determine the outputs: seemingly, very little has been thrown away! Thus, it would be interesting to extend our minimal model to understand further how the nonlinear activation functions %
are able to
decode relevant eigenvectors and extend the power-law regime in the test loss.\footnote{Note that Ref.~\cite{wei2022more} has a toy model of scaling laws that incorporates an eigenvector dependence.}

\section{Conclusion and Featured Directions}
\label{sec:future-directions}

In this paper we presented a solvable model of neural scaling laws. After exploring natural datasets and their behavior under nonlinear feature maps (\S\ref{sec:overview}), we constructed a \emph{minimal} statistical model that contains the full set of neural scaling behavior exhibited by large language models in practice (\S\ref{sec:derivation}).
We also explained why our model is minimal by showing that even simpler data or feature models (\S\ref{sec:other-models}) do not have the right behavior.
By solving our statistical model using techniques from random matrix theory (\S\ref{sec:data-and-feature-averages}), we were able to better understand the mechanism driving these practically relevant neural scaling laws.

With our solution, we were then able to explore a regime of our model $(\nl < \nf, \nA)$ that goes beyond the regimes explored experimentally for LLMs (\S\ref{sec:break-down-of-neural-scaling-laws}). This is a nice benefit of theoretical analyses such as ours; in addition to trying to understand experimentally observed phenomena in contexts that we've accessed experimentally, we can also use our models to probe new contexts. Relatedly, now we can ask the experimentalists, can we measure the scale $\nl$ for LLMs? Perhaps we can use the fact that the prediction of the test loss fit is better when you include $\nl$, as per our analysis, cf. \eqref{eq:phenomenogical-model-better}? Knowledge of the size of the latent space for language would have practical benefit by elucidating, in some sense,  how much better LLMs can get using power-law performance gains before transitioning into a noise-dominated regime.

Finally, we hope that the large-$\nf$ diagrammatic approach to RMT calculations (\S\ref{sec:data-and-feature-averages}) and the differential operator approach to simplifying expressions (\S\ref{sec:other-models}) will be useful theoretical tools for simplifying machine learning calculations.

\sbreak

\noindent{}We close by featuring a future direction.

\subsubsection*{Representation Learning?}

In the paper, we have studied generalized linear regression models with random feature maps, and we have seen that the ability of such maps to extend the power law present in the spectrum of input data is the mechanism that leads to the power law scaling of the test loss as the number of random features increases. A natural extension of this analysis is to move beyond linear regression with random feature maps and consider models that \emph{learn} representations of their inputs. A natural setting for this is the \textbf{quadratic models} framework of \cite{PDLT-2022}, as they are  minimal models of representation learning.\footnote{
    As we discussed in our \emph{\emph{Aside:} Random Feature Maps vs. DNNs} in \S\ref{sec:data-properties}, certain DNNs in the infinite-width limit correspond to linear regression models. In contrast, nonlinear models, of which \emph{quadratic models} are the simplest example, can instead correspond to more realistic finite-width networks \cite{PDLT-2022}.
}

A quadratic model extends the generalized linear model, \eqref{eq:linear-model-def}, in the following way:
\be\label{eq:quadratic-model}
      z(x;\param) \equiv \sum_{j=1}^{\nf} \param_{j} \fea_{j}(x) + \frac{\qeps}{2} \sum_{j_1, j_2=1}^{\nf}  \param_{j_1}\param_{j_2} \featwo_{j_1j_2}(x)   \, ,
\ee
where each 
$\featwo_{j_1j_2}(x)$ is a \textbf{meta feature function}, which is a (potentially) nonlinear transformation of an input $x$ that is distinct from the feature functions $\fea_{j}(x)$, $\qeps$ is an adjustable hyperparameter that controls the importance of the quadratic deformation of the linear model, 
and, for convenience, we assume that the model outputs a scalar $z$. Due to the constraint tying the parameters in the linear term, $\param_{j}$, to the product of parameters in the quadratic term, $\param_{j_1}\param_{j_2}$, the quadratic model
cannot be written in the form of a generalized linear model \eqref{eq:linear-model-def}; it is this nonlinear-in-the-parameters property that enables it to learn nontrivial representations of inputs $x$ over the course of training.\footnote{See \S11.4 of \cite{PDLT-2022} for the details of how this works.} 

Following \S11.4  of \cite{PDLT-2022}, if we optimize the MSE loss, \eqref{eq:def-mse-loss},  with the quadratic model, 
\eqref{eq:quadratic-model}, 
we can find a solution in perturbative expansion in the deformation parameter $\qeps$:
\be\label{eq:quadratic-model-solution-decomposition}
\param^\star \equiv \paramL + \qeps\,\paramNL \, ,
\ee
where
the \emph{linear} part, $\paramL$, is the linear regression solution with $\qeps = 0$, \eqref{eq:linear-regression-solution}, and the \emph{nonlinear} part is given in terms of the linear solution to leading order as
\be
\paramNL_{j} = - \frac{1}{2}\sum_{j_1,j_2,j_3=0}^{\nf} \paramL_{j_1} \paramL_{j_2}  \le[ \sum_{\alpha=1}^{\nA}  \featwo_{j_1 j_2}(x_\alpha) \,\fea_{j_3;\alpha}\ri] \res_{j_3 j} + \o{\qeps}\, ,
\ee
and would have a more complicated expression were we to substitute in $\paramL$. With this solution, test predictions can similarly be decomposed as
\be
\widehat{z}^\star \equiv \zL + \qeps\, \zNL\, ,
\ee
with $\zL$ the standard \emph{linear} regression inference, \eqref{eq:test-predictions}, and $\zNL$ the  correction coming from the \emph{nonlinear} quadratic term.

Now to the point: if we
substitute these predictions into the test loss,
\begin{align}\label{eq:quadratic-model-test-error}
    \L_\B(\param^\star) &= \frac{1}{2 \nB }\norm{\widehat{z}^\star-\widehat{y}}^2 \,, %
\end{align}
we can try to evaluate the performance of the quadratic model given a particular data model. 
In particular,  
we can evaluate if the nonlinear contribution -- and therefore if nontrivial representation learning -- improves the model's performance
by computing
\be\label{eq:dervative-of-quadratic-model-test-error}
\frac{d\L_\B(\param^\star)}{d\qeps} =  \frac{1}{\nB}\sum_{\beta = 1}^{\nB}\zNL_{\beta} \le(\zL_\beta - \widehat{y}_\beta\ri)  + \o{\qeps} \, .
\ee
This quantity captures the linear shift in the model's performance due to the nonlinear deformation; here we see that it depends on how nonlinear part of the test prediction, $\zNL_{\beta}$, aligns with the linear part of the prediction error, $\le(\zL- \widehat{y}\ri)$, when averaged over the test set.
If these quantities are anti-aligned for a given data model,
\be
\frac{d\L_\B(\param^\star)}{d\qeps} < 0 \, ,
\ee
then we can conclude that the quadratic model
improves the performance on that data, at least for small enough values of $\qeps$.\footnote{
    If we wanted to go further and find the optimal 
    value of the hyperparameter $\qeps= \qeps^\star$, 
    we'd need to compute 
    $\param^\star$
    to order $\qeps^2$ in order to get a nontrivial equation 
    when solving $d\L_\B(\param^\star)/ d\qeps = 0$.
} 
The qualification ``for a given data model'' is important, as it's easy to construct simple datasets that give either sign for $d\L_\B(\param^\star)/ d\qeps$!

To assess a realistic setting of interest, we can use our generative data model from \S\ref{sec:derivation}, modeling the data 
with a power law covariance spectrum. Then, we can use our RMT tools to find the average of \eqref{eq:dervative-of-quadratic-model-test-error} over the data and random features.\footnote{To do so, we would need to determine the statistics of $\featwo$;
    as per (11.152) in \cite{PDLT-2022}, $\featwo$ should have a nontrivial cross correlation with $\fea$. (N.B. the hyperparameter $\qeps$ is defined differently in that reference.)
} 
Altogether this may lead to a more complex, if not even more realistic, model of neural scaling laws: the form of the quadratic model, \eqref{eq:quadratic-model}, breaks the data-parameter duality 
in the average test loss; perhaps because of this we may find a much richer phenomenology for test loss, e.g. unequal scaling exponents $\alpha_{\nf}, \alpha_{\nA}$ as was originally observed in \cite{kaplan2020scaling}.

\section*{Acknowledgments}

We are grateful to Ben Adlam, Yasaman Bahri, Ethan Dyer, Sam Gershman, Ekdeep Lubana, Eva Silverstein, Hidenori Tanaka, Lechao Xiao for discussions. We are especially grateful to Sho Yaida for collaboration in the early stages of this project.
A.M. is supported in part by the Simons Foundation Grant No. 385602 and the Natural Sciences and Engineering Research Council of Canada (NSERC), funding reference number SAPIN/00047-2020.  
D.R. acknowledges support from the National Science Foundation under Cooperative Agreement PHY-2019786
(the NSF AI Institute for Artificial Intelligence and Fundamental Interactions, \url{http://
iaifi.org/}), and would also like to announce the birth of Simon
 ``large-$N$'' Roberts.
This paper has been brought to you by the letter $N$, in the planar limit after averaging over many realizations.

\appendix

\section{Linear Models}
\label{sec:other-models}

In 
\S\ref{sec:derivation},
we studied a joint generative data model and random feature model that minimally captured the rich neural scaling phenomenology present in the test loss of LLMs.
In this appendix, we will study some even simpler models that capture some, but not all, of that behavior. These models 
are instructive in explaining the origin of different parts of the test-loss curve by 
highlighting the different
properties 
of a statistical model 
that lead to their absence or presence.
Although these models have been described in detail in literature, here we will 
attempt to give slightly different derivations in order to connect to the methods used 
the main body of this paper
as well as 
to illustrate the simplicity of our approach.

In particular, in this appendix we will study (non-generalized) linear models for different generative data models. This means that we will learn a model of the form
\be\label{eq:non-general-linear-model-def}
     z_{i}(x;\param) \equiv \sum_{j=1}^{\nf} \param_{ij} x_j \, ,
\ee
where $\param \equiv \param_{ij}$ is a set of learnable parameters, and the input data $x$ lives in an $\nf$-dimensional feature space
with statistics
\be\label{eq:feature-feature-covariance-definition-appendix}
    \expval{x_j} = 0 \, , \qquad \expval{x_{j_1} x_{j_1}} = \covl_{j_1j_2} \, .
\ee
Our analysis will proceed with general covariance matrices $\covl$, and we will later evaluate our solution for different data models.
Analogously, our labels are defined as
\be\label{eq:def-teacher-labels-appendix}
    y_{i} = \sum_{j=1}^{\nf} \w_{ij}x_j , \quad \text{with} \quad i= 1, \ldots, \nout\, , 
\ee
where $\w \equiv \w_{ij}$ is an $\nout$-by-$\nf$-dimensional weight matrix
whose elements we will take to be
independent and drawn from a zero-mean Gaussian distribution:  
\be\label{eq:teacher-weights-statistics-appendix}
    \expval{\w_{ij}} = 0\,, \qquad \expval{\w_{i_1j_1} \w_{i_2j_2}} = \frac{\wvar}{\nf} \delta_{i_1i_2} \delta_{j_1j_2} \, .
\ee
Comparing \eqref{eq:non-general-linear-model-def} and the rest of this setup with the generalized linear model, \eqref{eq:linear-model-def}, in the main text, the main difference here is that there's no separate latent feature space, $\nl \to \nf$, and no random feature model, $\fea(x) \to x$.
Importantly, unlike the labels in the main text, \eqref{eq:def-teacher-labels}, here the complexity of the label, \eqref{eq:def-teacher-labels-appendix}, increases as we increase the number of features in the linear model: this means that we have no way to increase the capacity of our model without also making the task harder. %
On the plus side, the analysis will be much simpler since we will no longer have to average over random features, and the only nontrivial average will be over the training set $x$.

Using the same linear regression setup and same notational conventions discussed in \S\ref{sec:derivation-notation}, we can write down a training loss with a ridge parameter,
\begin{align}\label{eq:def-mse-lossa-appendix}
    \L_\A(\param) &\equiv  \frac{1}{2}\norm{\param x-y - \noise}^2 + \frac{\ridge}{2}\norm{\param}^2\,, \notag
\end{align}
where the random noise is defined the same as before \eqref{eq:def-noise-statistics},
\be\label{eq:def-noise-statistics-reprint}
\expval{\noise_{i;\alpha}} = 0\,, \qquad \expval{\noise_{i_1;\alpha_1} \noise_{i_2;\alpha_2}} = \vare \delta_{i_1i_2} \delta_{\alpha_1\alpha_2}  \, ,
\ee
and straight away use the well-known linear-regression solution for the parameters \eqref{eq:linear-regression-solution}:
\begin{align}
    \theta^\star &\equiv    (y+\noise)x^T\res \, ,
\label{eq:linear-regression-solution-appendix}
\end{align}
with the feature-feature resolvent now defined as
\be\label{eq:resolvent-feature-feature-appendix}
    \res\equiv   \frac{1}{ \ridge \IN + x x^T } \, .%
\ee
Finally, using this solution for inference on our test set $\widehat{x}$ as
\be\label{eq:test-predictions-appendix}
\widehat{z}^\star \equiv \param^\star \widehat{x},
\ee
we define the test loss,
\begin{align}\label{eq:def-test-loss-appendix}
    \L_\B(\param^\star) \equiv \frac{1}{2 \nB }\norm{\widehat{z}^\star-\widehat{y}}^2 =\frac{1}{2 \nB }\norm{(\w x +\noise)x^T\res  \widehat{x}-\w \widehat{x} }^2 
    \, , 
\end{align}
where on the final expression we've substituted in the test predictions, \eqref{eq:test-predictions-appendix}, the optimal parameters \eqref{eq:linear-regression-solution-appendix},  and the labels, \eqref{eq:def-teacher-labels-appendix}.
Our goal now is to average $\L_\B(\param^\star)$ over the training inputs $x$, the test inputs $\widehat x$, the label weights, $\w$, and the random noise $\noise$. This calculation is analogous to the one studied in Refs.~\cite{bordelon2020spectrum,Canatar:etal} via replica methods.

The first part of the calculation proceeds identically to the first steps in \S\ref{sec:derivation-notation} under the subheading \emph{Average Goals}: since they are both quadratic, the average over the random noise, cf. \eqref{eq:test-loss-noise-average}, and the subsequent average over the label weights, cf. \eqref{eq:test-loss-averaged-noise-weights}, can be performed trivially. Thus, our nontrivial point of departure is the noise-and-label-averaged test loss expression:
\begin{align}\label{eq:appendix-test-loss-point-of-departure}
    \expval{\L_\B(\param^\star)}_{\noise,\w} &=\frac{\nout\wvar}{2 \nB \nf}\norm{x x^T\res  \widehat{x}-  \widehat{x}}^2 + \frac{\nout \vare }{2 \nB }\norm{x^T\res  \widehat{x}}^2 \, ,
\end{align}
which we again will refer to as the \emph{label term} and \emph{noise term}, respectively. As in the main text, from this point forward we will set $\nout = 1$ as it's just a trivial rescaling.

Now, we will need to recall our simple identities, \eqref{eq:identity-squared} and \eqref{eq:identity-inverse}, which we will reprint here for convenience:
\be\label{eq:identities-q-reprint}
    (\ridge \IN + x x^T ) \res = \IN \, , \qquad \res(\ridge)^2 = - \frac{\partial}{\partial \ridge} \res(\ridge) \, .
\ee
Rearranging the first identity, note that we can rewrite the label term as
\be
\norm{x x^T\res  \widehat{x}-  \widehat{x}}^2 = \ridge^2 \norm{q \widehat{x}}^2 \,,
\ee
which eliminates the explicit dependence on $x$.
Next, using \eqref{eq:feature-feature-covariance-definition-appendix}, let's perform the average over the test set. Since both terms are quadratic in $\widehat{x}$, and remembering the test set and training set don't correlate since each sample is drawn independently, we find
\be
\expval{\L_\B(\param^\star)}_{\noise,\w, \widehat x} = \frac{\wvar \ridge^2}{2  \nf} \tr{ \covl \res^2} + \frac{ \vare }{2  }\Big(\tr { \covl \res} - \ridge\tr { \covl \res^2}\Big) \, ,
\ee
where to simplify the noise term we again made use of a rearrangement of the first identity in \eqref{eq:identities-q-reprint} to eliminate $x x^T$. 
While in principle the training set average of $\tr {\res^2  \covl }$ would be complicated given the $\res^2$, luckily we have our second identity in \eqref{eq:identities-q-reprint} 
which lets us 
write the test loss as
\begin{align}
    \expval{\L_\B(\param^\star)}_{\noise,\w,\widehat{x}} &=\frac{1}{2} \left[ -\frac{\wvar}{\nf} \ridge^2\frac{\partial}{\partial \ridge} + \vare \left(1+\ridge \frac{\partial}{\partial \ridge} \right) \right] \tr{\covl \res} \, .
\end{align}
This is determined entirely by the training set average of a single quantity: $\tr{\covl \res}$. 

Now, since this expression is linear in $\res$,  we can ``take'' the trainset average by making the replacement $\res \to \resb$.
Doing so, 
we immediately recognize the resulting quantity,
\be\label{eq:delta-self-consistent-reprint}
\tr{\covl \resb} \equiv  \secG(\nA,\nf) = \tr{\frac{\covl}{ \ridge \IN + \frac{\nA \covl}{1+\secG(\nA,\nf)} } } \,,
\ee 
as the nontrivial object
that appears in the coupled set of equations that determine $\resb$. Here, as the right-most expression, we've written the self-consistent equation satisfied by $\secG$, cf. \eqref{eq:self-consistent}.\footnote{
    The calculation that gives the self-consistent equation for $\secG(\nA,\nf)$ follows identically to the calculation in \S\ref{sec:the-noise-term}, with the following substitutions: $\fea \to x$ and $\covf \to \covl$. These reflect the fact that \emph{this} resolvent is built from $x x^T$, cf. \eqref{eq:resolvent-feature-feature-appendix}, and has statistics \eqref{eq:feature-feature-covariance-definition-appendix}, respectively.
}
This means that we can express our final answer as a differential operator acting on this quantity: 
\begin{align}\label{eq:result-appendix-a}
    \expval{\L_\B(\param^\star)}_{\noise,\w,\widehat{x},x} &=\frac{1}{2 } \left[ -\frac{\wvar}{\nf} \ridge^2\frac{\partial}{\partial \ridge} + \vare \left(1+\ridge\frac{\partial}{\partial \ridge} \right) \right] \secG \, .
\end{align}
Finally, in the ridgeless limit, $\ridge\to 0$, we can expand $\secG$ as
\be\label{eq:Delta-expansion-ansatz-prepreint}
\secG(\nA,\nf) \equiv \frac{\secm}{\ridge} + \secG_{0} + \secG_{1} \ridge + \dots \, ,
\ee
cf. \eqref{eq:Delta-expansion-ansatz} in Appendix~\ref{sec:delta}, and associate these differentials to specific terms in the expansion:
\begin{align}
   \lim_{\ridge \rightarrow 0} \le(-\ridge^2\frac{\partial}{\partial \ridge} \ri) \secG(\nA,\nf)  = \secm \,, \qquad \lim_{\ridge \rightarrow 0} \left(1+\frac{\partial}{\partial \ridge} \right) \secG(\nA,\nf)  = \secG_{0} \, . 
\end{align}
Using these identities, the test loss takes an extremely simple form
\begin{align}\label{eq:linear-model-averaged-test-loss}
    \expval{\L_\B(\param^\star)}_{\noise,\w,\widehat{x},x} &=\frac{1}{2 } \left[ \frac{\wvar}{\nf} \secm + \vare \secG_{0} \right] \, ,
\end{align}
and $\secm$ and $\secG_{0}$ can be very simply determined given an underlying data spectrum.

On the one hand, this solution for the linear model in the ridgeless limit is similar to the random feature model in the main text: in that case, the noise term was proportional to $\secF_0\equiv\secF_0(\nA,\nf)$, cf. \eqref{eq:second-to-final-noise-term}, and the label term was proportional to  $\secG_{-1}(\nf,\nl)$ or $\secG_{-1}(\nA,\nl)$ depending on whether the model was under- or overparameterized, cf. \eqref{eq:statistical-model-answer}. On the other hand, a key difference is the lack of separation of scales between the number of random features, $\nf$, and the number of latent features, $\nl$: 
this means that 
there's no
duality 
in the exchange of training set size and number of features, $\nf \leftrightarrow \nA$, 
no phase transition that exchanges  
$\secG_{-1}(\nf,\nl)$ 
with 
$\secG_{-1}(\nA,\nl)$
as the model crosses $\nA = \nf$,
and, consequently, the test loss is not symmetric under $\nf \leftrightarrow \nA$.
Instead, 
the linear model 
has a different phase transition solely in the function $\secG_{-1}(\nA,\nf)$ at the point of $\nf = \nA$; this phase transition is analogous only to the other phase transition in the random feature model that occurs when $\nA$ or $\nf$ crosses $\nl$, cf., e.g., \eqref{eq:breakdown-no-noise}, and leads to different overall behavior.

To probe these properties of the linear model further, let's explicitly evaluate the ridgeless test loss, \eqref{eq:linear-model-averaged-test-loss}, for two different data models: first we'll consider completely structureless data with Marchenko-Pastur statistics (\S\ref{sec:linear-mp-model}), and then we'll make a direct comparison to the main text by considering power-law data (\S\ref{sec:linear-power-law-model}).

\subsection{Marchenko-Pastur Data}
\label{sec:linear-mp-model}

As the simplest possible data model, let's take the covariance of the input features 
to be isotropic,
\be\label{eq:MP-spectrum}
    \covl = \varx \IN \, ,
\ee
which means that each component of the input is independent and identically distributed. 
The distribution of empirical covariance matrices of a finite-sized dataset is the Marchenko-Pastur (MP) distribution \cite{Mar_enko_1967}. The test loss of linear models with MP data have been studied extensively in the literature. %

For MP data, the calculation of $\secm$ and $\secG_{0}$ are both trivial because the sum over eigenvalues doesn't depend on their index. Starting with $\secm$, plugging the expansion \eqref{eq:Delta-expansion-ansatz-prepreint} into the self-consistent equation \eqref{eq:delta-self-consistent-reprint}, converting the sum into a trace, and substituting in the spectrum \eqref{eq:MP-spectrum}, we have find a simple equation that determines  $\secm$: 
\be\label{eq:leading-mp-self-energy-equation}
\secm  = \sum_{i=1}^{\nf} \frac{\secm }{\nA + \secm \stdx^{-2} } =  \frac{\nf\secm }{\nA + \secm \stdx^{-2} }\, .
\ee
Solving this trivial equation gives two solutions,
one valid for when the model when the model is under(over)-parameterized:
\be\label{eq:delta-minus-one-mp}
\secm(\nA,\nf) =     
    \begin{cases}
        (\nf -\nA)\varx \, , & \nA < \nf \, ,\\
        0 \, , & \nA > \nf \,.
    \end{cases}
\ee
The reason for this assignment is as follows: for $\nA < \nf$, the empirical covariance has zeros, which means that the resolvent, $\res$, and its average, $\resb$, must blow up as $\ridge^{-1}$ as $\ridge \to 0$, cf. \eqref{eq:resolvent-feature-feature-appendix}, and thus $\secm(\nA,\nf) \neq 0$; for $\nA > \nf$, the empirical covariance is full rank, there's no pole in $\res$ as $\ridge \to 0$, and thus $\secm(\nA,\nf) = 0$.

For calculation of $\secG_0$, we have to consider $\nA < \nf$ and $\nA > \nf$ separately. For $\nA < \nf$, starting with the $\o{\ridge^0}$ term in the power-series equation of the self-consistent equation \eqref{eq:delta-self-consistent-reprint}, 
\be\label{eq:delta-0-equation-preprint}
    \secG_0 = (1+\secG_0)\sum_{I=1}^\nl \frac{\nG \eig_I^2}{\left(\nG \eig_I + \secm\right)^2} \, ,
\ee
plugging in $\secm=(\nf -\nA)\varx $ and $\covl = \varx \IN$, and performing the trivial sum, we find
\be
\secG_0 =(1+\secG_0) \frac{\nA}{\nf}  \quad \implies \quad \secG_0(\nA,\nf) = \frac{1}{\nf/\nA -1} \, .
\ee
For $\nf < \nA$, our ansatz \eqref{eq:Delta-expansion-ansatz-prepreint} instead should begin at $\o{\ridge^0}$: plugging that expansion into the self-consistent equation \eqref{eq:delta-self-consistent-reprint}, we find
\be
    \secG_0 
    = \sum_{i=1}^{\nf} \frac{1+ \secG_{0}}{\nA} = (1+\secG_0) \frac{\nf}{\nA}  \quad \implies \quad \secG_0(\nA,\nf) = \frac{1}{\nA/\nf -1}\, . %
\ee
Altogether, we have
\begin{equation}\label{eq:delta-zero-mp}
    \secG_0(\nA,\nf) = 
    \begin{cases}
        \frac{1}{\nf/\nA-1} \, ,   & \nA < \nf\,, \\
        \frac{1}{\nA/\nf-1}   \,,          & \nA > \nf\,,
    \end{cases}
\end{equation}
independent of the scale of the size of the ``signal'' $\varx$.\footnote{
It's interesting to compare this expression, \eqref{eq:delta-zero-mp}, with our formula for $\secG_0$ with power-law data, which is computed in the next appendix,
\eqref{eq:delta-zero-complete}: 
as
the first argument of $\secG_0$ approaches the second, these formula agree completely; however, the MP formula, \eqref{eq:delta-zero-mp}, is \emph{exact}, while the coincident regime of the power-law formula, \eqref{eq:delta-zero-complete}, is only the leading part of the $\varepsilon$ 
expansion. The reason for this agreement is that as two scales become equal, the tail of the spectrum of the empirical covariance 
is
universally determined by MP statistics for any underlying $\covl$.
}

\begin{figure}[ht]
    \begin{center}
     \includegraphics[width=0.49\linewidth]{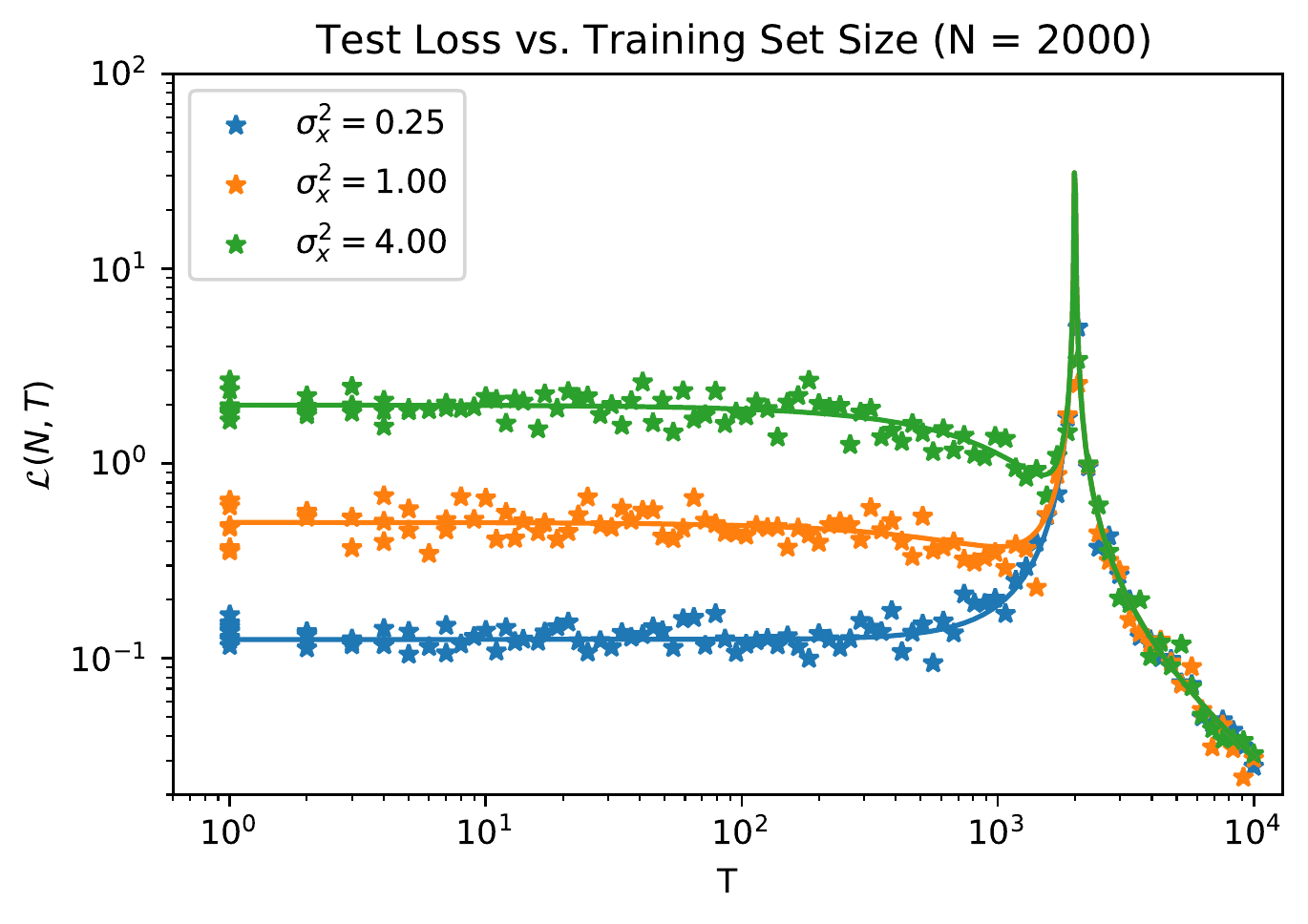} \includegraphics[width=0.49\linewidth]{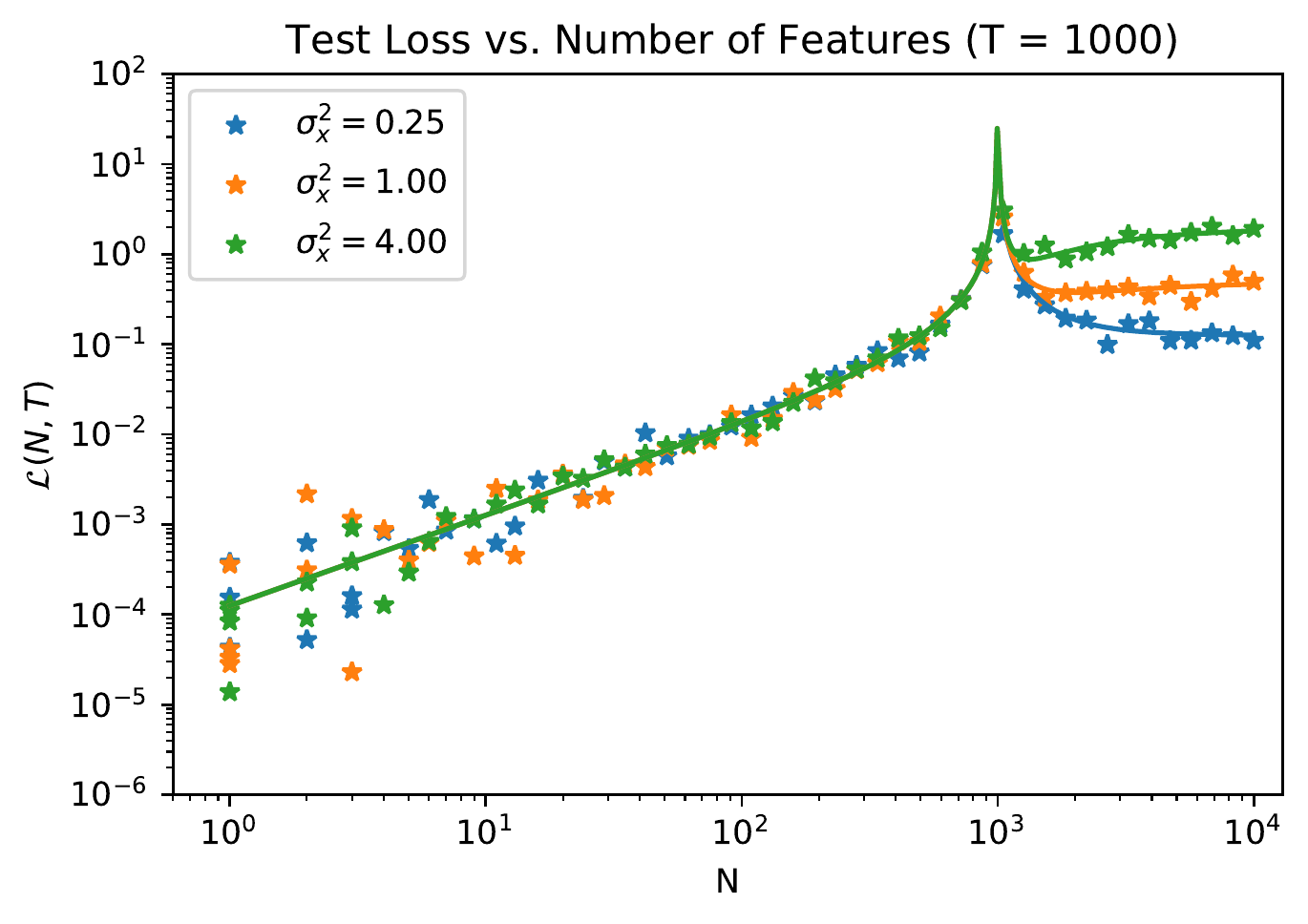}
    \end{center}
    \caption{Test loss from numerical simulations (stars) and our analytical expression (solid lines) of a linear model trained on MP data ($\wvar = 1$, $\vare = 0.25$). As the two panels show, the test loss of this model is not symmetric under the exchange $\nA \leftrightarrow \nf$.
    \textbf{Left:} The size of the training set, $\nA$, is varied for a few different sized signals, $\varx$, while the number of features is held fixed ($\nf = 2000$); note that the linear decrease of the test loss $(1-\nA/\nf)$ with increasing $\nA$ is not easily perceptible on a log-log scale.
    \textbf{Right:} The number of features, $\nf$, is varied for a few different sized signals, $\varx$, while the size of the training set is held fixed ($\nA = 1000$). 
}
    \label{fig:simple-model-gauss}
\end{figure}

Since we've evaluated $\secm$ and $\secG_{0}$ exactly, we can write an \emph{exact} formula for the test loss in the ridgeless limit ($\ridge \to 0$): trivially substituting \eqref{eq:delta-minus-one-mp} and \eqref{eq:delta-zero-mp} into \eqref{eq:linear-model-averaged-test-loss}, we get
\be\label{eq:test-loss-mp-data}
   \L_{\text{MP}}(\nf, \nA) \equiv \expval{\L_\B(\param^\star)}_{\noise,\w,\widehat{x},x} =\frac{1}{2 } 
    \begin{cases}
        
        \wvar \varx \le(1 - \frac{\nA}{\nf}\ri) + \frac{\vare}{\nf/\nA-1} \,, & \nA< \nf \,,\\  
        \frac{\vare}{\nA/\nf - 1} \,,  & \nA > \nf\,.
    \end{cases}
\ee
In Fig.~\ref{fig:simple-model-gauss}, we plot some example test loss curves using numerical simulations of MP data for different signal strengths $\varx$ and compare with our analytical formula \eqref{eq:test-loss-mp-data}: considering both fixed number of features while varying the training set size (left panel) and fixed training set size while varying the number of features (right panel), we see an excellent fit to the experimental data. 

Overall, we can characterize the performance of the model according to its different regimes:
\bi
\item When the model is \emph{overparameterized}, $\nA < \nf$, the test loss decreases linearly towards zero as $(1-\nA/\nf)$ with increasing $\nA$; each new training point gives the model access to an additional eigenvalue of the covariance matrix, each of which is equal size.
\item At the point of \emph{equiparameterization}, $\nA = \nf$, the performance formally diverges a universal way,  overfitting to the nonzero noise. (In the noiseless limit, $\noise \to 0$, the test loss would be identically zero.)
\item When the model is \emph{underparameterized}, $\nA > \nf$, the only contribution to the test loss comes from the noise; this eventually falls off with increasing $\nA$ as the noise is distributed independently for each training sample and ultimately averages away as $\nA \to \infty$.
\ei

This phenomenology is quite different than the random feature model from the main text, \eqref{eq:statistical-model-answer}:
\bi
\item[\emph{(i)}] The random feature model  has a regime that exhibits power-law scaling law of the test loss. 
This cannot happen for MP data since there's no power-law in the spectrum to translate into the test loss; essentially, MP \emph{data} is itself structureless noise that cannot capture the rich behavior of the LLMs.
\item[\emph{(ii)}] (For small noise, $\vare$) the test loss of the random feature model in the main text was \emph{symmetric} under the exchange of the training set size and the number of features $(\nA \leftrightarrow \nf)$, while the MP-data linear model is \emph{not}. (This is clear from inspection of the analytical solution, \eqref{eq:test-loss-mp-data}, and illustrated by the two panels of Fig.~\ref{fig:simple-model-gauss}.) The problem with the linearly model is that as we increase the number of features, $\nf$, we simultaneously increase the number of features used to determine the label, cf. \eqref{eq:def-teacher-labels-appendix}. In other words, 
the learning problem 
of
training a small number of features on a large training set is fundamentally easier
than
training many features on a small training set.
\ei
To fix the first problem, we need 
a more realistic generative data model (\S\ref{sec:linear-power-law-model}). To fix the second problem, we need 
to decouple the size of the model used in the regression from the complexity of the learning problem (\S\ref{sec:derivation}).

\subsection{Power-Law Data}
\label{sec:linear-power-law-model}

To address that first problem, let's now use the same underlying data model as in \S\ref{sec:derivation}: we'll assume the spectrum of 
$\covl$
has a number density that's well-approximated by 
\be\label{eq:eig-density-almost-reprint}
    n(\eig) d\eig = {\nf (\beta-1) \eigmin^{\beta-1}} \eig^{-\beta} \theta(\eig-\eigmin) d\eig\, ,
\ee
where here $\eigmin$ is the minimum eigenvalue, $\beta$ is an exponent that characterizes the tail of the distribution, and the constants are chosen such that the density integrates to $\nf$. Recall from our earlier discussion that we can instead characterize the spectrum by a maximum eigenvalue, $\eigmax$ ,and exponent, $\alpha$, which are related to $\eigmin$ and $\beta$ by,  cf. \eqref{eq:def-eig-max} and \eqref{eq:def-of-alpha},
\be
\eigmax \equiv \eigmin \nl^{1+\alpha} \, , \qquad \alpha \equiv \frac{2- \beta}{\beta-1}\, ,
\ee
with the range $1 < \beta < 2$ translated to $0 < \alpha < \infty$.

\begin{figure}
    \begin{center}
     \includegraphics[width=0.49\linewidth]{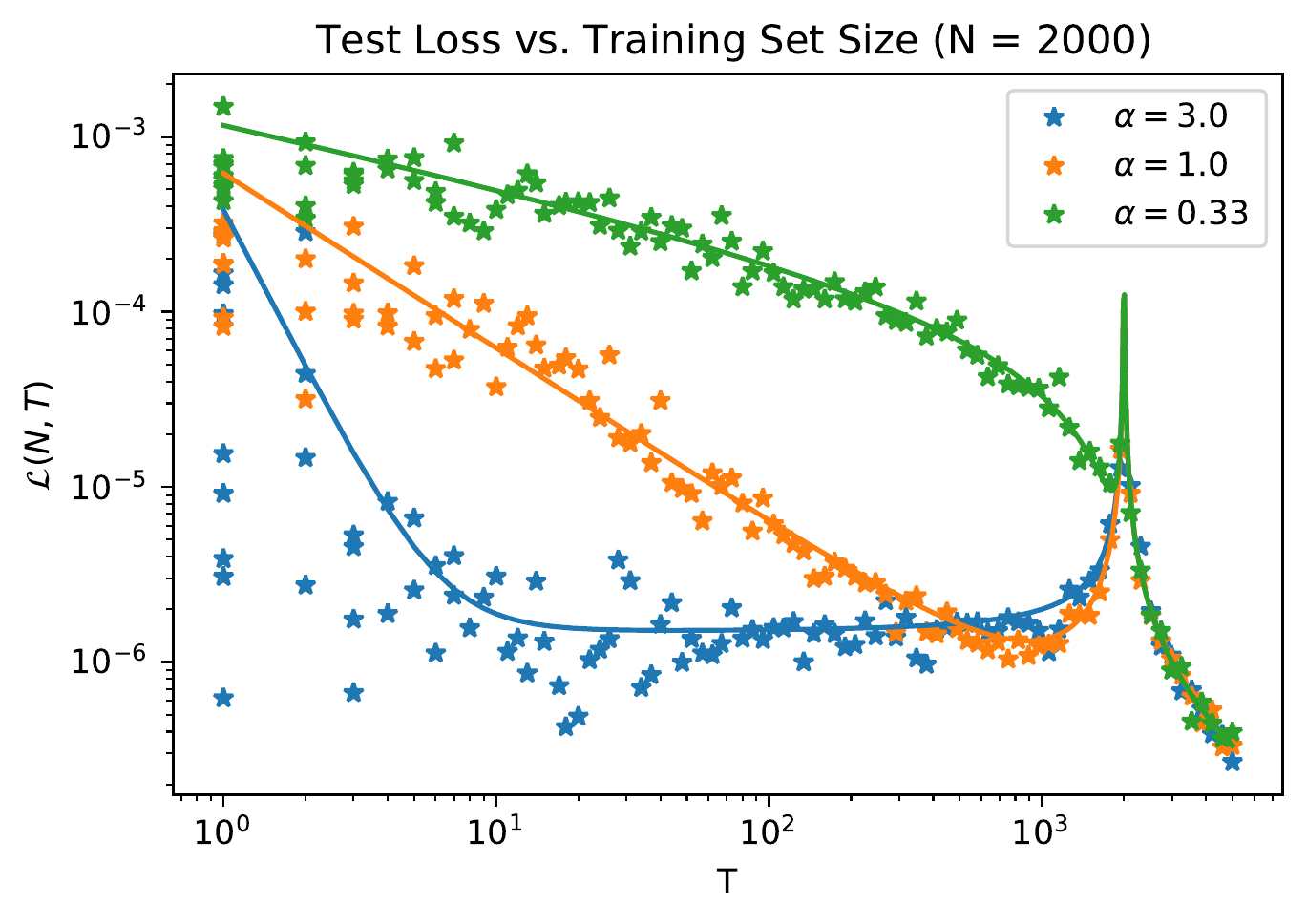} \includegraphics[width=0.49\linewidth]{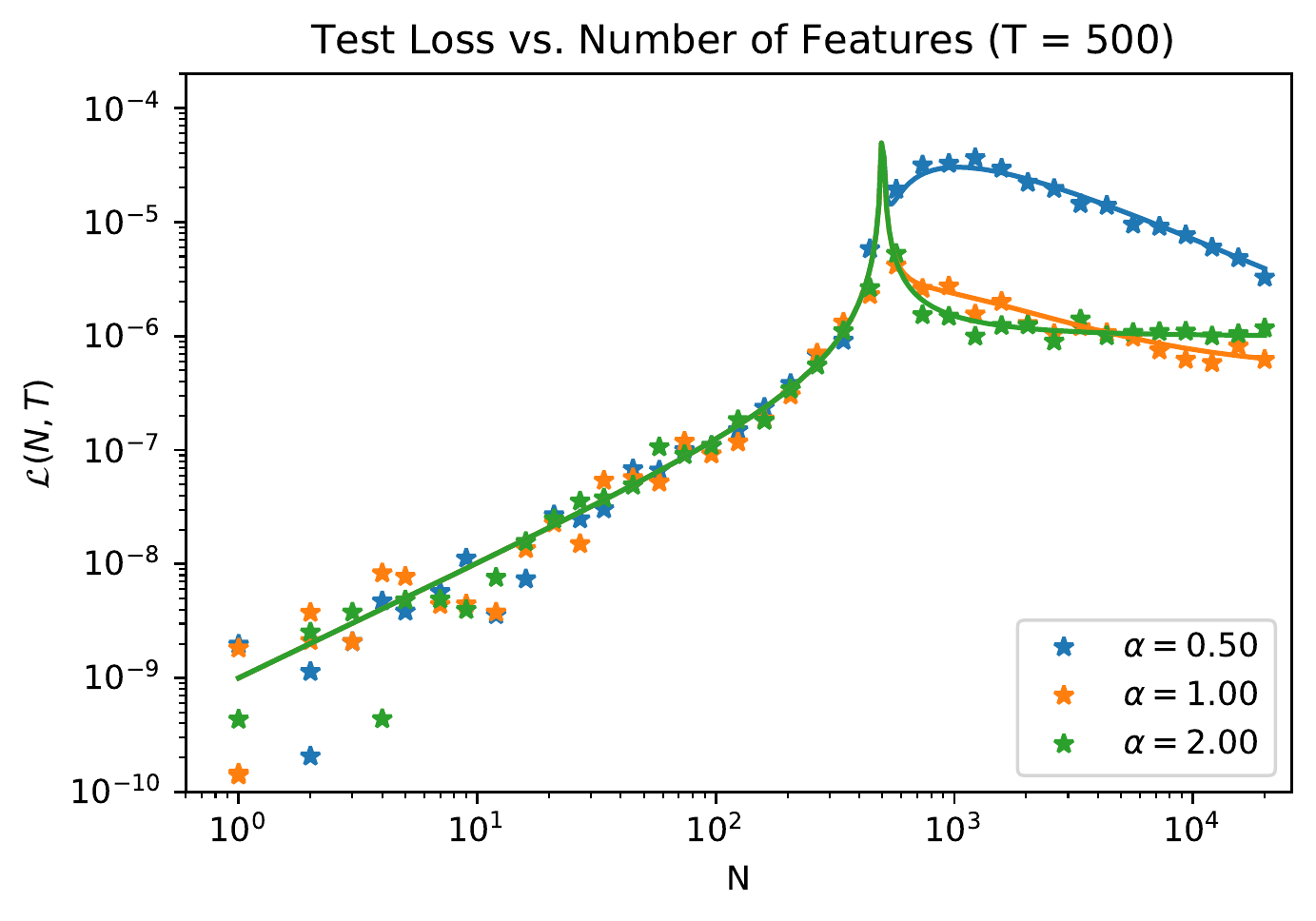} \\
     \includegraphics[width=0.49\linewidth]{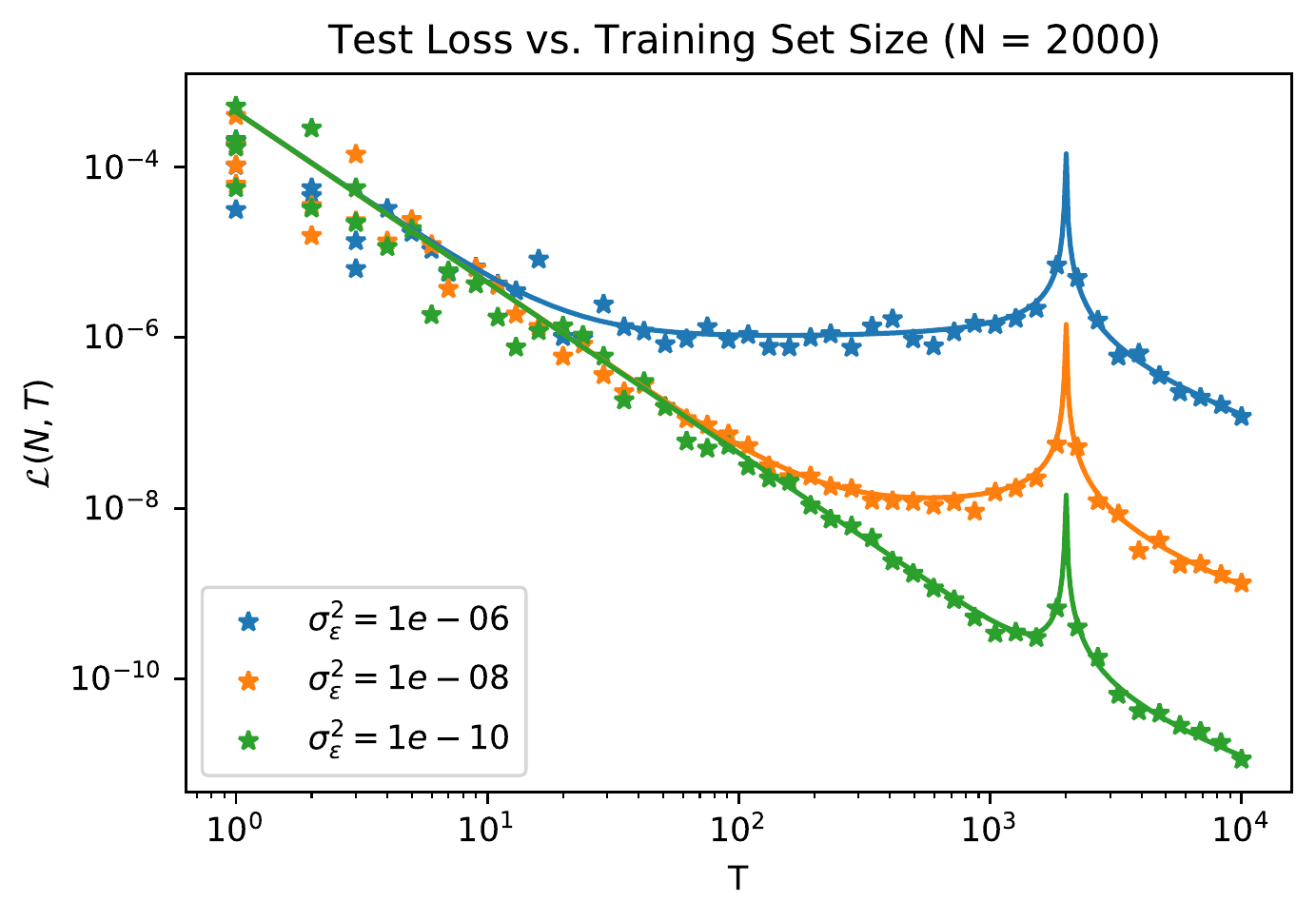}
    \end{center}
    \caption{Test loss from numerical simulations (stars) and our analytical expression (solid lines) of a linear model trained on power-law data ($\eigmax=1$, $\wvar = 1$). The top panels illustrate that the test loss of this model is not symmetric under the exchange $\nA \leftrightarrow \nf$.
\textbf{Top Left:} The size of the training set, $\nA$, is varied for a few different power-law exponents, $\alpha$, while the number of features and the noise variance are  held fixed ($\nf = 2000$, $\vare=10^{-6}$); for $\nA \ll \nf$, the test loss behaves as a power law in the training set size, $\L_{\text{PL}}(\nf,\nA) \sim \nA^{-\alpha}$, but for $\nA > \nf$ there's no plateau behavior.
\textbf{Top Right:} The number of features, $\nf$, is varied for a few different power-law exponents, $\alpha$, while the size of the training set and the noise variance are held fixed ($\nA = 1000$, $\vare=10^{-6}$); for $\nf\ll \nA$, the test loss actually \emph{grows} with increasing features as $\L_{\text{PL}}(\nf,\nA) \sim \nf$.
\textbf{Bottom:} The number of features, $\nf$ is varied for several different values of the noise variance, $\vare$, while the power-law exponent and training set size are held fixed ($\alpha=1.0$, $\nf =2000$).
}
\label{fig:simple-model-power-law}
\end{figure}

For power-law data, the calculations of $\secm(\nA,\nf)$ and $\secG_{0}(\nA,\nf)$ are performed in Appendix~\ref{sec:delta}. Importing the results of those calculations,
\eqref{eq:delta-minus-one-explicit-form-both-regimes} and \eqref{eq:delta-zero-complete}, and substituting into our expression for the test loss, \eqref{eq:linear-model-averaged-test-loss}, we can write a formula for the test loss of the linear model:
\begin{align}\label{eq:test-loss-pl-data}
   &\L_{\text{PL}}(\nf, \nA) \equiv \expval{\L_\B(\param^\star)}_{\noise,\w,\widehat{x},x}\, \\
    =&\frac{1}{2 } 
    \begin{cases}
        \frac{\eigmax\wvar}{\nf^{\alpha+1} }\le\{
        k \le[\le(\frac{\nf}{\nA}\ri)^\alpha -1 \ri] + \le[2+ \alpha(1-k)\ri] \le(1 - \frac{\nA}{\nf} \ri) 
    \ri\}   + \vare\le(\alpha + \frac{1}{\nf/\nA-1} \ri) \,, & \nA< \nf \,,\\  
        \frac{\vare}{\nA/\nf - 1} \,,  & \nA > \nf\,. \notag
    \end{cases}
\end{align}
Comparison with the MP formula \eqref{eq:test-loss-mp-data} highlights the way that near the coincident regime, $\nA \to \nf$, the loss behaves as if the data were structureless MP data. 
In Fig.~\ref{fig:simple-model-power-law}, we plot some example test loss curves using numerical simulations of power-law data and compare with our analytical formula \eqref{eq:test-loss-pl-data}. This shows an excellent fit of this formula to experiment.

On the one hand,
from the expression of the test loss, \eqref{eq:test-loss-pl-data}, and the figure, we can see that with power-law data there is a scaling regime, $\nA \ll \nf$, where the test loss has power-law behavior, $\L_{\text{PL}}(\nf, \nA) \sim \nA^{-\alpha}$, analogous to the random feature model in the main text. On the other hand, as discussed in the previous subsection, there cannot be a dual regime where the model behaves as a power-law in the number of features. In other words,  from the perspective of increasing $\nA$, there's no plateau behavior as exhibited by \eqref{eq:phenomenological-loss-original}; 
instead, for $\nA > \nf$, %
we only get the universal noise contribution.

Thus, we conclude that a simple linear model does not have the same rich phenomenology exhibited by the LLMs of \cite{kaplan2020scaling}. To make a sufficient minimal model of neural scaling phenomenology, we need to distinguish the latent features of the data from the model's features. We do this in the main body of the paper using the random feature model presented in \S\ref{sec:derivation}.

\section{Explicit Solutions for \texorpdfstring{$\Delta$}{Delta}}
\label{sec:delta}

We would like to self-consistently solve the equation
\be\label{eq:full-generic-self-energy-equation}
\secG(\nG, \nl) = \tr{\frac{\covl}{ \ridge \IM + \frac{\nG \covl}{1+\secG(\nG, \nl)} } }  \, ,
\ee
in the limit where the ridge parameter is vanishing, $\ridge \to 0$. Here, $\nG$ is a generic scale that can either represent the number of features, $\nf$, or the size of the training set, $\nA$. We will proceed to solve this equation order-by-order. 

\subsection{Explicit Solution for \texorpdfstring{$\Delta_{-1}$}{Delta Minus One}}
\label{sec:delta-minus-one}

First, let us consider the parameter regime where $\nG < \nl$.  Then, consideration of the above expression as $\secG(\nG, \nl)$ becomes large suggests that it should diverge as $1/\ridge$ as $\ridge \to 0$. Thus, to find the leading behavior, we can insert an ansatz expansion
\be\label{eq:Delta-expansion-ansatz}
\secG(\nG, \nl) \equiv \frac{\secm}{\ridge} + \secG_{0} + \secG_{1} \gamma + \dots \, ,
\ee
and rearrange to find:
\be\label{eq:leading-generic-self-energy-equation-pretrace}
1=\tr{\frac{ \covl}{\secG_{-1} \, \IM+ \nG \covl}}\, + \o{\ridge}\, .
\ee
Covering the trace into a sum over eigenvalues, we get
\be\label{eq:leading-generic-self-energy-equation}
1 = \sum_{I=1}^{\nl} \frac{1}{\nG + \secm \eig_I^{-1} }  + \o{\ridge} \, ,
\ee
where $\eig_I$ is the spectrum of $\covl$. %
Our goal is to find an explicit expression for $\secm(\nG, \nl)$.
We will proceed by considering two asymptotic regimes and then by constructing a solution that naturally interpolates between them.

The first regime concerns the limit where the scale, $\nG$, approaches the size of the latent space, $\nl$. It's easy to see that in the exact limit of $\nG \to \nl$, the equation \eqref{eq:leading-generic-self-energy-equation} is satisfied by $\secm = 0$. This suggests that in this \emph{coincident} regime, we might find a solution in terms of an expansion
\be\label{eq:coincident-expansion}
\secm(\varepsilon) \equiv a \varepsilon + a' \varepsilon^2 + \dots \, , 
\ee
with the closeness parameter defined as
\be\label{eq:def-of-closeness-parameter}
\varepsilon \equiv  1 - \frac{\nG}{ \nl}  \, .
\ee
Inverting this definition, \eqref{eq:def-of-closeness-parameter}, for $\nG$ and substituting both that and the expansion \eqref{eq:coincident-expansion} into the summation equation, \eqref{eq:leading-generic-self-energy-equation}, we find:
\be
1 = \frac{1}{\nl}\sum_{I=1}^{\nl}\le[1 + \le(1- \frac{a}{\nl \eig_I} \ri)\varepsilon + \o{\varepsilon^2}\ri]\,.
\ee
From this, we see that zeroth order term is satisfied by construction and that the constant $a$ must satisfy
\be\label{eq:generic-solution-coincident-limit}
a = \frac{\nl^{2} }{\sum_{I=1}^\nl \eig_I^{-1}}\,.
\ee
If desired, this can be systematically continued to find the higher-order coefficients in the expansion \eqref{eq:coincident-expansion}, but for our needs the first order term is sufficient.

Up to this point this expression, \eqref{eq:generic-solution-coincident-limit}, is completely universal given a spectrum $\eig_I$. To simplify further, let us insert our latent space power law \eqref{eq:exact-power-law-latent-generative}:
\be\label{eq:eig-fixed-spectrum-appendix-reprint}
\eig_I = \eigmax \le(\frac{1}{I}\ri)^{1+\alpha} \, ,
\ee
where recall $\eigmax$ represents the largest eigenvalue, and $\alpha > 0$ controls the exponent of the power-law limit of the scaling law.\footnote{Alternatively, we could instead average over a smooth density of eigenvalues, \eqref{eq:eig-density}: regardless, at large $\nl$ the smooth density will behave similarly to the fixed spectrum.} 
Inserting this spectrum into \eqref{eq:generic-solution-coincident-limit}, then 
approximating the sum as an integral and taking the large-latent-space limit, $\nl \to \infty$, we find:
\be
a = \frac{(2+\alpha)\eigmax}{\nl^\alpha} + \o{M^{-1-\alpha}} \, .
\ee

The second regime is a \emph{scaling} regime: to find another solution we essentially reverse the order of limits we just took; for the coincident regime we first expanded around $\nG \to \nl$ and then took $\nl \to \infty$, but in the scaling regime we will first take $\nl \to \infty$ and then look for a power-law scaling solution in $\nG$ as $\nG \to \infty$. 
In this limit, we can safely approximate the sum as an integral from the start,
\be
\sum_{I=1}^{\nl} \frac{1}{\nG + \secm \eig_I^{-1} }  \to \int_{1}^\infty\frac{dI}{ \nG + \secm  I^{1+\alpha} / \eigmax}  \,,
\ee
where importantly we have taken the limit of the integral to infinity as $\nl \to \infty$, and we have also substituted in for the power-law spectrum \eqref{eq:eig-fixed-spectrum-appendix-reprint}.\footnote{
    N.B. while the coincident limit has the universal form, \eqref{eq:generic-solution-coincident-limit}, being determined by a simple function of the underlying spectrum, the scaling limit is specific to the power law spectrum.
}
While this integral can be evaluated directly in terms of a hypergeometric function (as was done in \cite{bahri2021explaining}) instead let's instead substitute in a scaling ansatz:
\be\label{eq:scaling-ansatz-Delta-minus-one}
\secm(\nG) \equiv \frac{\eigmax k}{\nG^{\alpha}} \, , 
\ee
with a constant $k$ assumed to be independent of $\nG$, and with the function $\secm(\nG)$ independent of $\nl$ since we've taken $\nl \to \infty$. In addition to that substitution, changing integration variables as
\be\label{eq:change-of-variables}
I = \frac{s \nG}{k^{1/(1+\alpha)}} \, ,
\ee
and setting the integral equal to unity as per \eqref{eq:leading-generic-self-energy-equation}, we can find an integral expression for the constant $k$:
\be
k = \le[ \int_{k^{1/(1+\alpha)}\nG^{-1}}^\infty\frac{ds}{1 +  s^{1+\alpha} } \ri]^{1+\alpha}\,.
\ee
Since in our scaling limit, $\nG \to \infty$, the constant $k$ in our scaling ansatz, \eqref{eq:scaling-ansatz-Delta-minus-one}, is assumed to be independent of $\nG$, we can now safely take the second limit of $\nG \to \infty$ and evaluate the dimensionless integral in the  square brackets in terms of some gamma functions:
\be
\int_{0}^\infty\frac{ds}{1 + s^{1+\alpha} } = \Gamma\!\le(\frac{\alpha}{1+\alpha} \ri) \Gamma\!\le(\frac{2+\alpha}{1+\alpha} \ri) \,.
\ee
This is just some number, and using a gamma function identity, we can rewrite this number to give an expression for $k$:
\be\label{eq:def-k}
k = \left[\frac{\frac{\pi}{1+\alpha } }{\sin\!\left(\frac{\pi }{1+\alpha}\right)}\right]^{1+\alpha} \,.
\ee

Lastly, having solved \eqref{eq:leading-generic-self-energy-equation} in two non-commuting limits, as a final step we need to find a solution that interpolates between these two regimes. Consider a final ansatz
\be\label{eq:interpolating-ansatz-delta-minus-one}
\secm(\nG, \nl) \equiv \eigmax k\le(\nG^{-\alpha} - \nl^{-\alpha}\ri) + b \varepsilon \, ,
\ee
for some constant $b$ that we assume to be independent of $\nG$. To make sure that we match the correct behavior in the coincident regime, 
we expand \eqref{eq:interpolating-ansatz-delta-minus-one} as $\nG \to \nl$,
\be
\secm(\nG, \nl) = \le(b + \alpha \eigmax k \ri) \varepsilon + \o{\varepsilon^2}\, ,
\ee
and set it equal to \eqref{eq:coincident-expansion}, giving
\be
b = \le[2+ \alpha(1-k)\ri]\frac{\eigmax}{M^\alpha} \, .
\ee
Then, substituting this back in our ansatz \eqref{eq:interpolating-ansatz-delta-minus-one}, we see that in the scaling regime we will match onto our solution \eqref{eq:scaling-ansatz-Delta-minus-one} as we take $\nl \to \infty$:
\be
\secm(\nG, \infty) \to \secm(\nG) = \frac{\eigmax k}{\nG^{\alpha}}\, .
\ee

\begin{figure}[ht]
\begin{center}
 \includegraphics[width=0.6\linewidth]{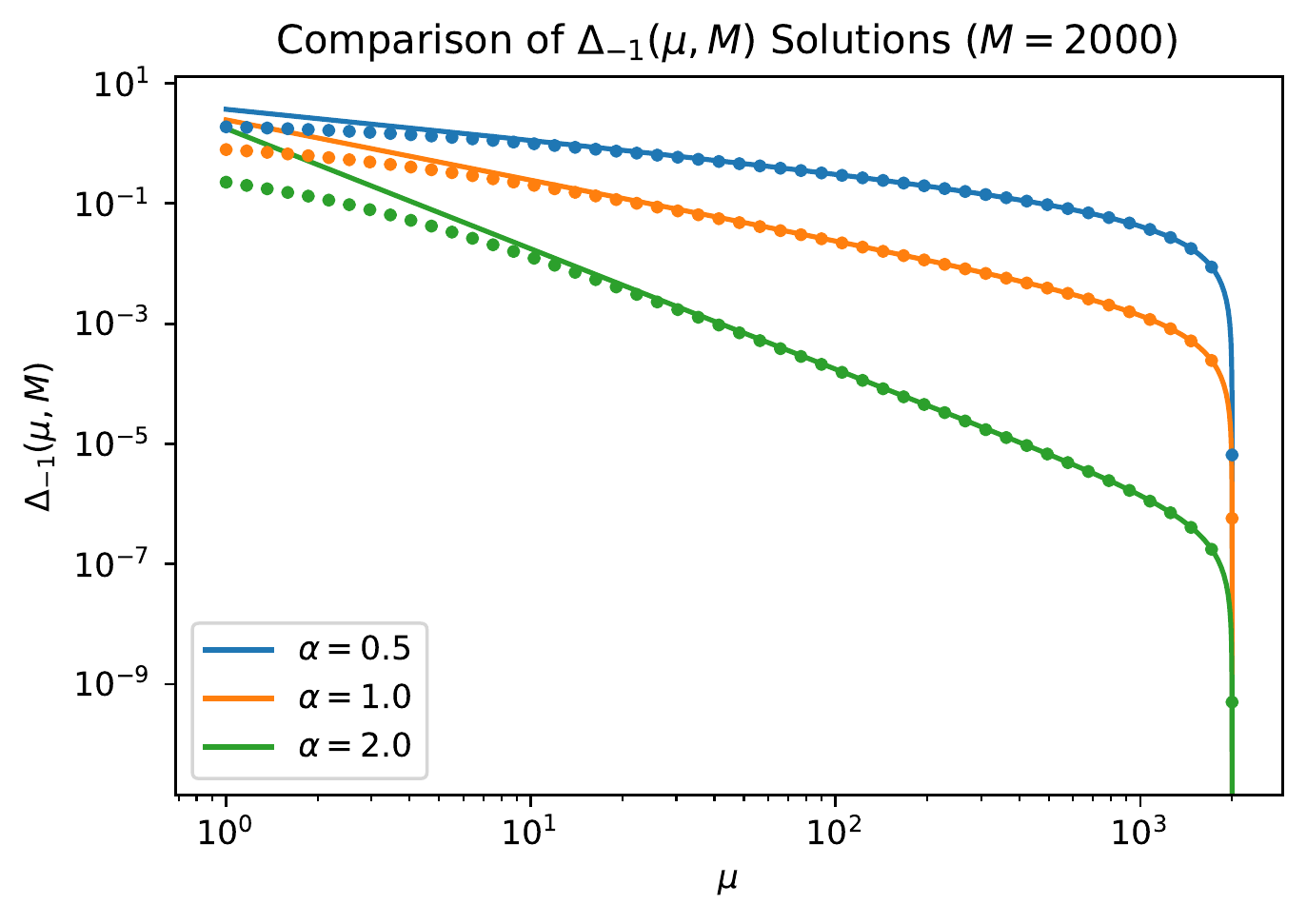}
\end{center}
\caption{Comparison analytical and numerical solutions for $\secm(\nG,\nl)$ for different exponents $\alpha$ and fixed 
$\nG$: the numerical solution of \eqref{eq:leading-generic-self-energy-equation} (dots) closely matches our analytical solution \eqref{eq:delta-minus-one-explicit-form} (solid lines) except for small small $\nG \lesssim 10$, where we also expect the random matrix theory analysis of the paper to deviate.
}
\label{fig:Delta-minus-one}
\end{figure}

Overall, we can write our explicit interpolating solution as
\be\label{eq:delta-minus-one-explicit-form}
\secm(\nG, \nl) = \frac{\eigmax}{\nl^\alpha}\le\{
    k \le[\le(\frac{\nl}{\nG}\ri)^\alpha -1 \ri] + \le[2+ \alpha(1-k)\ri] \le(1 - \frac{\nG}{\nl} \ri)
\ri\} \, ,
\ee
with the dimensionless constant $k$ defined in \eqref{eq:def-k}. To test our solution, we can also solve for $\secm(\nG,\nl)$ numerically using \eqref{eq:leading-generic-self-energy-equation}. In Fig.~\ref{fig:Delta-minus-one} we plot our numerical solution against our analytical solution \eqref{eq:delta-minus-one-explicit-form} 
as a function of $\nG$, for different $\alpha$ and $\nl$. Overall we see excellent match everywhere except for small $\nG \lesssim 10$, where our scaling solution would receive corrections and where we expect our random matrix theory analysis of \S\ref{sec:derivation}  also to be subject to corrections.

Finally, let us note that for $\nG > \nl$ the empirical covariance is full rank, so there's no pole in the resolvent as $\ridge \to 0$: this means that there's no $\ridge \to 0$ pole in $\secG$ either. 
Thus, overall we have:
\be\label{eq:delta-minus-one-explicit-form-both-regimes}
\secm(\nG, \nl) = \begin{cases} 
\frac{\eigmax}{\nl^\alpha}\le\{
    k \le[\le(\frac{\nl}{\nG}\ri)^\alpha -1 \ri] + \le[2+ \alpha(1-k)\ri] \le(1 - \frac{\nG}{\nl} \ri)
\ri\} \, , & \nG < \nl \, , \\
0 \,, &  \nG > \nl \, .
\end{cases}
\ee

\subsection{Explicit Solution for \texorpdfstring{$\secG_{0}$}{Delta Zero}}
\label{sec:delta-zero}

We will need to divide the analysis into the cases of $\nG < \nl$ and $\nG > \nl$. %

\subsubsection*{$\boldsymbol{\nG < \nl}$} 

Let's continue with our expansion of $\secG$ in the parameter regime where $\nG >\nl$
to find the NLO contribution $\secG_0$. The $\o{\ridge^0}$ term in our power series expansion of \eqref{eq:full-generic-self-energy-equation}
gives
\be\label{eq:delta-0-equation}
    \secG_0 = (1+\secG_0)\sum_{I=1}^\nl \frac{\nG \eig_I^2}{\left(\nG \eig_I + \secm\right)^2} \, ,
\ee
which can be solved for $\secG_0$ in terms of $\secm$. To begin, first note that we can rewrite \eqref{eq:delta-0-equation} as
\be\label{eq:def:kappa-and-sum}
  \kappa = \sum_{I=1}^\nl  \frac{\nG}{\left(\nG + \secm \eig_I^{-1}\right)^2} \,, \qquad \secG_0 = \frac{\kappa}{1-\kappa} \, . \qquad 
\ee
Given our solution for $\secm$, we  can evaluate the sum in the left expression in terms of known quantities and then easily find  $\secG_0$.
As before, let us consider two asymptotic regimes for the sum and then construct an interpolating solution. 

In the coincident regime ($\nG \to \nl$), 
substituting in $\secm = a \varepsilon + \o{\varepsilon^2}$ 
and $\mu = \nl(1-\varepsilon)$, and using \eqref{eq:generic-solution-coincident-limit}, we can evaluate the sum to linear order an find
\be
\kappa = 1 - \varepsilon + O(\varepsilon^2)\, ,
\ee
which gives 
\be\label{eq:delta-0-overparam-coincident}
    \secG_0
    = \frac{1}{\nl/\nG-1}  + \ldots \, . 
\ee
Interestingly, this leading order answer is completely independent of the spectrum. As mentioned before, we can systematically continue this  $\varepsilon$ expansion to find more terms if needed.

In the scaling regime, we can follow the same procedure as in the previous subsection: 
first, we substitute in $\secm = \eigmax k \nG^{-\alpha}$ and $\eig_I = \eigmax I^{-1-\alpha}$ for the leading order scaling solution and spectrum, respectively; 
next, we approximate the sum in \eqref{eq:def:kappa-and-sum} as an integral; then, we change variables in \eqref{eq:change-of-variables}. Finally, we safely take the $\nG \to \infty$ limit to get
\be\label{eq:delta-0-overparam-scaling}
    \kappa  = %
    \frac{1}{k^{1/(1+\alpha)}}\int_{0}^\infty\frac{ds}{\left(1 +  s^{1+\alpha} \right)^2 } = \frac{\alpha}{1+\alpha} \quad \implies \quad \secG_0 = \alpha 
\ee
in the scaling regime.
Putting \eqref{eq:delta-0-overparam-coincident} and \eqref{eq:delta-0-overparam-scaling} together, it's easy to see that 
\be\label{eq:delta-zero}
    \secG_0(\nG, \nl) = \alpha + \frac{1}{\nl/\nG-1} \,
\ee
behaves as a simple interpolating solution between these two regimes.

\subsubsection*{$\boldsymbol{\nG > \nl}$} 
As we discussed in previously,
when we assume
$\nl< \nG$, then
$\secG$ no longer diverges as $1/\ridge$ as $\ridge \to 0$, but instead approaches a constant. Thus, to find the new leading behavior, we should use 
a different ansatz,
\be\label{eq:second-ansatz}
\secG(\nG, \nl) \equiv \secG_{0} + \secG_{1} \gamma + \dots \, ,
\ee
for which we will only need to evaluate the first term. Inserting this expansion into  
the self-consistent equation,
we find
\be
    \secG_0 
    = \sum_{I=1}^{\nl} \frac{1+ \secG_{0}}{\nG}\, .%
\ee
This can be simply summed and rearranged to give
\be
    \secG_0 = \frac{1}{\nG/\nl-1}\,,
\ee
another universal answer independent of the spectrum.

\begin{figure}[ht]
    \begin{center}
     \includegraphics[width=0.6\linewidth]{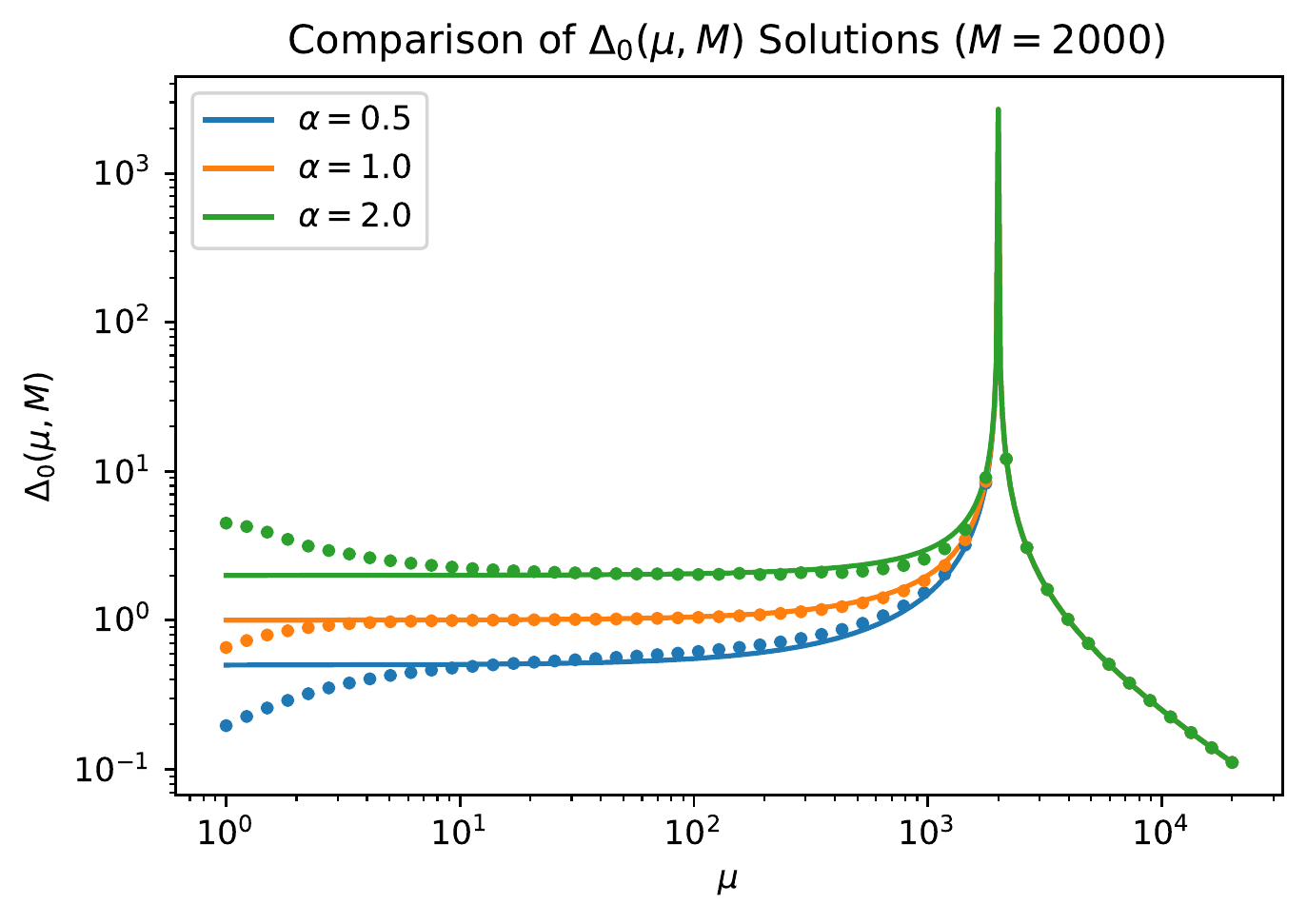}
    \end{center}
    \caption{Comparison analytical and numerical solutions for $\secG_0(\nG,\nl)$ for different exponents $\alpha$ and fixed 
    $\nG$: the numerical solution (dots) closely matches our analytical solution \eqref{eq:delta-zero} (solid lines) except for small $\nG \lesssim 10$ where we also expect the random matrix theory analysis of the paper to deviate. %
    }
    \label{fig:Delta-zero}
\end{figure}

Thus, altogether we have:
\begin{equation}\label{eq:delta-zero-complete}
    \secG_0 = 
    \begin{cases}
        \alpha + \frac{1}{\nl/\nG-1} \, ,   & \nG < \nl\,, \\
        \frac{1}{\nG/\nl-1}   \,,          & \nG > \nl\,.
    \end{cases}
\end{equation}
To verify this, in Fig.~\ref{fig:Delta-zero} we plot a comparison of this analytical approximation against a numerical evaluation of  $\secG_0$. As before, we see deviations for very small $\nG \lesssim 10$, where our scaling solution would receive corrections.
Additionally, we also note that for $\nG < \nl$, the fit in the crossover between the scaling and coincident regions 
has some very small error: this could be reduced by solving the self-consistent equations for $\secm$ and $\secG_0$ to order $\varepsilon^2$ in the coincident region.

\subsubsection*{$\secG_0$ for the Truncated Power-Law Spectrum}

In \S\ref{sec:the-noise-term}, we needed to calculate $\secF_0$,
which is determined by the 
self-consistent 
equation~\eqref{eq:self-consistent}:
\be\label{eq:self-consistent-reprint}
    \secF(\nA,\nf) = \tr{\frac{\covf}{\ridge \IN + \frac{\nA \covf}{1+\secF(\nA,\nf) } }} \, .
\ee
This is similar to the self-consistent equation \eqref{eq:full-generic-self-energy-equation} that we just solved, with trivial substitutions $\nG \to \nA$ and $\nf \to \nl$, and, less trivially, with
the covariance $\covl$, and its strict power-law spectrum \eqref{eq:eig-fixed-spectrum-appendix-reprint},
replaced by a random projection of that spectrum, $\covf=\fw \covl \fw^T$. 

Let's first consider the typical case where our number of features is less than the dimension of the latent space, $\nf < \nl$. 
Then, for a fixed $\fw$, the spectrum of $\covf$ will have a power-law bulk and then 
terminate in a non-power-law tail with a rapid decline to zero. 

When our model is also underparameterized, $\nf < \nA$, or overparameterized but near the coincident region, $\nf \gtrsim \nA$, we know from our analysis in the previous subsection that the answer is universal, independent of the spectrum: thus, automatically our calculation from before \emph{almost} carries over to this setting. On the other hand, we might expect a deviation when $\nA \ll \nf$, since $\secG_0$ was not universal in the scaling region, cf. \eqref{eq:delta-0-overparam-scaling}. 
However, 
in this limit there are only significant deviations from a strict power-law spectrum at at very small eigenvalues; since these small eigenvalues will make a negligible contribution to the integrals determining $\secF_{-1}$ and $\secF_0$, we expect this non-power-law tail to have negligible effect on the calculation of $\secF_0$. (We also expect the eigenvalues to shift by a multiplicative factor, but this scaling  does not appear in $\secF_0$ and won't contribute.)

\begin{figure}[ht]
    \begin{center}
     \includegraphics[width=0.6\linewidth]{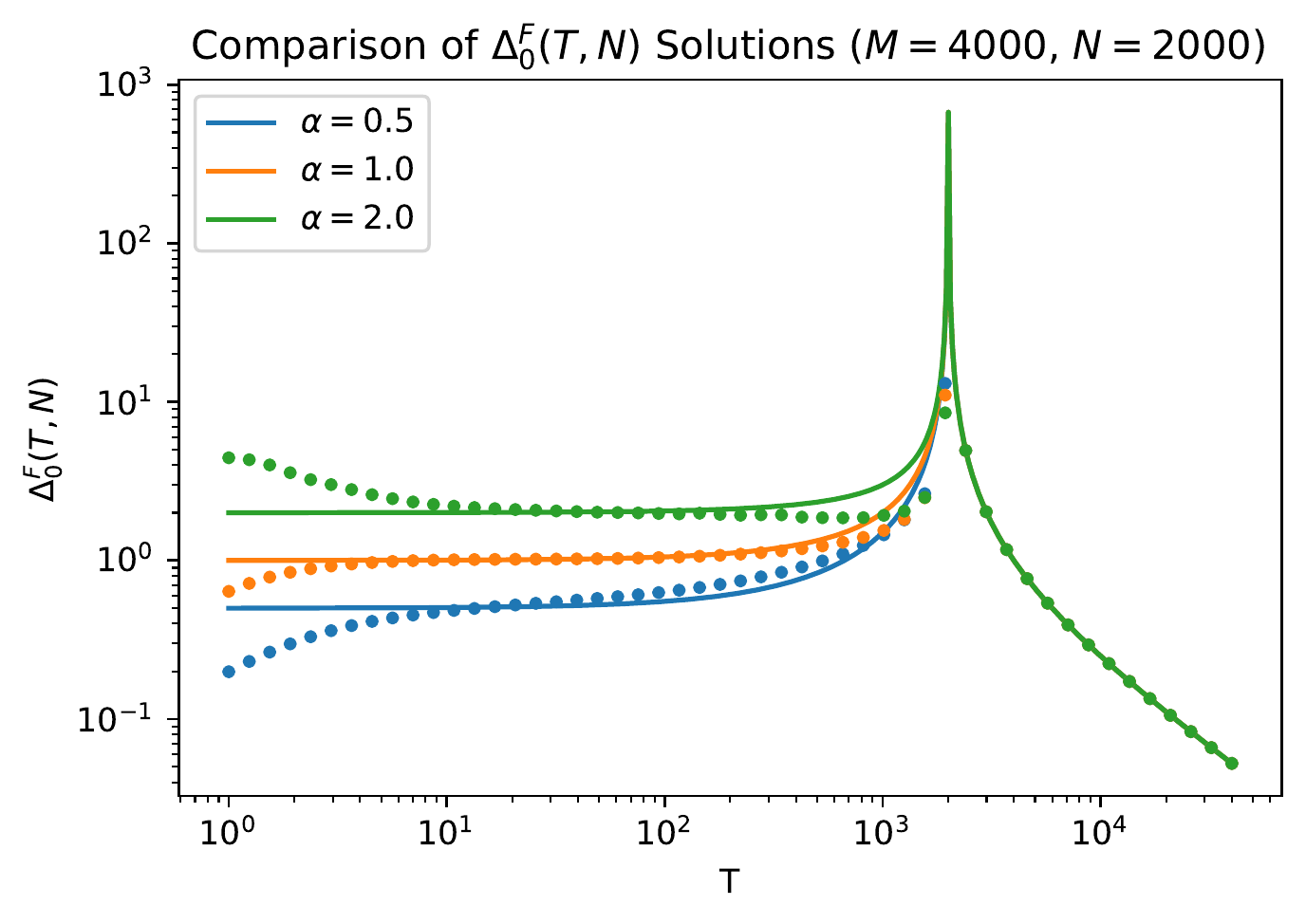}
    \end{center}
    \caption{Comparison of analytical and numerical solutions for $\secF_0(\nA,\nf)$, determined by a truncated power-law spectrum ($\lambda_+ = 2$, $\sigma_u^2 =1$), for different exponents $\alpha$ and fixed number of latent and random features ($\nl = 4000$, $\nf=2000$):
    the numerical solution (dots) 
    matches our analytical solution \eqref{eq:delta-zero-semi-complete-projected} (solid lines) except for small $\nA \lesssim 10$, where we also expect the random matrix theory analysis of the paper to deviate. For larger $\alpha$, we also see some non-negligible deviations 
    in the crossover from the scaling regime to the coincident regime; this 
    could be improved by analyzing the next term in the $\varepsilon$ expansion.
    }
    \label{fig:Delta-zero-truncated}
\end{figure}

Thus, as analogous to the answer in previous subsection \eqref{eq:delta-zero-complete}, even in this projected-spectrum case we expect
\be\label{eq:delta-zero-semi-complete-projected}
    \secF_0(\nA,\nf) = 
    \begin{cases}
        \alpha + \frac{1}{\nf/\nA-1} \,,   & \nA < \nf \,, \; \nf<\nl \,, \\
        \frac{1}{\nA/\nf-1} \,,            & \nA > \nf \,, \; \nf<\nl \, .
    \end{cases}
\ee
In Fig.~\ref{fig:Delta-zero-truncated}, we confirm this analysis in the overparameterized setting by comparing to a numerical evaluation of $\secF_0$. In this case, we find
a reasonably good fit: we see the expected deviations for very small $\nA \lesssim 10$, and we also see some
more 
deviations in the crossover region for larger $\alpha$ as compared to our fit in Fig.~\ref{fig:Delta-zero}; in this region, we expect that the higher-order spectrum-dependent terms are less negligible. 
Nevertheless, these deviations are overall insignificant as the precise crossover behavior is not particularly important and could always be improved by analyzing additional terms in the $\varepsilon$ expansion. 

We can also consider the somewhat less typical case of $\nf > \nl$, which is of interest in \S\ref{sec:break-down-of-neural-scaling-laws} where we considered the breakdown of neural scaling phenomenology. In this situation, for a typical fixed $\fw$, the spectrum of $\covf$ will have the complete power-law of $\covl$ plus the addition of $(\nf - \nl)$ 
zero eigenvalues to fill out the rest of the spectrum. Considering the self-consistent  equation for $\secF$, 
\eqref{eq:self-consistent-reprint},
it's clear that a similar analysis follows except that the sum now truncates at $\nl$:
\begin{equation}\label{eq:self-consistent-reprint-truncated}
    \secF(\nA,\nf) = \sum_{i=1}^{\nf} {\frac{\eig_i}{\ridge + \frac{\nA \eig_i}{1+\secF(\nA,\nf) } }} = \sum_{i=1}^{\nl} {\frac{\eig_i}{\ridge + \frac{\nA \eig_i}{1+\secF(\nA,\nf) } }} \, .
\end{equation}
This means that in this regime we have
\begin{equation}
    \secF(\nA,\nf) \to \secG(\nA,\nl) \, ,
\end{equation}
and the complete table of values for $\secF$ is given by
\be\label{eq:delta-zero-complete-projected}
    \secF_0(\nA,\nf) = 
    \begin{cases}
        \alpha + \frac{1}{\nf/\nA-1} \,,   & \nA < \nf \,, \;\; \nf<\nl \,, \\
        \frac{1}{\nA/\nf-1} \,,            & \nA > \nf \,, \;\; \nf<\nl \, ,\\
        \alpha + \frac{1}{\nl/\nA-1} \,,   & \nA < \nl \,, \;\; \nf>\nl \,, \\
        \frac{1}{\nA/\nl-1} \,,            & \nA > \nl \,, \;\; \nf>\nl \, .
    \end{cases}
\ee
Interestingly, though the self-consistent equation, \eqref{eq:self-consistent-reprint}, depends explicitly on 
$\fw$ through $\covf = \fw\covl\fw^T$,  we see here that $\secF_0(\nA,\nf)$ is actually \emph{independent} of the particular $\fw$, instead only caring about its size, $\nf$.

\mciteSetMidEndSepPunct{}{\ifmciteBstWouldAddEndPunct.\else\fi}{\relax}
\bibliographystyle{utphys}
\bibliography{scaling}{}

\end{document}